\documentclass[lettersize,journal]{IEEEtran}
\usepackage{amsmath,amsfonts}
\usepackage{algorithmic}
\usepackage{algorithm}
\usepackage{array}
\usepackage[caption=false,font=normalsize,labelfont=sf,textfont=sf]{subfig}
\usepackage{textcomp}
\usepackage{stfloats}
\usepackage{url}
\usepackage{verbatim}
\usepackage{graphicx}
\usepackage{cite}
\usepackage[switch]{lineno}
\usepackage{csquotes}
\usepackage{multirow}
\usepackage{xurl}
\usepackage{tikz}
\usepackage{array}
\newcolumntype{P}[1]{>{\centering\arraybackslash}p{#1}} % Centered fixed width column
\usepackage[hidelinks]{hyperref}
\usepackage{xcolor}
\definecolor{caribbeangreen}{rgb}{0.0, 0.8, 0.6}
\definecolor{coralred}{rgb}{1.0, 0.25, 0.25}
\definecolor{royalblue(web)}{rgb}{0.25, 0.41, 0.88}
\definecolor{jade}{rgb}{0.0, 0.66, 0.42}
\definecolor{jasper}{rgb}{0.84, 0.23, 0.24}

\newcommand\eg{\textit{e}.\textit{g}.}
\newcommand\ie{\textit{i}.\textit{e}.}
\newcommand\cf{\textit{c}\textit{f}.}

\newcommand\Mat[1]{{\mathbf{#1}}}
\newcommand\Vect[1]{{\mathbf{#1}}}

\newcommand\entropyII[1]{\mathcal{S}_{\uppercase\expandafter{\romannumeral 2\relax}}\left(#1\right)}

\newcommand\mrae{{\mathrm{\mathbf{MRAE}}}}
\newcommand\rmse{{\mathrm{\mathbf{RMSE}}}}
\newcommand\psnr{{\mathrm{\mathbf{PSNR}}}}
\newcommand\sam{{\mathrm{\mathbf{SAM}}}}

\begin{document}

\title{Limitations of Data-Driven Spectral Reconstruction:\\An Optics-Aware Analysis}

\author{Qiang Fu*, Matheus Souza*, Eunsue Choi, Suhyun Shin, Seung-Hwan Baek, Wolfgang Heidrich~\IEEEmembership{Fellow,~IEEE}
        % <-this % stops a space
\thanks{Q. Fu, M. Souza, and W. Heidrich are with King Abdullah University of Science and Technology (KAUST), Thuwal, Saudi Arabia. E. Choi, S. Shin, and S. Baek are with Pohang University of Science and Technology (POSTECH), Pohang, Korea. Corresponding author: Q. Fu (qiang.fu@kaust.edu.sa). 
This paper has supplementary downloadable material 
%available at http://ieeexplore.ieee.org, 
provided by the author. The material includes additional experimental results to provide further evidence for the major findings in this work. This material is 14.2 MB in size.
}
        % <-this % stops a space
\thanks{* Joint first authors.}
%\thanks{Manuscript received XXX xx, 2025; revised XXX xx, 2025.}
}

%\markboth{Transactions on Computational Imaging,~Vol.~xx, No.~x, XXX~2025}%
\markboth{}%
{Fu \MakeLowercase{\textit{et al.}}: Limitations of Data-Driven Spectral Reconstruction}

% \IEEEpubid{0000--0000/00\$00.00~\copyright~2025 IEEE}

\maketitle

% \linenumbers

\begin{abstract}
Hyperspectral imaging empowers machine vision systems with the
distinct capability of identifying materials through recording their
spectral signatures. Recent efforts in data-driven spectral
reconstruction aim at extracting spectral information from RGB images
captured by cost-effective RGB cameras, instead of dedicated
hardware. Published work reports exceedingly high numerical
scores for this reconstruction task, yet real-world performance lags
substantially behind.

In this paper we {\em systematically analyze} the performance of
such methods with three groups of dedicated experiments. First, we
evaluate the practical overfitting limitations with respect to current
datasets by training the networks with less data, validating the
trained models with unseen yet slightly modified data, and
cross-dataset validation. Second, we reveal {\em fundamental
limitations} in the ability of RGB to spectral methods to deal with
metameric or near-metameric conditions, which have so far gone largely
unnoticed due to the insufficiencies of existing datasets. We achieve
this by validating the trained models with metamer data generated by
metameric black theory and re-training the networks with various forms
of metamers. This methodology can also be used for data augmentation
as a partial mitigation of the dataset issues, {\em although the RGB
to spectral inverse problem remains fundamentally ill-posed}.

Finally, we analyze the potential for modifying the problem setting to
achieve better performance by exploiting some form of optical encoding
provided by either incidental optical aberrations or some form of
deliberate optical design. Our experiments show that such approaches
do indeed provide improved results under certain circumstances,
however their overall performance is limited by the same dataset
issues as in the plain RGB to spectral scenario. We therefore
conclude that {\em future progress on snapshot spectral imaging will
heavily depend on the generation of improved datasets which can then
be used to design effective optical encoding strategies}.
Code can be found at \url{https://github.com/vccimaging/OpticsAwareHSI-Analysis}. 
\end{abstract}


%---------- main texts ----------%
\section{Introduction}
\label{sec:intro}

\IEEEPARstart{H}yperspectral imaging is a method that involves recording the light in a scene in the form of many, relatively narrow, spectral bands, rather
than projected into three broadband RGB color channels. Where
RGB imaging utilizes the trichromaticity theory of human color vision,
spectral imaging provides additional information that can help
discriminate between different materials and lighting conditions that
are hard to tell apart in RGB images.
For example, red stains in a crime scene could be blood, or paint, or a dyed cloth, which cannot be distinguished from their RGB colors. Skin tumors could not be diagnosed from surrounding tissues of the same color. It is difficult to spot and sort out plastic leaves from living plants by their greenish colors. Therefore, spectral imaging has been applied in many fields, including computer graphics~\cite{kim20133d}, machine vision~\cite{fauvel2012advances,rangnekar2022semi}, healthcare~\cite{lu2014medical}, agriculture~\cite{dale2013hyperspectral}, and environment~\cite{banerjee2020uav}, to name just a few.

However, conventional hyperspectral cameras require scanning mechanisms~\cite{lodhi2019hyperspectral,huang2022spectral} to acquire the 3D hyperspectral datacube with 2D sensors. To simplify the demanding hardware, extensive efforts have been made in the development of various snapshot hyperspectral cameras~\cite{wagadarikar2008single,saragadam2019krism,jeon2019compact}.

On the extreme end of these hardware simplification efforts, deep learning methods have emerged in recent years that attempt to solve the problem entirely in software by reconstructing spectral data from RGB images (RGB2HS). This has resulted in three CVPR-hosted NTIRE challenges~\cite{arad2018ntire,arad2020ntire,arad2022ntire} and various network architectures~\cite{shi2018hscnn+,li2020adaptive,zhao2020hierarchical,cai2022mst++,zhang2022survey}. Yet it remains unclear how these methods generalize to unseen data, how they deal with the difficult but important problem of resolving metamerism~\cite{hill1999color,ortega2020hyperspectral}, and how they depend on the optical system of both the RGB source and the spectral cameras used to capture the datasets.

At its core, estimating spectral information from RGB colors is an under-determined one-to-many mapping problem. As stated By Pharr et al.~\cite{pharr2023physically} (Ch.~4), ``any such conversion is inherently ambiguous due to the existence of metamers''. Intuition would therefore suggest that the achievable spectral fidelity of RGB to spectral methods is limited. However, this intuition flies in the face of very high numerical test results reported in recent NTIRE challenges~\cite{arad2018ntire,arad2020ntire,arad2022ntire}.

One possible explanation for the experimental success of spectral reconstruction from an RGB image is that the networks learn to exploit spatial structures, or scene semantics, to estimate spectral information. However, spectral images are usually employed when RGB images do not provide sufficient information for downstream tasks. Therefore, it is questionable to use scene semantics to resolve spectral ambiguities. Instead, in computational imaging, the idea is to {\em measure} the spectral information to help better understand the scene semantics, particularly in difficult scenarios. Clearly, these two methodologies feature reversed information flow. We argue that the computational imaging approach is more compatible with spectral imaging itself.

\begin{figure}[!htp]
\centering
\includegraphics[width=\columnwidth]{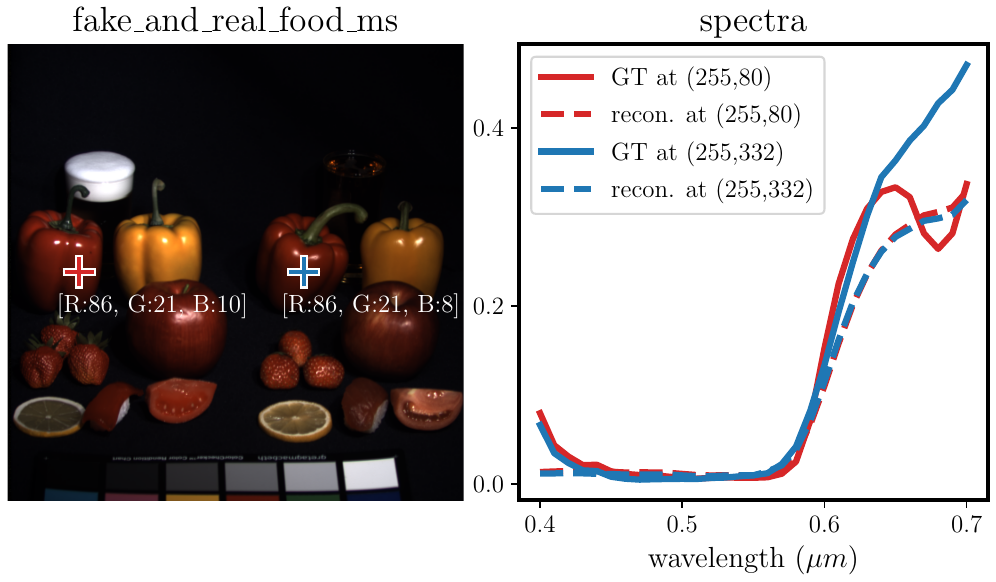}
\caption{
%\WH{Could you clean up the figure a bit? Too many different fonts in the plots vs, the labels; it doesn't look tidy.}
An example scene {fake\_and\_real\_food\_ms} from the CAVE dataset~\cite{yasuma2010generalized} consists of objects with visually similar colors, but actually different spectra. Left: Color image with highlighted points on the red peppers. Their RGB values are nearly the same. Right: Ground-truth and reconstructed spectra at the corresponding points show their spectral differences. The reconstructed spectra are predicted by the pre-trained MST++ model~\cite{cai2022mst++} on the ARAD1K dataset~\cite{arad2022ntire}. The neural network struggles to distinguish either the two spectra from each other, or from their true spectra.}
\label{fig:cave_metamers}
\end{figure}

In practice, metameric or near-metameric colors often occur in situations where the spatial structure is also similar, \eg, vein finding (the global geometry is a segment of a forearm, but vein structure is unknown), or biometrics (\eg, distinguishing real faces from masks or images). To illustrate this effect, we show an example from the smaller and older CAVE dataset~\cite{yasuma2010generalized} where several fake and real objects are presented in two groups, as shown in Fig.~\ref{fig:cave_metamers}. Two sample points from the two red peppers (one real and the other fake, but it's unclear which is which) have the RGB values (86, 21, 10) and (86, 21, 8), respectively. As can be seen, the real spectra of the real and fake red peppers differ substantially in the red part of the spectrum.  A pre-trained model (arbitrarily chosen from previous NTIRE challenges trained on the much larger and newer ARAD1K dataset~\cite{arad2022ntire}) predicts spectra that differ from the ground truth, but more importantly, the predicted spectra for the real and fake pepper are almost identical, illustrating the failure to adequately deal with metamers.

In this paper, we take a systematic and outside-the-box look at all the above aspects. To the best of our knowledge, we are the first to analyze, document, and discuss the inherent shortcomings of this research theme. We highlight realistic conditions under which recent efforts fall short, aiming to constructively instigate, debate, provide insights, and forge a new path regarding the physical phenomena that have been overlooked. By conducting a series of adversarial attacks and thorough analysis, we reveal a number of shortcomings in both current datasets and reconstruction methods. Specifically, we find that:
\begin{itemize}
\item Existing hyperspectral image datasets severely lack in diversity especially with respect to metameric colors but also other factors including nuisance parameters such as noise and compression ratios.
\item State-of-the-art methods suffer from {\em atypical} overfitting problems that arise from various factors in the image simulation pipeline, such as noise, RGB data format, and lack of optical aberrations.
\item Optical aberrations in RGB images, while currently ignored by all methods, are actually {\em beneficial} rather than harmful to spectral reconstruction \emph{if modeled accurately}.
\item Crucially, the limitations of the datasets that we document not only affect the RGB to spectral work, but also any other spectral reconstruction and processing that uses the same training data~\cite{jacome2023middle,xu2024wavelength,arguello2021shift}. We show how \emph{metameric augmentation} can be used to at least partially overcome the dataset issues.
\end{itemize}

A seemingly apparent observation from the results we show in this paper reinforces that it is impossible to distinguish metamers solely from RGB colors. Remarkably, this fundamental limitation has been largely overlooked within the research in this field. Through the evidence in this work, we contribute to a deepened understanding of the limitations of current datasets as well as of underlying sources that result in the limitations of spectral reconstruction accuracy. The results of the interplay between metameric spectra and optical aberrations open the door for new approaches for spectral recovery down the road.

\section{Related Work}
\label{sec:related}

%-------------------------------------------------------------------------
\subsection{Hyperspectral cameras}
Conventional hyperspectral imaging systems require filter wheels, liquid-crystal tunable filters, or mechanical motion (\eg, pushbroom)~\cite{lodhi2019hyperspectral,huang2022spectral} to scan the 3D hyperspectral datacube. To enable snapshot acquisition, coded-aperture snapshot spectral imager (CASSI)~\cite{wagadarikar2008single,choi2017high} has been proposed to achieve high spectral accuracy using spectrum-dependent coded patterns. Based on this hardware architecture, supervised learning (such as TSA-Net~\cite{meng2020end}, BiSRNet~\cite{cai2023binarized}, VmambaSCI~\cite{zhang2024vmambasci}, SpeCAT~\cite{yao2024specat}) and unsupervised learning (such as LRSDN~\cite{chen2024hyperspectral}, SAH-SCI~\cite{zeng2024sah}, CEINR~\cite{mei2024progressive}) algorithms have been proposed in recent years to address the inverse problem on hyperspectral datasets. Other variants, such as dual-camera CASSI~\cite{wang2016adaptive} and reconstruction algorithms have also been proposed. Another category of emerging methods also exploit spectrally encoded point spread functions (PSFs) to computationally reconstruct a hyperspectral image~\cite{jeon2019compact,baek2017compact,cao2011prism}. Various DOE designs for optimal spectral PSFs, such as equalization DOE~\cite{xu2023snapshot}, non-serial quantization-aware deep optics~\cite{wang2024non}, tunable phase encoding~\cite{zhang2025tunable}, Double-DOE~\cite{urrea2024dodo}, have been proposed along with corresponding reconstruction algorithms over the past few years. In general, great efforts have been made to simplify hyperspectral camera hardware by software reconstruction.

%-------------------------------------------------------------------------
\subsection{Spectral reconstruction from RGB images}
A recent trend to solve the snapshot hyperspectral imaging problem is to exploit hyperspectral data with deep neural networks to reconstruct spectral information from RGB images~\cite{arad2016sparse}. Owing to the wide availability of RGB cameras, this approach seems to be a promising candidate for hyperspectral imaging if successful. A large number of neural network architectures have been proposed in the past three NTIRE spectral recovery challenges ~\cite{arad2018ntire,arad2020ntire,arad2022ntire} and other venues afterwards. Our analysis in this paper focuses on this class of methods to gain insights on their strengths and limitations. In particular, we comprehensively evaluate 17 open-sourced neural networks to date. HSCNN+~\cite{shi2018hscnn+} is one of the first networks that employs CNN as the backbone for spectral reconstruction which won the first challenge in 2018~\cite{arad2018ntire}. It features dense residual blocks (HCNN-R) and densely-connected structures (HCNN-D). EDSR~\cite{lim2017enhanced} introduces an enhanced deep super-resolution network by removing unnecessary modules in conventional residual networks. It was originally designed for single-image super-resolution, and later extended for spectral reconstruction. HRNet~\cite{zhao2020hierarchical} employs a Hierarchical Regression Network that consists of 4 levels followed by PixelShuffle layers for inter-level interaction, followed by a residual dense block and a residual global block to reconstruct the hyperspectral images.  AWAN~\cite{li2020adaptive} utilizes a backbone stacked with multiple dual residual attention blocks for dual residual learning. The sensor response function is used as a finer constraint to improve the reconstruction quality. MIRNet~\cite{zamir2020learning} adopts a multi-scale residual block that learns contextual information from multiple scales to enhance spatial resolution for image restoration tasks, and later extended for spectral reconstruction. HINet~\cite{chen2021hinet} proposes a Half Instance Normalization Block that was originally designed to boost image restoration networks, and later extended for spectral reconstruction. MPRNet~\cite{zamir2021multi} is a multi-stage network to learn the projection from degraded measurements to the high-quality images, with a couple of manageable steps. It can also be extended to hyperspectral reconstruction with proper modifications.  HDNet~\cite{hu2022hdnet} proposes a dual domain learning network with a spatial-spectral attention module for pixel-level features, and a frequency domain learning to narrow the frequency domain discrepancy. Restormer~\cite{zamir2022restormer} was also designed for image restoration tasks initially and then extended for spectral reconstruction, with a focus on designing the building block with the Transformer architecture to capture long-range pixel interactions. MST~\cite{cai2022mask} and MST++~\cite{cai2022mst++} are Transformer-based networks that employ spectral-wise multi-head self-attention to fully make use of spatial sparsity and spectral self-similarity for efficient spectral reconstruction in a coarse-to-fine manner. HySAT~\cite{wang2023learning} employs an exhaustive correlation Transformer to simultaneously model spectral-wise similarity with a token-independent mapping mechanism and particularity with a spectral-wise re-calibration mechanism. HRPN~\cite{wu2023hprn} integrates comprehensive multisource priors, in particular the semantic prior of RGB inputs, to regularize and optimize the solution space with a Transformer-based holistic prior-embedded relation network. SSRNet~\cite{dian2023spectral} employs a model-guided network based on cross fusion that uses the image formation model and the sensor spectral response function to guide the training of a CNN-backed network. SSTHyper~\cite{xu2024ssthyper} introduces a sparse spectral transformer model to learn shallow and deep spatial-spectral priors and allows adaptive masking of non-significant details. Computational cost is reduced by a cross-level fusion network architecture. MSFN~\cite{wu2024multistage} is a multi-stage UNet structure that captures both spatial and spectral features in a multiscale manner. A feature alignment scheme is proposed to preserve spatial correlations and spectral self-similarities. GMSR~\cite{wang2024gmsr} builds on a more recently developed architecture Mamba~\cite{gu2023mamba} as the backbone and develops a lightweight model for global feature representation. Spatial gradient attention and spectral gradient attention are proposed to improve the spectral reconstruction. 

%-------------------------------------------------------------------------
\subsection{Multispectral and hyperspectral image fusion}
Another class of related spectral reconstruction methods is to fuse images with low spatial resolution but high spectral resolution with images that have high spatial resolution but low spectral resolution~\cite{dian2021recent,vivone2023multispectral}. Different from spectral reconstruction from RGB images, it requires two inputs and a final image with high spatial and spectral resolution is obtained. In recent years, neural networks have also been extensively employed in solving the image fusion problem. Similar datasets as well as remote sensing images are usually used to evaluate the performance. CNN and Transformer are popular backbones in the design of such networks~\cite{xie2019multispectral,cao2024unsupervised,liang2024fourier,cao2024unsupervisedmulti,yan2025spatial}. In this paper, however, we focus on data-driven methods for hyperspectral recovery from single RGB images, where the limitations arising from existing datasets are more prominent.

%-------------------------------------------------------------------------
\subsection{Dataset bias and data augmentation}

Deep neural networks are prone to suffer from data bias~\cite{torralba2011unbiased,fabbrizzi2022survey} and overfitting problems~\cite{bejani2021systematic}. Overfitting can lead to the inability of trained models to generalize in real-world applications~\cite{kornblith2019better}. Although overfitting can sometimes be detected by inspecting the training and validation performance over the course of training, it can often be imperceivable in challenging problems. A useful technique to detect overfitting is to use adversarial examples~\cite{werpachowski2019detecting} generated from the original dataset. On the other hand, it is important to address overfitting when the amount of data is limited. Data augmentation~\cite{shorten2019survey,rebuffi2021data} techniques are usually employed to improve the robustness of deep neural networks.

%-------------------------------------------------------------------------
\subsection{Metamerism}
Metamerism is a physical phenomenon where distinct spectra produce the same color~\cite{akbarinia2018color} as the high-dimensional spectral space is projected down to three dimensions of a trichromatic vision system (either the human eye or an RGB camera). This phenomenon has been studied in color science~\cite{thornton1998strong,foster2006frequency}, spectral rendering~\cite{jakob2019low,weidlich2021spectral,van2023metameric}, and hyperspectral imaging~\cite{hill1999color,foster2019hyperspectral}. In hyperspectral imaging, it is crucial in many applications to distinguish between metamers or near-metamers (\ie, different spectra that project to {\em similar} RGB values)~\cite{cukierski2010metamerism,ortega2020hyperspectral}. Indeed spectral imaging is usually employed when the RGB color differences between two materials or features are too small to reliably distinguish between them. Therefore, hyperspectral imaging systems require special attention in the system design to acquire accurate spectral signatures~\cite{hill2002optimization,ji2023mhealth}.

However, modeling metamerism is challenging in data-driven spectral reconstruction as metamerism is hard to capture since they are relatively rare in everyday environments~\cite{foster2006frequency}, although they are vital for many applications of spectral imaging. Previous works make use of illumination and camera spectral response as means of providing additional information to help improve the spectral reconstruction. For example,  Fu et al.~\cite{fu2020joint} propose to select optimal camera response functions from a dataset and recover hyperspectral images with a CNN-based network. This is equivalent to observer metamerism~\cite{fascione2024study} where different cameras observe the spectra differently. Another technique is to explore the structure of illuminant space~\cite{akbarinia2018color}. For example, Baek et al.~\cite{baek2021single} employ 29 CIE standard illuminants to augment the hyperspectral dataset for joint hyperspectral and depth reconstruction. Cao et al.~\cite{cao2024unsupervisedrgb} develop an unsupervised network to recover spectral information under two lighting conditions. Although such methods enrich the spectral content of hyperspectral datasets, they model metamersim indirectly. Our work differs from these techniques since we directly generate metamers using the metameric black theory~\cite{van1994metameric,finlayson2005metamer,zhang2016metamer}. This guarantees the underrepresented metamerism phenomena (the same color from different spectra) in existing datasets can be modeled more effectively.

\section{Fundamentals}
\label{sec:fundamentals}

\begin{figure*}[!htp]
\includegraphics[width=0.98\textwidth]{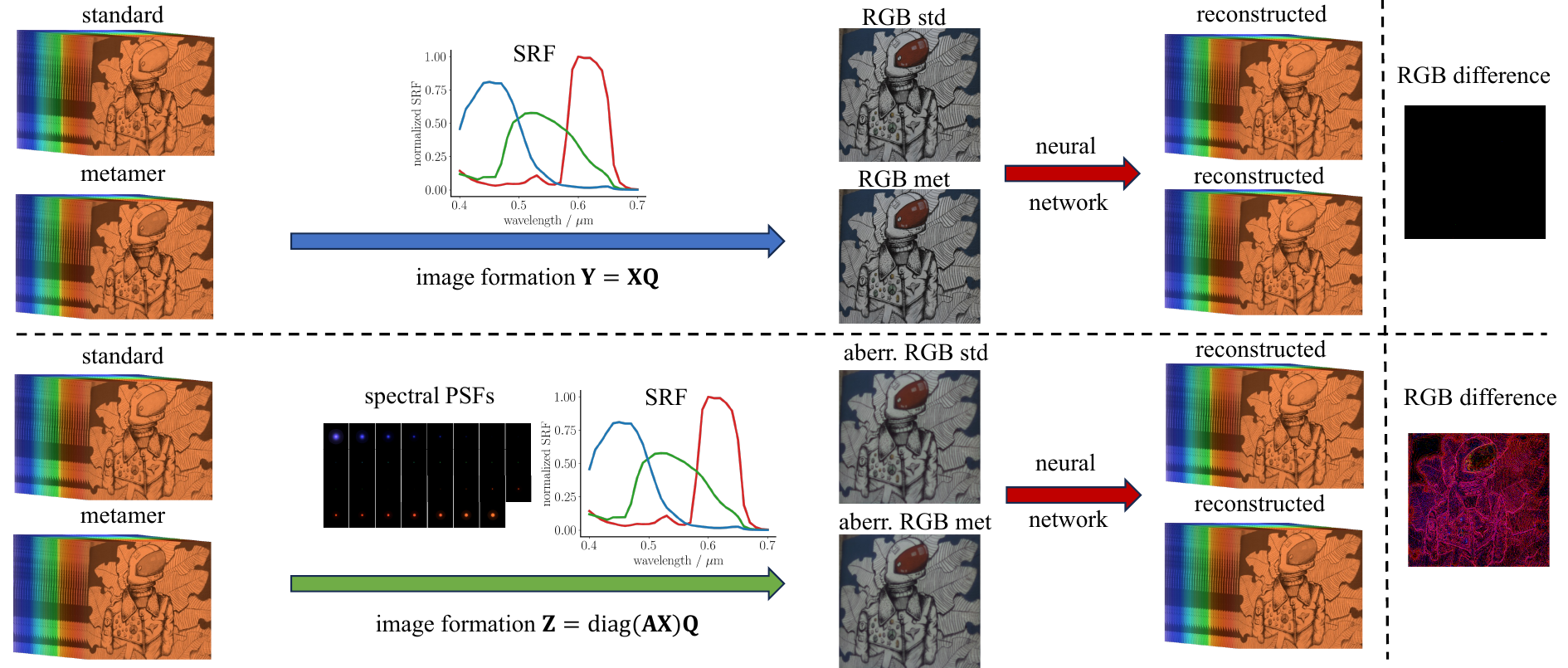}
\caption{Spectral image formation models used in the analysis in this work. Top: In the NTIRE spectral recovery challenges, an RGB image is considered as a linear projection from a high-dimensional hyperspectral datacube to a 3D color image. The existence of metamerism results in identical RGB images for different spectra. The neural network trained in this way cannot distinguish their corresponding spectra. Bottom: A possible mitigation to this problem is to include the optical aberrations of the lens in the image formation model. Spectral information is encoded into the aberrated RGB images, enabling the neural network to tell the difference between metamers. In both cases, the RGB image differences are shown on the right (intensity enhanced for better visualization).}
  \label{fig:image_formation}
\end{figure*}

%-------------------------------------------------------------------------
\subsection{Spectral image formation}

We denote the hyperspectral image as a matrix $\Mat{X} \in \mathbb{R}^{MN \times K}$, where $M, N$ are the number of pixels, and $K$ is the number of spectral bands. Note that we model the spectral radiance here, not spectral reflectance. Illumination spectrum is included. We have stacked the 2D spatial dimensions in rows of $\Mat{X}$. The spectral response function (SRF) of the camera can be expressed as a matrix $\Mat{Q} \in \mathbb{R}^{K \times 3}$. Therefore, the spectrum-to-color projection results in a color image
\begin{equation}
    \Mat{Y} = \Mat{X} \Mat{Q},
    \label{eq:color_formation}
\end{equation}
where $\Mat{Y} \in \mathbb{R}^{MN \times 3}$. This is the color formation model in the NTIRE 2022 challenge~\cite{arad2022ntire}. The inverse problem is to recover $\Mat{X}$ from $\Mat{Y}$.

In the past NTIRE challenges~\cite{arad2018ntire,arad2020ntire,arad2022ntire}, optical aberrations have not been included in the image simulation pipeline. However, the optical system of the RGB camera inevitably introduces spectrally-varying blurs to the spectral images, which is modeled as PSFs. This optical process can be described by a linear matrix-vector product in each spectral band followed by a sum over the spectral dimension. The spectral images through the optical system are 
$\Mat{W} = \mathrm{diag} \left( \Mat{A} \Mat{X} \right)$, where $\Mat{A} \in \mathbb{R}^{K MN \times MN}$ is a block matrix that stacks the spectral PSF matrices vertically, and $\mathrm{diag}(\cdot)$ extracts the diagonal blocks. The final RGB image is then
\begin{equation}
    \Mat{Z} = \mathrm{diag} \left( \Mat{A} \Mat{X} \right) \Mat{Q},
    \label{eq:image_formation_matrix}
\end{equation}
where $\Mat{Z} \in \mathbb{R}^{MN \times 3}$. With the optical image formation model accounted for, the inverse problem is to recover $\Mat{X}$ from $\Mat{Z}$. It is evident that the optical property in $\Mat{A}$ spreads the spectral information to the RGB channels, offering side-channel information to help spectral reconstruction. See the Appendix for the full derivation.

The two spectral image formation models are illustrated in Fig.~\ref{fig:image_formation}. Without considering optical aberrations (Eq.~\eqref{eq:color_formation}), the neural network struggles to reconstruct the real hyperspectral images in the presence of metamers. The spectrally-varying PSFs (Eq.~\eqref{eq:image_formation_matrix}) are helpful to mitigate this issue since the aberrated RGB images from metamers are different. In the following sections, we will discuss the limitations of existing data-driven spectral reconstruction based on these two image formation models in detail.

%-------------------------------------------------------------------------
\subsection{Hyperspectral datasets and data diversity}

Compared to very large color (RGB) image datasets (\eg, ImageNet \cite{deng2009imagenet}, DIV2K~\cite{agustsson2017ntire}), hyperspectral datasets are far smaller in size, primarily limited by the unavailability of high-quality hyperspectral cameras and the difficulty in acquiring outside the lab with moving target scenes. The largest dataset so far is ARAD1K used in the NTIRE 2022 challenge~\cite{arad2022ntire}. In addition, we also include the CAVE~\cite{yasuma2010generalized}, ICVL~\cite{arad2016sparse}, and KAUST~\cite{li2021multispectral} datasets that share the same spectral range (400~nm to 700~nm) to extend our experiments. The datasets are summarized in Table~\ref{tab:datasets}. Although other datasets, such as Harvard~\cite{chakrabarti2011statistics}, KAIST~\cite{DeepCASSI:SIGA:2017}, and TokyoTech~\cite{monno2015practical} exist, they cover slightly different spectral bands (420~nm to 720~nm), making it difficult to directly compare and cross-validate results among different datasets. We therefore restrict our analysis to the datasets listed in the table. 

\begin{table*}[!htp]
\caption{Basic information of four existing hyperspectral datasets.}
\centering
\setlength{\tabcolsep}{12pt}
\begin{tabular}{l|l|l|c|c|l}
\hline
Dataset  & Spectra (nm) & Resolution (x, y, $\lambda$)   & Amount & Device                              & Scene   \\ \hline
CAVE~\cite{yasuma2010generalized}     & 400:10:700          & 512 $\times$ 512 $\times$ 31   & 32     & monochrome sensor + tunable filters & lab setup    \\
ICVL~\cite{arad2016sparse}     & 400:10:700          & 1392 $\times$ 1300 $\times$ 31 & 201    & HS camera (Specim PS Kappa DX4)     & outdoor \\
KAUST~\cite{li2021multispectral}    & 400:10:700          & 512 $\times$ 512 $\times$ 31   & 409    & HS camera (Specim IQ)               & outdoor \\
ARAD1K~\cite{arad2022ntire} & 400:10:700          & 482 $\times$ 512 $\times$ 31   & 1000   & HS camera (Specim IQ)               & outdoor \\ \hline
\end{tabular}
\label{tab:datasets}
\end{table*}

The difficulties in the data capture not only affect the size but also the {\em diversity} of the datasets. In particular, effects like metamerism, which are comparatively rare in everyday environments~\cite{foster2006frequency,prasad2015metrics}, yet crucial for many actual applications of spectral imaging, are under-represented in the datasets. While the CAVE dataset~\cite{yasuma2010generalized} contains some fake-and-real pairs of objects (\eg, Fig.~\ref{fig:cave_metamers}) to account for metamerism, the total amount of such data is still very low. We analyze the general data diversity issue in Section~\ref{sec:overfitting} and the specific case of metamerism in Section~\ref{sec:metamer_failure}.

%-------------------------------------------------------------------------
\subsection{Modeling metamerism}
Since there are not enough examples of metamerism in existing datasets, their effects in spectral reconstruction went unnoticed in prior works. On the one hand, we need adversarial examples to reveal the unexplored problems. On the other hand, we want to investigate how they can complement the current datasets. Therefore,  we propose a new form of data augmentation in our experiments. {\em Metameric augmentation} starts with existing spectral images and creates a new, {\em different} spectral image that however maps to the same RGB image (given a specific set of RGB spectral response functions). In Section~\ref{sec:metamer_failure}, we first use metameric augmentation as an adversarial example to reveal the previously omitted effects of metamerism on the performance discrepancy. In Section~\ref{sec:aberration_advantage}, we show metameric augmentation is beneficial to mitigate the raised problems along with an aberration-aware training strategy.
 
Note that data augmentation has proven to be effective in deep learning to mitigate data shortage. Color image augmentation techniques have been focusing mainly on geometric transformations and intensity adjustment. Although these techniques have been employed in prior methods, they only augment the spatial dimensions in hyperspectral images. To the best of our knowledge, metameric augmentation beyond RGB colors, accounting for the physical phenomenon of metamerism, has not been adopted before in spectral reconstruction problems. Metameric augmentation can greatly enrich the \textbf{spectral content} in existing datasets to mitigate the lack of diversity.

Interestingly, metamer generation from existing spectra has been studied in color science and spectral rendering to accurately model the scenes using various methods, \eg, metameric black~\cite{van1994metameric,finlayson2005metamer,alsam2007calibrating} and spectral uplifting~\cite{jakob2019low,van2023metameric,belcour2023one}. In this work, to support our analysis, we adopt the metameric black approach to generate metamers, whereas we note that other metamer generation methods can also be employed for the same purpose. 

A spectrum $\Vect{S}$ can be projected onto two orthogonal subspaces, one for the fundamental metamer $\Vect{S}^{*}$, and the other for metameric black $\Vect{B}$~\cite{cohen1982metameric,vienot2014verriest}. That means the original spectrum can be decomposed as $\Vect{S} = \Vect{S}^{*} + \Vect{B}$. The fundamental metamer is a particular solution to Eq.~\eqref{eq:color_formation}, and the metameric black always leads to zero tristimulus, \ie, no impact on the color appearance. The set of all possible metamers is called a metamer set~\cite{finlayson2005metamer}. Wyszecki~\cite{wyszecki1958evaluation} first introduces a decomposition technique to calculate the metameric black. To generate new metamers, it is possible to add a linear combination of metameric blacks to the fundamental metamer~\cite{finlayson2005metamer}. In a linear algebra perspective, metameric blacks lie in the null space (or kernel) of the camera SRF. Any scalar multiplication to a vector in the null space remains in the null space according to the scalar multiplication property~\cite{strang2022introduction}. Inspired by the metameric black theory and the mathematical properties, we propose a simple yet effective way to generate metamers by scaling the metameric black component. A new metameric spectrum $\Vect{S}'$ is then
\begin{equation}
    \Vect{S}' = \Vect{S}^{*} + \alpha \Vect{B},
    \label{eq:metameric_black}
\end{equation}
where $\Vect{S}^{*} = \Mat{Q} \left(\Mat{Q}^{T} \Mat{Q}\right)^{-1}
\Mat{Q}^{T} \Vect{S}$ and $\Vect{B} = \Vect{S} - \Vect{S}^{*}$. Since
adding metameric black does not alter the RGB color, we can vary the coefficient $\alpha$ to generate different spectra that are all metamers. To avoid negative spectral radiance, we clip the negative values in the generated spectra and re-calculate the RGB colors for the affected pixels. See the Appendix for the analysis on the effects of clipping to non-negative values while generating metamer data.

%-------------------------------------------------------------------------
\subsection{Performance evaluation metrics}

Consider a hyperspectral image $\Mat{X}_{k,i,j}$ and its estimate $\hat{\Mat{X}}_{k,i,j}$, where $k$ is the spectral index, and $i, j$ are spatial indices. The reconstruction quality can be evaluated in various ways. The NTIRE 2022 spectral reconstruction challenge~\cite{arad2022ntire} adopts two numerical metrics, Mean Relative Absolute Error (MRAE), 
\begin{equation}
    \mrae = \frac{1}{KMN} \sum_{k,i,j} \left|\hat{\Mat{X}}_{k,i,j} - \Mat{X}_{k,i,j} \right| / \Mat{X}_{k,i,j},
    \label{eq:mrae}\\
\end{equation}
and Root Mean Square Error (RMSE)
\begin{equation}
    \rmse = \sqrt{\frac{1}{KMN} \sum_{k,i,j} \left(\hat{\Mat{X}}_{k,i,j} - \Mat{X}_{k,i,j} \right)^2}.
    \label{eq:rmse}
\end{equation}

Another metric widely used in hyperspectral imaging is the Spectral Angle Mapper (SAM)~\cite{van2006effectiveness,kuching2007performance,park2007contaminant}, although it has not yet found its way into the relevant computer vision literature. SAM emphasizes the spectral accuracy compared to the previous metrics, which are more forgiving of large errors in individual spectral channels:
\begin{equation}
    \sam = \frac{1}{MN} \sum_{i,j} \cos^{-1} \left( \frac{\sum_{k} \hat{\Mat{X}}_{k,i,j} \Mat{X}_{k,i,j}}{\sqrt{\sum_{k} \hat{\Mat{X}}_{k,i,j}^2 \sum_{k} \Mat{X}_{k,i,j}^2}}\right).
    \label{eq:sam}
\end{equation}

Finally, we also inspect the spatial quality in individual spectral channels, and calculate the spectrally averaged Peak Signal-to-Noise Ratio (PSNR),
\begin{equation}
    \psnr = \frac{1}{K} \sum_{k} 20 \log_{10} \left( \frac{\textrm{MAX}}{\sqrt{\textrm{MSE}_{k}}} \right),
    \label{eq:psnr}
\end{equation}
where $\textrm{MAX}$ is the maximum possible value, and
$\textrm{MSE}_k$ is the mean squared error in the $k$-th spectral band. This metric complements RMSE to account for performance variation in individual spectral bands.

\subsection{Training details}
In our study, we conduct all the experiments across various datasets and network architectures. Following the methodologies proposed by the latest champion network MST++~\cite{cai2022mst++}, we employ their patch-wise training approach (patches of 128$\times$128 pixels). 

\subsubsection{Data Preparation}
Throughout the experiments in this work, we follow the same data format (Matlab-compatible \texttt{mat} files) for hyperspectral images in the ARAD1K dataset~\cite{arad2022ntire}. To be consistent, we also convert the raw hyperspectral datacubes in the CAVE~\cite{yasuma2010generalized}, ICVL~\cite{arad2016sparse}, and KAUST~\cite{li2021multispectral} datasets to this format. The data values are normalized by their respective bit-depths such that the data range is [0.0, 1.0]. The training and validation sets in ARAD1K are kept the same as offered in the NTIRE 2022 spectral recovery challenge~\cite{arad2022ntire}, \ie, 900 files for training, and 50 for validation. We split the CAVE, ICVL, and KAUST datasets by 90\% for training, and 10\% for validation. Following the training strategy in MST++~\cite{cai2022mst++}, we keep the training and validation lists fixed.

\subsubsection{Metamer Generation}
We adopt the metameric black method~\cite{van1994metameric,finlayson2005metamer,alsam2007calibrating} to generate metamers from the original hyperspectral data. By varying the coefficient of the metameric black term, we could generate metamers that project to the same RGB color. Note that the original datacube corresponds to $\alpha = 1$, and the fundamental metamer corresponds to $\alpha = 0$. For the experiments with fixed metamers, we use the fundamental metamers to complement the original standard data. This is sufficient to demonstrate our findings. Other arbitrary values would result in the same conclusions. A more aggressive setting is to vary $\alpha$ as a variable to account for the infinite possible metamers in a more realistic situation.

\subsubsection{Training and Validation Procedure}
Following the training strategies in MST++~\cite{cai2022mst++}, we sub-sample the hyperspectral datacubes and the corresponding RGB images into overlapping patches of 128$\times$128. Spatial augmentations, such as random rotation, vertical flipping, and horizontal flipping are randomly applied to the training patches. In the validation step, we calculate the evaluation metrics (MRAE, RMSE, PSNR, and SAM) on the full spatial resolution for the ARAD1K (482$\times$512), CAVE (512$\times$512), and KAUST (512$\times$512) datasets. Note that this is different from MST++~\cite{cai2022mst++}, where only the central 256$\times$256 regions are evaluated. The ICVL dataset has a very large spatial resolution (1300$\times$1392). To be consistent with other datasets, we evaluate ICVL only in the central 512$\times$512 regions.

Similar as MST++, in each epoch, we train the networks for 1000 iterations, with a total number of 300 epochs. All the reported results are evaluated at the end of the training epochs, \ie, 300k iterations. We find that the training iterations are sufficient to achieve convergence in all our proposed experiments. Hyperparameters, such as learning rate and batch size, are tuned to achieve the best performance for each network on each dataset. All the experiments are conducted on an NVIDIA A100 GPU (80~GB memory).
\section{Finding 1: Atypical Overfitting}
\label{sec:overfitting}

Although it is well-known that deep neural networks may suffer from overfitting problems, we find that the overfitting behavior in spectral reconstruction is atypical and difficult to notice with standard evaluations. Here we introduce minimalist changes to the ARAD1K dataset used in NTIRE 2022 challenge~\cite{arad2022ntire} in three experiments to demonstrate it. We exhaustively evaluate a total of 17 open-sourced neural network architectures to-date, namely MST++~\cite{cai2022mst++}, MST-L~\cite{cai2022mask}, MPRNet~\cite{zamir2021multi}, Restormer~\cite{zamir2022restormer}, MIRNet~\cite{zamir2020learning}, HINet~\cite{chen2021hinet}, HDNet~\cite{hu2022hdnet}, AWAN~\cite{li2020adaptive}, EDSR~\cite{lim2017enhanced}, HRNet~\cite{zhao2020hierarchical}, HSCNN+~\cite{shi2018hscnn+}, HySAT~\cite{wang2023learning}, HPRN~\cite{wu2023hprn}, SSTHyper~\cite{xu2024ssthyper}, MSFN~\cite{wu2024multistage},  GMSR~\cite{wang2024gmsr}, and SSRnet~\cite{dian2023spectral}. 

\subsection{Training with less data}

First, we make a simple change to the training of the participating networks in the NTIRE 2022 challenge~\cite{arad2022ntire}. While keeping all the training settings intact, we randomly choose only 50\% or 20\% of the original training data, respectively, to train the candidate networks and validate the performance on the original validation data. We illustrate the validation curves for MST++~\cite{cai2022mst++} in Fig.~\ref{fig:mstpp_percentage}. See the Supplementary Material for the results of other networks. We summarize the results for 100\% and 50\% training data in Table~\ref{tab:train_percent} for all networks. 

\begin{figure}[!htp]
    \centering
    \setlength{\tabcolsep}{1pt}
    \begin{tabular}{cc}
    \centering
    \includegraphics[width=0.49\columnwidth]{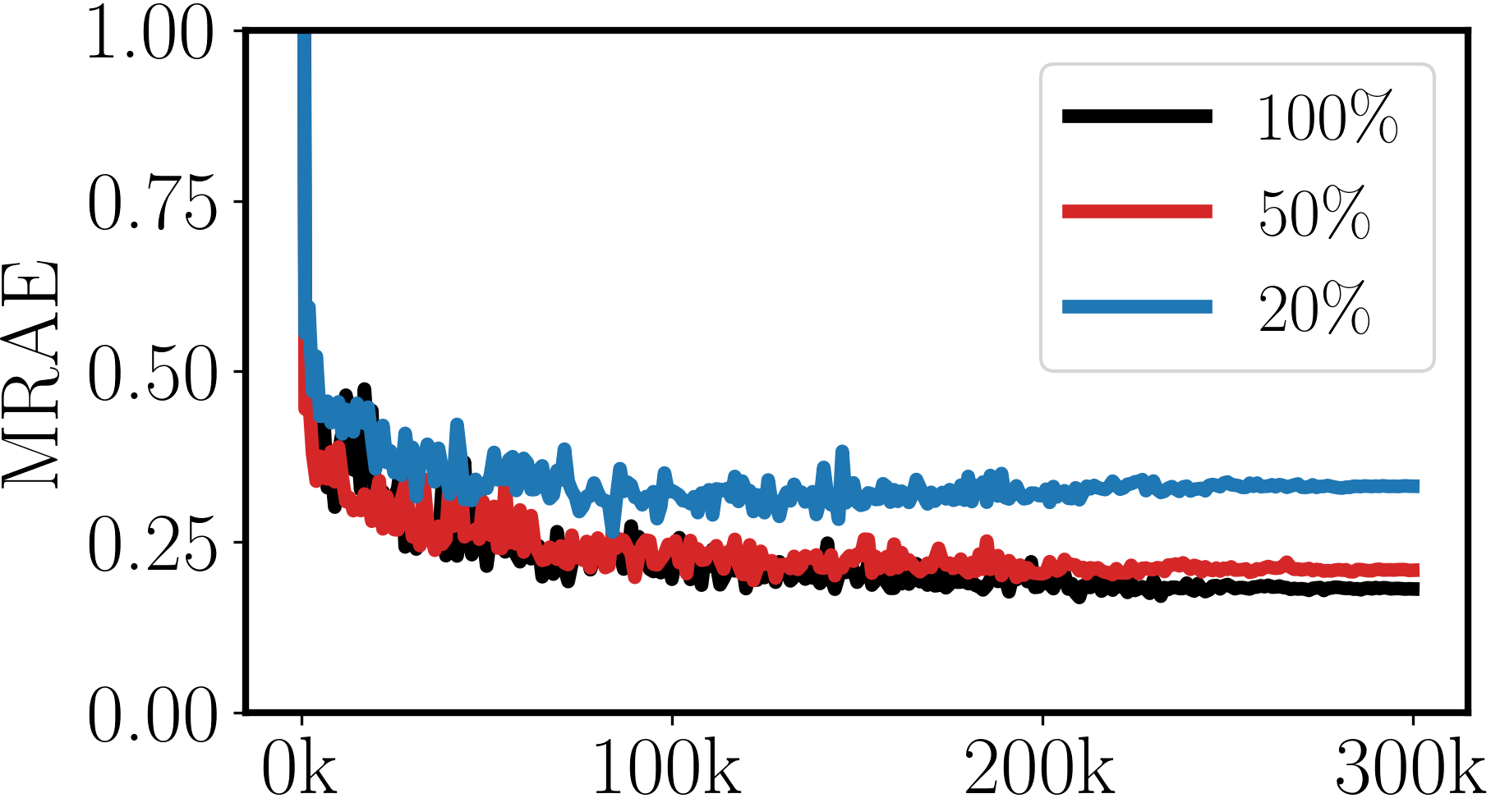} & \includegraphics[width=0.49\columnwidth]{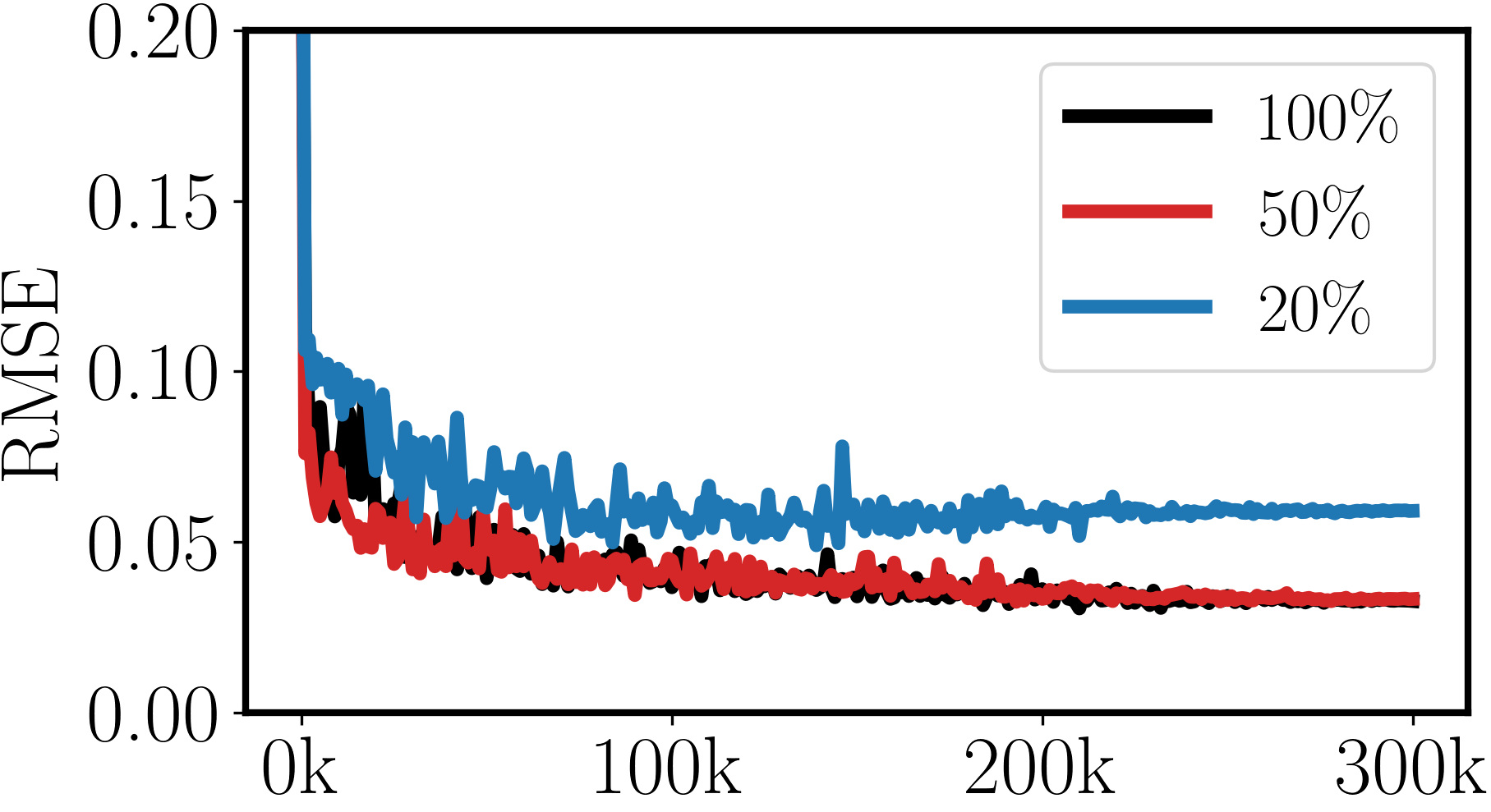} \\
    \includegraphics[width=0.49\columnwidth]{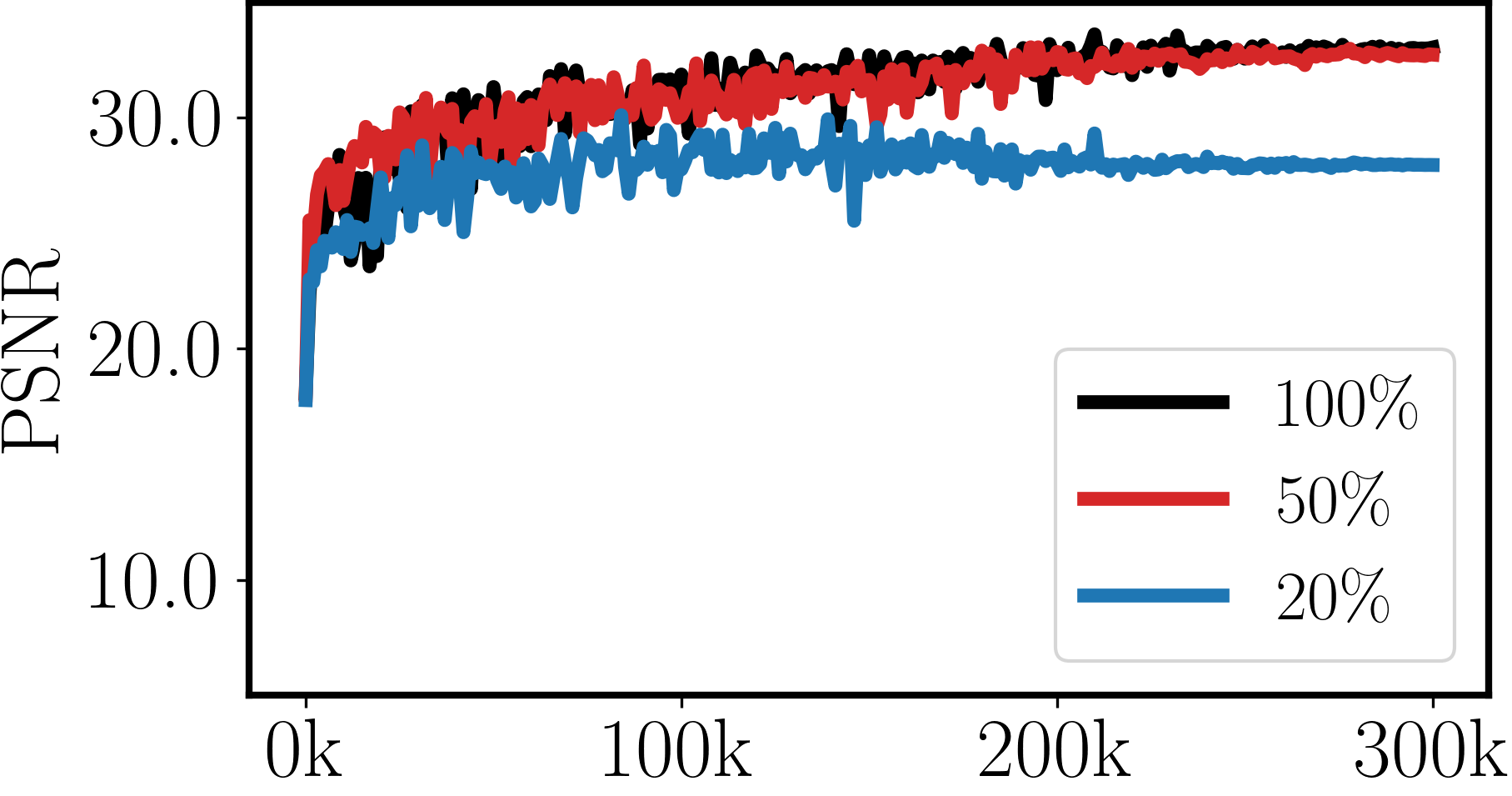} & \includegraphics[width=0.49\columnwidth]{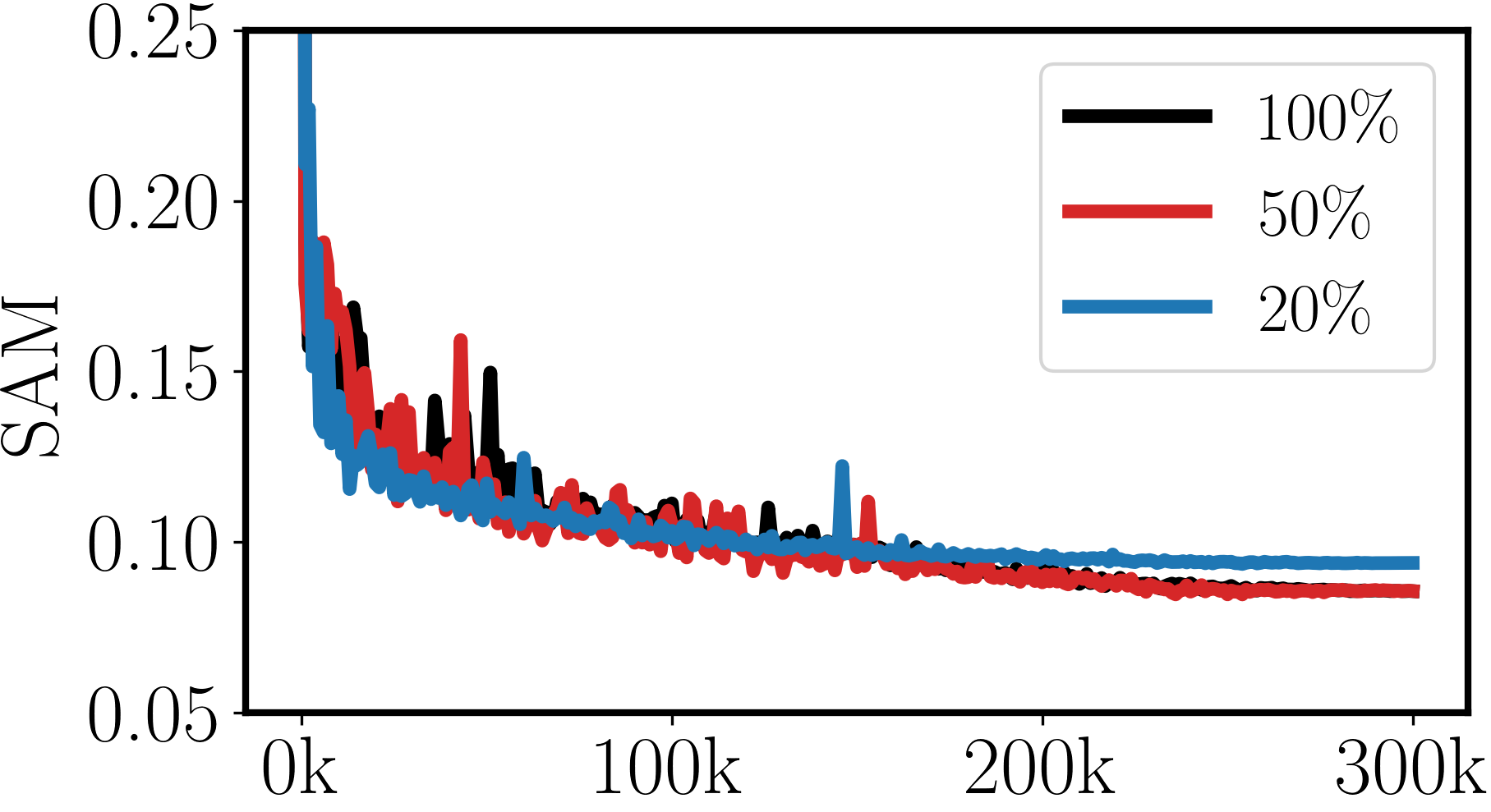}
    \end{tabular}
    \caption{Validation performance for MST++~\cite{cai2022mst++} with 100\%, 50\%, and 20\% of the original training data on ARAD1K~\cite{arad2022ntire}.}
    \label{fig:mstpp_percentage}
\end{figure}

Although the performance with less training data deviates mildly in MRAE, RMSE, and PSNR, the spectral accuracy SAM (highlighted in bold in Table~\ref{tab:train_percent}) is surprisingly less affected. In particular, some networks (\eg, MST++~\cite{cai2022mst++}, MIRNet~\cite{zamir2020learning}) achieve exactly the same SAM scores. MST-L~\cite{cai2022mask} (50\%) even improves SAM slightly, placing itself the best among all.

\begin{table}[!htp]
\caption{Performance comparison with 100\% and 50\% of training data for different methods on the original ARAD1K dataset.}
\label{tab:train_percent}
\setlength{\tabcolsep}{6pt}
\centering
\begin{tabular}{lccccc}
\hline
Network                    & Data       & MRAE$\downarrow$ & RMSE$\downarrow$ & PSNR$\uparrow$ & \textbf{SAM}$\downarrow$ \\ \hline
\multirow{2}{*}{MST++}     & 100\%      &0.182      &0.033      &33.0      &\textbf{0.086}     \\ \cline{2-6} 
                           & 50\%       &0.209      &0.033      &32.7      &\textbf{0.086}     \\ \hline
\multirow{2}{*}{MST-L}     & 100\%      &0.184      &0.031      &33.5      &\textbf{0.084}     \\ \cline{2-6} 
                           & 50\%       &0.253      &0.042      &30.6      &\textbf{0.080}     \\ \hline
\multirow{2}{*}{MPRNet}    & 100\%      &0.212      &0.034      &32.5      &\textbf{0.084}     \\ \cline{2-6} 
                           & 50\%       &0.293      &0.039      &31.3      &\textbf{0.091}     \\ \hline
\multirow{2}{*}{Restormer} & 100\%      &0.204      &0.033      &33.2      &\textbf{0.083}     \\ \cline{2-6} 
                           & 50\%       &0.304      &0.041      &31.4      &\textbf{0.092}     \\ \hline
\multirow{2}{*}{MIRNet}    & 100\%      &0.186      &0.030      &33.7      &\textbf{0.082}     \\ \cline{2-6} 
                           & 50\%       &0.214      &0.033      &32.6      &\textbf{0.082}     \\ \hline
\multirow{2}{*}{HINet}     & 100\%      &0.234      &0.036      &32.3      &\textbf{0.085}     \\ \cline{2-6} 
                           & 50\%       &0.267      &0.041      &30.7      &\textbf{0.090}     \\ \hline
\multirow{2}{*}{HDNet}     & 100\%      &0.223      &0.038      &31.2      &\textbf{0.095}     \\ \cline{2-6} 
                           & 50\%       &0.296      &0.047      &28.9      &\textbf{0.097}     \\ \hline
\multirow{2}{*}{AWAN}      & 100\%      &0.213      &0.034      &32.2      &\textbf{0.091}     \\ \cline{2-6} 
                           & 50\%       &0.273      &0.042      &30.5      &\textbf{0.095}     \\ \hline
\multirow{2}{*}{EDSR}      & 100\%      &0.358      &0.052      &27.3      &\textbf{0.095}     \\ \cline{2-6} 
                           & 50\%       &0.430      &0.059      &26.1      &\textbf{0.093}     \\ \hline
\multirow{2}{*}{HRNet}     & 100\%      &0.388      &0.057      &26.4      &\textbf{0.094}     \\ \cline{2-6} 
                           & 50\%       &0.413      &0.065      &25.5      &\textbf{0.096}     \\ \hline
\multirow{2}{*}{HSCNN+}    & 100\%      &0.428      &0.066      &25.4      &\textbf{0.098}     \\ \cline{2-6} 
                           & 50\%       &0.462      &0.068      &25.0      &\textbf{0.101}     \\ \hline
\multirow{2}{*}{HySAT} & 100\% &0.176 &0.028 &34.6 &\textbf{0.085}\\ \cline{2-6} 
                           &50\% &0.254 &0.037 &31.6 &\textbf{0.089} \\ \hline
\multirow{2}{*}{HPRN} & 100\% &0.257 &0.044 &30.4 &\textbf{0.098}\\ \cline{2-6} 
                           & 50\% &0.261 &0.041 &30.8 &\textbf{0.098} \\ \hline
\multirow{2}{*}{SSTHyper} & 100\% &0.181 &0.030 &33.6 &\textbf{0.083} \\ \cline{2-6} 
                           & 50\% &0.241 &0.039 &31.1 &\textbf{0.083} \\ \hline
\multirow{2}{*}{MSFN} & 100\% &0.226 &0.038 &31.9 &\textbf{0.084} \\ \cline{2-6} 
                           & 50\% &0.271 &0.044 &30.4 &\textbf{0.096} \\ \hline
\multirow{2}{*}{GMSR} & 100\% &0.308 &0.056 &27.5 &\textbf{0.113} \\ \cline{2-6} 
                           & 50\% &0.393 &0.077 &25.6 &\textbf{0.127} \\ \hline
\multirow{2}{*}{SSRnet} & 100\% &0.270 &0.048 &29.5 &\textbf{0.097} \\ \cline{2-6} 
                           & 50\% &0.307 &0.052 &29.0 &\textbf{0.108} \\ \hline
\end{tabular}
\end{table}

We therefore paradoxically find that despite the small size of hyperspectral datasets, the data already seems to be redundant. This serves as a first indication that the {\em diversity} of the datasets is severely lacking. We analyze this effect in more depth in the following experiments.

\subsection{Validation with unseen data}

\begin{table*}[!htp]
\caption{Evaluation of the pre-trained MST++ model on synthesized validation data from the ARAD1K dataset with different noise levels, compression quality, and realistic optical aberrations.}
\setlength{\tabcolsep}{10pt}
\centering
\begin{tabular}{l|cccc|cccc}
\hline
  &\multicolumn{4}{c|}{Data property} 
  &\multirow{2}{*}{MRAE $\downarrow$} 
  &\multirow{2}{*}{RMSE $\downarrow$} 
  &\multirow{2}{*}{PSNR $\uparrow$} 
  &\multirow{2}{*}{SAM $\downarrow$} \\ \cline{2-5}
  &\multicolumn{1}{c|}{Data source}
  &\multicolumn{1}{c|}{Noise (npe)}
  &\multicolumn{1}{c|}{RGB format}
  & Aberration &        &       &      &       \\ \hline
1 &\multicolumn{1}{c|}{NTIRE 2022} &\multicolumn{1}{c|}{unknown} &\multicolumn{1}{c|}{jpg (Q unknown)} &None &0.170 &0.029 &33.8 &0.084 \\ \hline
2 &\multicolumn{1}{c|}{\multirow{3}{*}{Synthesized}} &\multicolumn{1}{c|}{0} &\multicolumn{1}{c|}{jpg (Q = 65)} &None &0.460  &0.049 &29.2 &0.094 \\ \cline{1-1} \cline{3-9} 
3 &\multicolumn{1}{c|}{} & \multicolumn{1}{c|}{0} &\multicolumn{1}{c|}{png (lossless)} &None &0.362  &0.057 &28.7 &0.087 \\ \cline{1-1} \cline{3-9} 
4 &\multicolumn{1}{c|}{} & \multicolumn{1}{c|}{1000} &\multicolumn{1}{c|}{png (lossless)} &CA* &0.312 &0.055 &28.4 &0.118 \\ \hline 
\end{tabular}
\flushleft{\quad \quad \quad \footnotesize{*CA: chromatic aberration, from a patent double Gauss lens (US20210263286A1).}}
\label{tab:unseen}
\end{table*}

To further scrutinize the underlying issue, we validate existing pre-trained models with  ``unseen'' data synthesized from the original dataset used in the NTIRE challenge. The challenge organizers state that \blockquote{the exact noise parameters and JPEG compression level used to generate RGB images for the challenge was kept confidential}~\cite{arad2022ntire}. Only the spectrum-to-color projection was considered, and no aberrations of the optical system were simulated.

In our experiments, we generate new RGB images using the same methodology and calibration data, but different noise and compression settings. Specifically, we use the SRF data for a Basler ace 2 camera (model A2a5320-23ucBAS) known to the networks, and simulate Poisson noise at varying noise levels by controlling the number of photon electrons (npe). We adopt the same rudimentary in-camera image signal processing pipeline. As an illustrative example, we use MST++~\cite{cai2022mst++} in Table~\ref{tab:unseen}, Row~1 as a reference for comparison; results for other networks can be found in the Supplemental Material.

First, as a baseline, we consider a noiseless (npe = 0) and aberration-free case with moderate JPEG compression quality (Q = 65), shown in Table~\ref{tab:unseen}, Row~2. The results show significant drops in all the performance metrics. Note that the only differences here compared to the challenge dataset are the noise level and compression quality -- the base images are identical! This indicates that the network overfits both the noise and JPEG compression parameters.

Second, in Row~3, we generate noiseless RGB images, but in lossless PNG format, as opposed to the JPEG (Q = 65) in Row~2. Note that JPEG compression is not necessary for the core inverse problem in hyperspectral imaging, since raw data could be readily obtained from the sensors. This results in paradoxical reconstruction performance. MRAE and SAM improve compared to Row~2 (but are still worse than Row~1), while RMSE and PSNR deteriorate further. Considering this only eliminates image compression, and the networks were trained on MRAE~\cite{cai2022mst++}, we can confirm that the network indeed overfits the specific unknown JPEG compression used in the challenge~\cite{arad2022ntire}.

Third, we consider a more realistic imaging scenario, in which we eliminate the impact of unnecessary compression by employing the lossless PNG format to save the RGB images (equivalent to using raw camera data). We adopt moderate noise levels (npe = 1000) and realistic optical aberrations from a recent double Gauss lens patent~\cite{ichimura2021optical} to mimic a real photographic camera. We can observe a further performance drop in Row~4, which provides additional evidence that the network overfits the unknown parameters in the image simulation pipeline~\cite{arad2022ntire}. When used under realistic imaging conditions, the performance degrades significantly.

\subsection{Cross-dataset validation}
In addition, we inspect the effects of different datasets (\cf~Table~\ref{tab:datasets}) on the performance. We train the MST++ network on the four datasets with the same image simulation parameters. To eliminate the impact of other factors, we choose the ideal noiseless and aberration-free condition without compression. In the validation, we use our trained model on ARAD1K dataset to validate on the other three datasets, respectively. In Table~\ref{tab:cross_validation}, we compare the performance with the models both trained and validated on the original datasets. Results for other networks can be found in the Supplementary Material. They all illustrate the same difficulties in generalization.

\begin{table}[!htp]
\caption{Cross-dataset validation using MST++~\cite{cai2022mst++}.}
\setlength{\tabcolsep}{5pt}
\centering
\begin{tabular}{l|l|cccc}
\hline
Trained on & Validated on & MRAE$\downarrow$   & RMSE$\downarrow$ & PSNR$\uparrow$ & SAM$\downarrow$ \\ \hline
CAVE     & CAVE     &0.237 &0.034 &31.9  &0.194 \\ \hline
ARAD1K   & CAVE     &1.626 &0.074 &24.4  &0.376 \\ \hline \hline
ICVL     & ICVL     &0.079 &0.019 &38.3  &0.024 \\ \hline
ARAD1K   & ICVL     &0.627 &0.091 &22.0  &0.110 \\ \hline \hline
KAUST    & KAUST    &0.069 &0.013 &44.4  &0.061 \\ \hline
ARAD1K   & KAUST    &1.042 &0.100 &22.0  &0.370 \\ \hline
\end{tabular}
\label{tab:cross_validation}
\end{table}

Even though the imaging conditions are the same and ideal, the network trained on one dataset experiences significant performance drops in all metrics when validated on other datasets. This indicates that the contents of the datasets, as well as the acquisition devices used to capture the datasets, play important roles.

We also point out that the CAVE dataset~\cite{yasuma2010generalized}, although smaller and older than the others, is more difficult to train for better performance. This is probably due to the fact that CAVE consists of several challenging scenes of real and fake objects that appear in similar colors, but other datasets comprise less aggressive natural scenes.

\subsection{Discussion}
The experiments conducted in this section clearly highlight several shortcomings with respect to the existing datasets. {\bf (1) They lack diversity in nuisance parameters} such as noise and compression ratios. Our experiments show that when the RGB images are slightly modified by changing the nuisance parameters embedded in the image formation (\eg, noise level, compression factor, and optical aberrations), a significant drop in the performance is observed. {\bf (2) They lack scene diversity.} Training modern deep neural networks via supervised learning typically demands large-scale datasets. However, our experiments reveal that reducing the training data volume leads to only a marginal or even no drop in spectral accuracy. This suggests that the dataset lacks diversity in its content. The cross-dataset validation experiments further show that each dataset has its own statistics. Neural networks trained on a single dataset exhibit limited generalization to other datasets, suggesting the lack of sufficient scene diversity in individual datasets. In summary, our experiments provide clear evidence that the performance degradation primarily stems from insufficient scene diversity in existing datasets. While we show here the results for the largest available dataset (ARAD1K), the Supplementary Material shows consistent results for all the datasets. Both of these aspects result in over-fitting and prevent the networks from learning the general spectral image restoration task. Next, we specifically analyze the effect of metamerism; the analysis of the impact of optical aberrations will be deepened in Section~\ref{sec:aberration_advantage}.

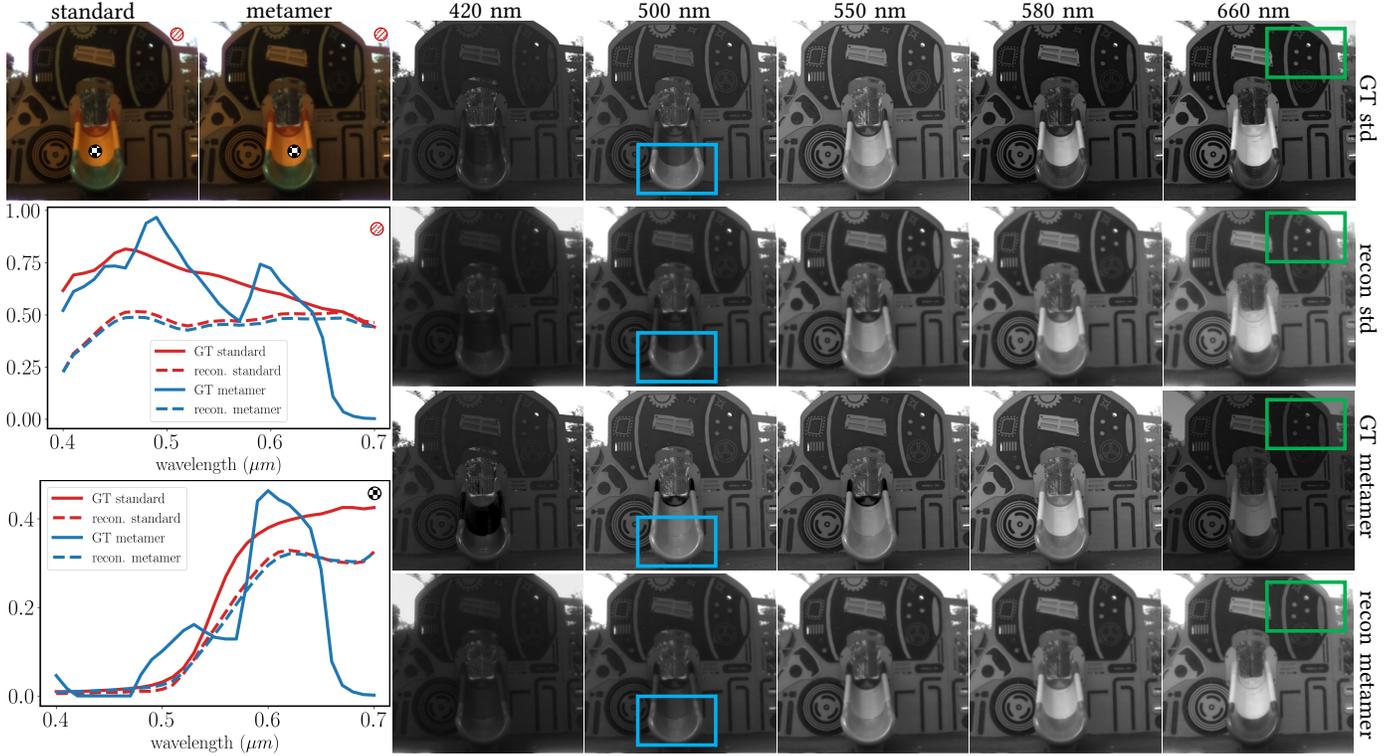
\begin{figure*}[!htp]
    \centering
    \begin{tikzpicture}
        \node[anchor=south west, inner sep=0] at (0,0) {\includegraphics[width=0.98\textwidth]{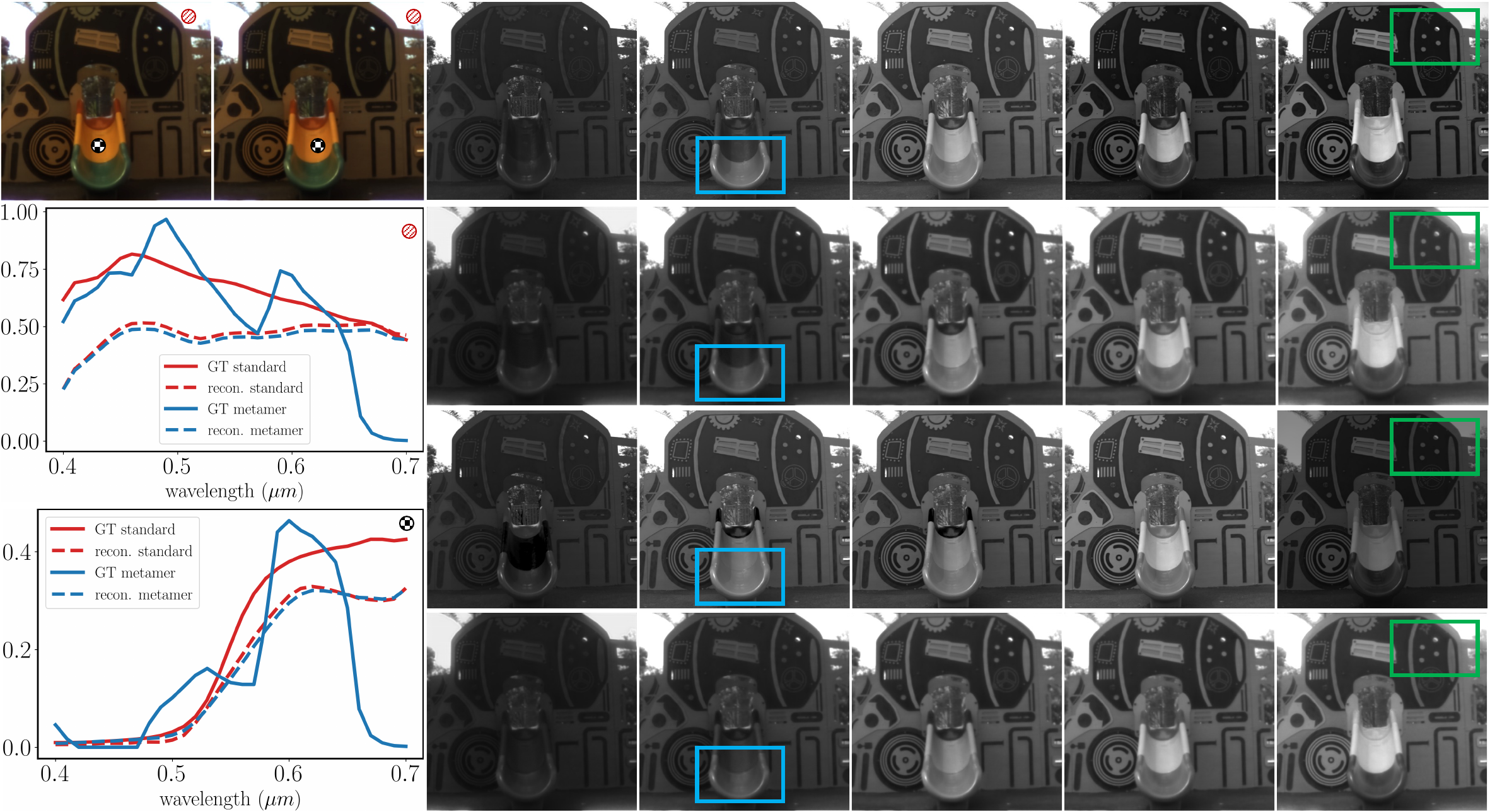}};
        \draw (2.5,9.9) node[text width=4cm,align=left] {\textcolor{black}{standard}};
        \draw (5.1,9.9) node[text width=4cm,align=left] {\textcolor{black}{metamer}};
        \draw (7.8,9.9) node[text width=4cm,align=left] {\textcolor{black}{420~nm}};
        \draw (10.3,9.9) node[text width=4cm,align=left] {\textcolor{black}{500~nm}};
        \draw (12.9,9.9) node[text width=4cm,align=left] {\textcolor{black}{550~nm}};
        \draw (15.4,9.9) node[text width=4cm,align=left] {\textcolor{black}{580~nm}};
        \draw (18.0,9.9) node[text width=4cm,align=left] {\textcolor{black}{660~nm}};
        \draw (19.9,8.6) node[text width=4cm,align=left] {\rotatebox[origin=c]{-90}{GT std}};
        \draw (19.9,6.2) node[text width=4cm,align=left] {\rotatebox[origin=c]{-90}{recon std}};
        \draw (19.9,3.7) node[text width=4cm,align=left] {\rotatebox[origin=c]{-90}{GT metamer}};
        \draw (19.9,1.2) node[text width=4cm,align=left] {\rotatebox[origin=c]{-90}{recon metamer}};
    \end{tikzpicture}
     \caption{Validation with metamers for MST++~\cite{cai2022mst++}. An example Scene {ARAD\_1K\_0944} is shown to visualize the standard and metamer datacubes. Top left: the standard and metamer data result in similar color images. Bottom left: ground-truth and reconstructed spectra from two labeled points. Right: ground-truth and reconstructed spectral images in 420~nm, 500~nm, 550~nm, 580~nm, and 660~nm.}
    \label{fig:validate_metamers}
\end{figure*}

\section{Finding 2: Metameric Failure}
\label{sec:metamer_failure}
In this section, we inspect the performance of existing methods using metamer as an adversary to validate as well as re-train the neural networks for performance analysis.

\subsection{Validation with metamers}
We generate metamer datacubes (metamer data) from the original ARAD1K dataset (standard data) using the metameric black method~\cite{finlayson2005metamer}. For this set of experiments, we fix the coefficient $\alpha = 0$ in Eq.~\eqref{eq:metameric_black}. We choose a realistic imaging condition as used in Table~\ref{tab:unseen}, Row~4, and keep it the same for both cases. The validation results on the ARAD1K dataset for existing pre-trained networks are summarized in Table~\ref{tab:validate_metamers}. We also visualize the reconstructed spectral images in five arbitrary bands (420~nm, 500~nm, 550~nm, 580~nm, and 660~nm) and spectra of two points in Fig.~\ref{fig:validate_metamers} for Scene {ARAD\_1K\_0944} from the validation set.

\begin{table}[!htp]
\caption{Validation performance for different pre-trained models on standard (std) data and metamer (met) adversary synthesized from the ARAD1K dataset~\cite{arad2022ntire}.}
\centering
\setlength{\tabcolsep}{8pt}
\begin{tabular}{lccccc}
\hline
Network & Data &MRAE$\downarrow$ &RMSE$\downarrow$ &PSNR$\uparrow$ &SAM$\downarrow$ \\ \hline
\multirow{2}{*}{MST++} &std &0.312 &0.055 &33.8 &0.084\\ \cline{2-6} 
                                           &met &\textbf{52.839} &0.091 &26.0 &\textbf{0.580}\\ \hline
\multirow{2}{*}{MST-L}  &std &0.327 &0.055 &28.0 &0.118\\ \cline{2-6} 
                                           &met &\textbf{51.321} &0.090 &25.9 &\textbf{0.579}\\ \hline
\multirow{2}{*}{MPRNet} &std &0.661 &0.066 &26.1 &0.125 \\ \cline{2-6} 
                                           &met &\textbf{145.981} &0.122 &23.3 &\textbf{0.547}\\ \hline
\multirow{2}{*}{Restormer} &std &0.510 &0.066 &25.5 &0.126\\ \cline{2-6} 
                                           &met &\textbf{79.705} &0.116 &23.4 &\textbf{0.567}\\ \hline
\multirow{2}{*}{MIRNet} &std &0.404 &0.077 &24.8 &0.124\\ \cline{2-6} 
                                           &met &\textbf{38.252} &0.089 &24.8 &\textbf{0.570}\\ \hline
\multirow{2}{*}{HINet} &std &0.450 &0.063 &26.5 &0.120\\ \cline{2-6} 
                                           &met &\textbf{67.148} &0.096 &24.8 &\textbf{0.552}\\ \hline
\multirow{2}{*}{HDNet}  &std &0.450 &0.082 &23.9 &0.126\\ \cline{2-6} 
                                           &met &\textbf{34.429} &0.095 &23.8 &\textbf{0.570}\\ \hline
\multirow{2}{*}{AWAN}&std &0.424 &0.080 &24.6 &0.119\\ \cline{2-6} 
                                           &met &\textbf{39.854} &0.095 &24.4 &\textbf{0.558}\\ \hline
\multirow{2}{*}{EDSR}&std &0.421 &0.066 &25.5 &0.132 \\ \cline{2-6} 
                                           &met &\textbf{49.435} &0.100 &23.8 &\textbf{0.564}\\ \hline
\multirow{2}{*}{HRNet}&std &0.514 &0.078 &23.9 &0.128\\ \cline{2-6} 
                                           &met &\textbf{43.726} &0.112 &22.7 &\textbf{0.560}\\ \hline
\multirow{2}{*}{HSCNN+}&std &0.508 &0.075 &24.4 &0.148\\ \cline{2-6} 
                                           &met &\textbf{42.274} &0.098 &23.1 &\textbf{0.556}\\ \hline
\multirow{2}{*}{HySAT} &std &0.326 &0.047 &29.4 &0.127\\ \cline{2-6} 
                                &met &\textbf{61.516} &0.095 &26.2 &\textbf{0.594}\\ \hline
\multirow{2}{*}{HPRN} &std &0.524 &0.104 &22.0 &0.130\\ \cline{2-6} 
                               &met &\textbf{33.439} &0.102 &22.8 &\textbf{0.574}\\ \hline
\multirow{2}{*}{SSTHyper} &std &0.314 &0.058 &27.7 &0.117\\ \cline{2-6} 
                                   &met &\textbf{48.427} &0.089 &25.8 &\textbf{0.575}\\ \hline
\multirow{2}{*}{MSFN} &std &0.328 &0.055 &28.3 &0.119\\ \cline{2-6} 
                               &met &\textbf{51.846} &0.090 &26.0 &\textbf{0.573}\\ \hline
\multirow{2}{*}{GMSR} &std &0.484 &0.075 &24.9 &0.138\\ \cline{2-6} 
                               &met &\textbf{53.063} &0.109 &23.6 &\textbf{0.547}\\ \hline
\multirow{2}{*}{SSRNet} &std &0.419 &0.075 &25.8 &0.130\\ \cline{2-6} 
                                 &met &\textbf{38.225} &0.104 &24.5 &\textbf{0.564}\\ \hline
\end{tabular}
\label{tab:validate_metamers}
\end{table}

From the numerical results in Table~\ref{tab:validate_metamers}, it is
apparent that all the existing methods experience catastrophic
performance drop in terms of MRAE and SAM in the presence of metamers,
which we call metameric failure. The MRAE (\cf~Eq.~\eqref{eq:mrae})
may yield large values when large errors occur for dark ground-truth
pixels (see the exemplary spectra in
Fig.~\ref{fig:validate_metamers}). The SAM values become large when
the spectra are essentially dissimilar with each other. RMSE
and PSNR do not capture the spectral differences as well, since they
average out differences in the spatial and spectral dimensions. 

The visual results in Fig.~\ref{fig:validate_metamers} show that the reconstruction results are very close to each other for both standard and metamer data, because the input RGB images are quite similar. However, distinct differences exist in the scene for certain spectral bands, \eg, the intensities of the yellow and green parts of the slide (blue box) in 500~nm band vary in the standard data, but remain identical in the metamer data. The reconstructions fail to reflect this important difference. All spectral images are displayed on the same global intensity scale, so the brightness differences (green box) in corresponding images reflect the reconstruction artifacts.

\subsection{Training with metamers}
The pre-trained models were not explicitly trained to cope with metamers. This raises the question whether it is possible to improve the performance by training the networks with metamer data. 

As a first step, we use both the standard and metamer data ($\alpha = 0$) generated from the ARAD1K dataset to train various networks. To eliminate the impact of other factors, we simulate the RGB images in a noiseless, aberration-free condition, and without compression.  

However, it is not sufficient to consider only a pair of standard and fixed metamer data. In reality, there are infinite metamers that project to the same color. As a second variant, we train the neural networks with random metamers generated on-the-fly as a spectral augmentation to enhance the spectral content of existing datasets. We vary the coefficient for the metameric black by setting $\alpha$ as a uniformly distributed random number in the range $[-1, 2]$. During validation, we use both the standard validation data and their corresponding metamer data with fixed $\alpha = 0$, which doubles the amount of the original validation data.

\begin{figure}[!htp]
    \centering
    \setlength{\tabcolsep}{1pt}
    \begin{tabular}{c@{}c}
    \centering
    \includegraphics[width=0.49\columnwidth]{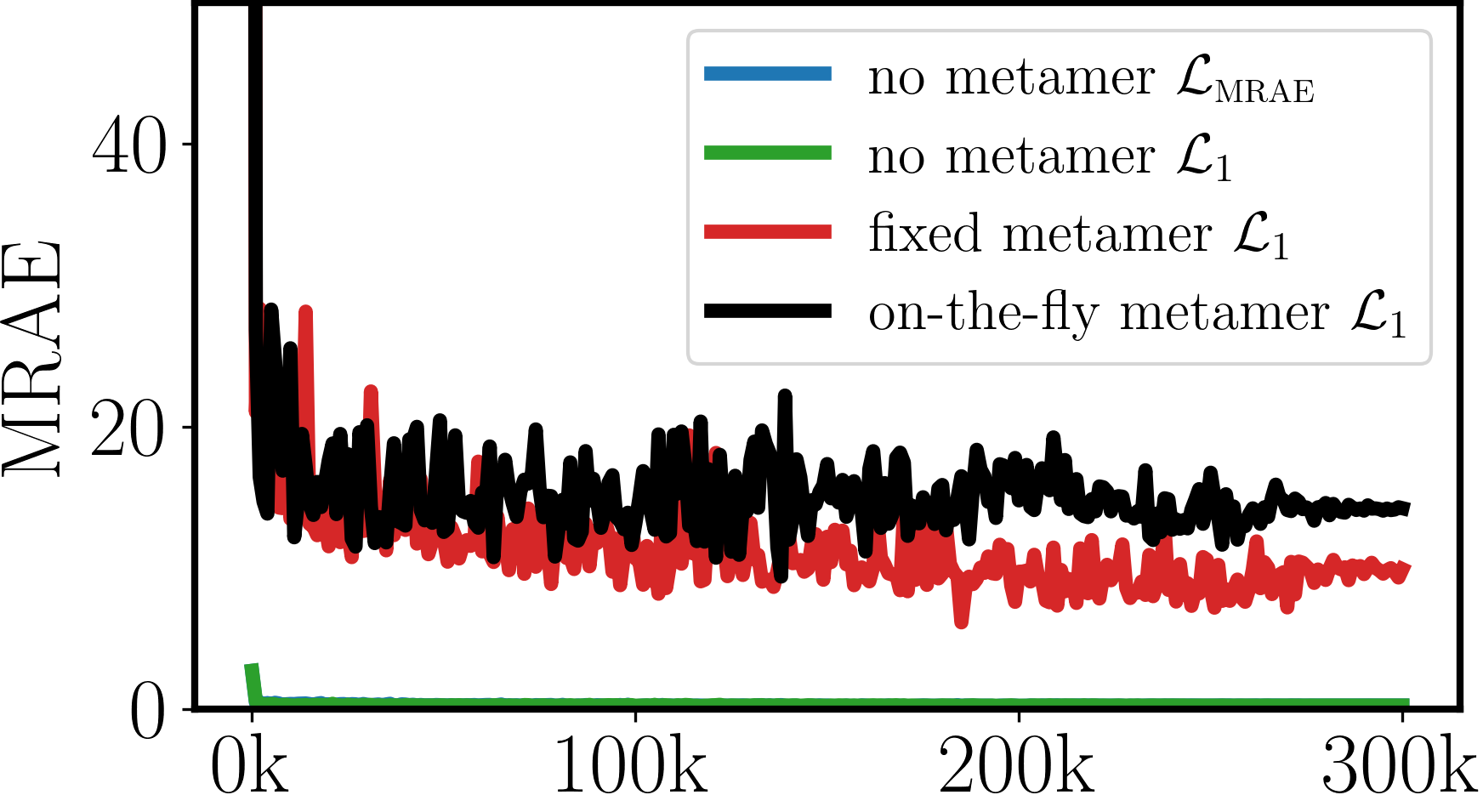} 
    & \includegraphics[width=0.49\columnwidth]{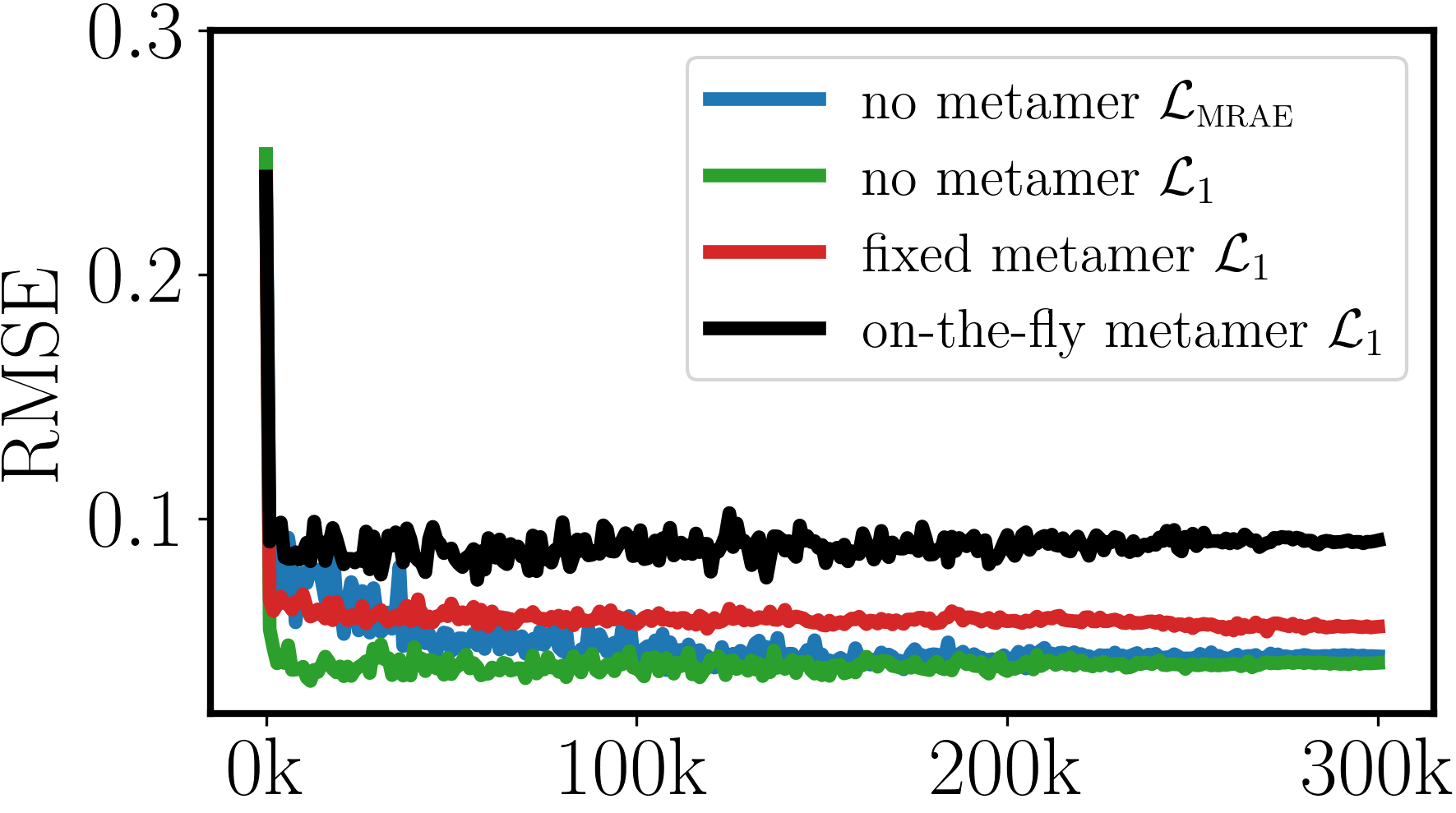} \\
    \includegraphics[width=0.49\columnwidth]{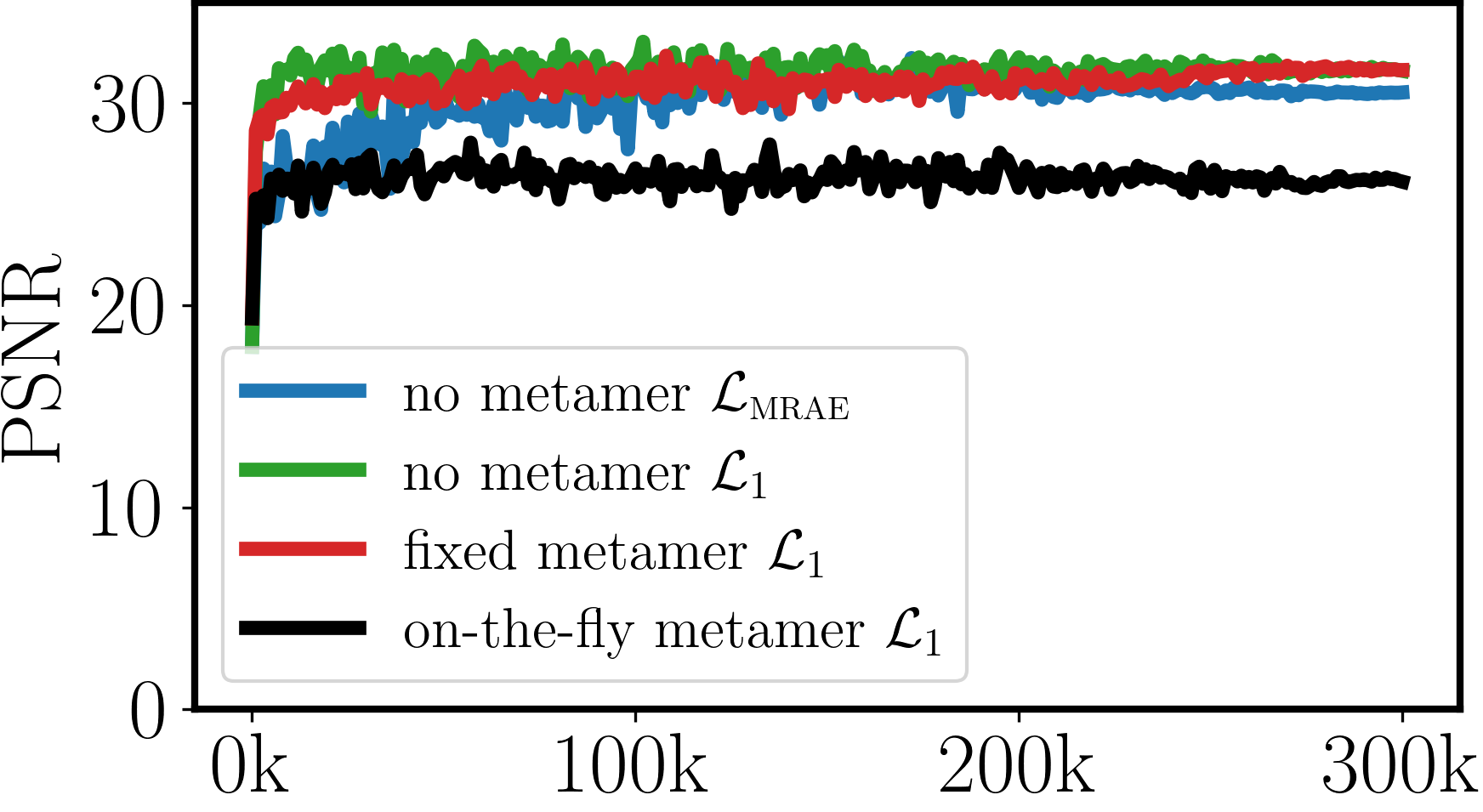} 
    & \includegraphics[width=0.49\columnwidth]{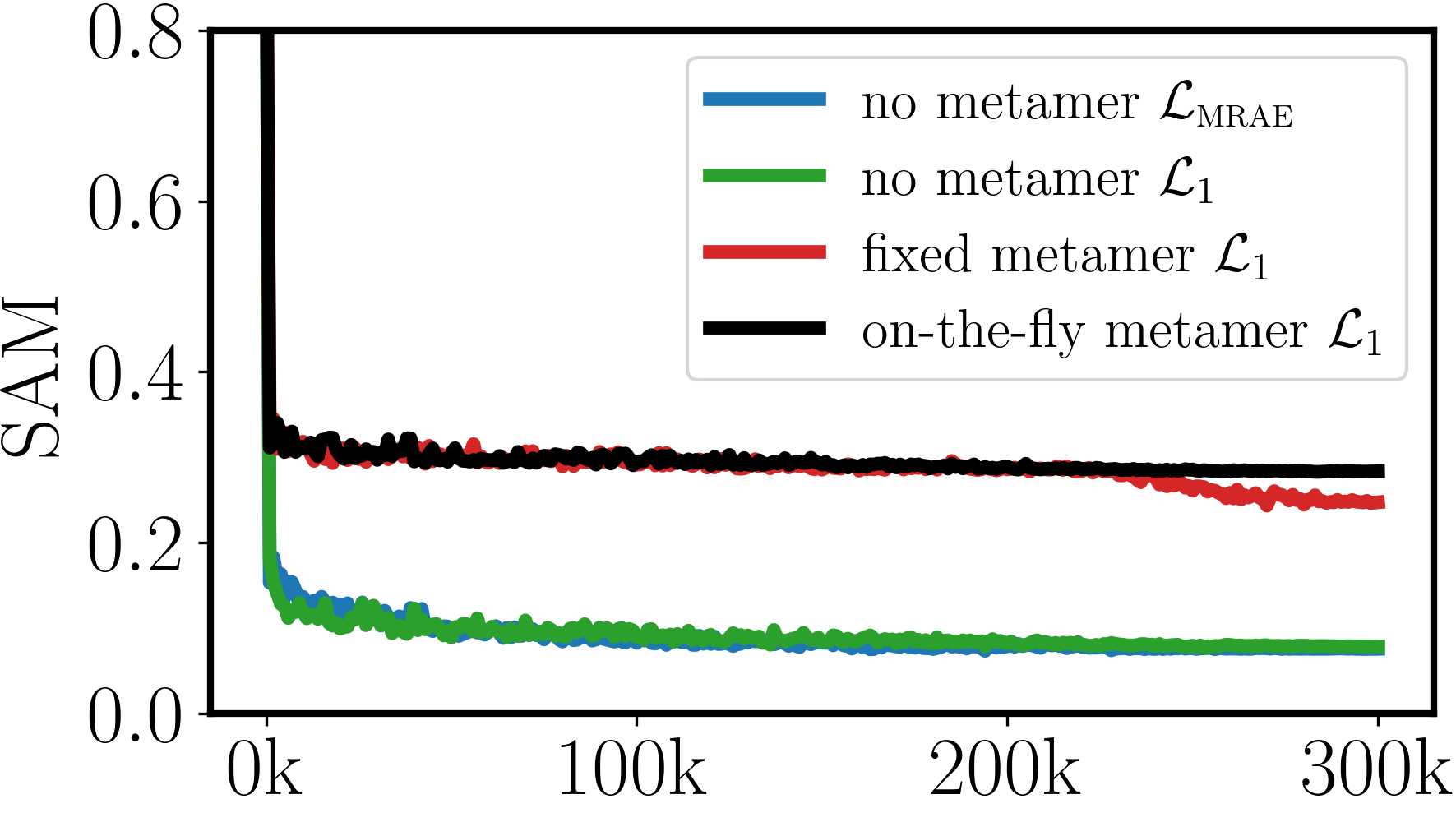}
    \end{tabular}
    \caption{Training MST++ with metamers. It fails to combat fixed metamers and on-the-fly metamers, in particular on the spectral accuracy SAM.}
    \label{fig:train_metamers}
\end{figure}

As an example, we train MST++ and evaluate its validation performance over the training process. In Fig.~\ref{fig:train_metamers}, we show that it is no longer a good choice to use MRAE as the loss function~\cite{cai2022mst++} and evaluation metric~\cite{arad2022ntire}, because it is completely overwhelmed by metamers. Instead, we find that L1 loss is a more stable loss function.
\begin{equation}
    \mathcal{L}_1 \left(\hat{\Mat{X}}, \Mat{X}\right) = \frac{1}{KMN} \sum_{k,i,j} \left|\hat{\Mat{X}}_{k,i,j} - \Mat{X}_{k,i,j} \right|.
\end{equation}
We then train the network with L1 loss for three cases, no metamer (easy), fixed metamer (medium), and on-the-fly metamer (difficult). Nevertheless, we can see in Fig.~\ref{fig:train_metamers} that the network fails in particular for the spectral accuracy SAM.

\begin{table}[!htp]
\caption{Performance comparison for training various networks with fixed and on-the-fly metamers.}
\setlength{\tabcolsep}{6pt}
\centering
\begin{tabular}{llcccc}
\hline
Network & Metamer & MRAE$\downarrow$ & RMSE$\downarrow$ & PSNR$\uparrow$ & SAM$\downarrow$ \\ \hline
\multirow{3}{*}{MST++} & no &0.270 &0.041  &31.6  &0.079  \\ \cline{2-6} 
                                          & fixed &9.912 &0.056  &31.6  &0.247 \\ \cline{2-6} 
                                          & on-the-fly &14.224 &0.091  &26.1  &0.284  \\ \hline
\multirow{3}{*}{MST-L}  & no &0.269  &0.040  &32.2  &0.081 \\ \cline{2-6} 
                                          & fixed &11.398  &0.061  &30.2 &0.289  \\ \cline{2-6} 
                                          & on-the-fly &12.155  &0.061  &27.0  &0.258 \\ \hline
\multirow{3}{*}{MPRNet} & no &0.346  &0.051  &29.9  &0.076 \\ \cline{2-6} 
                                          & fixed &7.359  &0.059  &30.6  &0.224 \\ \cline{2-6} 
                                          & on-the-fly &13.492  &0.087  &26.6  &0.264 \\ \hline
\multirow{3}{*}{Restormer} & no &0.286  &0.041  &31.5  &0.068 \\ \cline{2-6} 
                                          & fixed &8.129  &0.059  &31.0  &0.242 \\ \cline{2-6} 
                                          & on-the-fly &10.186  &0.089  &26.7  &0.264 \\ \hline
\multirow{3}{*}{MIRNet} & no &0.258  &0.040  &32.2  &0.083 \\ \cline{2-6} 
                                                & fixed &9.555  &0.061  &30.5  &0.289 \\ \cline{2-6} 
                                                & on-the-fly &13.205  &0.088  &26.4  &0.290 \\ \hline
\multirow{3}{*}{HINet}& no &0.315  &0.056  &28.0  &0.081 \\ \cline{2-6} 
                                          & fixed &8.322  &0.068  &29.6  &0.296 \\ \cline{2-6} 
                                          & on-the-fly &14.238  &0.090  &25.5  &0.288 \\ \hline
\multirow{3}{*}{HDNet}  & no &0.287  &0.045  &30.2  &0.087 \\ \cline{2-6} 
                                          & fixed &8.884  &0.064  &30.2  &0.296 \\ \cline{2-6} 
                                          & on-the-fly &18.270  &0.087  &26.4  &0.299 \\ \hline
\multirow{3}{*}{AWAN}& no &0.240  &0.039  &32.0  &0.073 \\ \cline{2-6} 
                                          & fixed &8.789  &0.068  &29.6  &0.294 \\ \cline{2-6} 
                                          & on-the-fly &14.406  &0.090  &26.0  &0.264 \\ \hline
\multirow{3}{*}{EDSR} & no &0.415  &0.061  &26.1  &0.084 \\ \cline{2-6} 
                                            & fixed &9.717  &0.073  &26.6  &0.297 \\ \cline{2-6} 
                                            & on-the-fly &11.723  &0.101  &23.2  &0.290 \\ \hline
\multirow{3}{*}{HRNet}& no &0.430  &0.065  &25.6  &0.085 \\ \cline{2-6} 
                                                 & fixed &8.142  &0.076  &26.0  &0.299 \\ \cline{2-6} 
                                                 & on-the-fly &13.098  &0.103  &23.0  &0.294 \\ \hline
\multirow{3}{*}{HSCNN+} & no &0.516  &0.077  &24.1  &0.082 \\ \cline{2-6} 
                                            & fixed &9.362  &0.085  &24.8   &0.297 \\ \cline{2-6} 
                                                & on-the-fly &13.311  &0.102  &22.9  &0.286 \\ \hline
\multirow{3}{*}{HySAT} &no &0.295  &0.036  &32.6 &0.078 \\ \cline{2-6} 
                                 &fixed &7.450  &0.063  &29.9 &0.270 \\ \cline{2-6} 
                                 &on-the-fly &12.806 &0.087 &26.5 &0.283 \\ \hline
\multirow{3}{*}{HPRN} &no &0.249  &0.038  &32.7 &0.076 \\ \cline{2-6} 
                                 &fixed &5.633  &0.060  &31.6 &0.221 \\ \cline{2-6} 
                                 &on-the-fly &12.197 &0.080 &26.7 &0.263 \\ \hline
\multirow{3}{*}{SSTHyper} &no &0.261  &0.036  &32.4 &0.083 \\ \cline{2-6} 
                                 &fixed &10.070  &0.060  &30.5 &0.285 \\ \cline{2-6} 
                                 &on-the-fly &12.194 &0.083 &27.3 &0.276 \\ \hline
\multirow{3}{*}{MSFN} &no &0.283  &0.043  &30.4 &0.078 \\ \cline{2-6} 
                                 &fixed &8.609  &0.065  &29.1 &0.286 \\ \cline{2-6} 
                                 &on-the-fly &12.198 &0.096 &24.9 &0.288 \\ \hline
\multirow{3}{*}{GMSR} &no &0.333  &0.049  &29.0 &0.114 \\ \cline{2-6} 
                                 &fixed &10.612  &0.065  &29.4 &0.302 \\ \cline{2-6} 
                                 &on-the-fly &11.674 &0.096 &24.9 &0.300 \\ \hline
\multirow{3}{*}{SSRNet} &no &0.354  &0.052  &29.0 &0.098 \\ \cline{2-6} 
                                 &fixed &6.678  &0.072  &28.9 &0.286 \\ \cline{2-6} 
                                 &on-the-fly &13.359 &0.102 &25.0 &0.291 \\ \hline
\end{tabular}
\label{tab:train_metamers}
\end{table}

We also train all other candidate networks with fixed and on-the-fly metamers. The results are summarized in Table~\ref{tab:train_metamers}.  Again, the same performance drop applies to all networks. Finally, we show the results of the top-performing network, MST++ on the CAVE, ICVL, and KAUST datasets in Table~\ref{tab:train_metamers_other_datasets}. (See Supplementary Material for more results). As before, the performance drops similarly in the presence of metamers.

\subsection{Discussion}
The experiments conducted in this section clearly highlight the difficulties that the data-driven spectral recovery methods face with metamers: (1) {\bf lack of sufficient metameric data} in current datasets,  (2) {\bf training with metamers} alone cannot mitigate the issue when the problem is formulated by Eq.~\eqref{eq:color_formation}, and (3) {\bf spectral estimation from RGB data} is indeed limited in the presence of metamers.

\begin{table}[!htp]
\caption{Training with metamers for MST++ on the CAVE~\cite{yasuma2010generalized}, ICVL~\cite{arad2016sparse}, and KAUST~\cite{li2021multispectral} datasets.}
\setlength{\tabcolsep}{6pt}
\centering
\begin{tabular}{llcccc}
\hline
Dataset & Metamer & MRAE$\downarrow$ & RMSE$\downarrow$ & PSNR$\uparrow$ & SAM$\downarrow$ \\ \hline
\multirow{3}{*}{CAVE}& no &1.014 &0.038  &29.9  &0.192  \\ \cline{2-6} 
                     & fixed &38.26 &0.053  &29.6  &0.229 \\ \cline{2-6} 
                     & on-the-fly &226.0 &0.078  &25.2  &0.451  \\ \hline
\multirow{3}{*}{ICVL} & no &0.067  &0.016  &40.1  &0.027 \\ \cline{2-6} 
                      & fixed &1.454  &0.041  &34.8 &0.229  \\ \cline{2-6} 
                      & on-the-fly &2.615  &0.087  &24.3  &0.268 \\ \hline
\multirow{3}{*}{KAUST} & no &0.082  &0.016  &43.2  &0.076 \\ \cline{2-6} 
                       & fixed &2.033  &0.022  &39.0 &0.217  \\ \cline{2-6} 
                       & on-the-fly &1.874  &0.032  &33.7  &0.245 \\ \hline
\end{tabular}
\label{tab:train_metamers_other_datasets}
\end{table}

The limitations of spectral estimation from RGB data are ultimately
not overly surprising -- after all the projection from the high
dimensional spectral space to RGB invariably destroys scene
information that can be difficult to recover. Spatial context from
underrepresented data does not contribute to the spectral estimation,
because such information remains the same for metamers. However, our
experiments show that this is indeed an issue faced by the
state-of-the-art methods, which so far went unnoticed due to the
under-representation of metamers in the datasets. This shortcoming
will also affect other uses of the same datasets, for example in the
training of reconstruction methods for spectral computational
cameras~\cite{jeon2019compact,baek2017compact,cao2011prism}. A
metameric adversary helps to identify this overlooked issue and avoid
unrealistically high numerical scores for existing systems. It also
underscores that, without side-channel information, no intrinsic
property exists in RGB images to distinguish between metamers, even
when they are augmented. Note that this does not downplay the effects
of metameric augmentation, since the problem formulation of
reconstructing spectral information from RGB images in
Eq.~\eqref{eq:color_formation} is fundamentally limited. Once the
problem is formulated in Eq.~\eqref{eq:image_formation_matrix},
metameric augmentation contributes to improving the network
robustness. We will explore it further in the next section.

\section{Finding 3: The Aberration Advantage and Effective Spectral Encoding}
\label{sec:aberration_advantage}

% Experiments with aberrations
\subsection{Aberration-Aware Training with Metameric Augmentation}
As shown so far, the existing methods have difficulties distinguishing metamers in the ideal noiseless and aberration-free condition. In this section, we analyze what effect (if any) optical aberrations have on this situation, \ie, aberration-aware training~\cite{yang2023aberration}. To this end, we train the networks in a realistic imaging condition with moderate noise level (npe = 1000), lossless PNG format, and aberrations from the same double Gauss lens as before~\cite{ichimura2021optical}.
In short, we simulate, through spectral ray tracing, the effect that an imperfect (\ie, aberrated) optical system has on the RGB image measured when observing a specific spectral scene. The details of this simulation can be found in the Supplementary Material.

In Fig.~\ref{fig:aberration_advantage}, we show an example with MST++ for the validation on SAM in two situations, one with fixed metamers, and the other with on-the-fly metamers. As a reference, we also show the standard validation without metamers as done in previous works (thin dashed black lines). As we can see, the realistic optical aberrations of the lens actually {\em improve} the spectral estimation in the presence of metamers as long as metamers are modeled in the training. With chromatic aberrations combined with metameric augmentation, the network can already distinguish fixed metamer pairs, achieving similar accuracy as the standard case. In the more aggressive case of on-the-fly metamers, chromatic aberrations also improve the spectral accuracy, compared with their no-aberration counterparts. Again, this aberration advantage holds for all datasets (Table~\ref{tab:aberration_advantage_other_datasets}). See Supplementary Material for details.

\begin{table}[!htp]
\caption{SAM metrics for MST++ on CAVE~\cite{yasuma2010generalized}, ICVL~\cite{arad2016sparse}, and KAUST~\cite{li2021multispectral}.}
\setlength{\tabcolsep}{6pt}
\centering
\begin{tabular}{lcc|cc}
\hline
\multicolumn{1}{c}{\multirow{2}{*}{Dataset}} & \multicolumn{2}{c|}{Fixed metamers} & \multicolumn{2}{c}{On-the-fly metamers} \\ \cline{2-5} 
\multicolumn{1}{c}{} & \multicolumn{1}{c|}{no aberration} & aberration & \multicolumn{1}{c|}{no aberration} & aberration \\ \hline
CAVE & \multicolumn{1}{c|}{0.251} &0.135  & \multicolumn{1}{c|}{0.380} &0.167  \\ \hline
ICVL & \multicolumn{1}{c|}{0.028} &0.077  & \multicolumn{1}{c|}{0.240} &0.085  \\ \hline
KAUST & \multicolumn{1}{c|}{0.212} &0.113  & \multicolumn{1}{c|}{0.221} & 0.113 \\ \hline
\end{tabular}
\label{tab:aberration_advantage_other_datasets}
\end{table}

\begin{figure}[!htp]
    \centering
    \setlength{\tabcolsep}{1pt}
    \begin{tabular}{cc}
    \includegraphics[width=0.49\columnwidth]{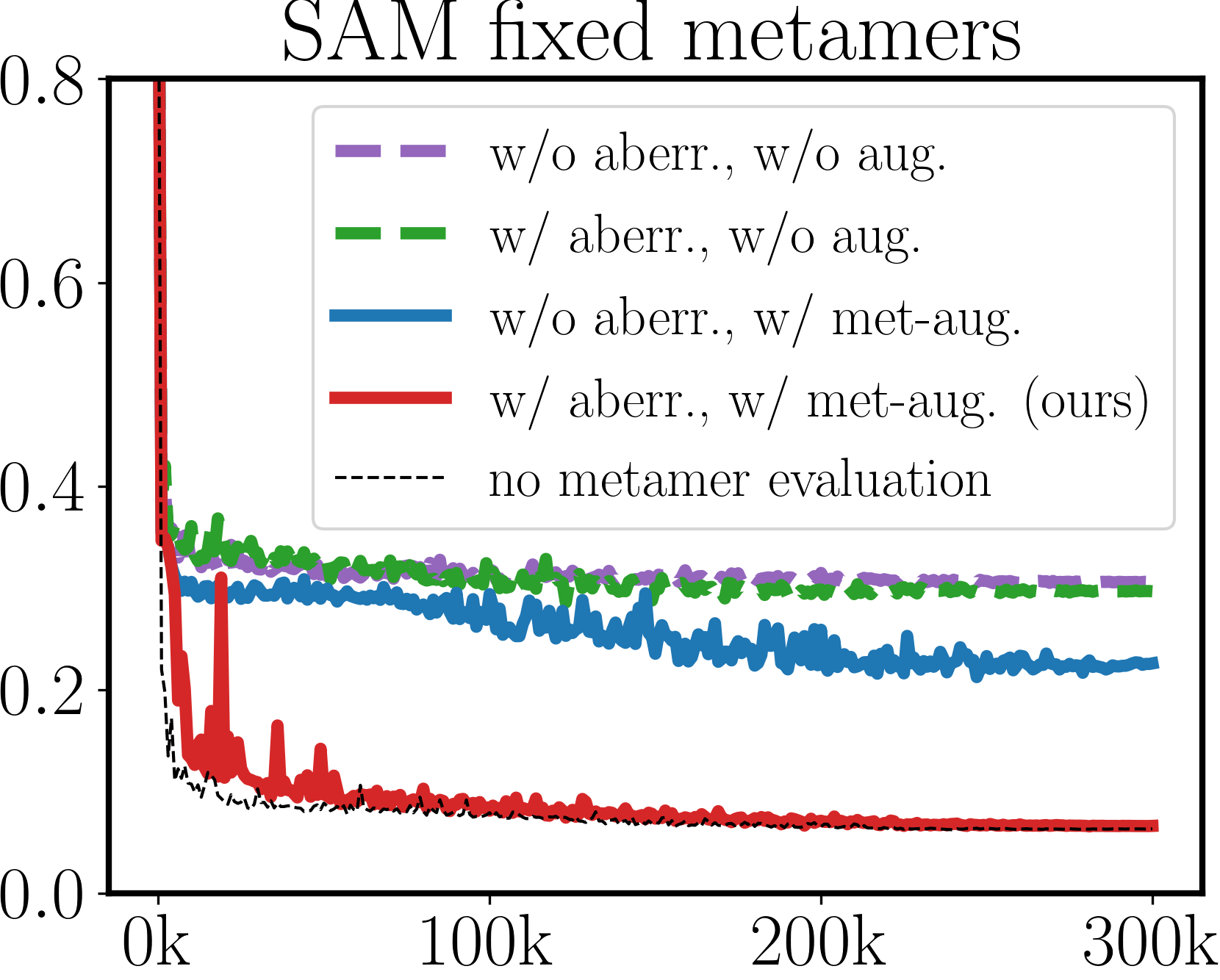} & \includegraphics[width=0.49\columnwidth]{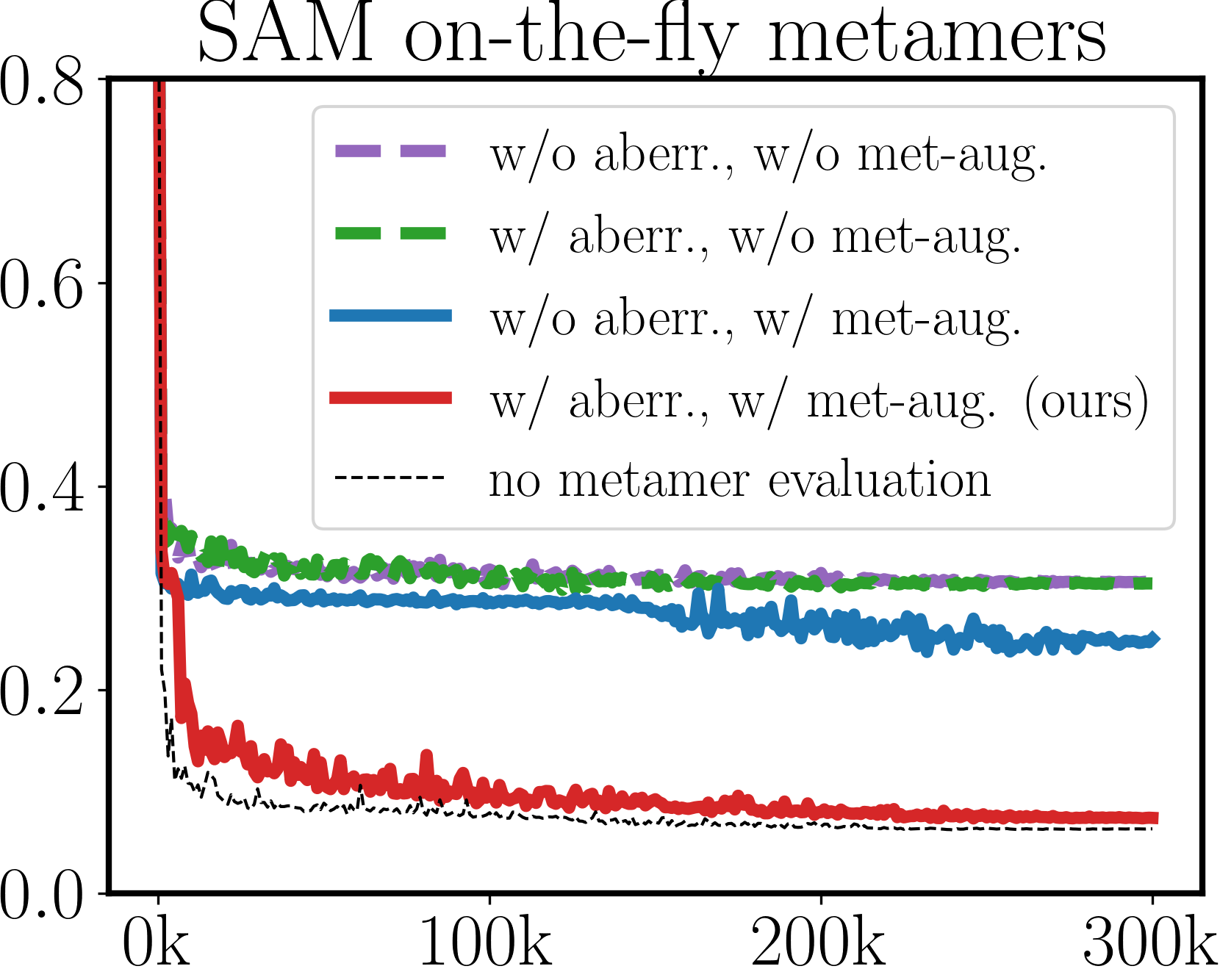}\\
    \end{tabular}
    \caption{Chromatic aberrations improve spectral accuracy. Left: fixed metamers. Right: on-the-fly metamers. aberr.: aberrations; met-aug.: metameric augmentation.}
    \label{fig:aberration_advantage}
    \vspace{-4mm}
\end{figure}

We carry out an ablation study to examine all possible combinations of aberrations and metameric augmentation, as shown in Fig.~\ref{fig:aberration_advantage}. This leads to 4 situations: (1) Training without aberrations, and without metameric-augmentation (purple lines). It is the existing training method that fails to combat metamers, as we have demonstrated in Section~\ref{sec:metamer_failure}. (2) Training with aberrations, but without metameric augmentation (green lines). It corresponds to simply a more blurry RGB image without considering metamers, which fails similarly. (3) Training without aberrations, but with metameric augmentation (blue lines). It slightly improves the spectral reconstruction, owing to the fact that noise may introduce slight differences in the RGB images. (4) Training with both aberrations and metameric augmentation. With the combination of aberrations and metameric augmentation, the network can learn the differences in RGB images between metamers, achieving similar performance as existing methods where metamers are actually not evaluated. These results clearly demonstrate again that the spectral reconstruction solely from RGB images fails for metamer data. The problem can be more effectively formulated only when spectral information, such as optical aberrations, is taken into account.

\begin{figure}[!htp]
    \centering
    \setlength{\tabcolsep}{1pt}
    \footnotesize
    \begin{tabular}{ccc}
    \centering
    standard & metamer & color difference\\
    \includegraphics[width=0.32\columnwidth]{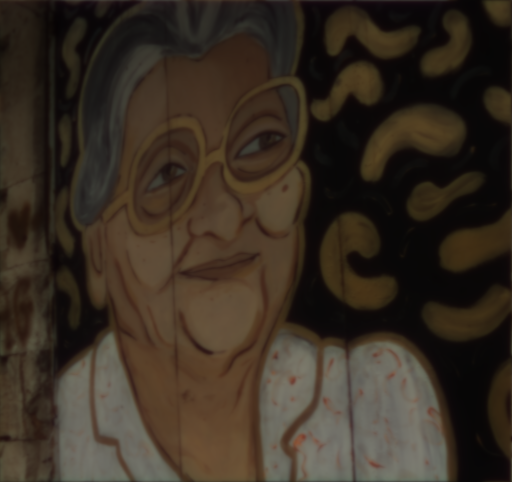} 
    & \includegraphics[width=0.32\columnwidth]{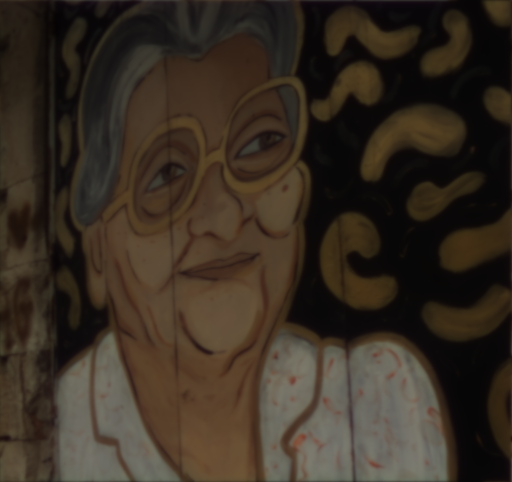} 
    & \raisebox{0mm}{%
    \begin{tikzpicture}[baseline={(0,0)}]
        \node[inner sep=0pt, anchor=south west] at (0,0) {\includegraphics[width=0.32\columnwidth]{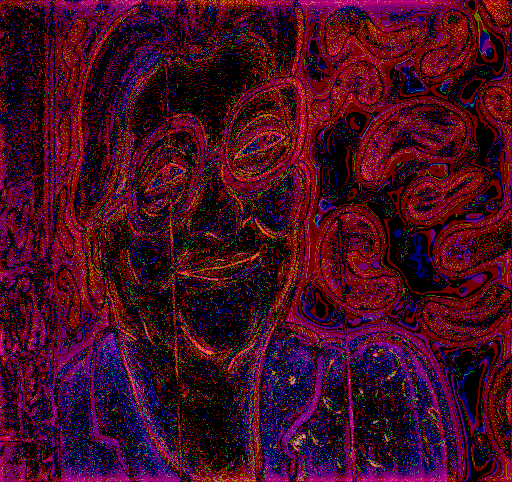}};
        % 0417, max 0.0549
        % 0402, max 0.0235
        % 0453, max 0.0157
        \node[anchor=center] at (1,0.2) {\textcolor{white}{$\Delta_{\textrm{max}} = 0.0235$}};
    \end{tikzpicture}} \\
    \end{tabular}
    \caption{Chromatic aberration induced informative color differences (right) as spectral cues for metamer pairs (left and middle).}
    \label{fig:color_difference}
\end{figure}

\begin{figure}[!htp]
    \centering
    \includegraphics[width=\columnwidth]{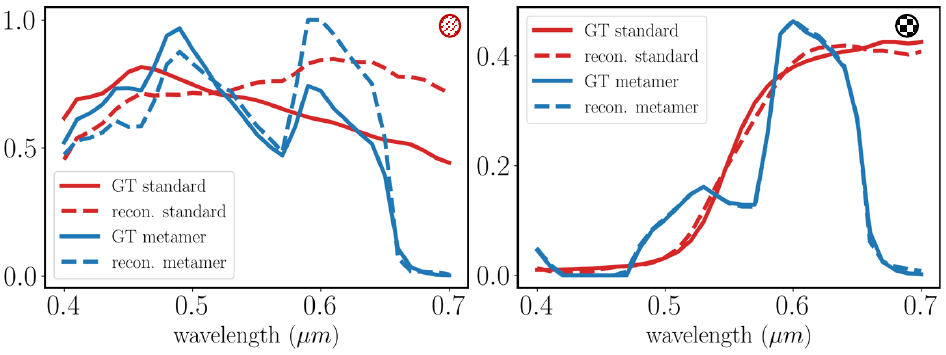}
    \caption{Reconstructed metameric spectra with MST++ trained with aberrations and metameric augmentation. The two example spectra are exactly the same as in Fig.~\ref{fig:validate_metamers}.}
    \label{fig:metamer_success}
\end{figure}

To understand why optical aberrations help improve the reconstruction, consider the simulated images in Fig.~\ref{fig:color_difference}. The left and middle images are simulations of RGB images for metameric scene pairs, with the difference image on the right. The different spectra of the two scenes are affected {\em differently} by the optical aberrations, and therefore, although the {\em scenes} are metamers of each other, the {\em RGB images} are in fact different. With optical aberrations, spectral information spreads out into adjacent pixels spatially at the cost of slightly making the RGB images blurry. The networks then see these color differences from metamers during training, and are able to correctly learn the mapping from metameric spectra to colors. We visualize the reconstructed spectra for the two example points in Fig.~\ref{fig:validate_metamers} with the MST++ model trained with both aberrations and on-the-fly metameric augmentation in Fig.~\ref{fig:metamer_success}. Now it is clear that when the network is trained with aberrations and metameric augmentation, it can tell the metameric spectra apart, which is not possible otherwise. In effect, the optical aberrations have {\em encoded} spectral information into the RGB image, which the networks can learn to distinguish, lending credibility to {\em PSF engineering} methods for hyperspectral encoding~\cite{jeon2019compact,baek2017compact,cao2011prism}. We also note that {\bf incidental aberrations of a lens are only a weak spectral encoding}. Better performance would require optimizing the spectral encodings deliberately.

\subsection{Effective Spectral Encoding}

\begin{figure}[!htp]
    \centering
    \includegraphics[width=0.98\columnwidth]{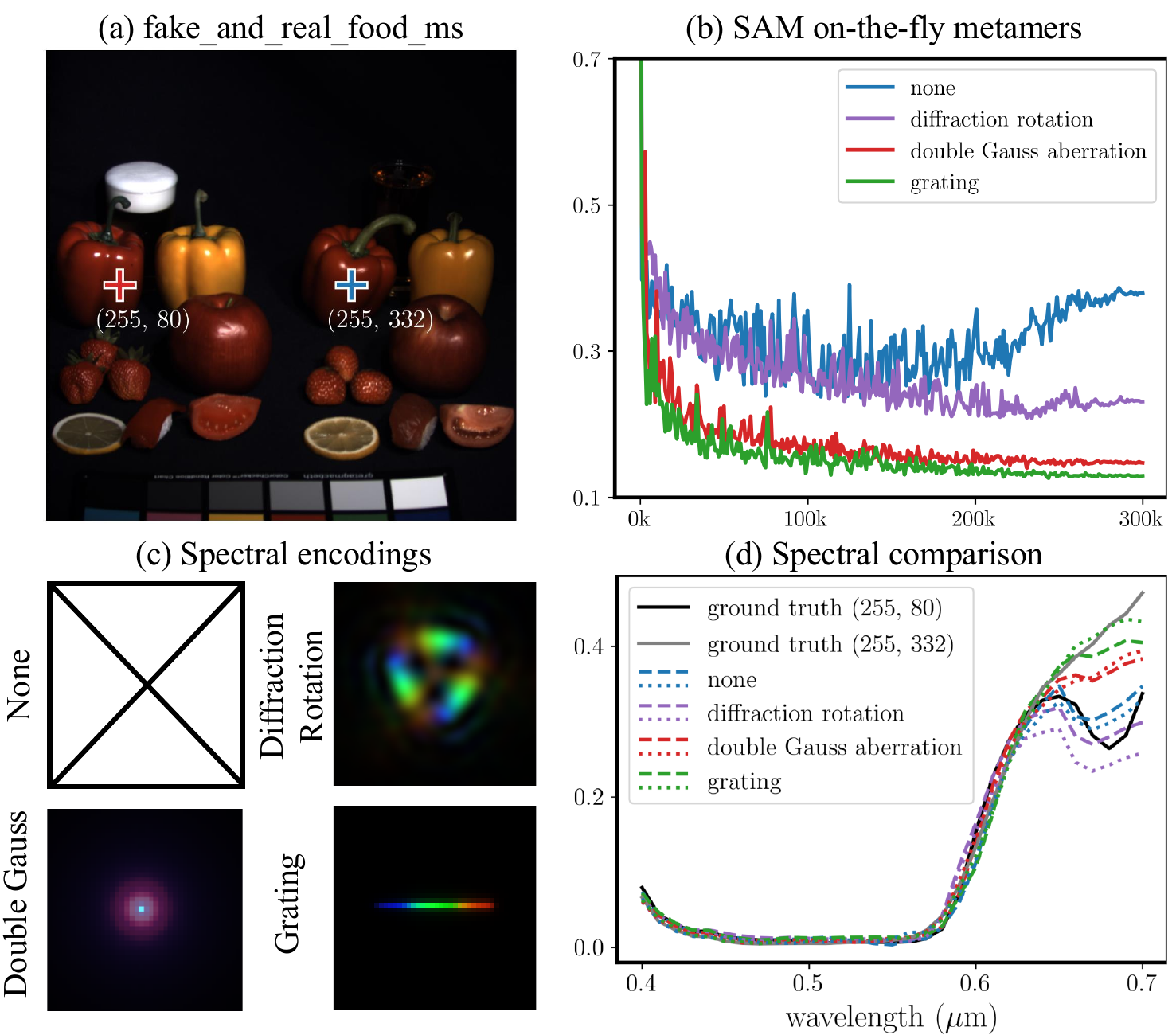}
    \caption{Comparison of different spectral encoding methods. (a) The {fake\_and\_real\_food\_ms} scene from CAVE. (b) Training performance in SAM for different encoding methods. (c) PSFs for different spectral encodings: none, diffraction rotation~\cite{jeon2019compact}, double Gauss aberration, and grating~\cite{baek2017compact}. (d) Reconstructed spectra for the two points denoted in (a).}
    \label{fig:comp_encodings}
\end{figure}

The above experiments point to a better formulation of the spectral reconstruction problem where spectral encoding plays a key role. Optical aberrations, however, are usually {\em minimized} in camera lenses, making them imperfect candidates for effective spectral encoding. Existing works have explored some deliberate use of dispersive optical elements for this purpose, including diffraction rotation~\cite{jeon2019compact} and grating~\cite{baek2017compact}. However, their performance has historically not been analyzed for metamers either. To this end, we test such spectral encoding schemes on a real challenging scene as shown earlier in Fig.~\ref{fig:cave_metamers}. In Fig.~\ref{fig:comp_encodings}, we compare four spectral encoding conditions: None (no encoding), Diffraction Rotation (used in~\cite{jeon2019compact}), Double Gauss aberrations (used above), and Grating (used in~\cite{baek2017compact}). We train MST++ with metameric augmentation for these spectral encodings, and the SAM results in Fig.~\ref{fig:comp_encodings}(b) show that spectral encodings indeed improve the overall spectral accuracy compared with no spectral encoding. The corresponding spectral PSFs are shown in Fig.~\ref{fig:comp_encodings}(c). The reconstructed spectra for the same two points in Fig.~\ref{fig:cave_metamers} are shown in Fig.~\ref{fig:comp_encodings}(d). Without spectral encoding, the spectral accuracy diverges, while all the spectral encodings improve the spectral quality. Interestingly, different spectral encoding schemes lead to varying spectral accuracy. Diffraction rotation tends to separate the metameric colors more, but the overall SAM is worse than aberrations and grating. We highlight that such challenging metameric spectra have not been properly evaluated in previous works. Although these spectral encoding schemes have been proposed, they are not optimized to deal with metamers yet. Again, the primary culprits are the dataset limitations we have pointed out -- metamers are highly underrepresented in existing datasets. While the CAVE dataset includes such examples, they are present in only limited quantities. To achieve better spectral reconstruction in such challenging situations, an effective spectral encoding and a powerful neural network should be trained on a large-scale and diverse dataset in which metamers are well represented. This constitutes a critical yet unresolved challenge in the field of data-driven spectral reconstruction.

\begin{table*}[!htp]
\caption{Generalization analysis on the effects of the proposed metameric augmentation for various spectral encoding schemes.}
\label{tab:generalization}
\centering
\setlength{\tabcolsep}{4pt}
\begin{tabular}{l|c|P{1cm}P{1cm}|P{1cm}P{1cm}|P{1cm}P{1cm}|P{1cm}P{1cm}|P{1cm}P{1cm}}
\hline
\multirow{3}{*}{Trained on} & \multirow{3}{*}{Validated on} & \multicolumn{4}{c|}{No metameric augmentation} & \multicolumn{6}{c}{With metameric augmentation} \\ \cline{3-12} 
                            &                               & \multicolumn{2}{c|}{RGB2HS} & \multicolumn{2}{c|}{CASSI} & \multicolumn{2}{c|}{RGB2HS} & \multicolumn{2}{c|}{RGB2HS + aberrations} & \multicolumn{2}{c}{CASSI} \\ \cline{3-12} 
                            &                               & PSNR & SAM & PSNR & SAM & PSNR & SAM & PSNR & SAM & PSNR & SAM \\ \hline
\multirow{6}{*}{ARAD1K}     & CAVE std                      & 29.9 & 0.359 & 30.6 & 0.270 & 22.4 & 0.692 & 31.1 & 0.363 & 31.7 & 0.257 \\ \cline{2-12} 
                            & CAVE met                      & 27.3 & 0.510 & 24.2 & 0.382 & 29.9 & 0.296 & 37.9 & 0.160 & 27.4 & 0.295 \\ \cline{2-12} 
                            & ICVL std                      & 24.5 & 0.090 & 39.7 & 0.048 & 35.5 & 0.168 & 34.4 & 0.164 & 39.9 & 0.046 \\ \cline{2-12} 
                            & ICVL met                      & 24.5 & 0.497 & 27.8 & 0.314 & 33.9 & 0.290 & 34.1 & 0.311 & 32.8 & 0.195 \\ \cline{2-12} 
                            & KAUST std                     & 23.5 & 0.512 & 41.9 & 0.099 & 32.7 & 0.253 & 35.5 & 0.247 & 44.3 & 0.078 \\ \cline{2-12} 
                            & KAUST met                     & 25.3 & 0.775 & 32.8 & 0.284 & 34.0 & 0.220 & 39.9 & 0.073 & 37.2 & 0.184 \\ \hline
\end{tabular}
\end{table*}

Other spectral encoding methods, such as CASSI, could also benefit from metameric augmentation. In Table~\ref{tab:generalization}, we show results on one of the best-performing networks, MST-L~\cite{cai2022mask}, trained on the ARAD1K dataset and validated on the other three datasets. All the experiments are carried out with Poisson noise npe = 1000, and aberrations from the double Gauss lens (US20210263286A1). We adopt the CASSI settings in ~\cite{cai2022mask}.

As indicated by the results, RGB2HS cannot maintain its performance in PSNR and SAM for data from other datasets, even with metameric augmentation to account for metamers during training (see the RGB2HS columns). With the aid of aberrations and metameric augmentation, the reconstruction performance could be boosted for both standard and metamer data. In addition, evident from the CASSI results without metameric augmentation, CASSI offers overall better reconstruction quality, thanks to its spatial-spectral encoding design rooted in the compressive sensing theory. When metameric augmentation is applied to CASSI, its performance has also been boosted, proving the effectiveness of our metameric augmentation scheme. Note that the reconstruction quality varies among datasets, owing to the different spectral content in each dataset.

\subsection{Discussion}
We have demonstrated through aberration-aware training with metameric augmentation that only when optical aberrations are considered in the image formation, the problem of spectral reconstruction from RGB images can be better formulated. Our ablation experiments further prove that it is not sufficient to only model metamers without employing spectral encodings. In addition, we highlight that optical aberrations are incidental but not deliberate spectral encodings, so their effect is still limited. We further examine alternative spectral encoding methods, such as PSF engineering approaches~\cite{jeon2019compact,baek2017compact,cao2011prism}, and compressed sensing methods like CASSI~\cite{wagadarikar2008single,choi2017high}. The results of our extensive experiments reiterate the credibility to such computational camera approaches. However, learned reconstruction methods for these approaches also suffer from the same dataset issues as the methods analyzed in this paper, making the collection of large-scale, diverse spectral image data a matter of urgency. These new datasets in turn will enable the design of improved optical encodings in computational spectral imaging systems without overfitting to specific scenarios.

\section{Conclusion}
\label{sec:conclusion}

In this work, we have comprehensively analyzed a category of data-driven spectral reconstruction methods from RGB images by reviewing the problem fundamentally from dataset bias to physical image formation, and to reconstruction networks. From an optics-aware perspective, we leverage both metamerism and optical aberrations to reassess existing methodologies.

The major findings of our study reveal important yet previously overlooked limitations in this research direction. {\bf(1)} The limitations of current datasets lead to overfitting to both nuisance parameters (noise, compression), as well as limited scene content. {\bf(2)} Metamerism in particular presents a challenge both in terms of under-representation in the datasets, and in terms of fundamental limitations of spectral reconstruction from RGB input. {\bf(3)} Metameric augmentation along with the targeted use of optical aberrations paves the way to combating the metamer issue, though more effective spectral encodings are demanded to solve the challenge.

Our results systematically demonstrate that it is impossible to accurately reconstruct spectra solely from RGB images. In order to realize the dream of spectral estimation from arbitrary RGB sources, it is necessary to coherently and jointly diversify the spectral contents in hyperspectral image datasets, adopt side-channel information from the optical system, and embrace versatile spectral data augmentation methods to fully enable the power of networks in adaptation to whole families of spectral encodings. We argue that addressing these foundational issues is imperative. Continuing to propose new network designs without rethinking the misdefined problem formulation will fall into the same fundamental shortcomings. Clarifying these limitations will enable the community to focus on solving the real challenges in snapshot spectral imaging.

The dataset limitations we point out in this work may also apply to other spectral reconstruction problems using the referenced datasets, such as CASSI, PSF engineering, and multispectral-hyperspectral fusion. In particular, the same metamerism issue has not yet been extensively evaluated in such domains either. Our findings underscore the broader importance of effective spectral encodings in such snapshot spectral imaging problems. The proposed metameric augmentation technique could inform future directions in optical design, network design, and, more importantly, their joint optimization to cope with metamers.

%---------- main texts ----------%

\section*{Acknowledgments}
This work was supported by the KAUST Individual Baseline Funding and Center of Excellence for Generative AI. Seung-Hwan Baek was supported by the National Research Foundation of Korean (NRF) grant funded by the Korea Goverment (MSIT) (RS-2024-00438532, RS-2023-00211658) and Basic Science Research Program through the NRF funded by the Ministry of Education (2022R1A6A1A03052954).

% {\appendix[Proof of the Zonklar Equations]
% Use $\backslash${\tt{appendix}} if you have a single appendix:
% Do not use $\backslash${\tt{section}} anymore after $\backslash${\tt{appendix}}, only $\backslash${\tt{section*}}.
% If you have multiple appendixes use $\backslash${\tt{appendices}} then use $\backslash${\tt{section}} to start each appendix.
% You must declare a $\backslash${\tt{section}} before using any $\backslash${\tt{subsection}} or using $\backslash${\tt{label}} ($\backslash${\tt{appendices}} by itself
%  starts a section numbered zero.)}

{\appendices
\section*{Appendix}

\subsection{Image Formation Model}
Mathematically, the physical image formation of a color image from the spectral radiance can be expressed by
\begin{equation}
    g_c\left(x, y\right) = \int_{\lambda_1}^{\lambda_2} \left(f\left(x, y, \lambda\right) * h\left(x, y, \lambda\right) \right) q_c\left(\lambda\right) d \lambda,
    \label{eq:image_formation}
\end{equation}
where $f\left(x, y, \lambda\right)$ is the spectral image, $h\left(x, y, \lambda\right)$ is the spectral point spread function (PSF) of the optical system, $q_c\left(\lambda\right)$ is the spectral response function (SRF) of the sensor, and $g_c\left(x, y\right)$ is the color image in color channel $c \in [R, G, B]$.

Let us denote the hyperspectral image as a matrix $\Mat{X} \in \mathbb{R}^{MN \times K}$, where $M, N$ are the number of pixels in spatial dimensions, and $K$ is the number of spectral bands in spectral dimension. Note that we have stacked the 2D spectral images in rows of $\Mat{X}$. Explicitly, we have
\begin{equation}
    \Mat{X} = \left[\Vect{x}_1, \Vect{x}_2, \dots, \Vect{x}_K  \right],
\end{equation}
where each column $\Vect{x}_k \in \mathbb{R}^{MN \times 1}$ is a vector for the spectral image in spectral channel $k$.
The SRF of the sensor is a matrix $\Mat{Q} \in \mathbb{R}^{K \times 3}$, \ie,
\begin{equation}
    \Mat{Q} = \left[\Vect{q}_1, \Vect{q}_2, \Vect{q}_3\right] = 
    \left[
    \begin{matrix}
        q_{11} & q_{21} & q_{31}\\
        q_{21} & q_{22} & q_{32}\\
        \vdots & \ddots & \vdots\\
        q_{K1} & q_{K2} & q_{K3}\\
    \end{matrix}
    \right],
\end{equation}
where each column $\Vect{q}_c \in \mathbb{R}^{K \times 1}$.
Therefore, the spectrum-to-color projection results in a color image
\begin{equation}
    \Mat{Y} = \Mat{X} \Mat{Q},
    \label{eq:supp_color_formation}
\end{equation}
where $\Mat{Y} \in \mathbb{R}^{MN \times 3}$ with three columns
\begin{equation}
    \Mat{Y} = \left[\Vect{y}_1, \Vect{y}_2, \Vect{y}_3\right],
\end{equation}
and each column $\Vect{y}_c \in \mathbb{R}^{MN \times 1}$ is a vector for the image in color channel $c \in [R, G, B]$.

When considering the spectral PSFs in each spectral channel, the optically blurred image can be expressed by 
\begin{equation}
    \Vect{w}_k = \Mat{A}_k \Vect{x}_k, \quad k \in [1, 2, \dots, K],
\end{equation}
where $\Mat{A}_k \in \mathbb{R}^{MN \times MN}$ is a matrix that represents the spectral PSF in channel $k$. Similar as $\Mat{X}$, we concatenate $\Mat{w}_k$ horizontally to obtain the spectral images through the optical system as
\begin{equation}
\begin{aligned}
    \Mat{W} &= \left[\Vect{w}_1, \Vect{w}_2, \dots, \Vect{w}_K\right]\\
            &= \left[\Mat{A}_1 \Vect{x}_1, \Mat{A}_2 \Vect{x}_2, \dots, \Mat{A}_K \Vect{x}_K\right].\\
\end{aligned} 
\end{equation}
We define a block matrix 
\begin{equation}
    \Mat{A} = 
    \left[
    \begin{matrix}
        \Mat{A}_1\\
        \Mat{A}_2\\
        \dots\\
        \Mat{A}_K\\
    \end{matrix}
    \right],
\end{equation}
which stacks the matrices $\Mat{A}_k$ vertically, and $\Mat{A} \in \mathbb{R}^{KMN \times MN}$. Therefore, we have 
\begin{equation}
    \Mat{W} = \mathrm{diag} \left(\Mat{A} \Mat{X}\right),
\end{equation}
where $\mathrm{diag}\left(\cdot\right)$ extracts the $K$ diagonal blocks and concatenate them horizontally,
\begin{equation}
    \begin{aligned}
        \Mat{A} \Mat{X} &=
        \left[
        \begin{matrix}
            \Mat{A}_1\\
            \Mat{A}_2\\
            \dots\\
            \Mat{A}_K\\
        \end{matrix}
        \right]
        \left[
        \Vect{x}_1, \Vect{x}_2, \dots, \Vect{x}_K
        \right]\\
        &=
        \left[
        \begin{matrix}
            \Mat{A}_1 \Vect{x}_1 & \Mat{A}_1 \Vect{x}_2 & \cdots & \Mat{A}_1 \Vect{x}_K\\
            \Mat{A}_2 \Vect{x}_1 & \Mat{A}_2 \Vect{x}_2 & \cdots & \Mat{A}_2 \Vect{x}_K\\
            \vdots               & \vdots               & \ddots & \vdots\\
            \Mat{A}_K \Vect{x}_1 & \Mat{A}_K \Vect{x}_2 & \cdots & \Mat{A}_K \Vect{x}_K\\
        \end{matrix}    
        \right].
    \end{aligned}
\end{equation}
Finally, the color image is 
\begin{equation}
\Mat{Z} = \Mat{W} \Mat{Q} = \mathrm{diag} \left(\Mat{A} \Mat{X} \right) \Mat{Q}.
\end{equation}
where $\Mat{Z} = \left[\Vect{z}_1, \Vect{z}_2, \Vect{z}_3\right] \in \mathbb{R}^{MN \times 3}$.

\subsection{Effect of clipping to non-negative values.}
It is necessary to clip negative values in the generated metamer data to ensure the resulting spectra are physically plausible (\ie, no negative spectral radiance). This may lead to slight deviations in the RGB values, and therefore images that are not {\em exact} metamers. However, we verify that the resulting difference is actually negligible by comparing the projected RGB images from the metamer pairs. For example, in the experiments of Table~5 in the main paper, 32.9\% of the generated metamers produce exactly the same RGB images (exact-metamers). Among the remaining 67.1\% that are affected by clipping (\ie, near-metamers), the average PSNR between the RGB pairs is 75.8~dB, with a standard deviation of $\pm$17.7~dB. This indicates that the effect of clipping the negative values in the metamer spectra is negligible.
}

\bibliographystyle{plain}
\bibliography{main}

\end{document}

\title{Limitations of Data-Driven Spectral Reconstruction: An Optics-Aware Analysis -- Supplementary Material}

\author{Qiang Fu*, Matheus Souza*, Eunsue Choi, Suhyun Shin, Seung-Hwan Baek, Wolfgang Heidrich~\IEEEmembership{Fellow,~IEEE}
        % <-this % stops a space
}

%\markboth{Transactions on Computational Imaging,~Vol.~xx, No.~x, XXX~2025}%
\markboth{}
{Fu \MakeLowercase{\textit{et al.}}: Limitations of Data-Driven Spectral Reconstruction}

% \IEEEpubid{0000--0000/00\$00.00~\copyright~2025 IEEE}
% \IEEEpubidadjcol 

\maketitle

% \linenumbers

%---------- main texts ----------%
\section{Aberrated Spectral PSFs}

In Supplemental Fig.~\ref{fig:supp_double_gauss}, we show the schematic optical layout of the double Gauss lens~\cite{ichimura2021optical} we use throughout the experiments. It consists of 6 lens elements with an aperture stop in the middle. The effective focal length is 50~mm, and the F-number is F/1.8. We model the spectral PSFs at each wavelength in the spectral range [400~nm, 700~nm] with a step size of 10~nm in the optical design software ZEMAX (v14.2) by spectral ray tracing. The sensor parameters are set according to the specifications of Basler ace 2 camera (model A2a5320-23ucBAS), as used in the NTIRE 2022 spectral recovery challenge~\cite{arad2022ntire}. In Supplemental Fig.~\ref{fig:supp_double_gauss}, we render the corresponding spectral PSFs in color with the SRF of that sensor. Although the lens is well designed to minimize all kinds of aberrations, clear chromatic aberrations can still be observed, in particular in the short (blue) and long (red) ends of the spectral bands. It is impossible to completely {\em eliminate} aberrations in photographic lenses~\cite{smith2008modern,fischer2000optical}. Note that all the spectral PSFs are normalized by their own maximum values for visualization purpose only.

% double Gauss lens and PSFs
\begin{figure*}[!htp]
\centering
\begin{tikzpicture}
\node[anchor=south west, inner sep=0] at (0,0) {\includegraphics[width=0.98\textwidth]{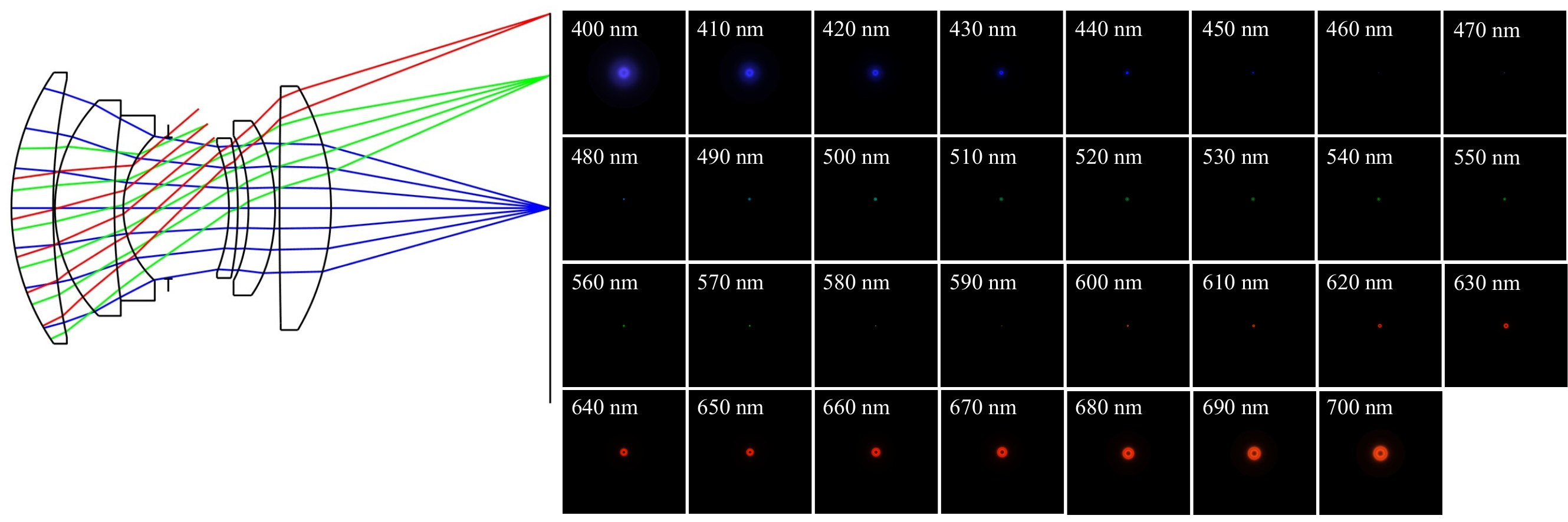}};
% lens info
\draw (2.2,0.5) node[text width=5cm,align=left] {\textcolor{black}{\small{Double Gauss Lens, f50~mm, F/1.8}}};
\end{tikzpicture}
\caption{Double Gauss lens layout (left) and its aberrated spectral PSFs (right). Lens data is obtained from~\cite{ichimura2021optical} (Numerical Example 2). Spectral PSFs are modeled in optical design software ZEMAX and rendered in color for the Basler ace 2 camera (model A2a5320-23ucBAS) sensor. Clear chromatic aberrations can be observed throughout the spectral range.}
\label{fig:supp_double_gauss}
\end{figure*}

\section{Results of Training with Less Data for Other Networks}

In Table~2 of the main paper, we summarize that the performance of all the candidate networks on the ARAD1K dataset is mildly affected by using only half of the training data. The detailed validation results over the course of training are shown in Supplemental Fig.~\ref{fig:supp_train_percent1} and Fig.~\ref{fig:supp_train_percent2}. Here we show extended experimental results for the effects of training with 100\%, 50\%, and 20\% of the full training data. All the results consistently support our indication of lack of diversity in the dataset.

% Supplementary results for training with less data, Part I
\begin{figure*}
    \centering
    \begin{tabular}{c@{\hspace{0.6em}}c@{}c@{}c@{}c}
         \raisebox{15mm}{\rotatebox[origin=c]{90}{MST++}}
         &\includegraphics[width=0.24\textwidth]{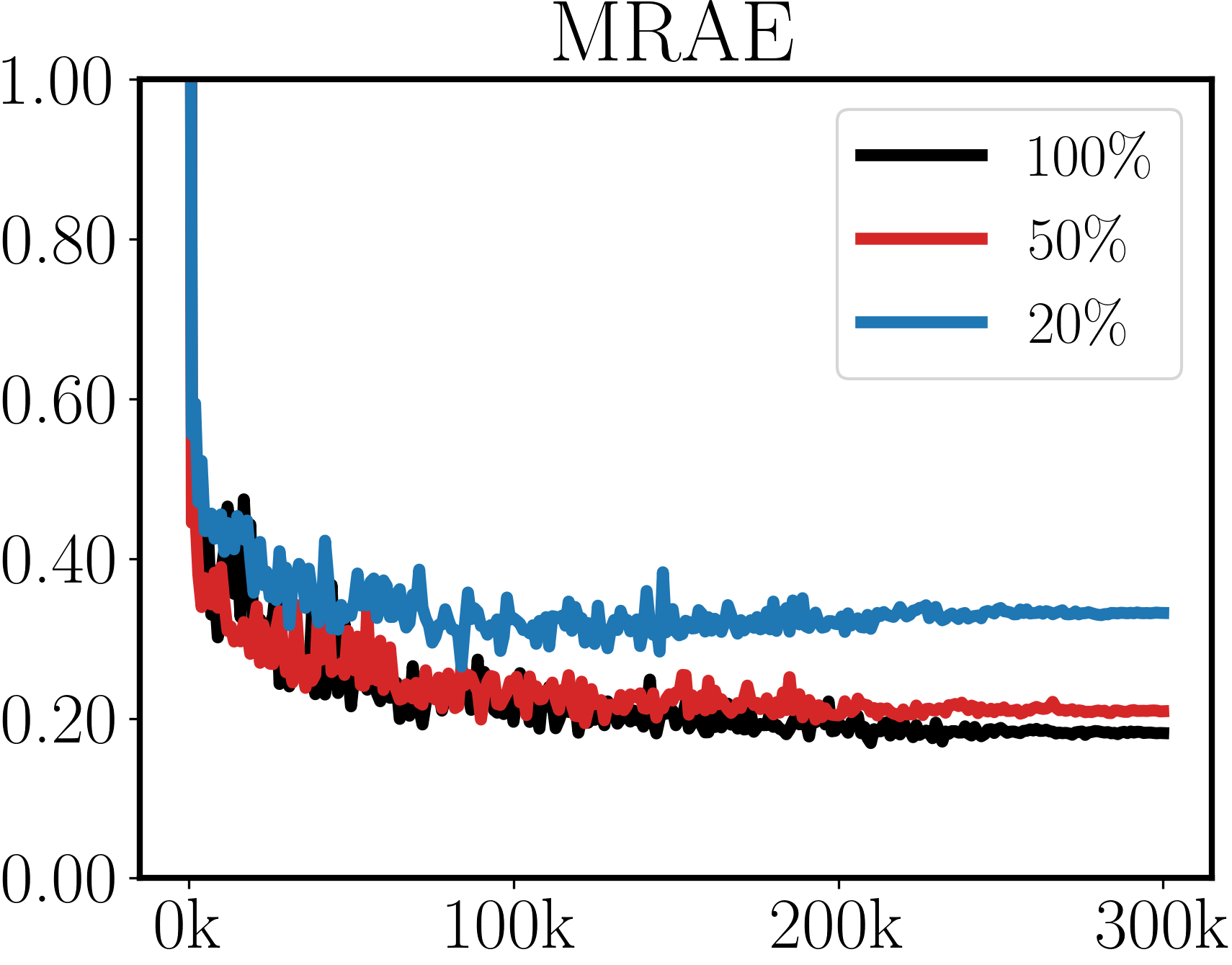}  
         &\includegraphics[width=0.24\textwidth]{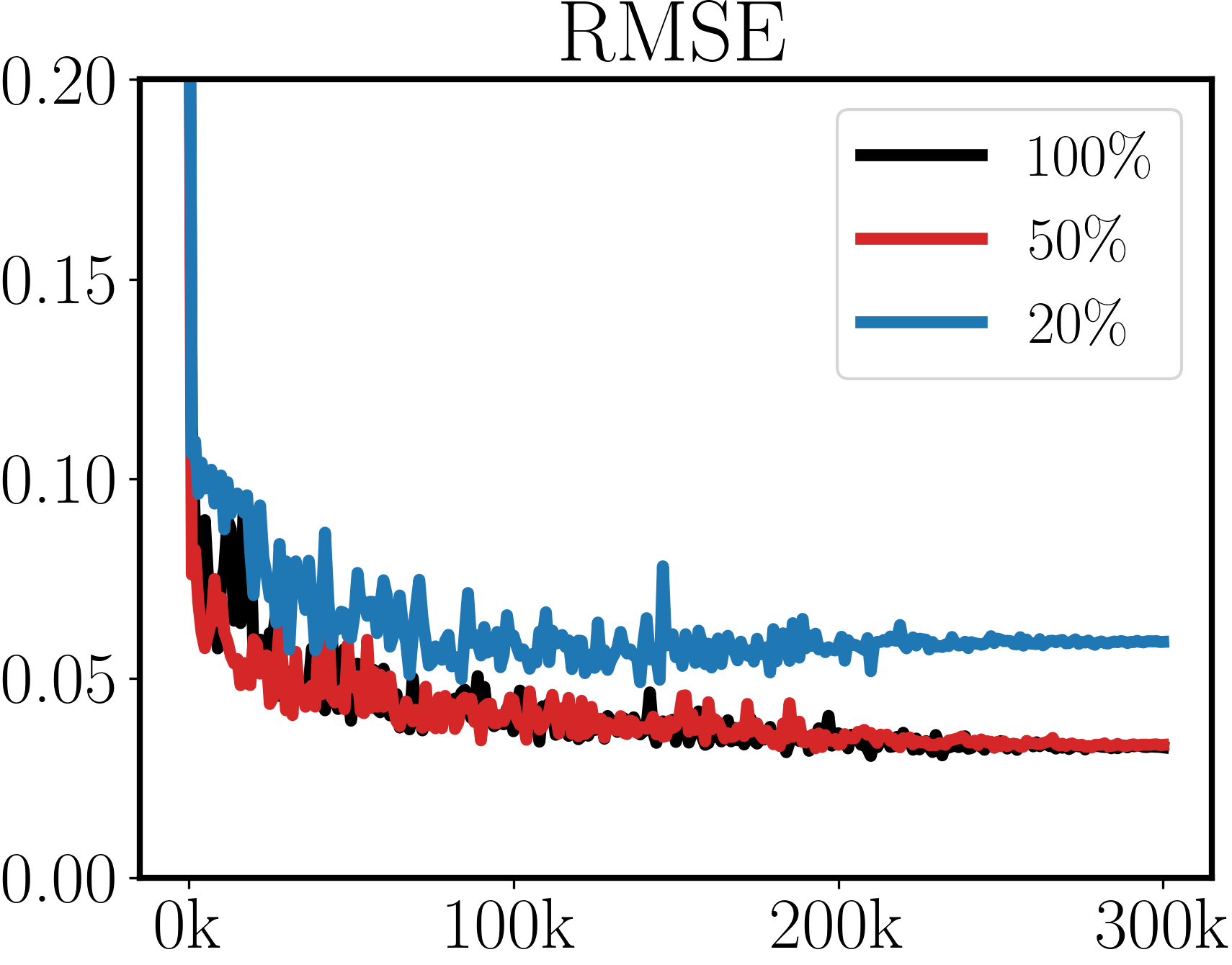}  
         &\includegraphics[width=0.24\textwidth]{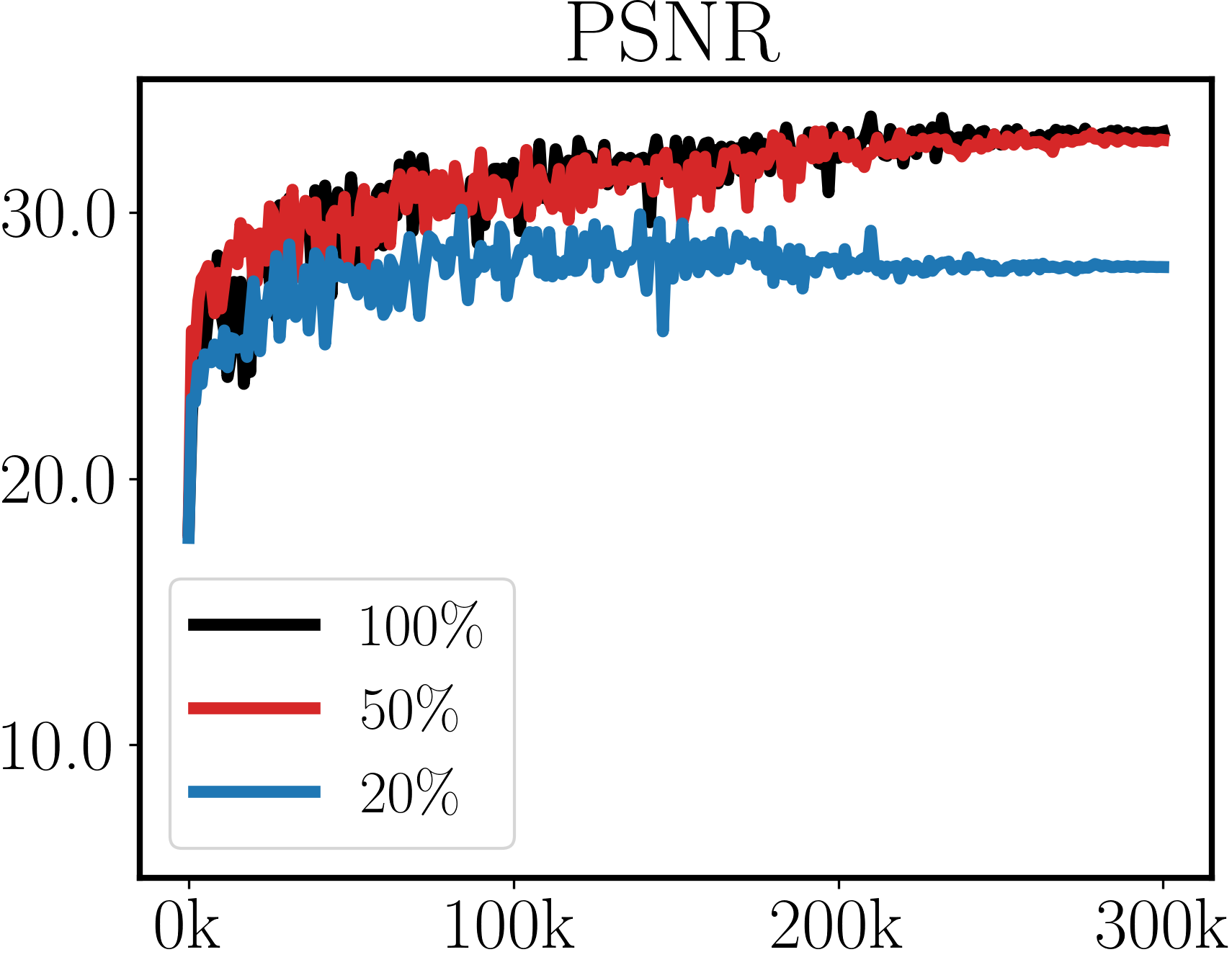}  
         &\includegraphics[width=0.24\textwidth]{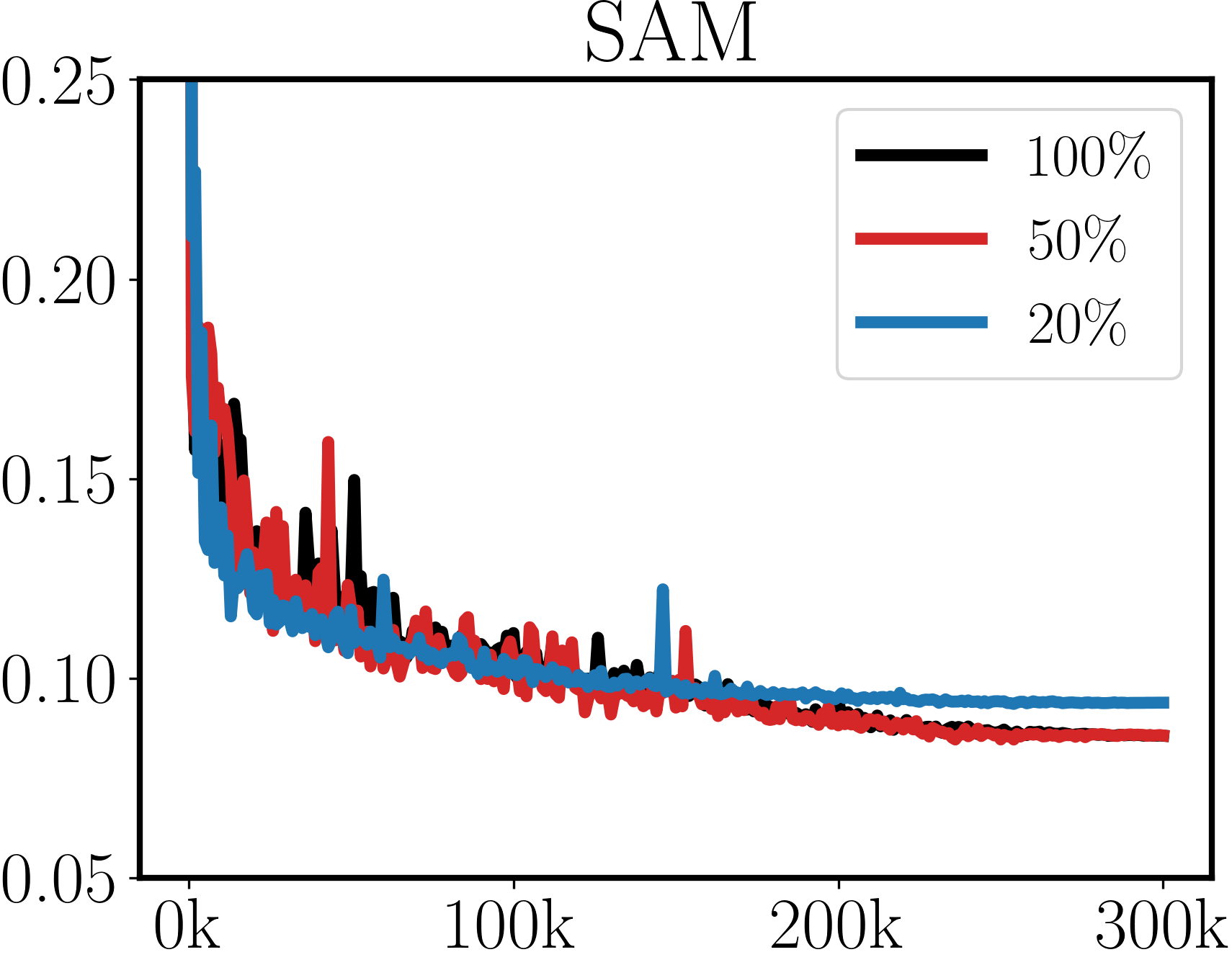}\\
         \raisebox{15mm}{\rotatebox[origin=c]{90}{MST-L}}
         &\includegraphics[width=0.24\textwidth]{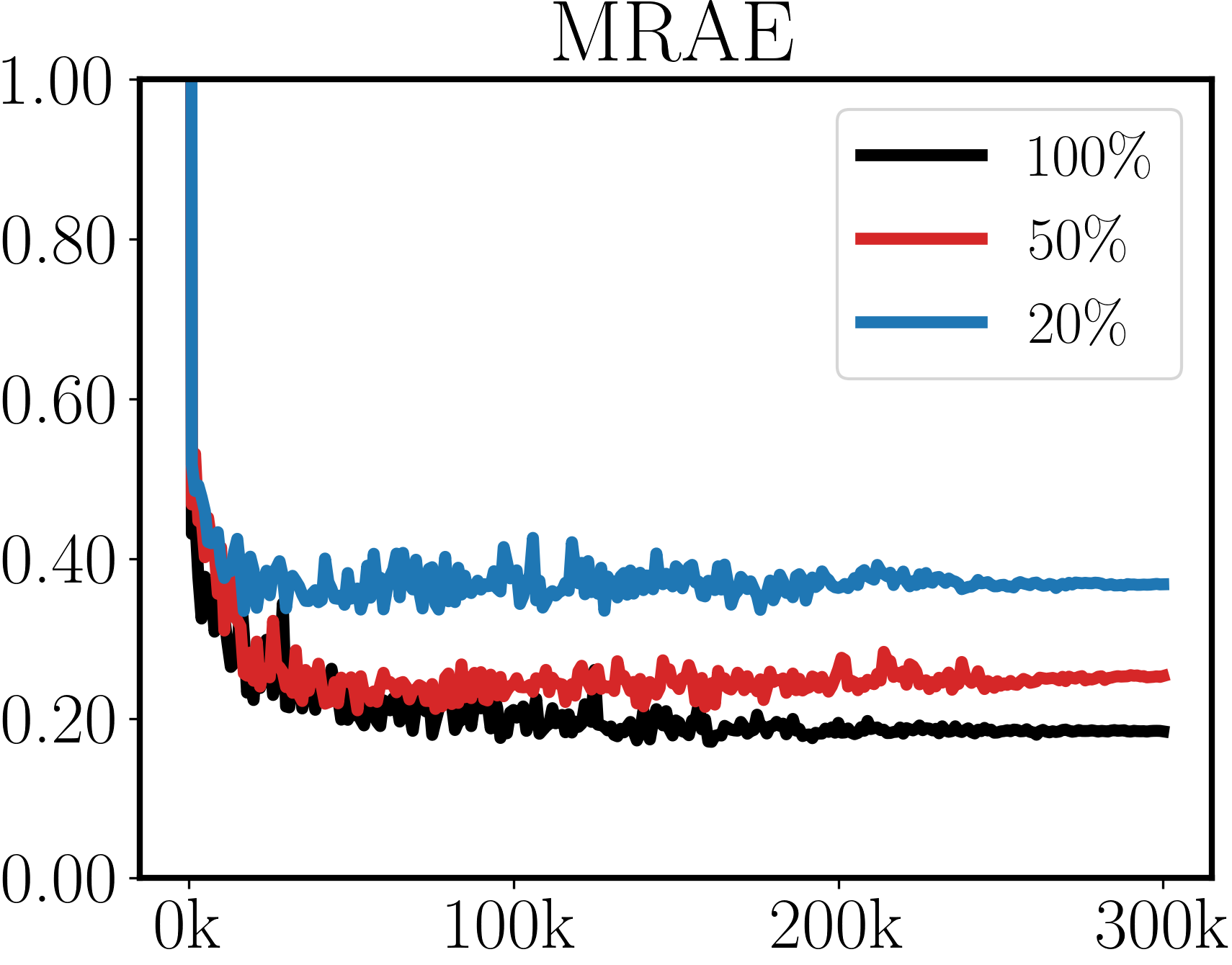}  
         &\includegraphics[width=0.24\textwidth]{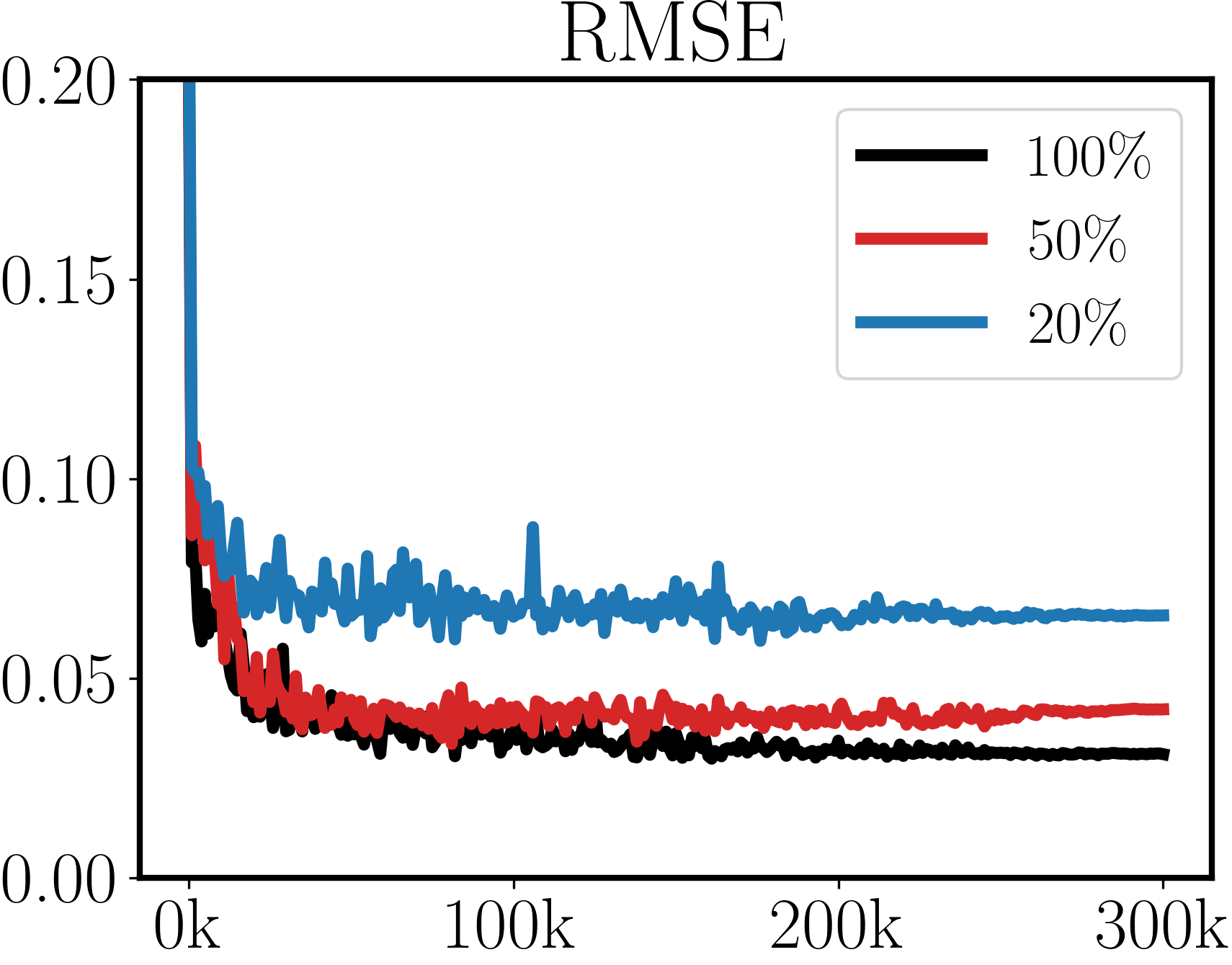}  
         &\includegraphics[width=0.24\textwidth]{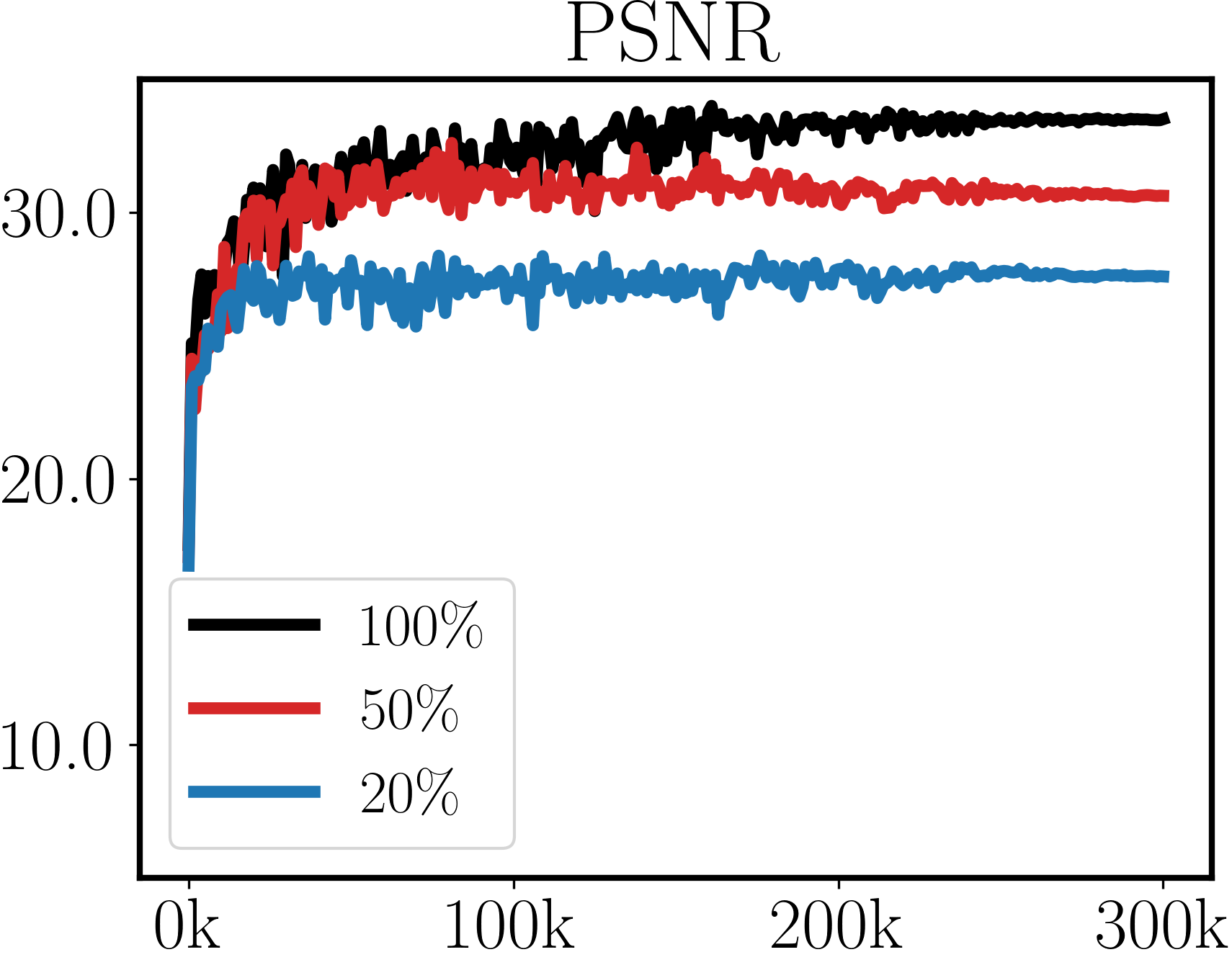}  
         &\includegraphics[width=0.24\textwidth]{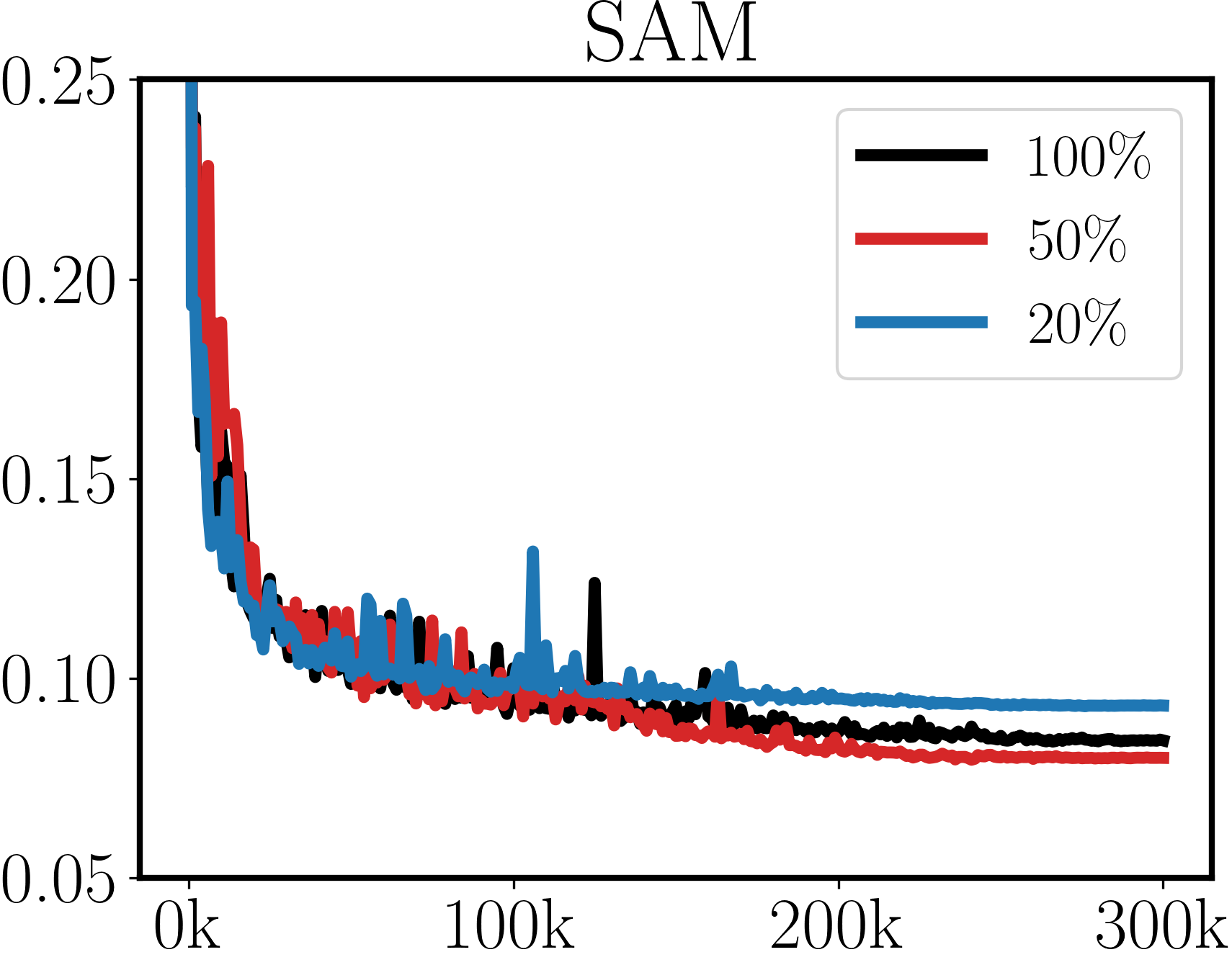}\\
         \raisebox{15mm}{\rotatebox[origin=c]{90}{MPRNet}}
         &\includegraphics[width=0.24\textwidth]{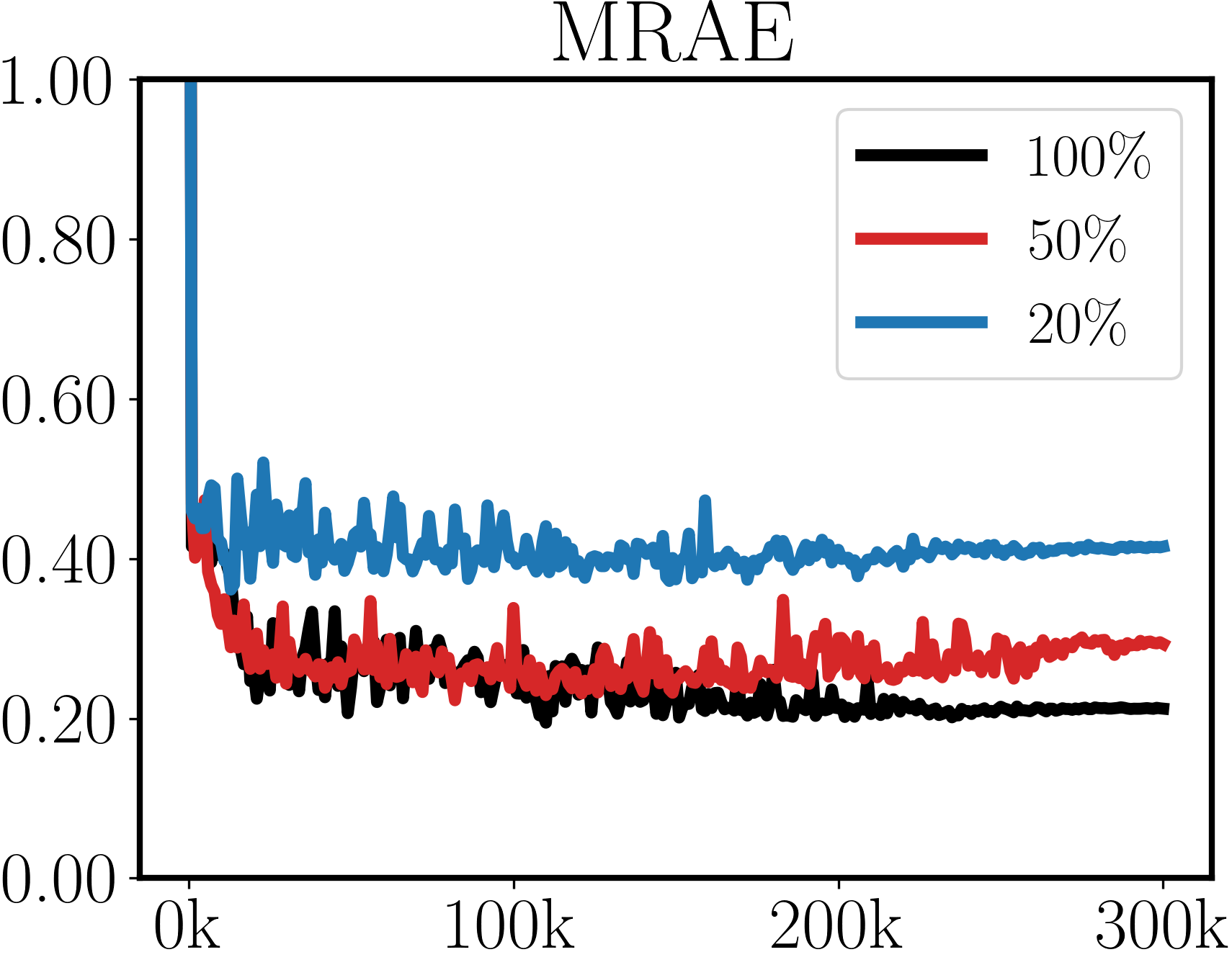}  
         &\includegraphics[width=0.24\textwidth]{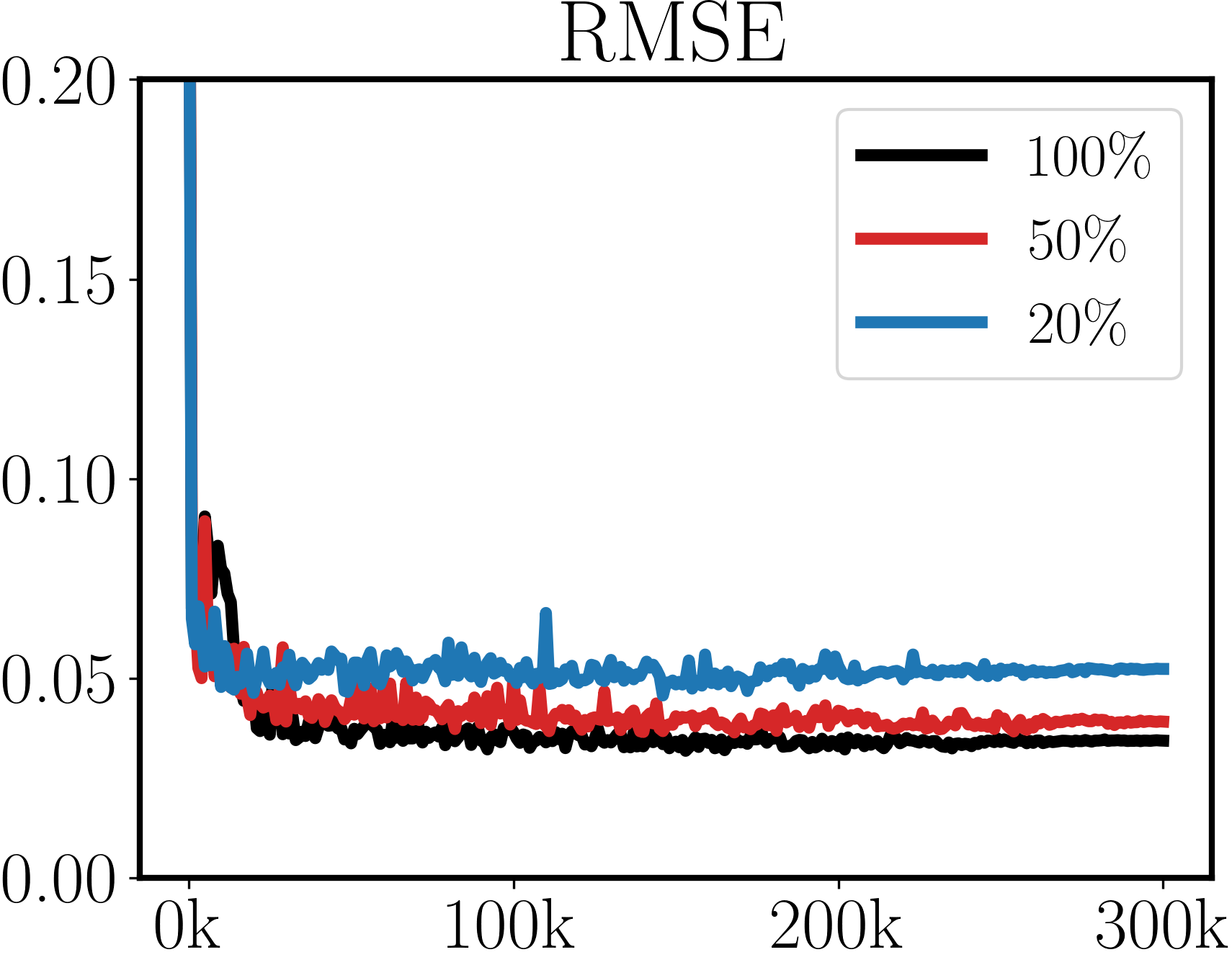}  
         &\includegraphics[width=0.24\textwidth]{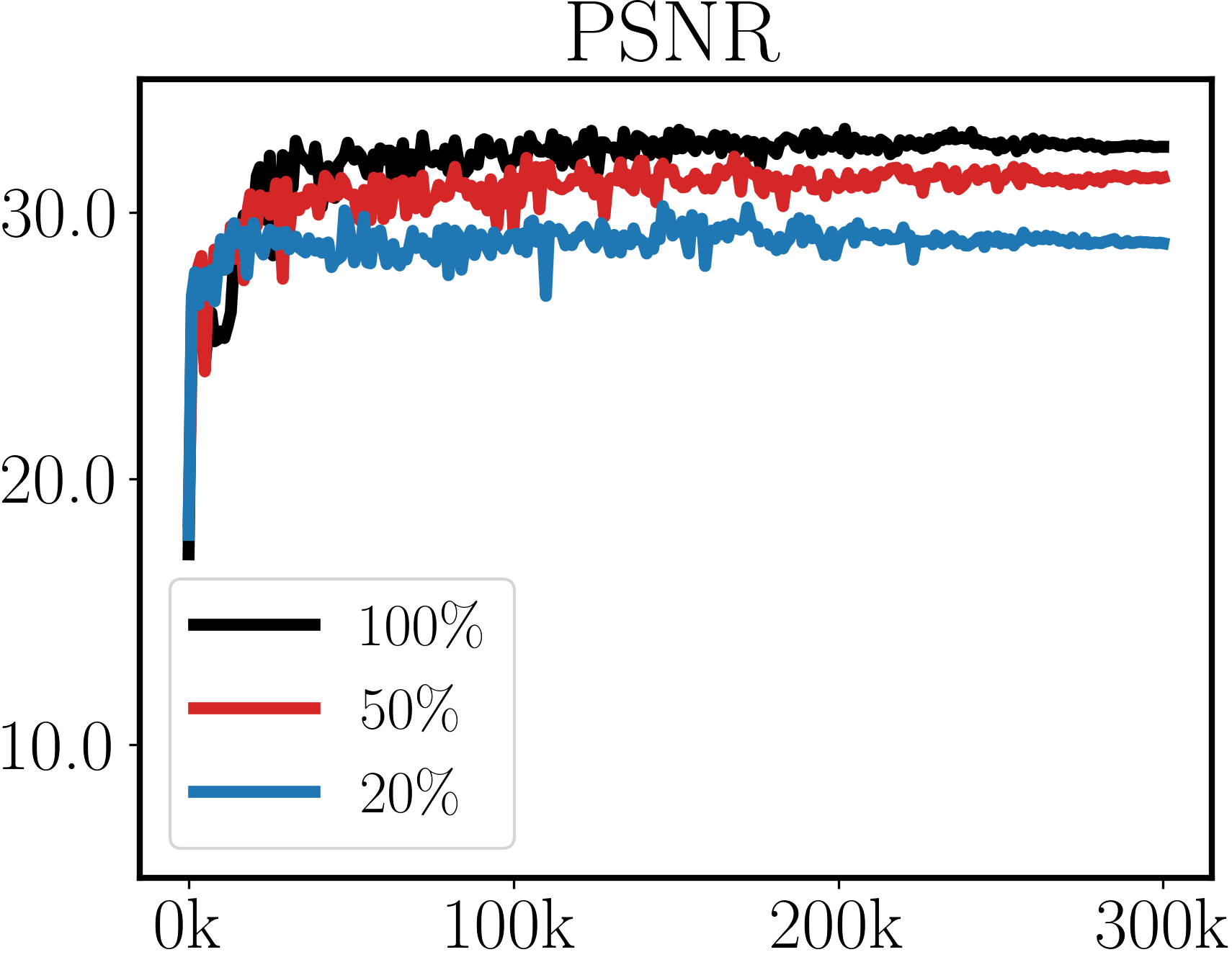}  
         &\includegraphics[width=0.24\textwidth]{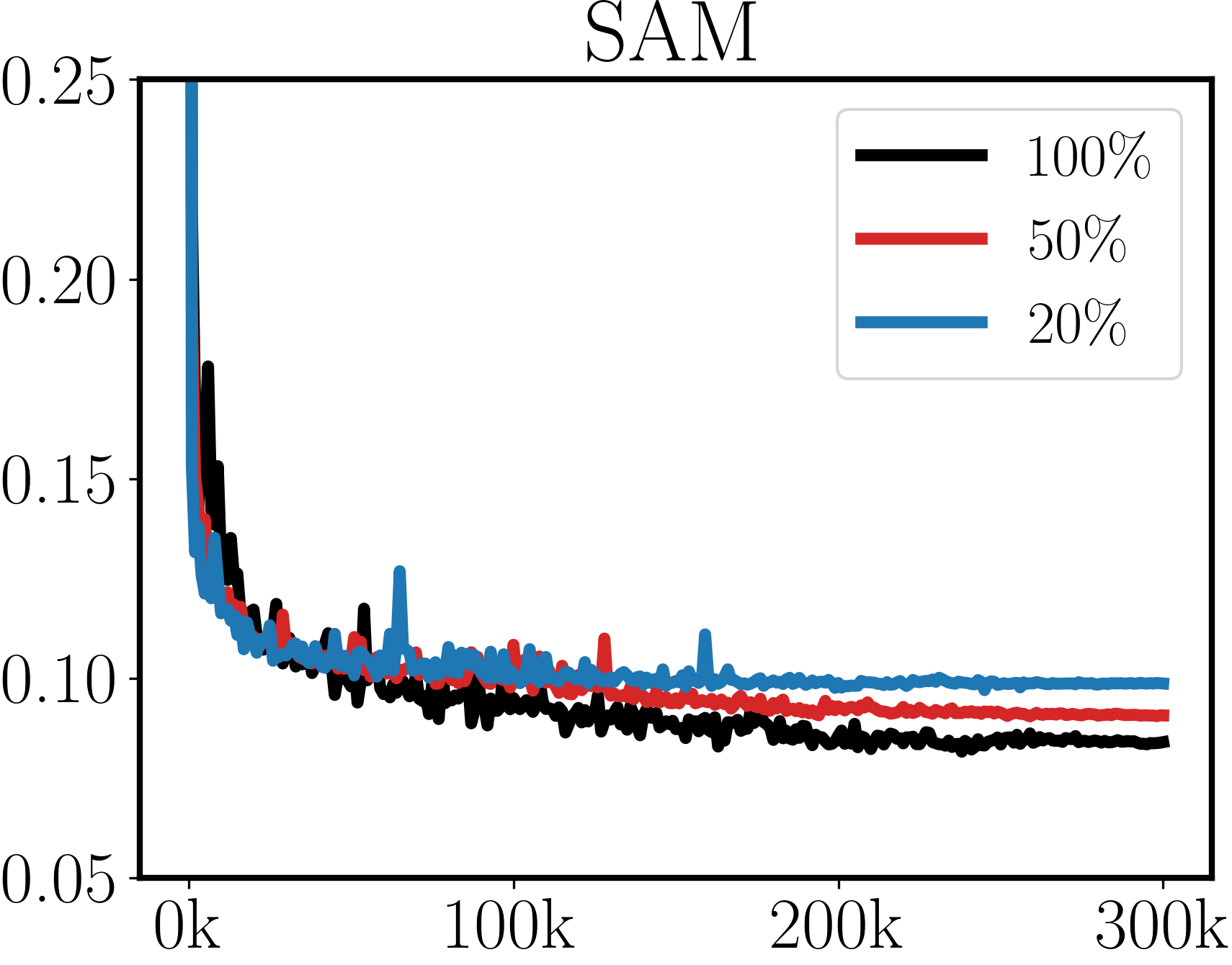}\\
         \raisebox{15mm}{\rotatebox[origin=c]{90}{Restormer}}
         &\includegraphics[width=0.24\textwidth]{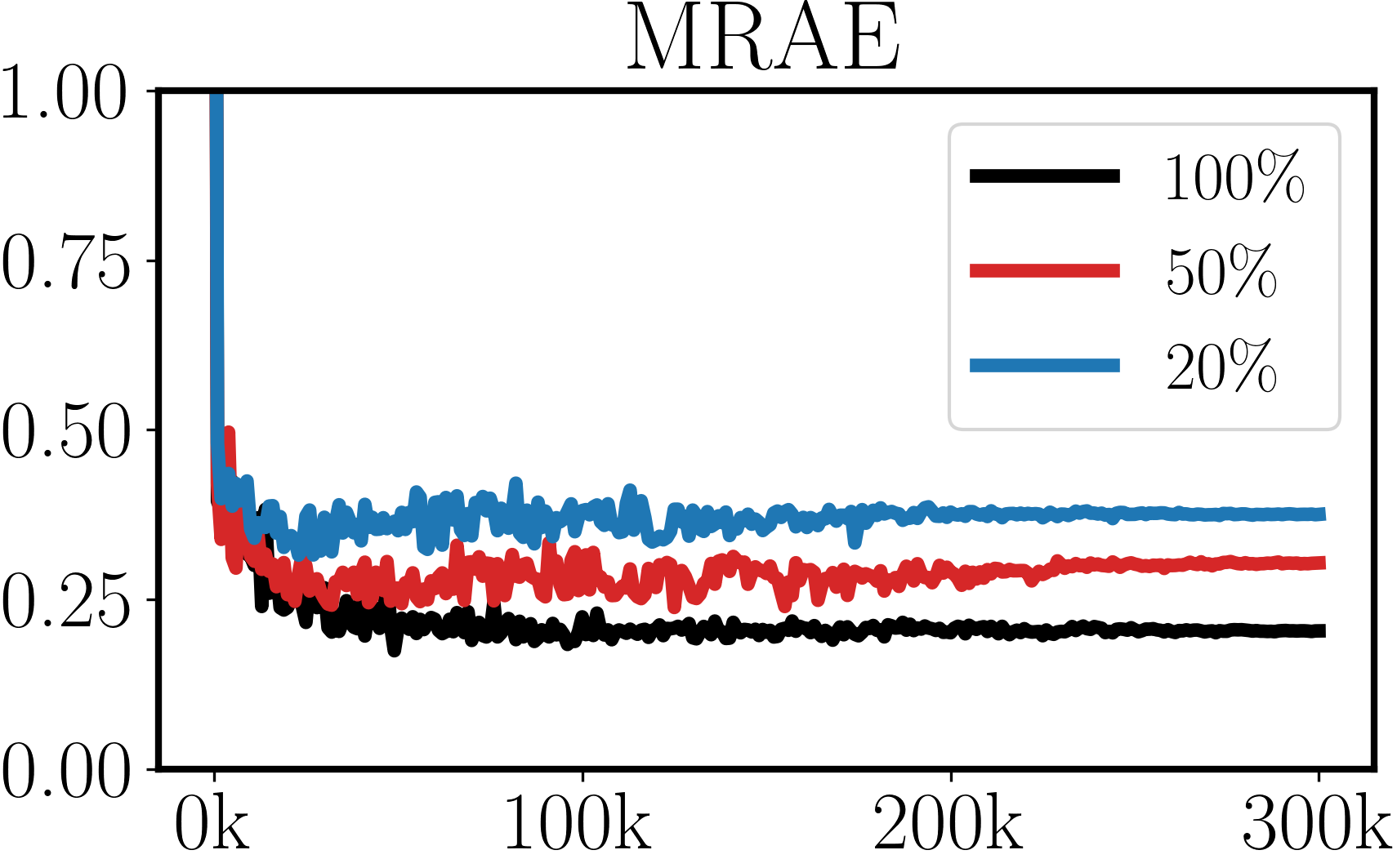}  
         &\includegraphics[width=0.24\textwidth]{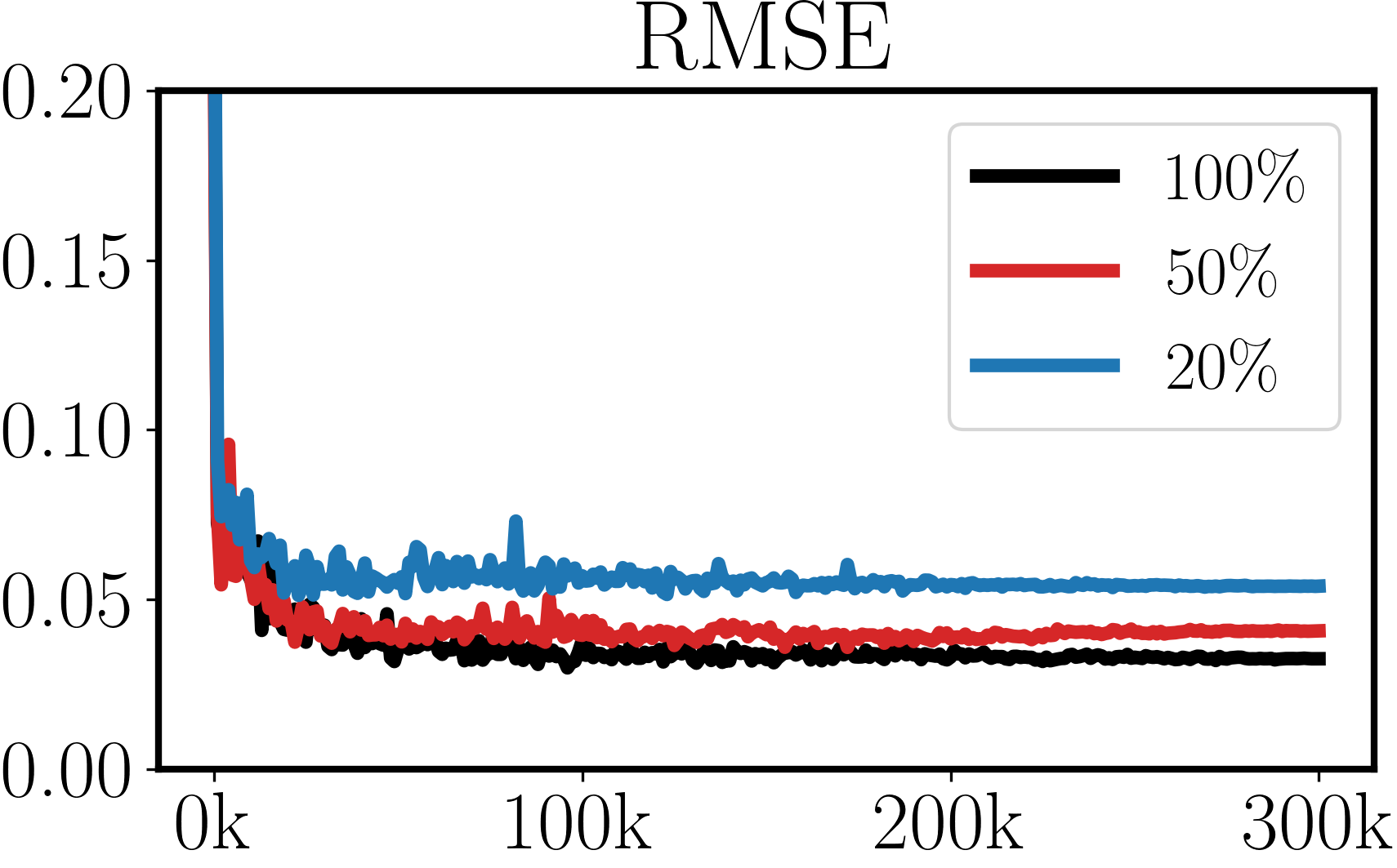}  
         &\includegraphics[width=0.24\textwidth]{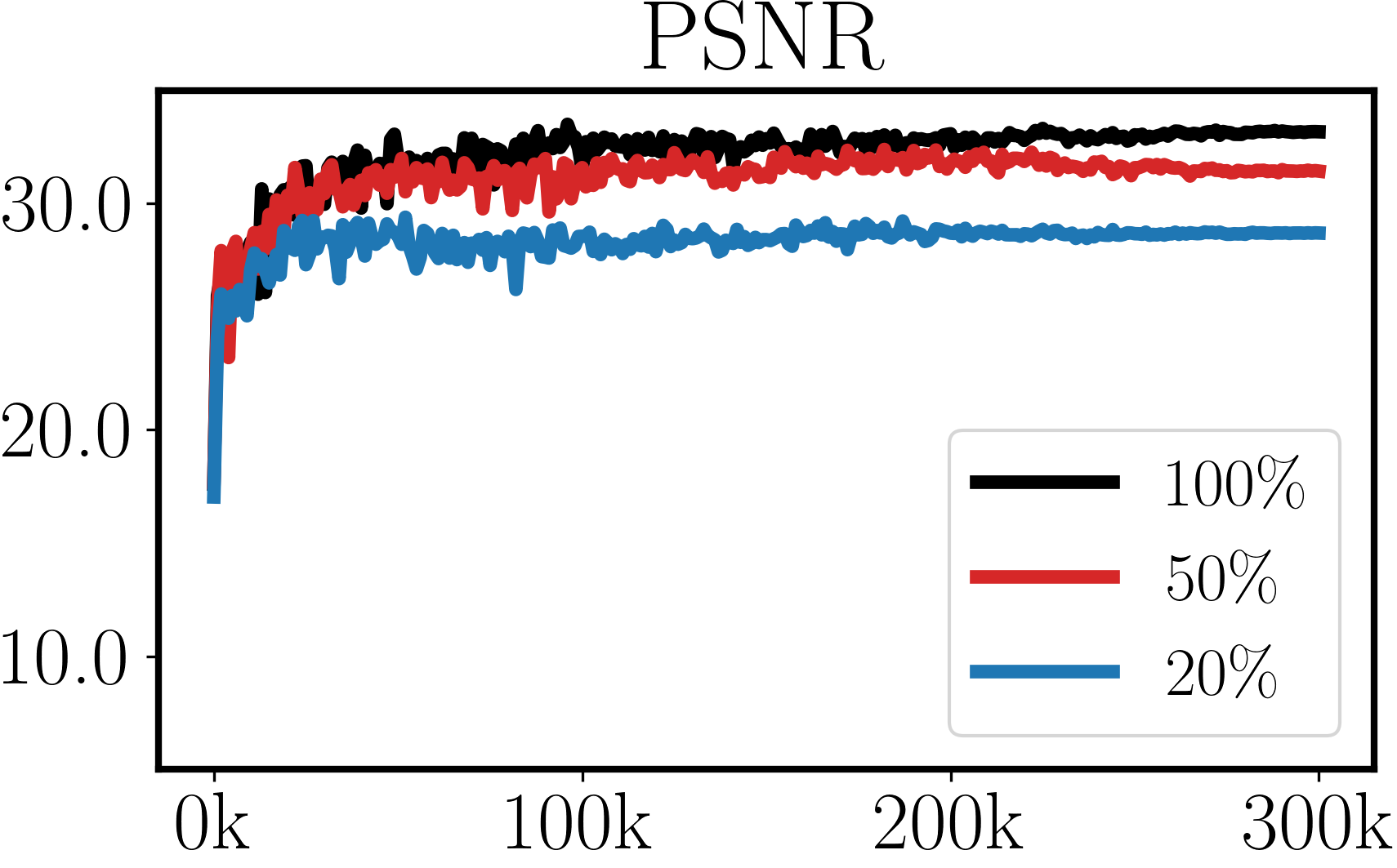}  
         &\includegraphics[width=0.24\textwidth]{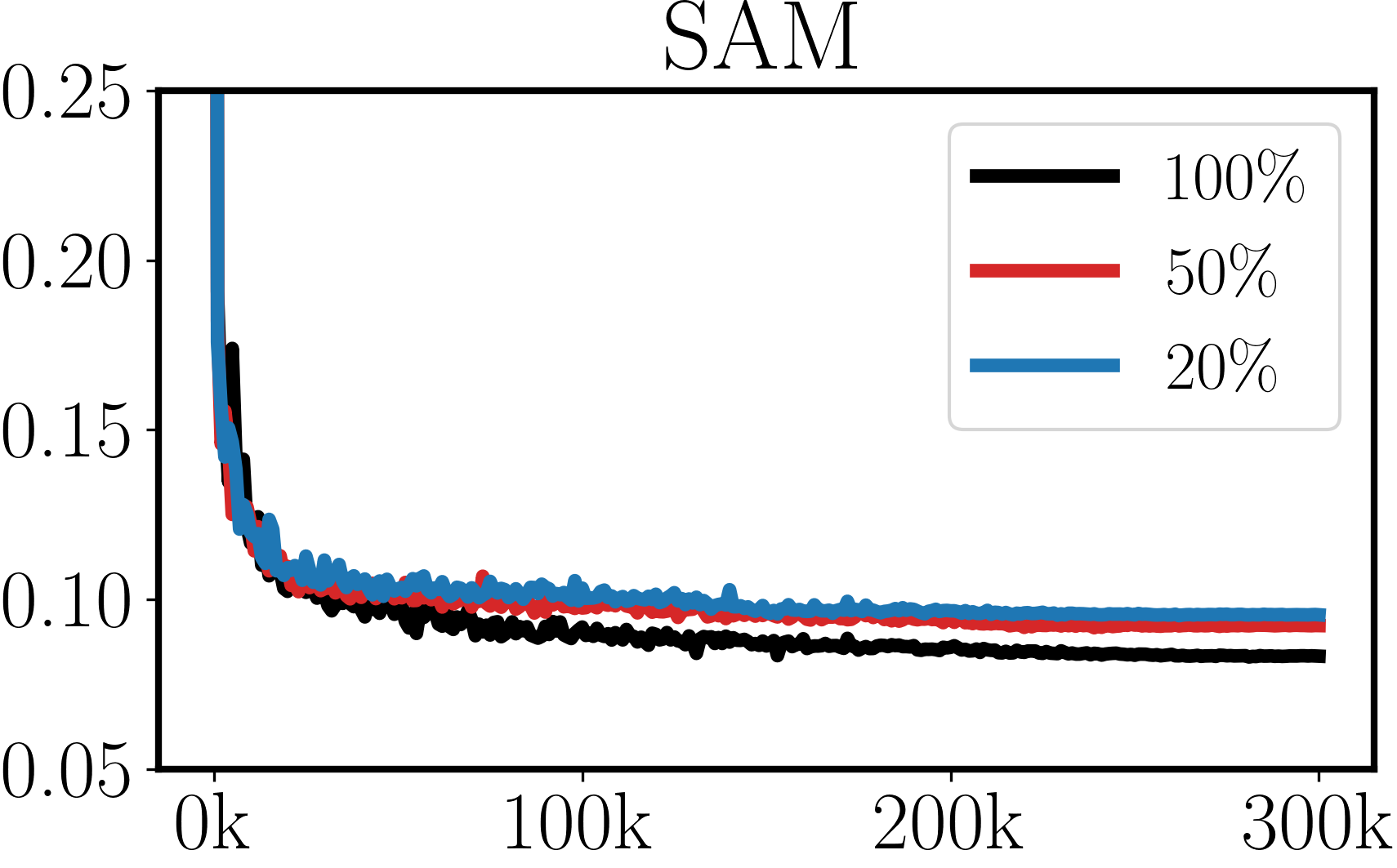}\\
         \raisebox{15mm}{\rotatebox[origin=c]{90}{MIRNet}}
         &\includegraphics[width=0.24\textwidth]{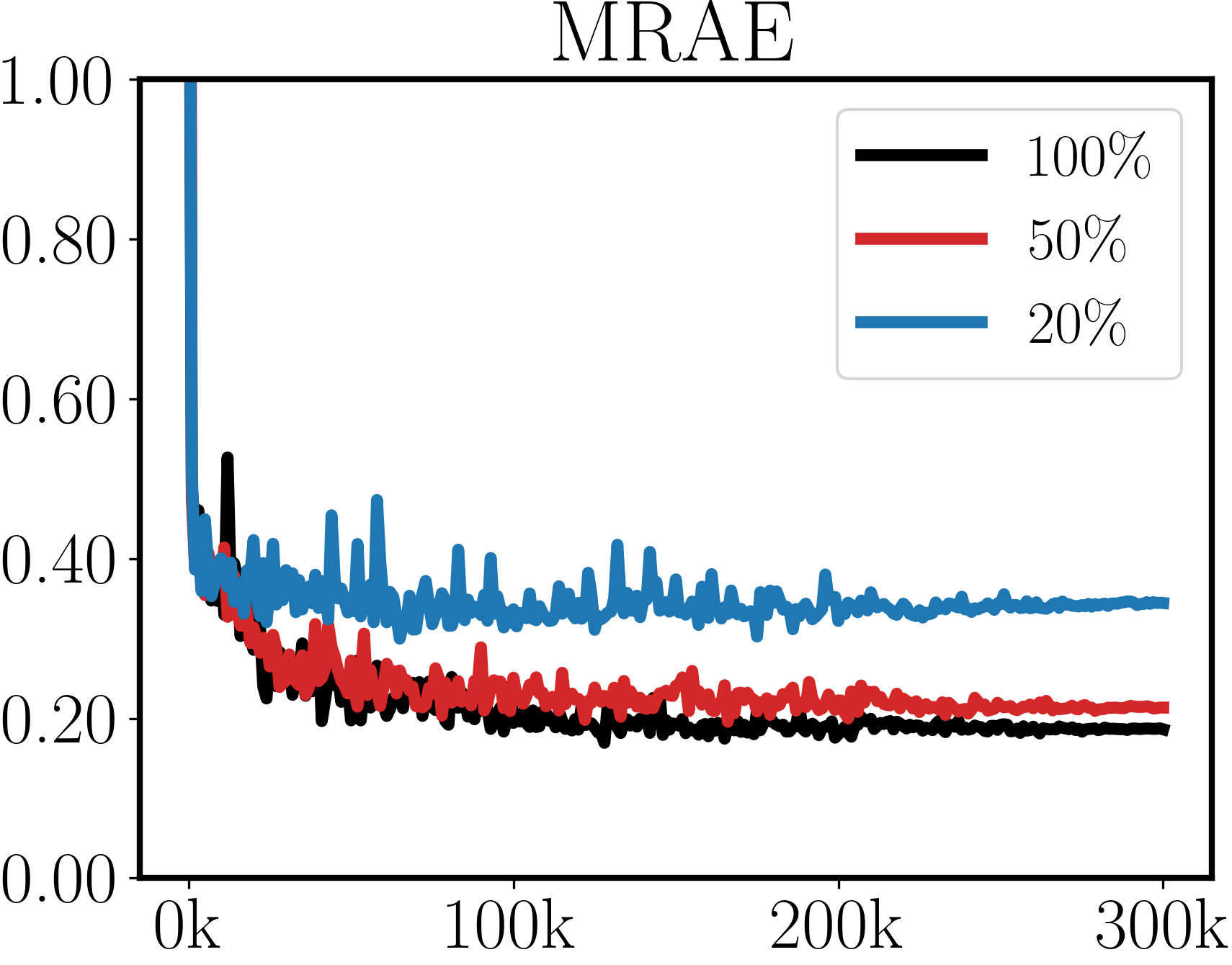}  
         &\includegraphics[width=0.24\textwidth]{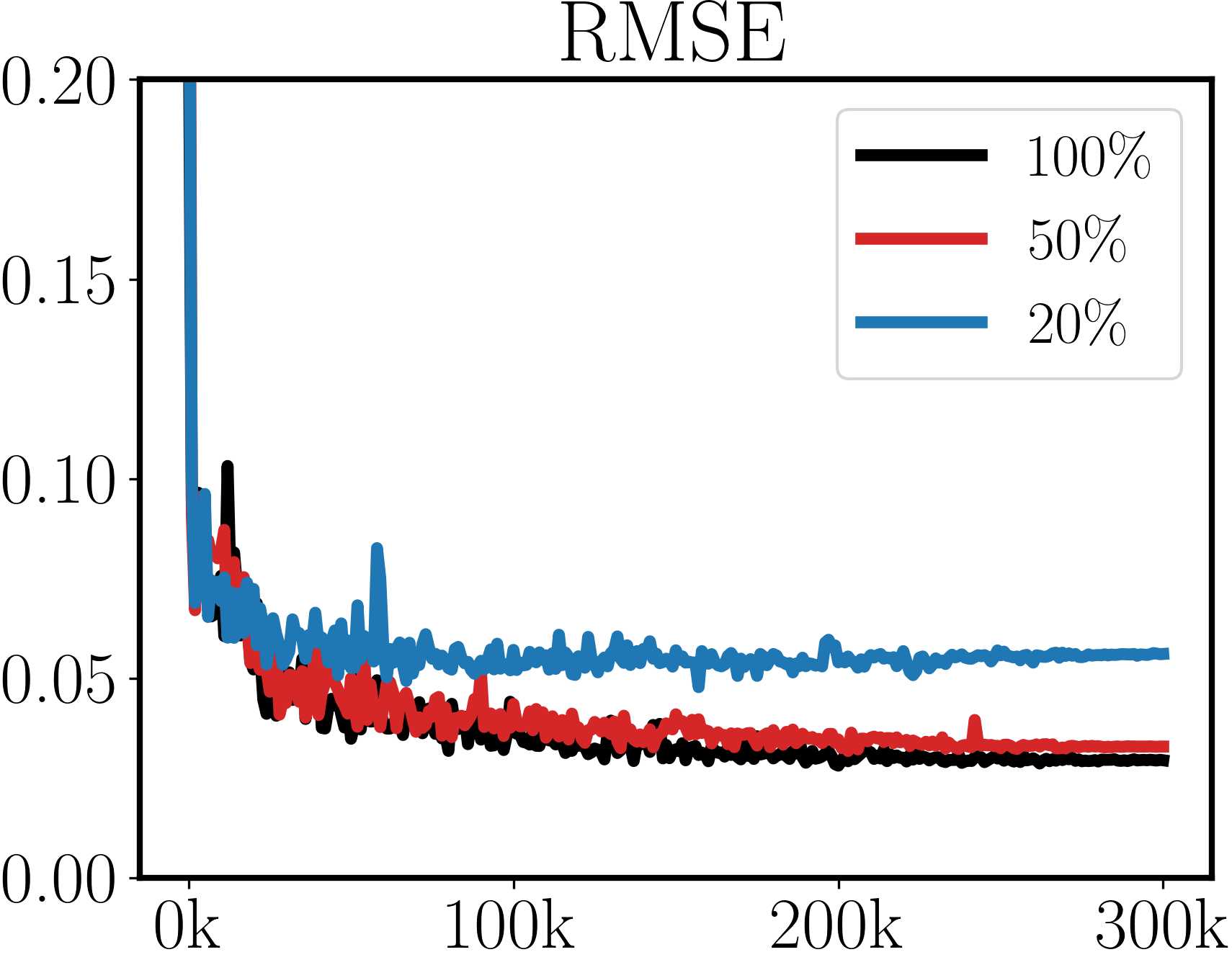}  
         &\includegraphics[width=0.24\textwidth]{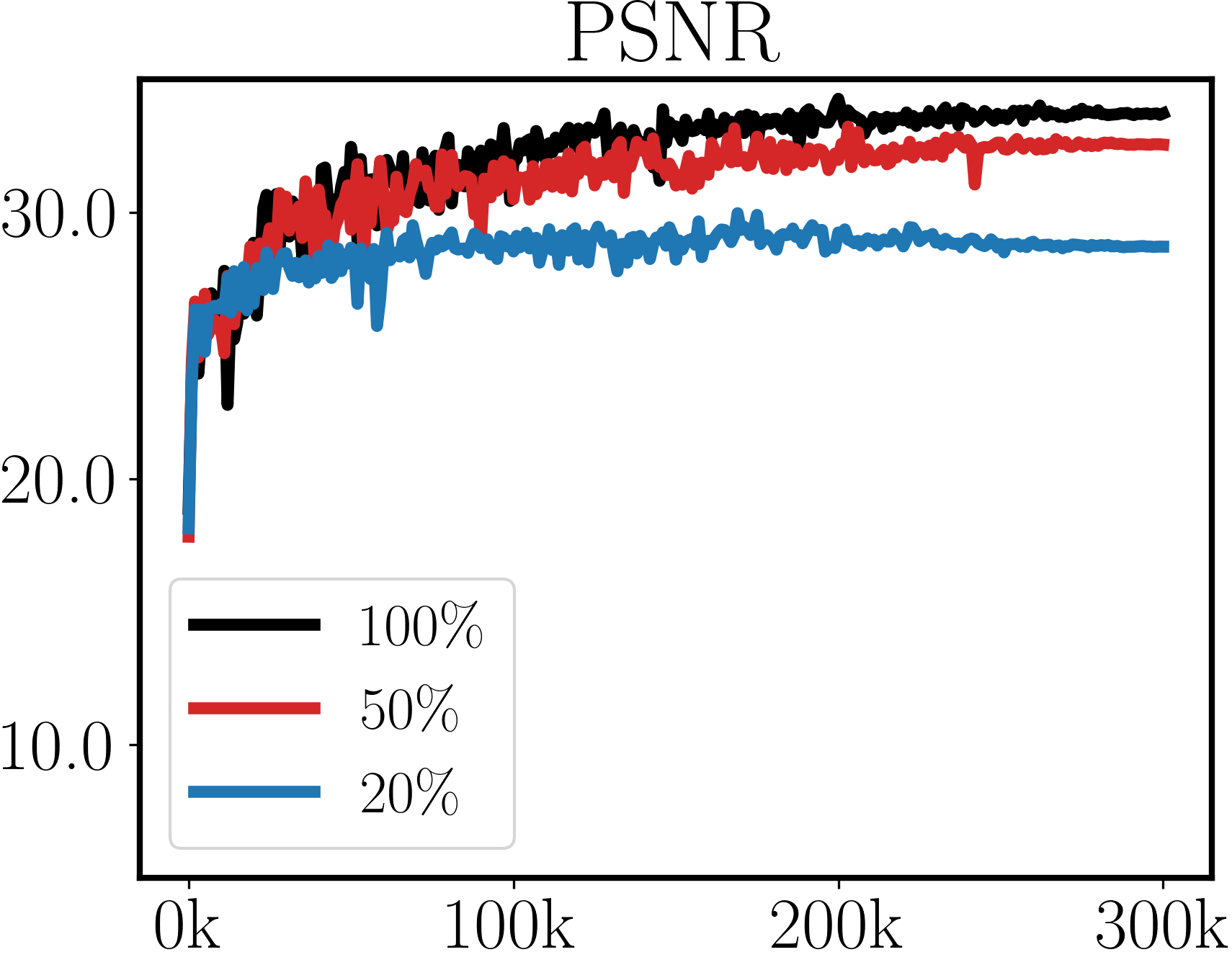}  
         &\includegraphics[width=0.24\textwidth]{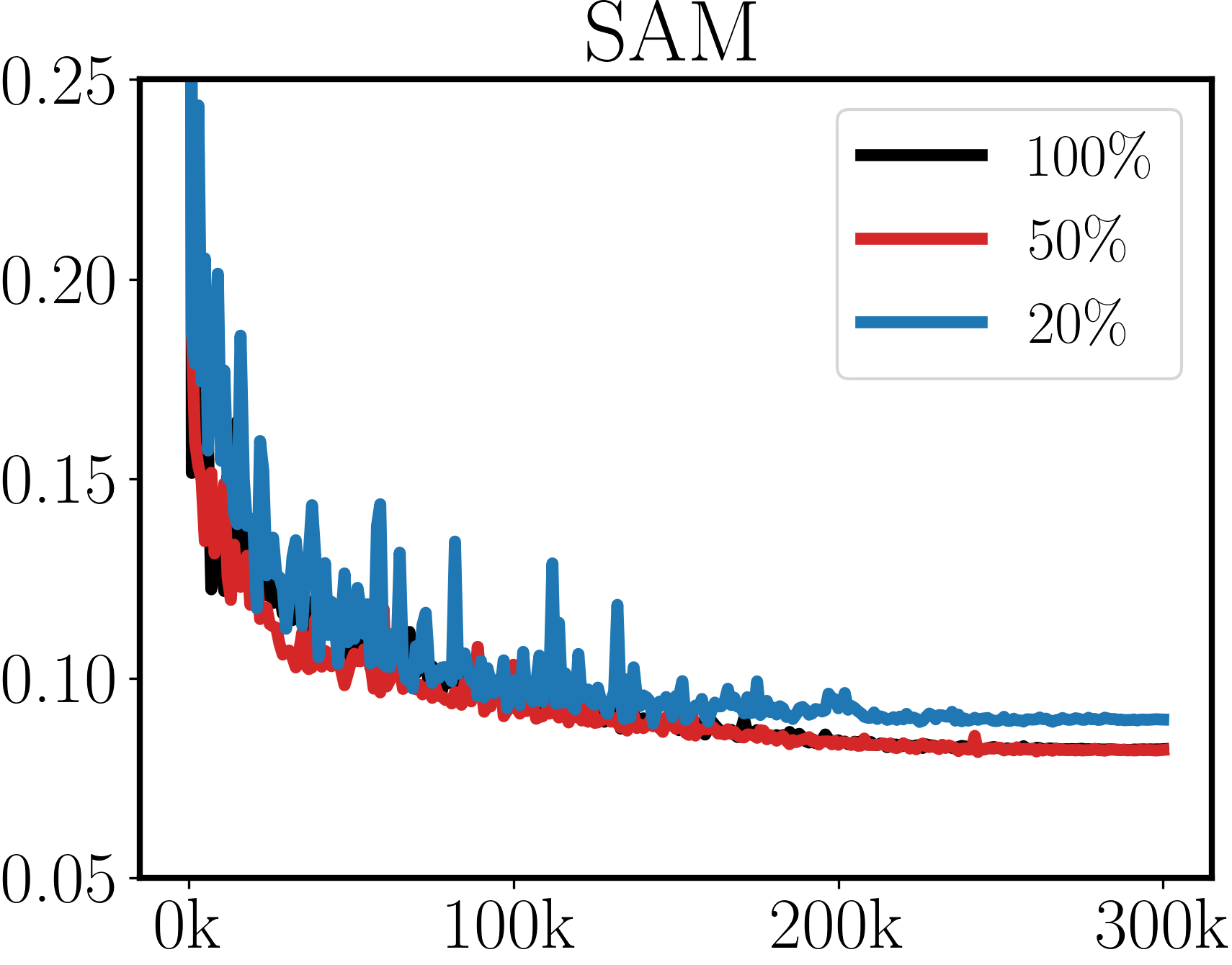}\\
         \raisebox{15mm}{\rotatebox[origin=c]{90}{HINet}}
         &\includegraphics[width=0.24\textwidth]{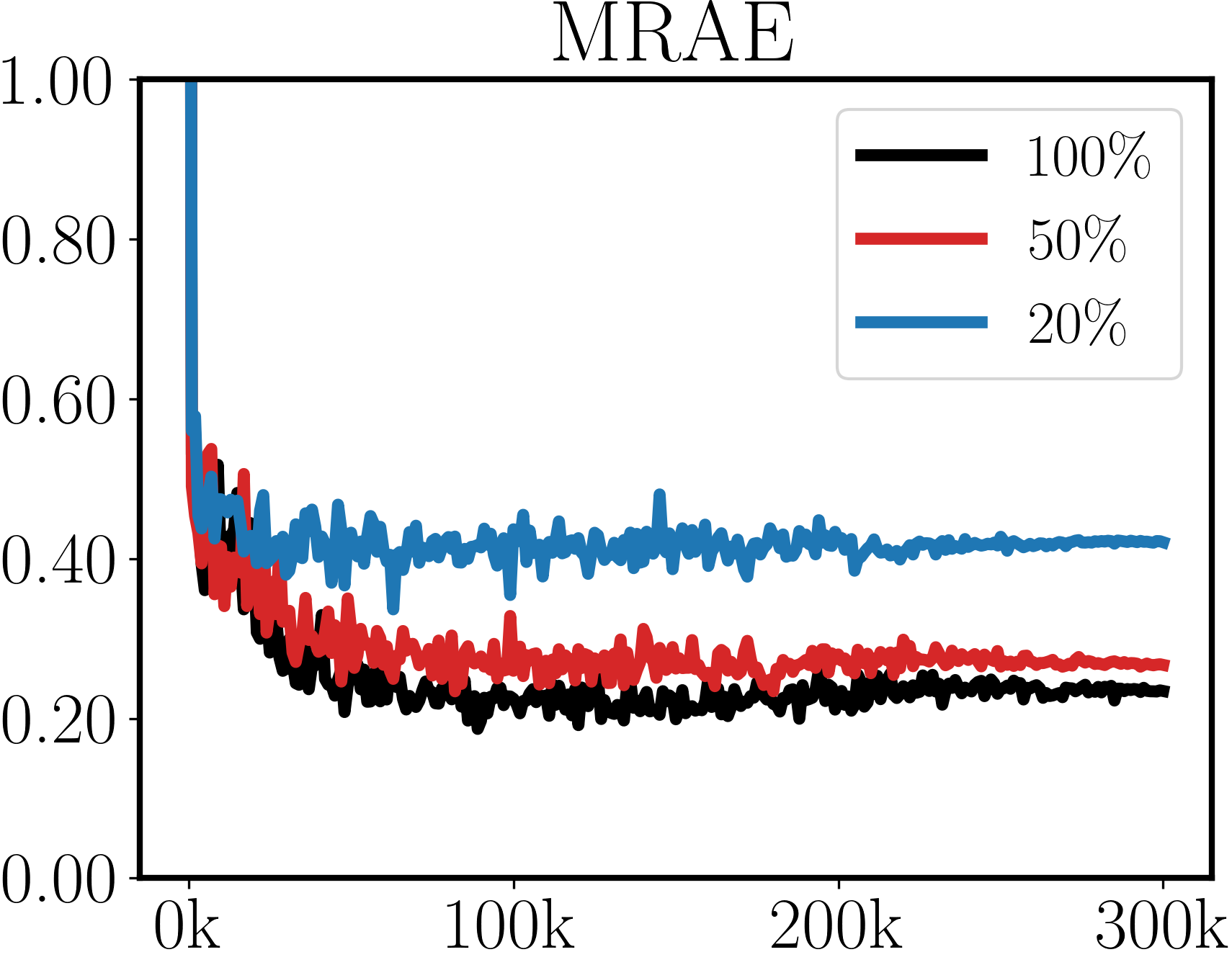}  
         &\includegraphics[width=0.24\textwidth]{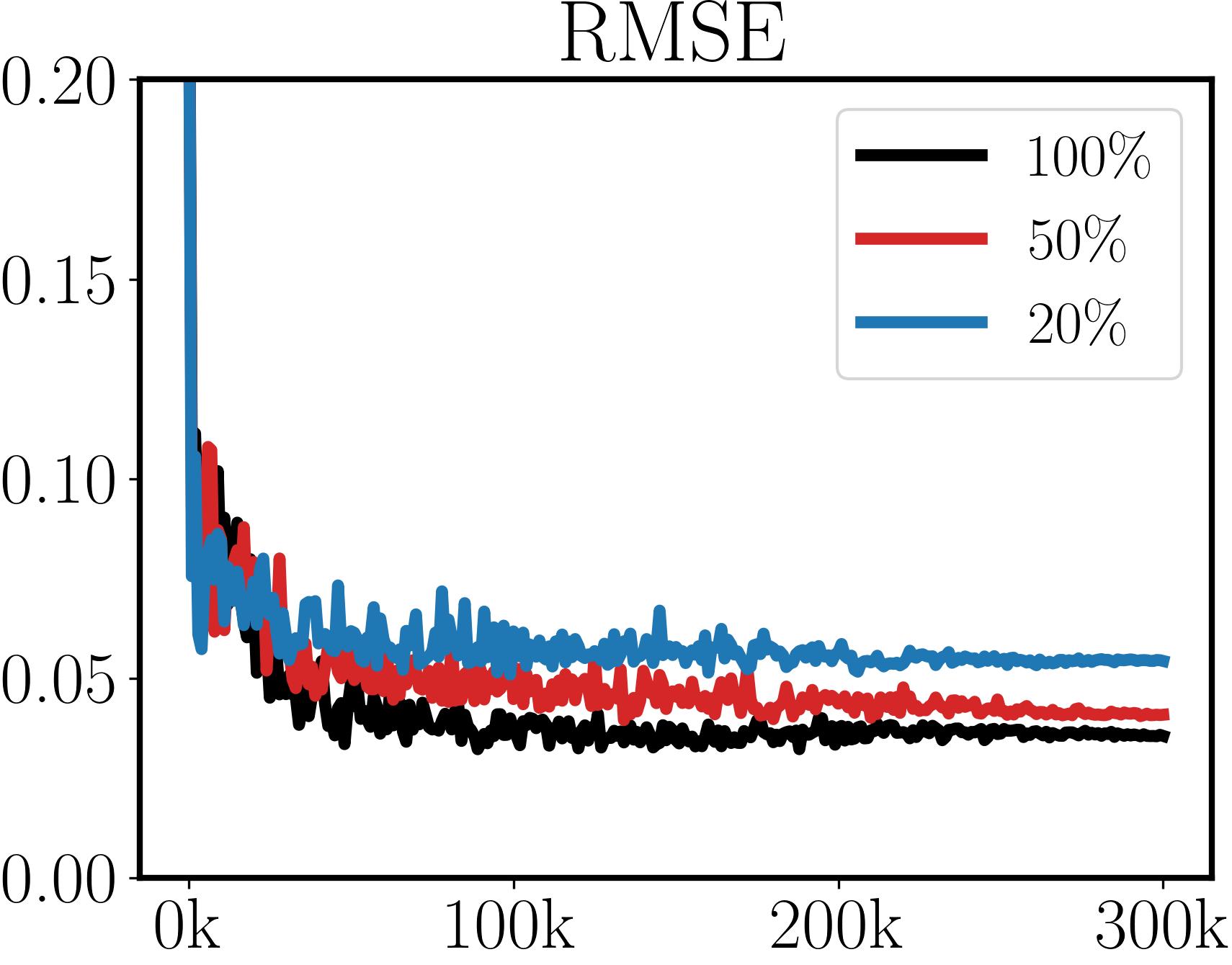}  
         &\includegraphics[width=0.24\textwidth]{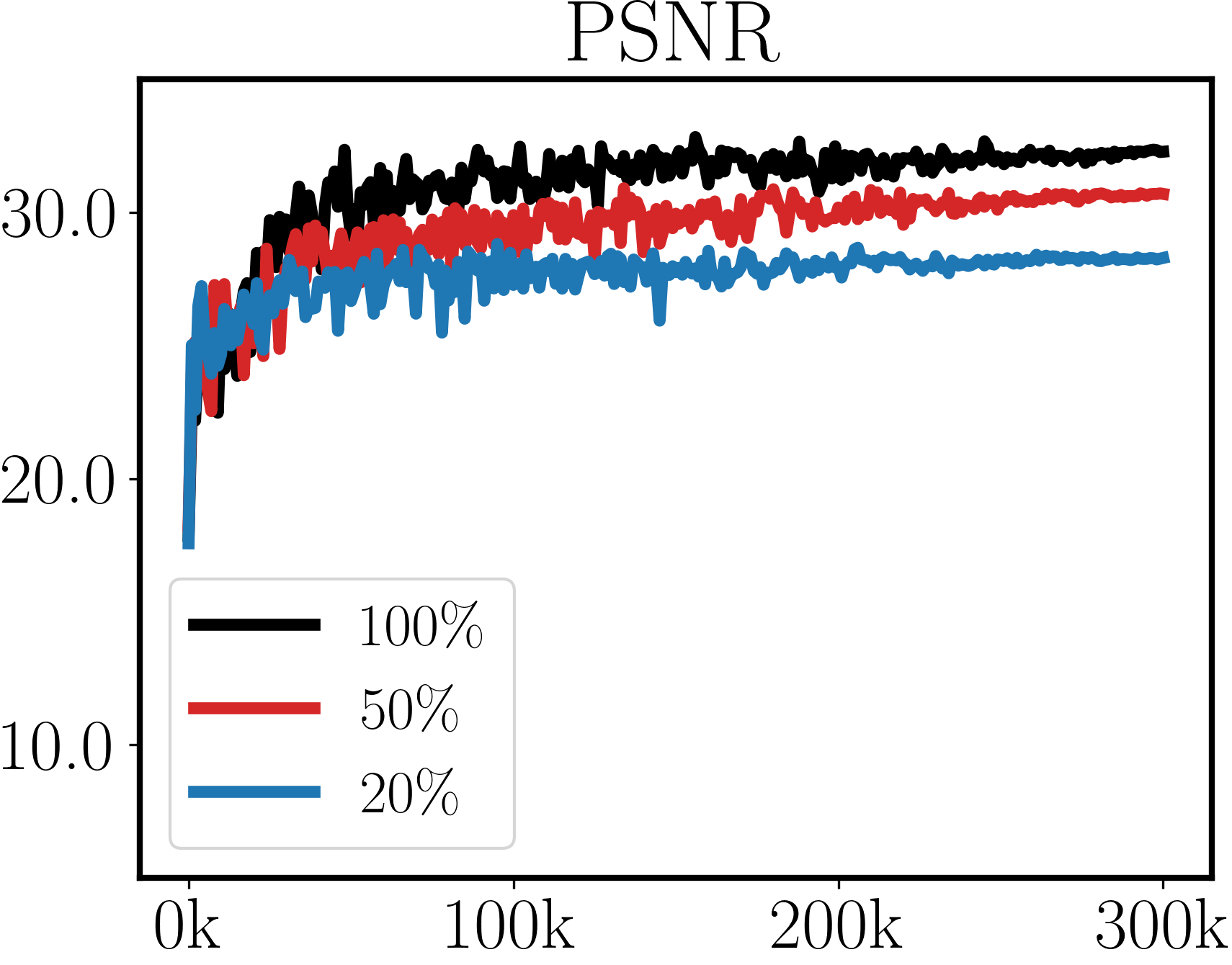}  
         &\includegraphics[width=0.24\textwidth]{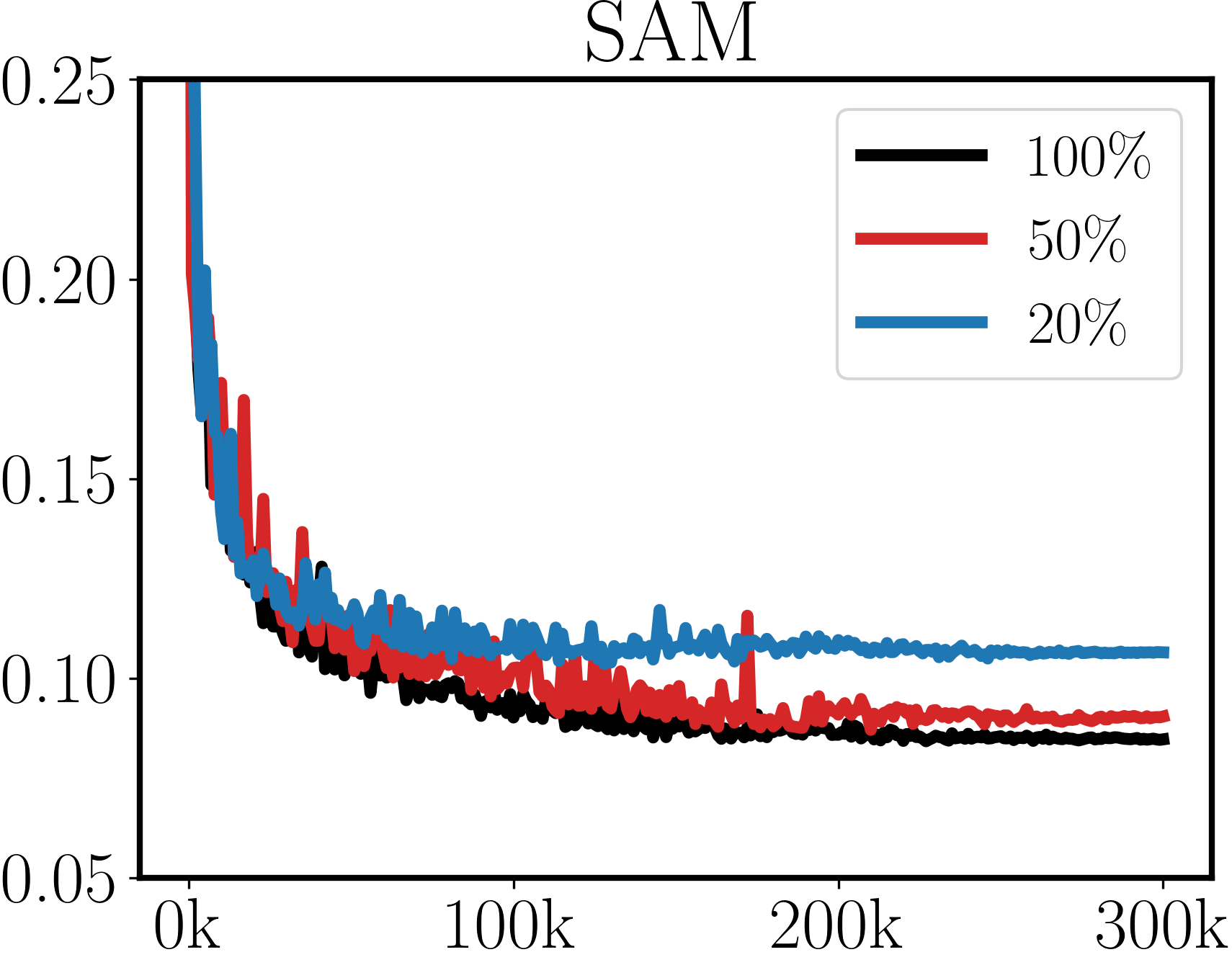}\\
    \end{tabular}
    \caption{Performance evaluation on MRAE, RMSE, PSNR, and SAM for MST++~\cite{cai2022mst++}, MST-L~\cite{cai2022mask}, MPRNet\cite{zamir2021multi}, Restormer\cite{zamir2022restormer}, MIRNet\cite{zamir2020learning}, and HINet\cite{chen2021hinet} with 100\%, 50\%, and 20\% of the original training data on the ARAD1K dataset.}
    \label{fig:supp_train_percent1}
\end{figure*}
% Supplementary results for training with less data, Part II
\begin{figure*}
    \centering
    \begin{tabular}{c@{\hspace{0.6em}}c@{}c@{}c@{}c}
         \raisebox{15mm}{\rotatebox[origin=c]{90}{HDNet}}
         &\includegraphics[width=0.24\textwidth]{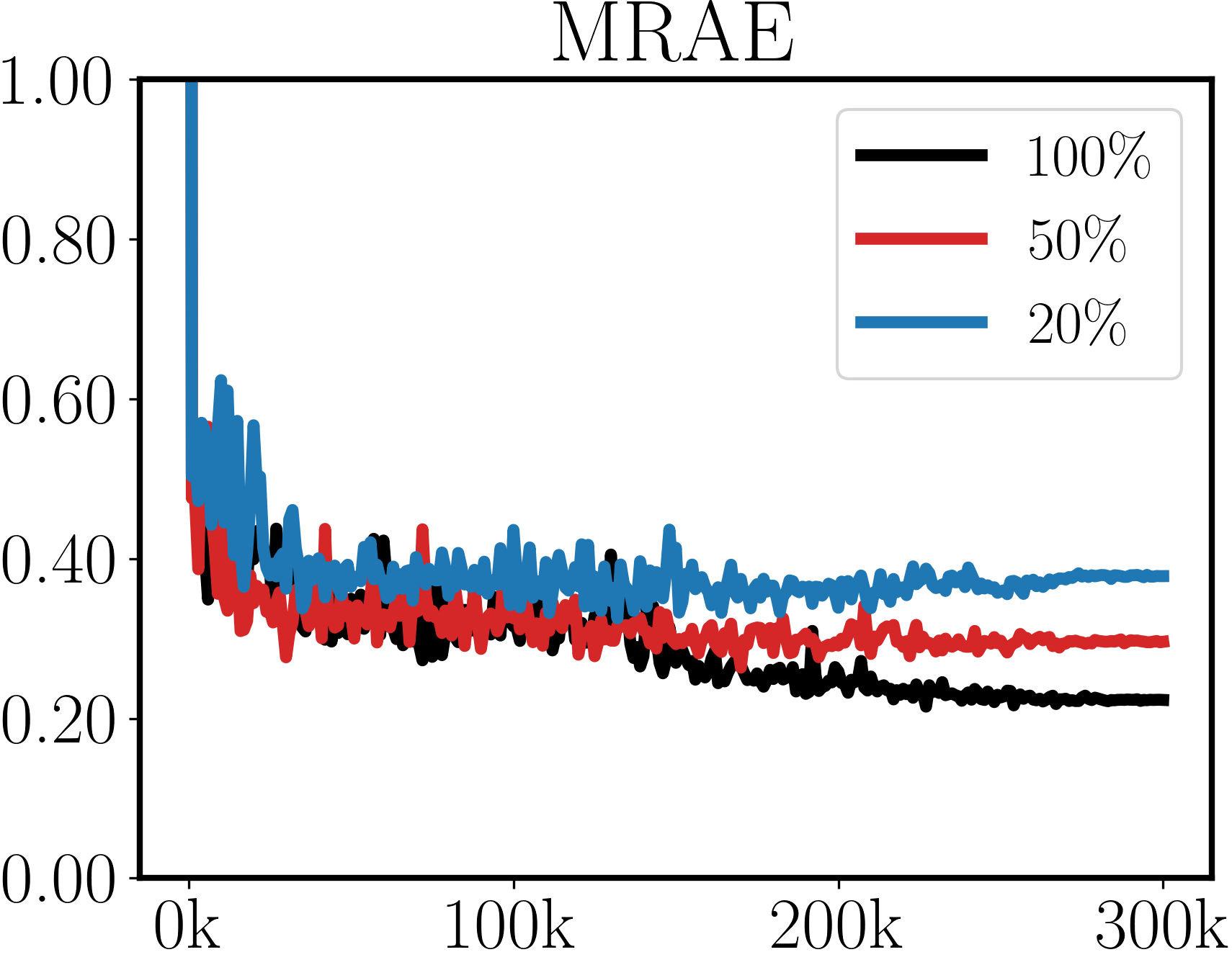}  
         &\includegraphics[width=0.24\textwidth]{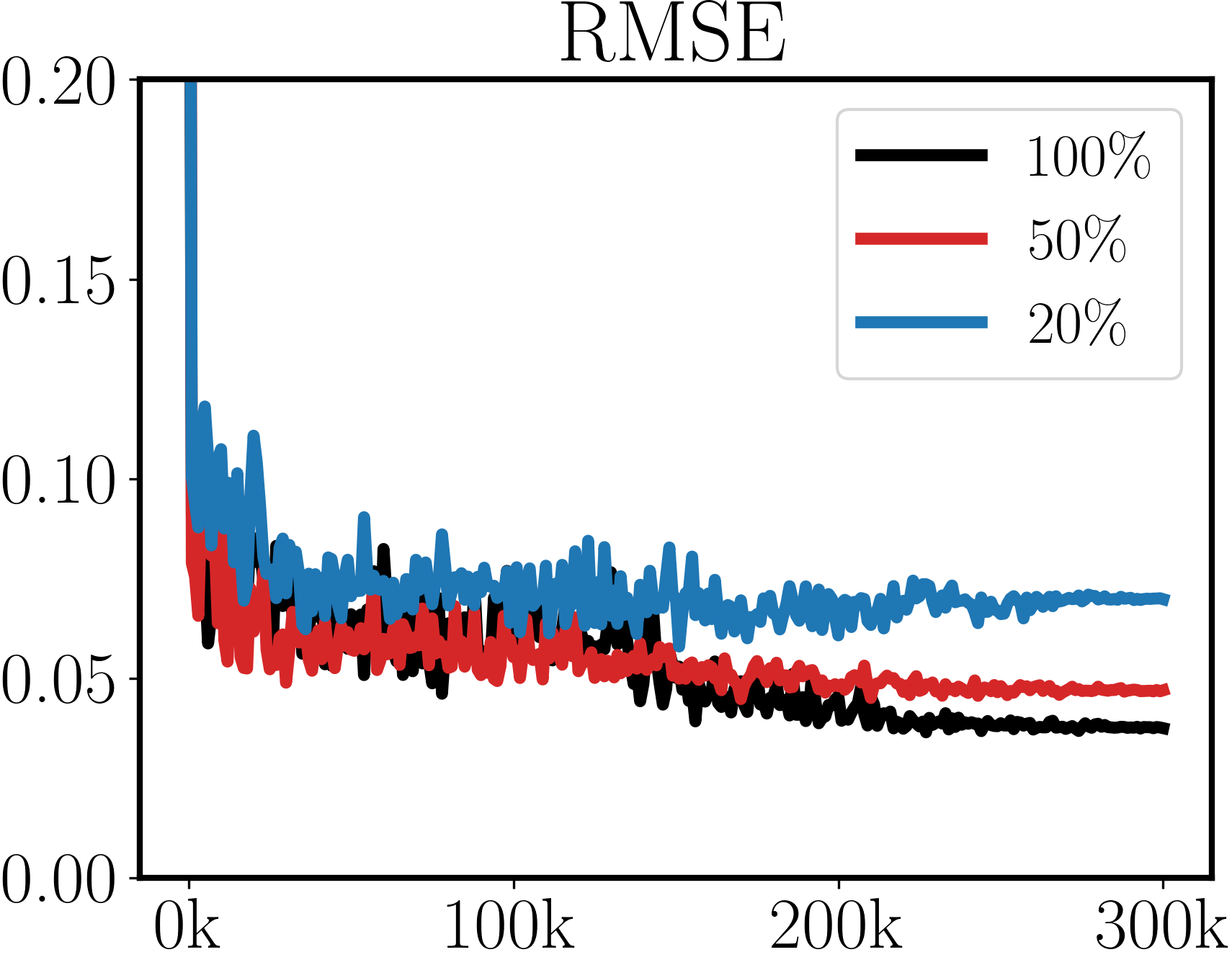}  
         &\includegraphics[width=0.24\textwidth]{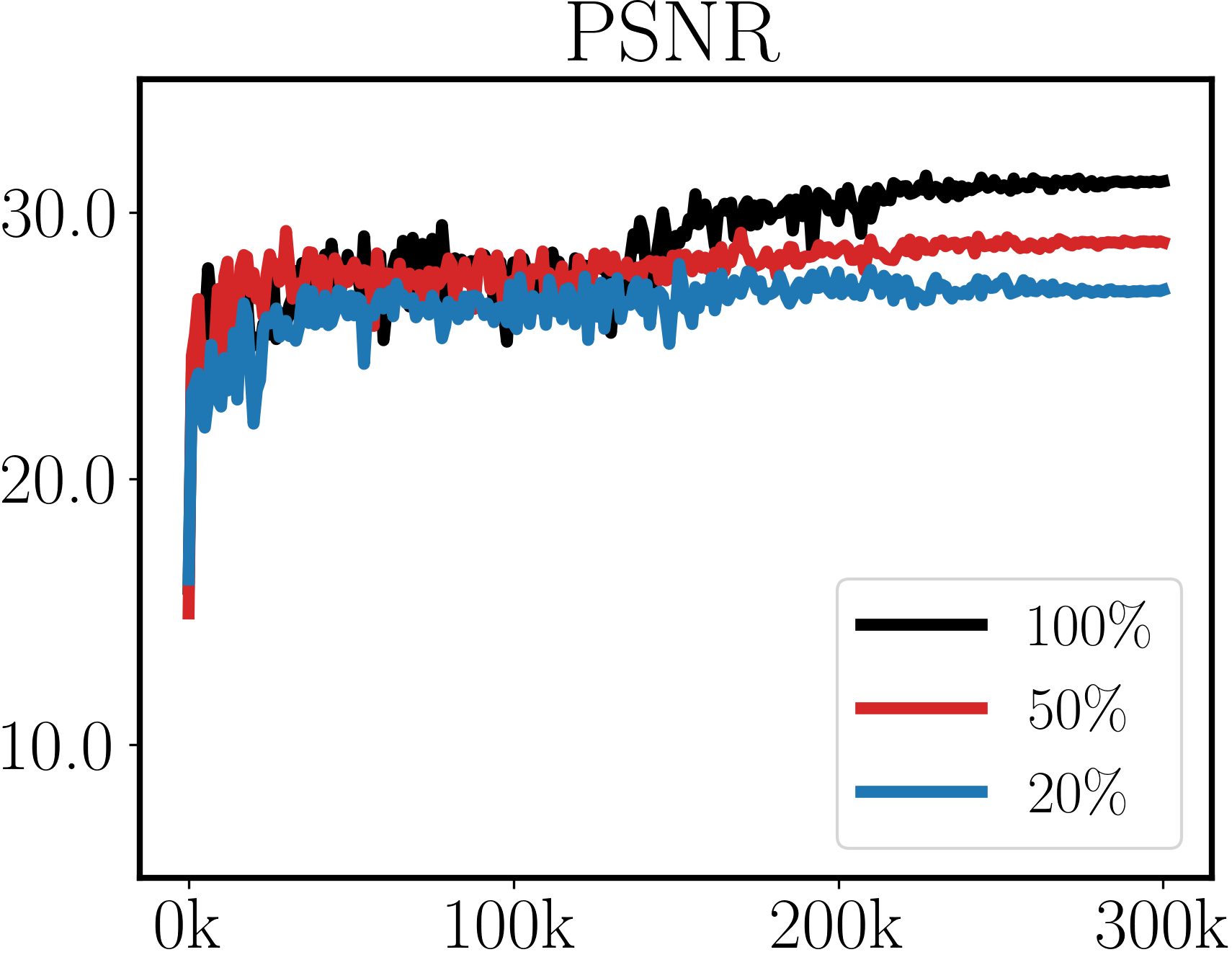}  
         &\includegraphics[width=0.24\textwidth]{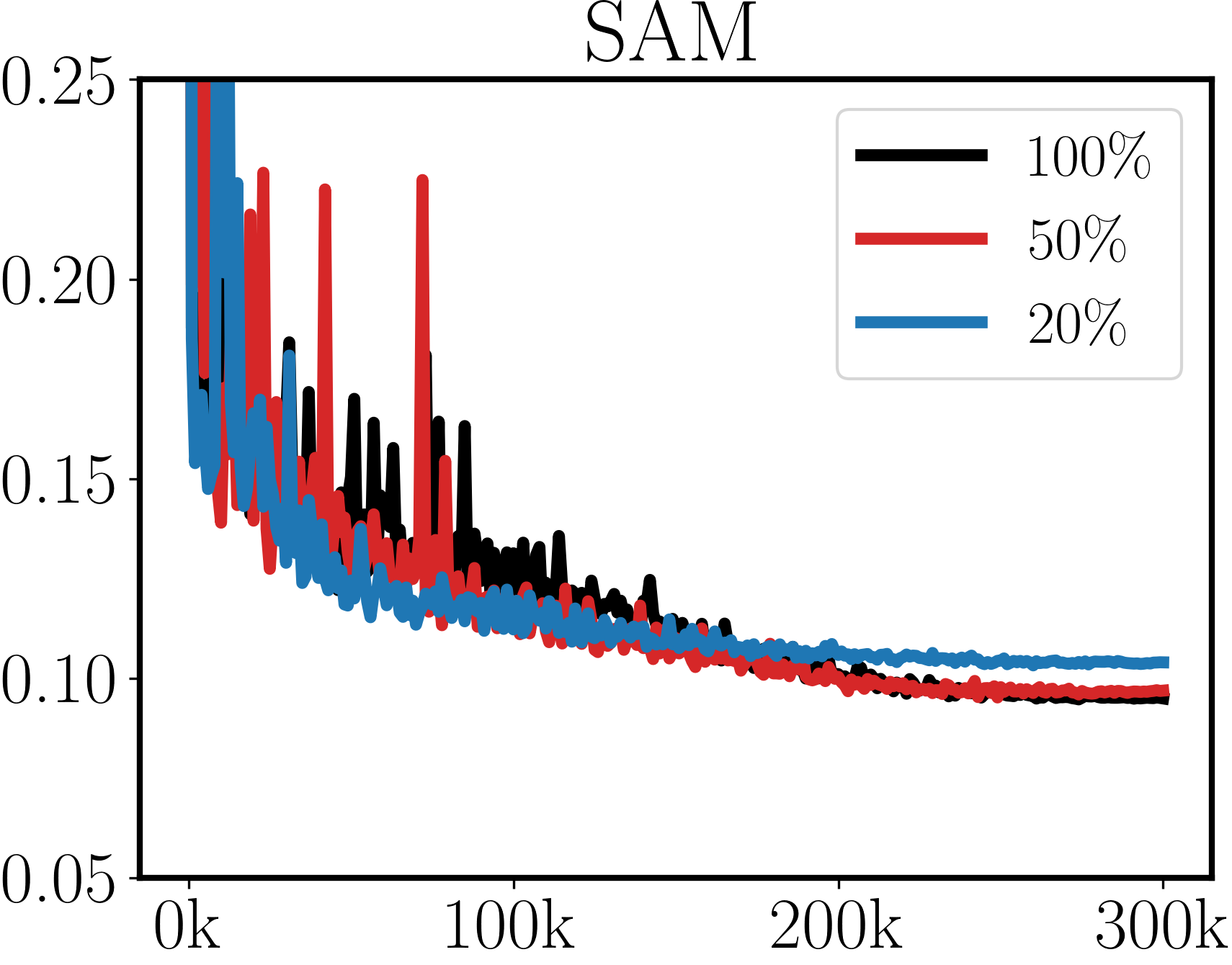}\\
         \raisebox{15mm}{\rotatebox[origin=c]{90}{AWAN}}
         &\includegraphics[width=0.24\textwidth]{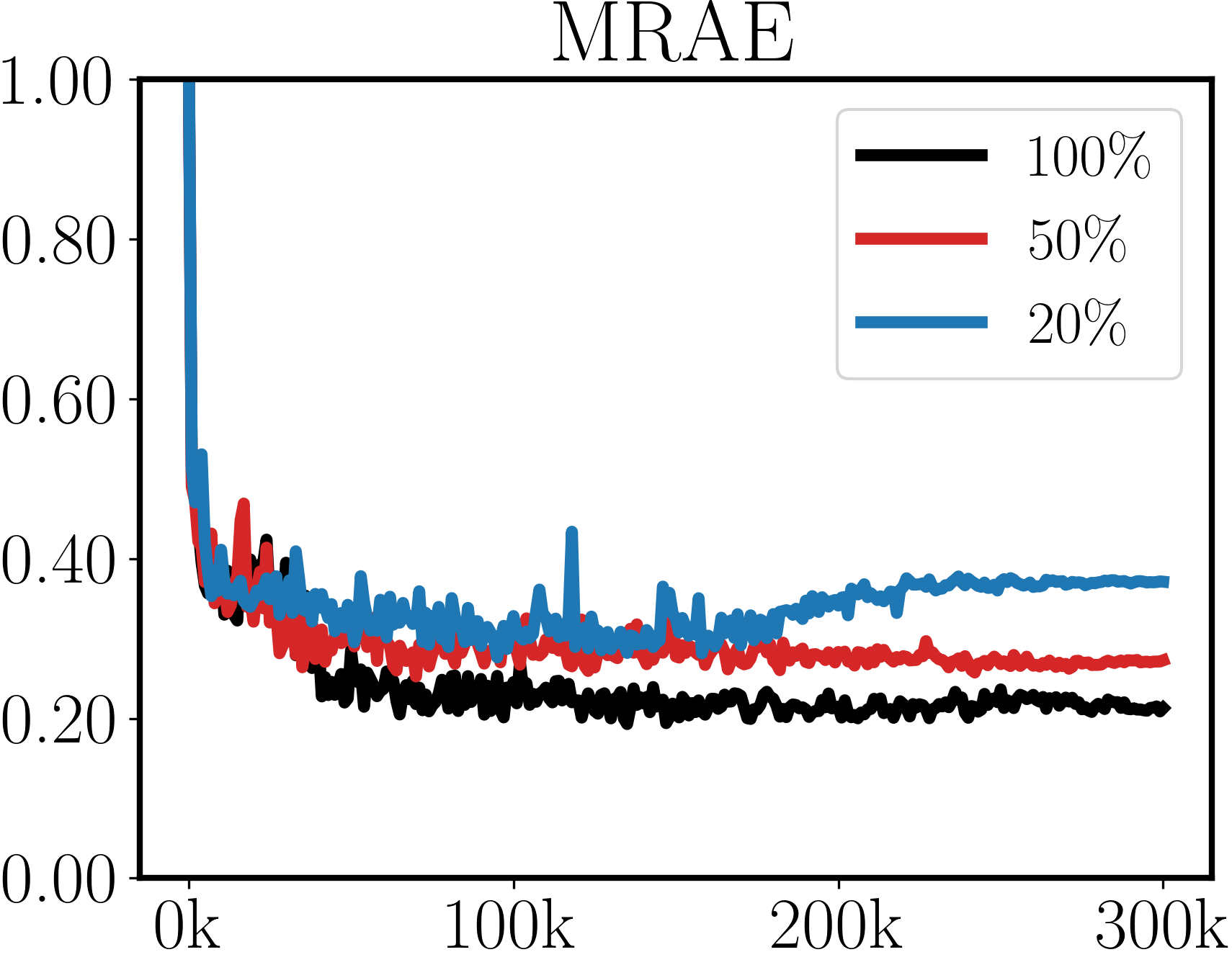}  
         &\includegraphics[width=0.24\textwidth]{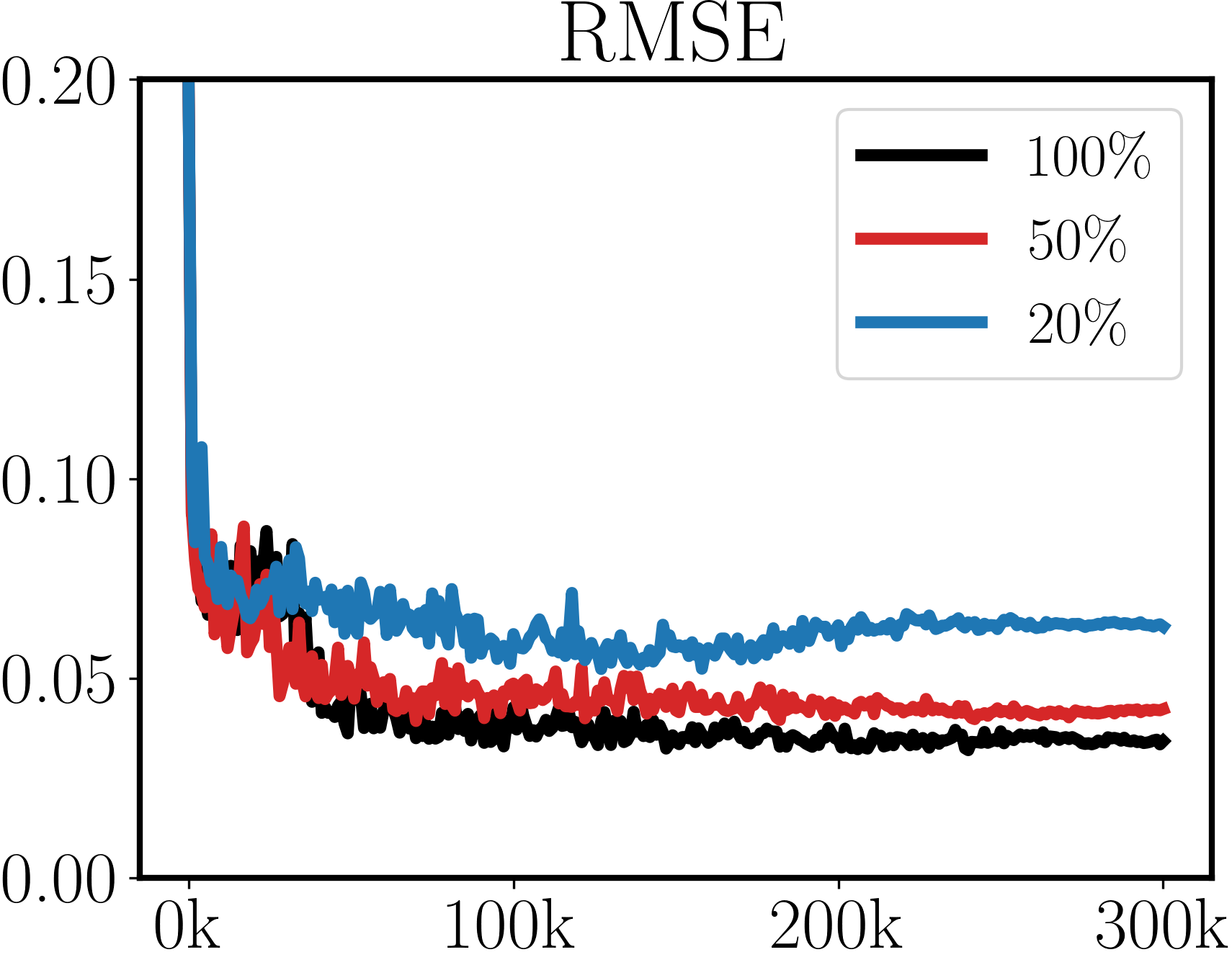}  
         &\includegraphics[width=0.24\textwidth]{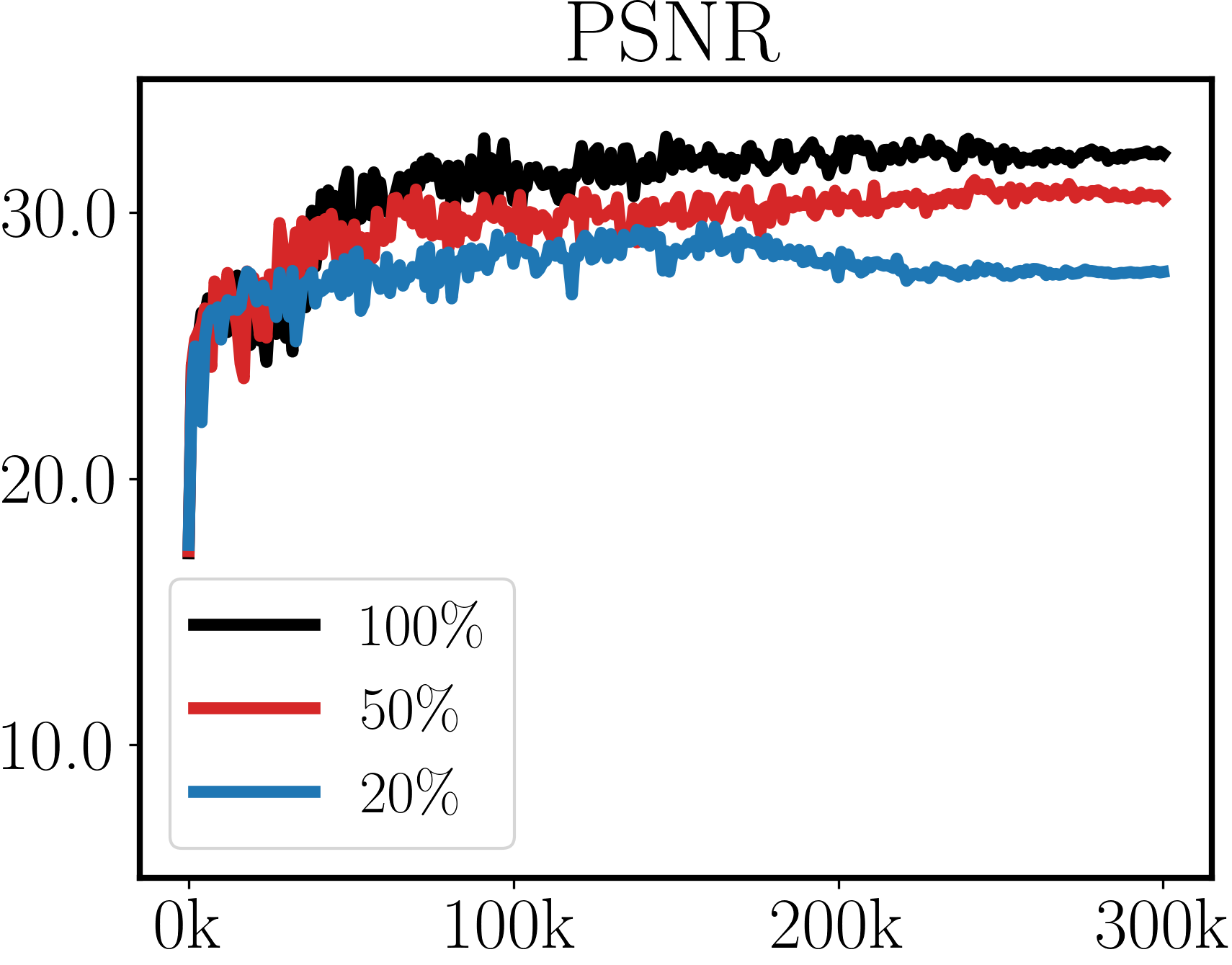}  
         &\includegraphics[width=0.24\textwidth]{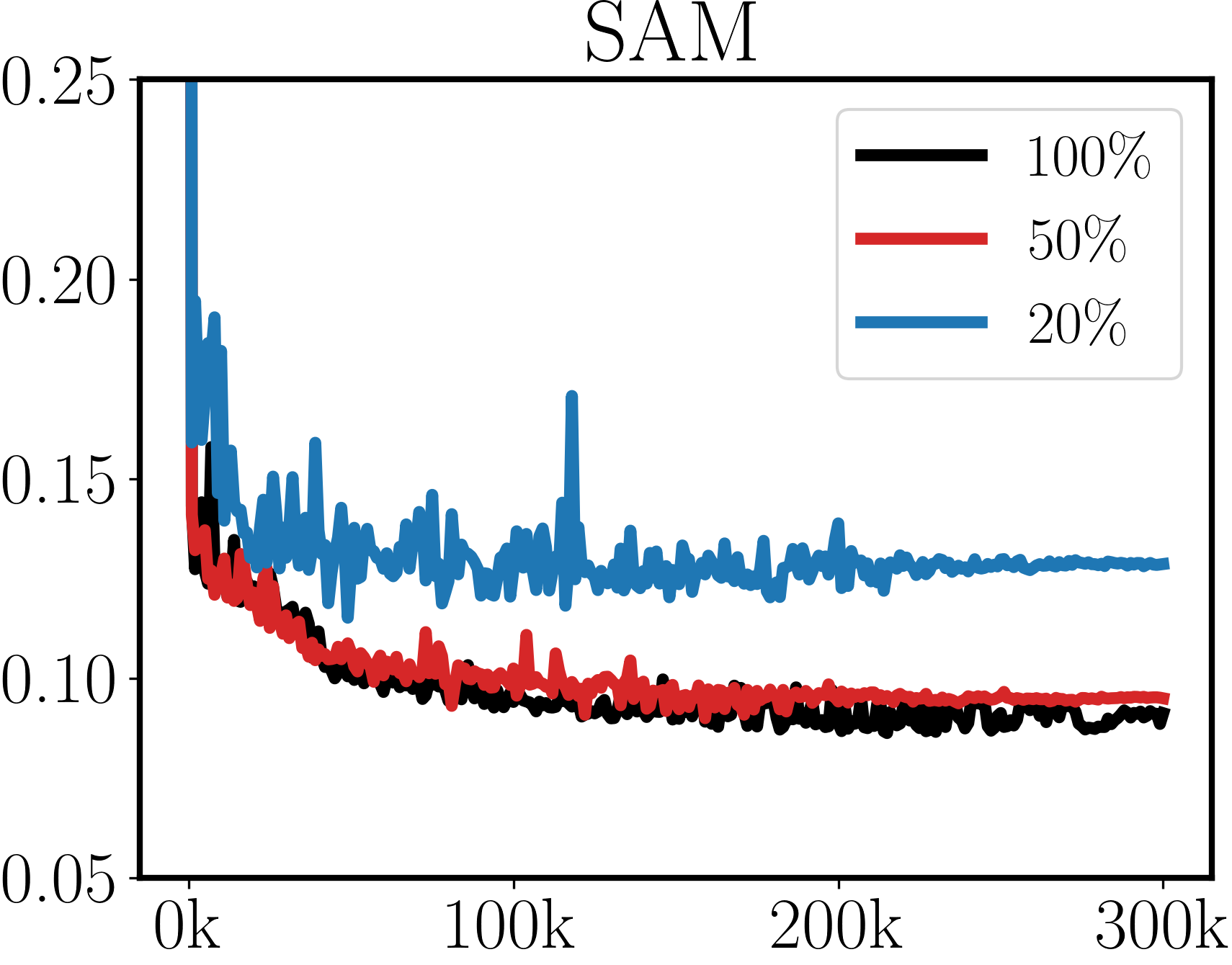}\\
         \raisebox{15mm}{\rotatebox[origin=c]{90}{EDSR}}
         &\includegraphics[width=0.24\textwidth]{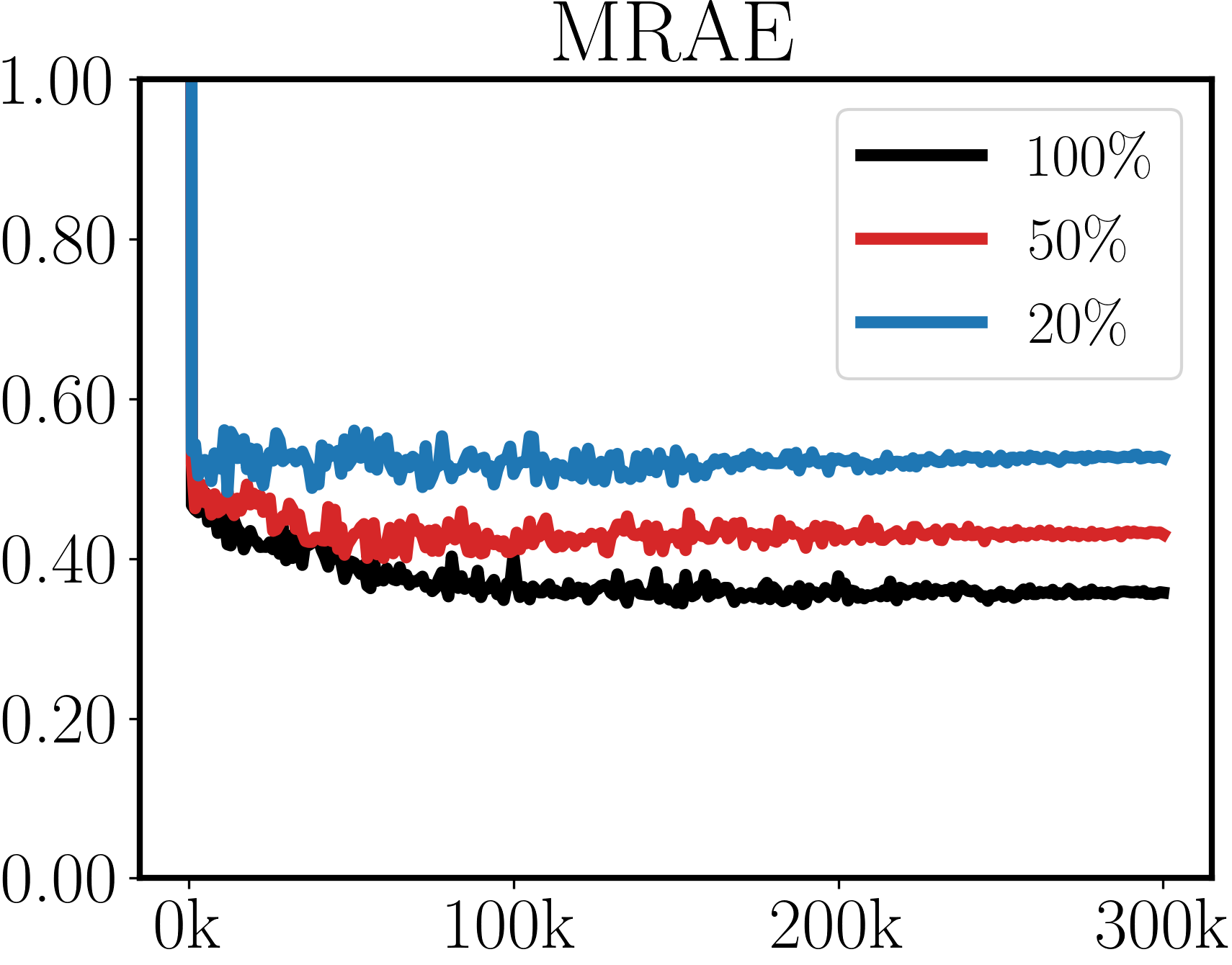}  
         &\includegraphics[width=0.24\textwidth]{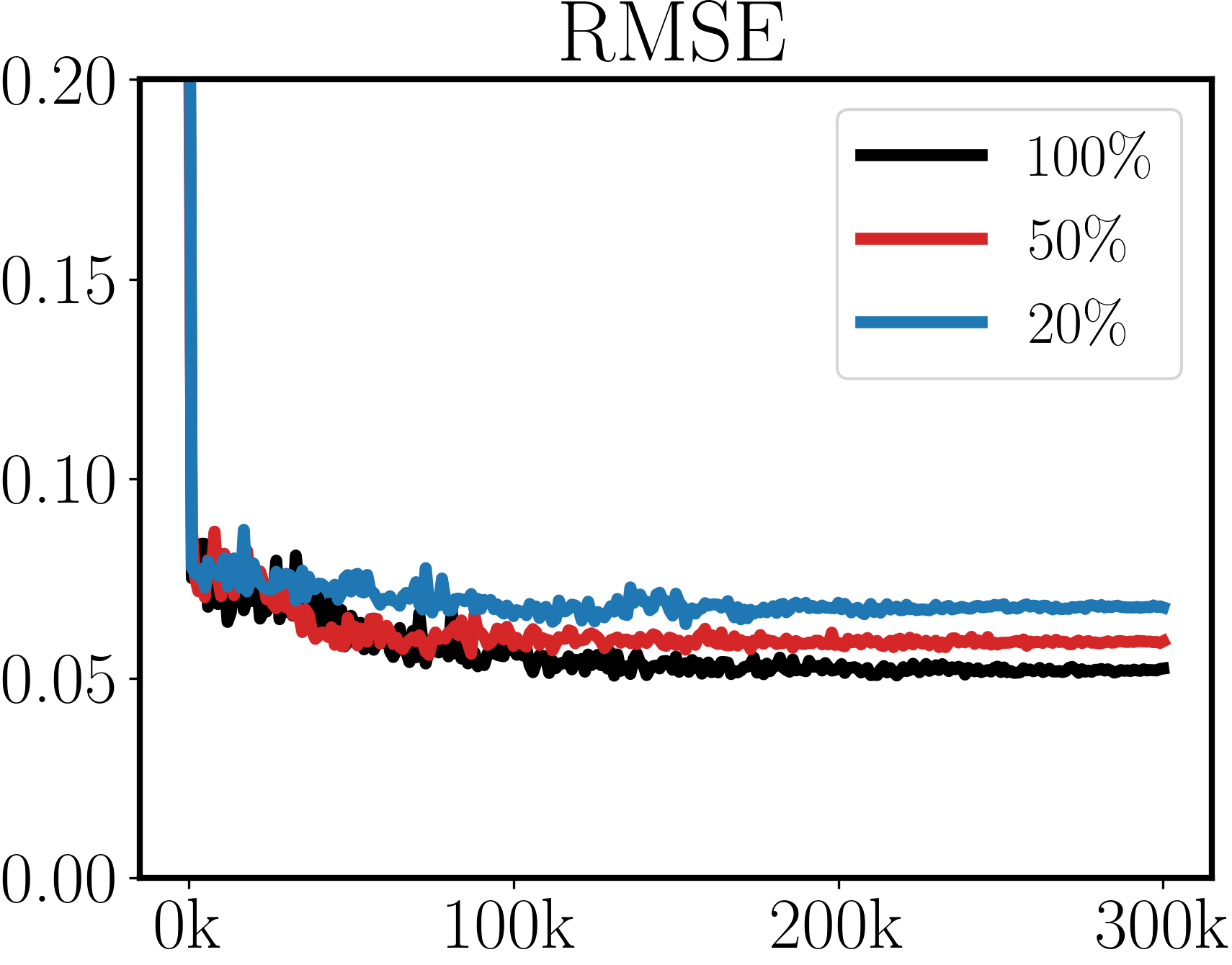}  
         &\includegraphics[width=0.24\textwidth]{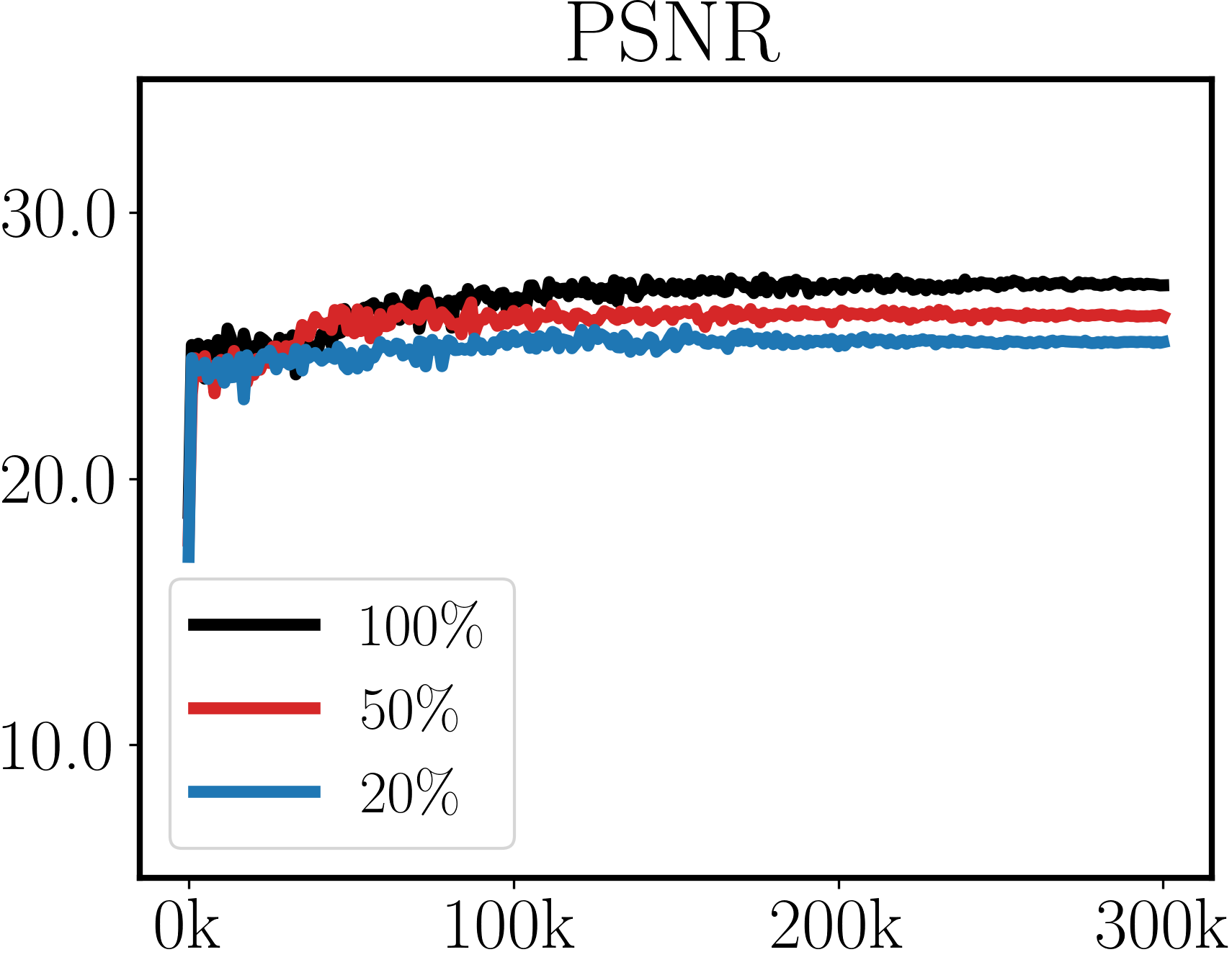}  
         &\includegraphics[width=0.24\textwidth]{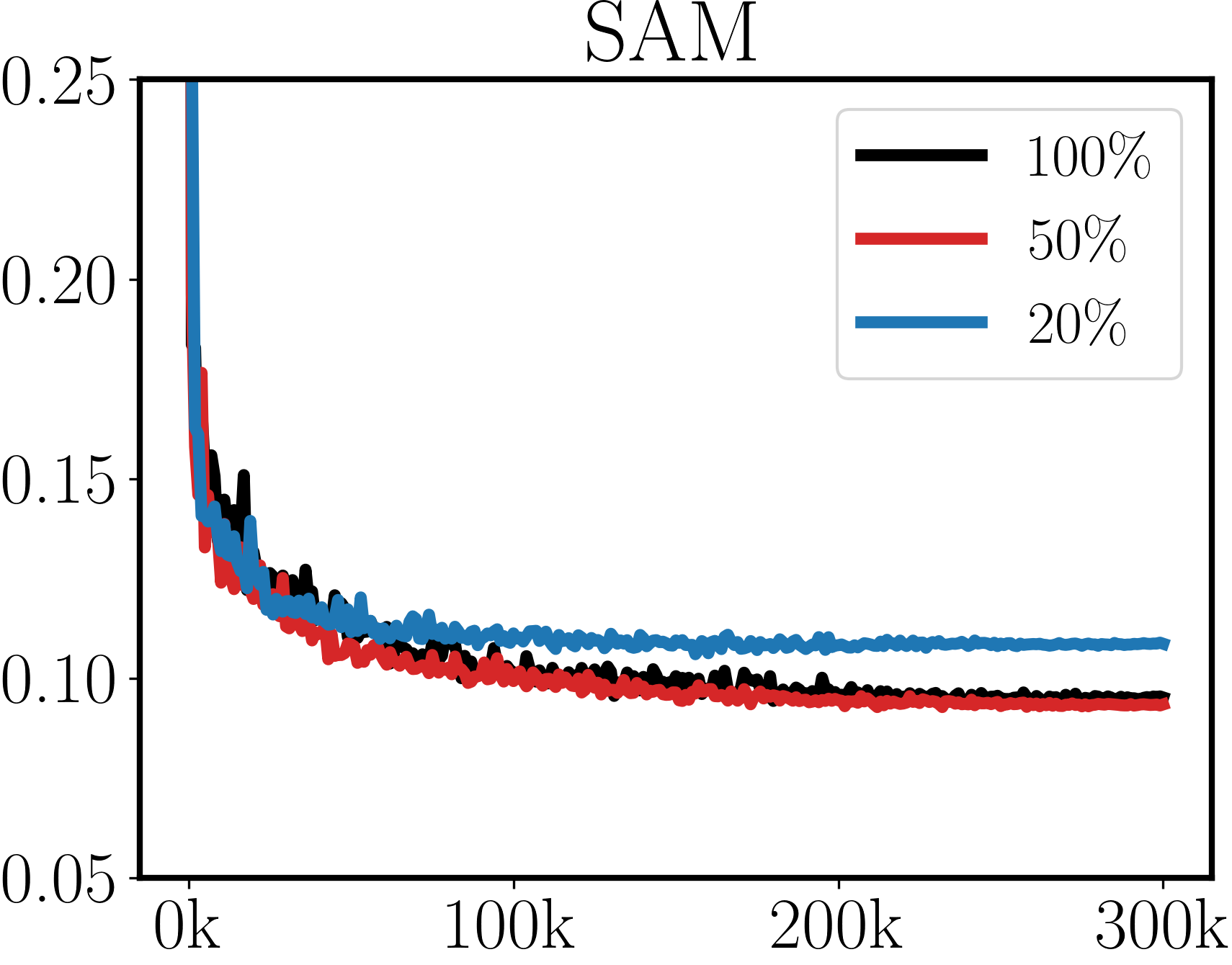}\\
         \raisebox{15mm}{\rotatebox[origin=c]{90}{HRNet}}
         &\includegraphics[width=0.24\textwidth]{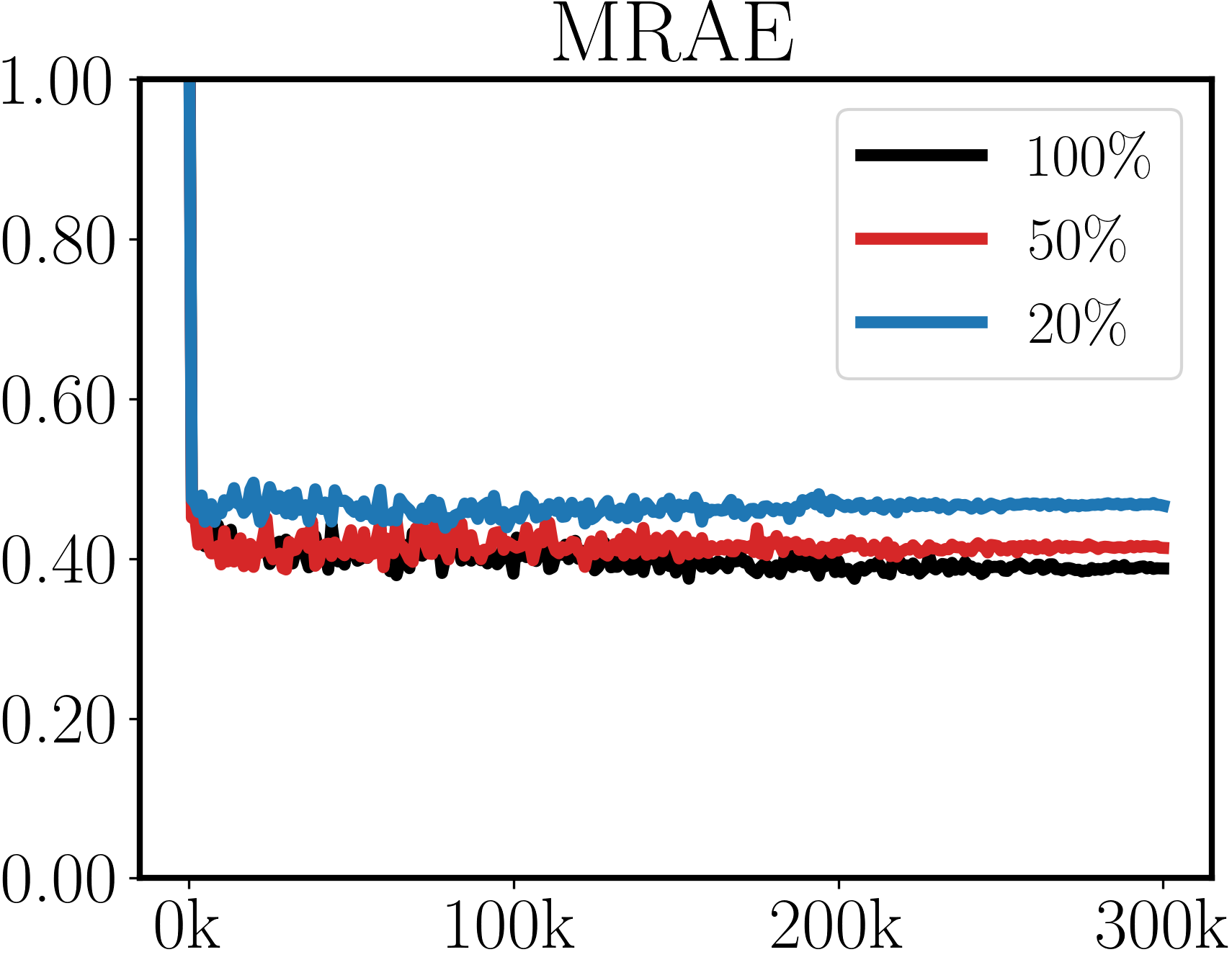}  
         &\includegraphics[width=0.24\textwidth]{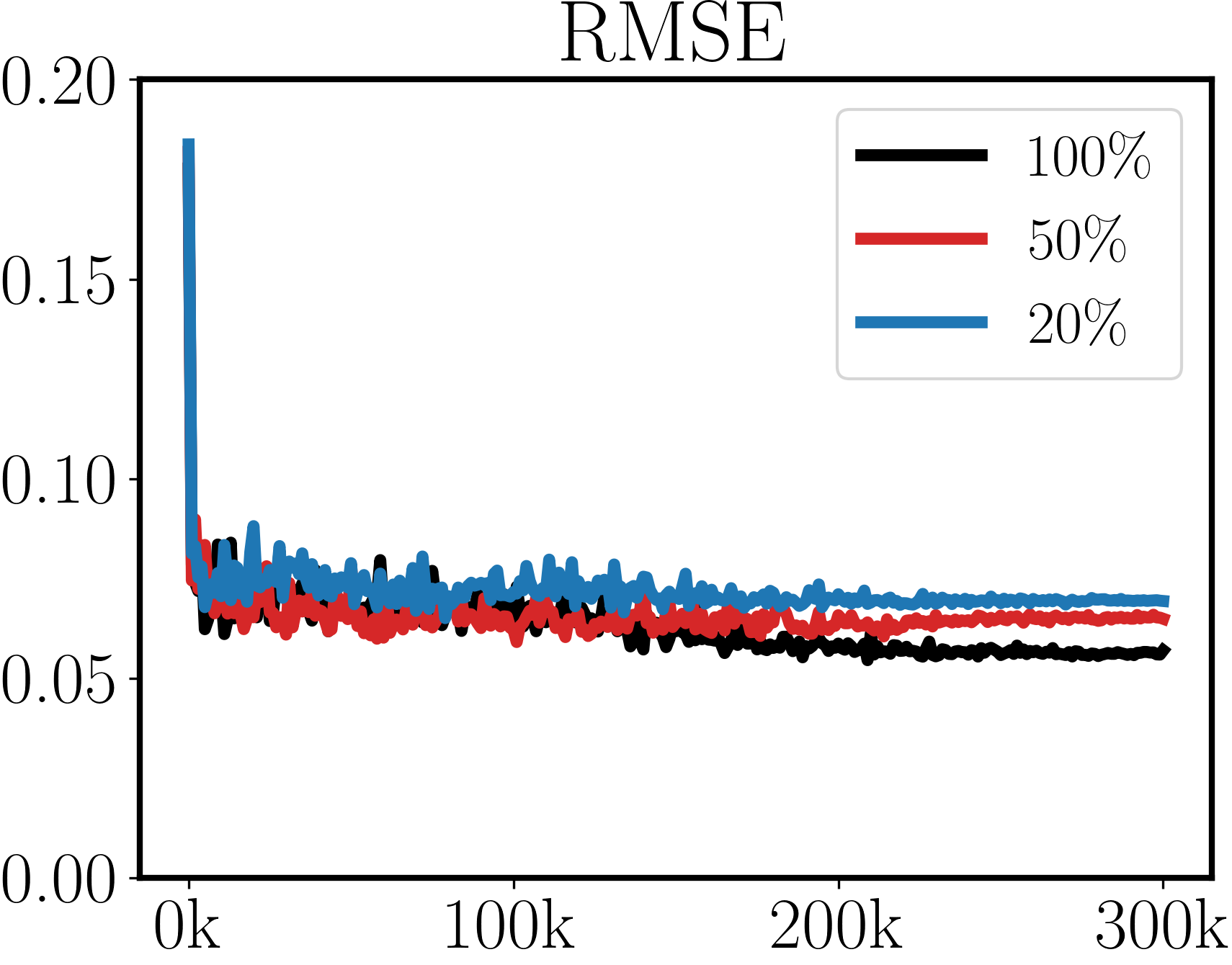}  
         &\includegraphics[width=0.24\textwidth]{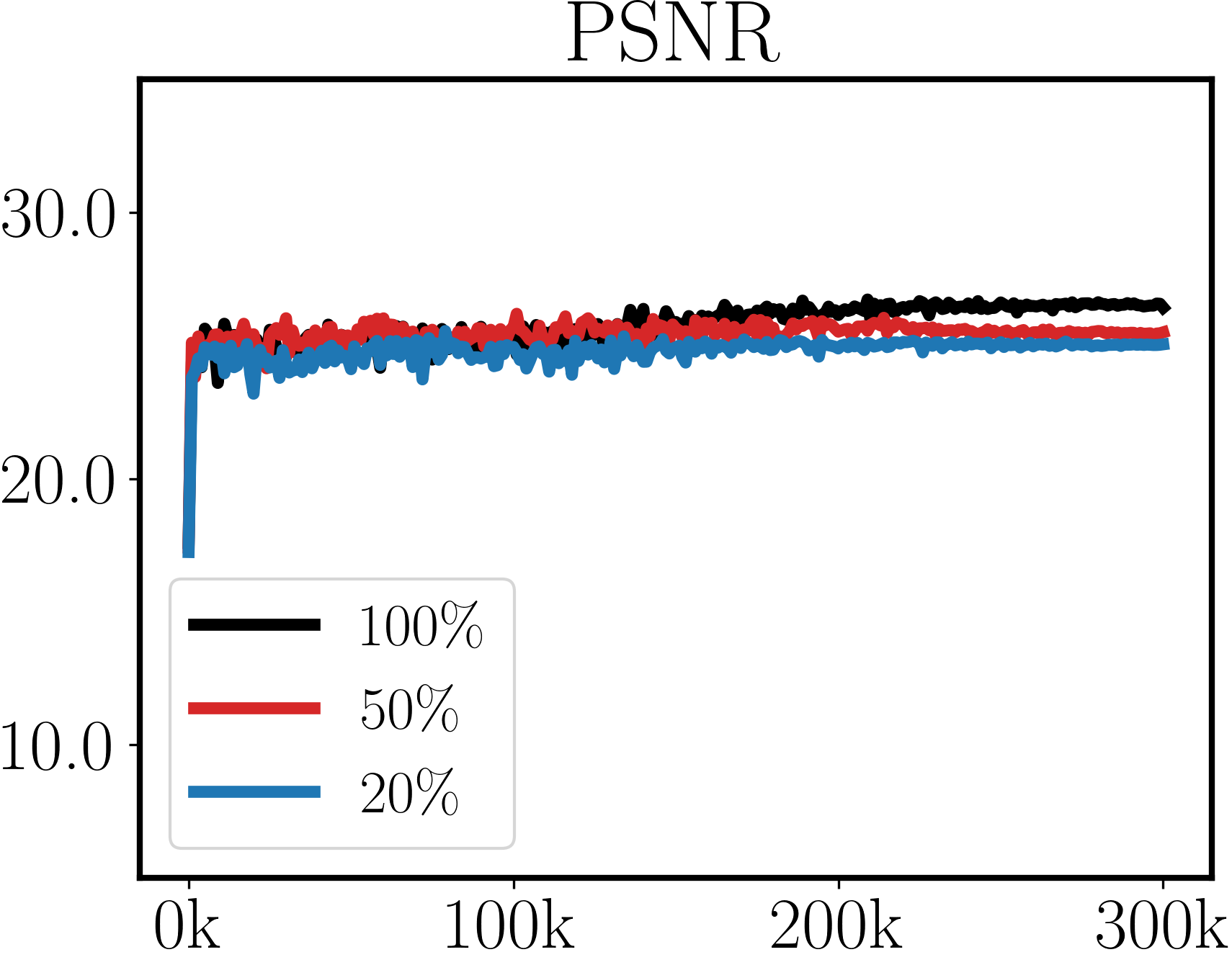}  
         &\includegraphics[width=0.24\textwidth]{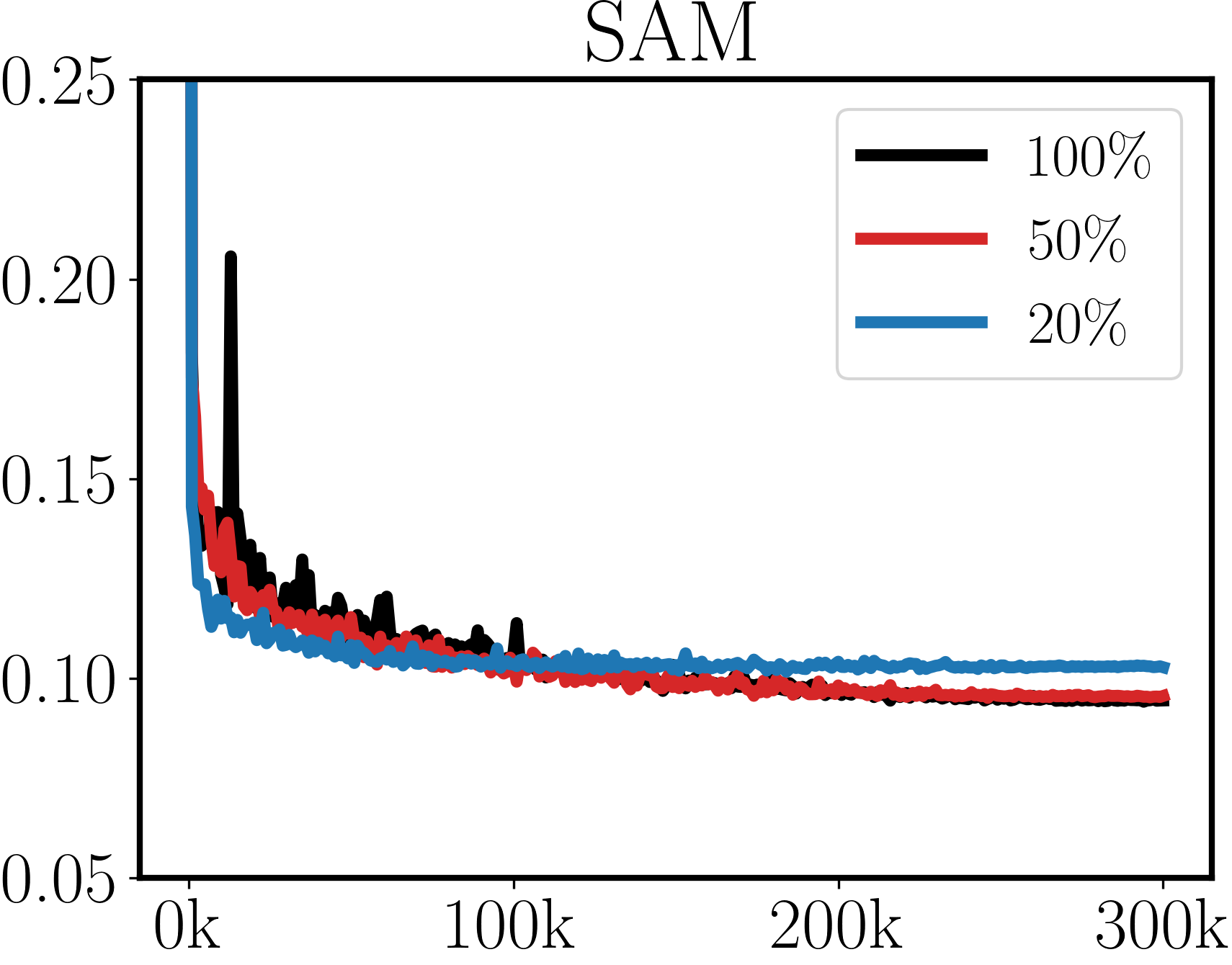}\\
         \raisebox{15mm}{\rotatebox[origin=c]{90}{HSCNN+}}
         &\includegraphics[width=0.24\textwidth]{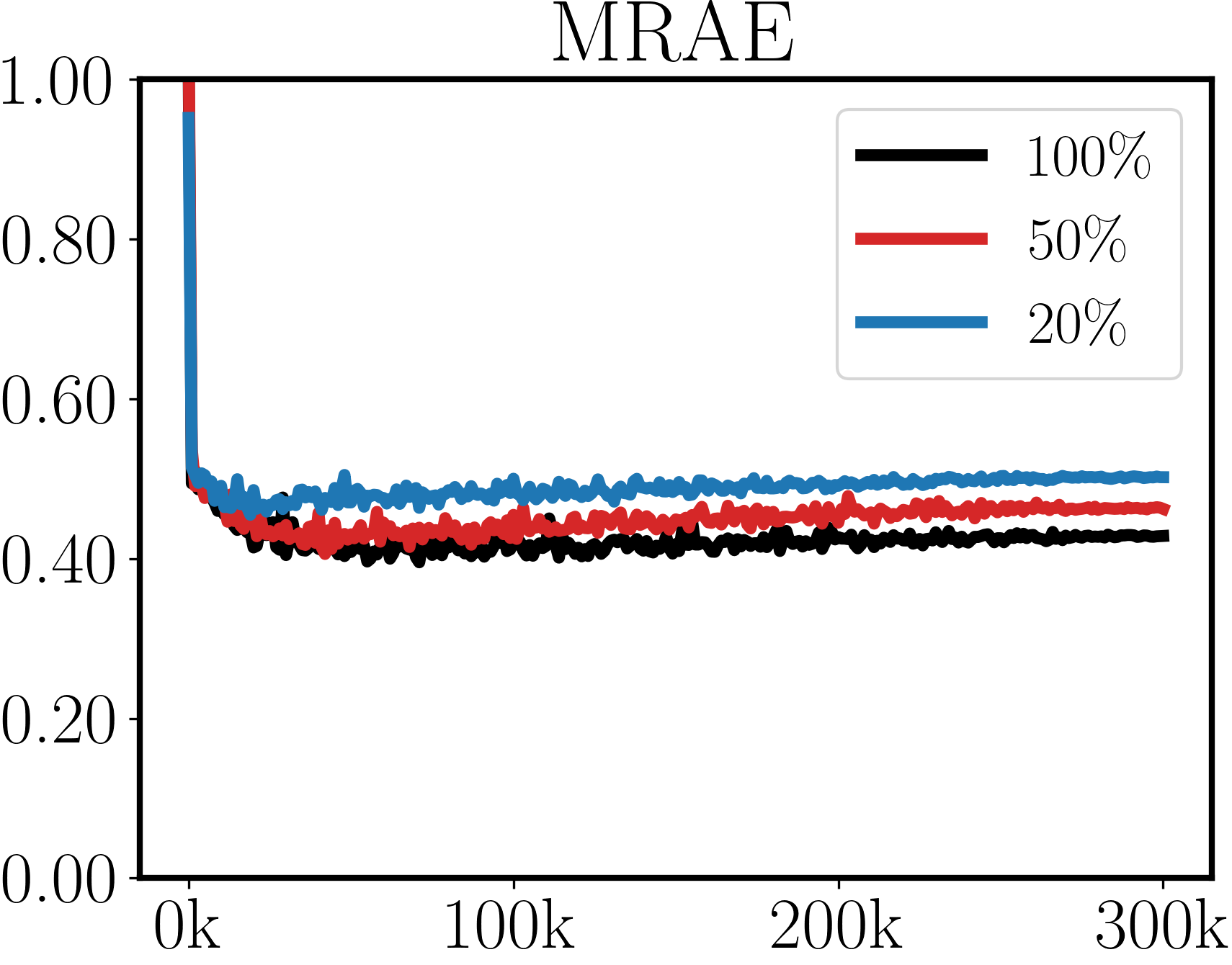}  
         &\includegraphics[width=0.24\textwidth]{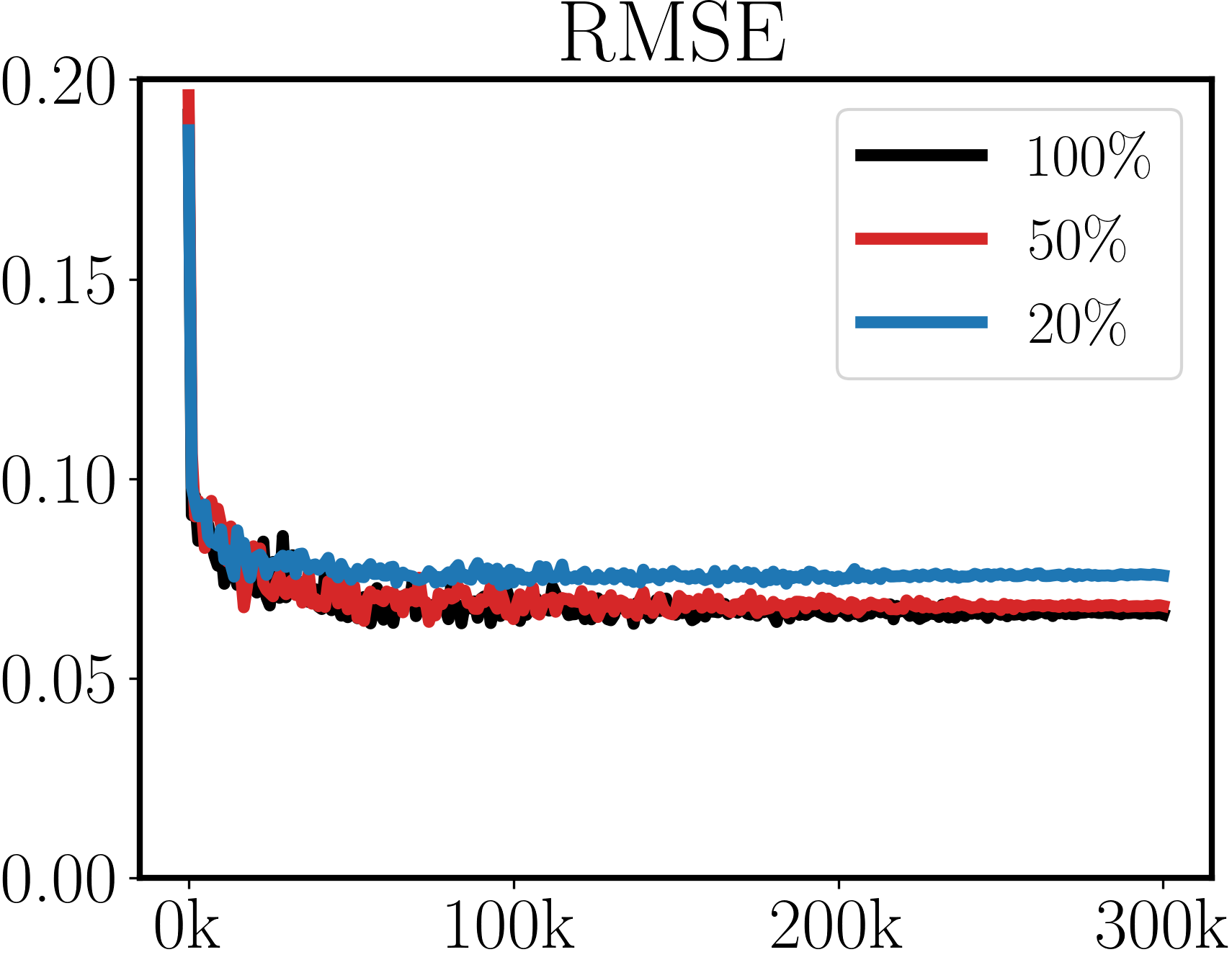}  
         &\includegraphics[width=0.24\textwidth]{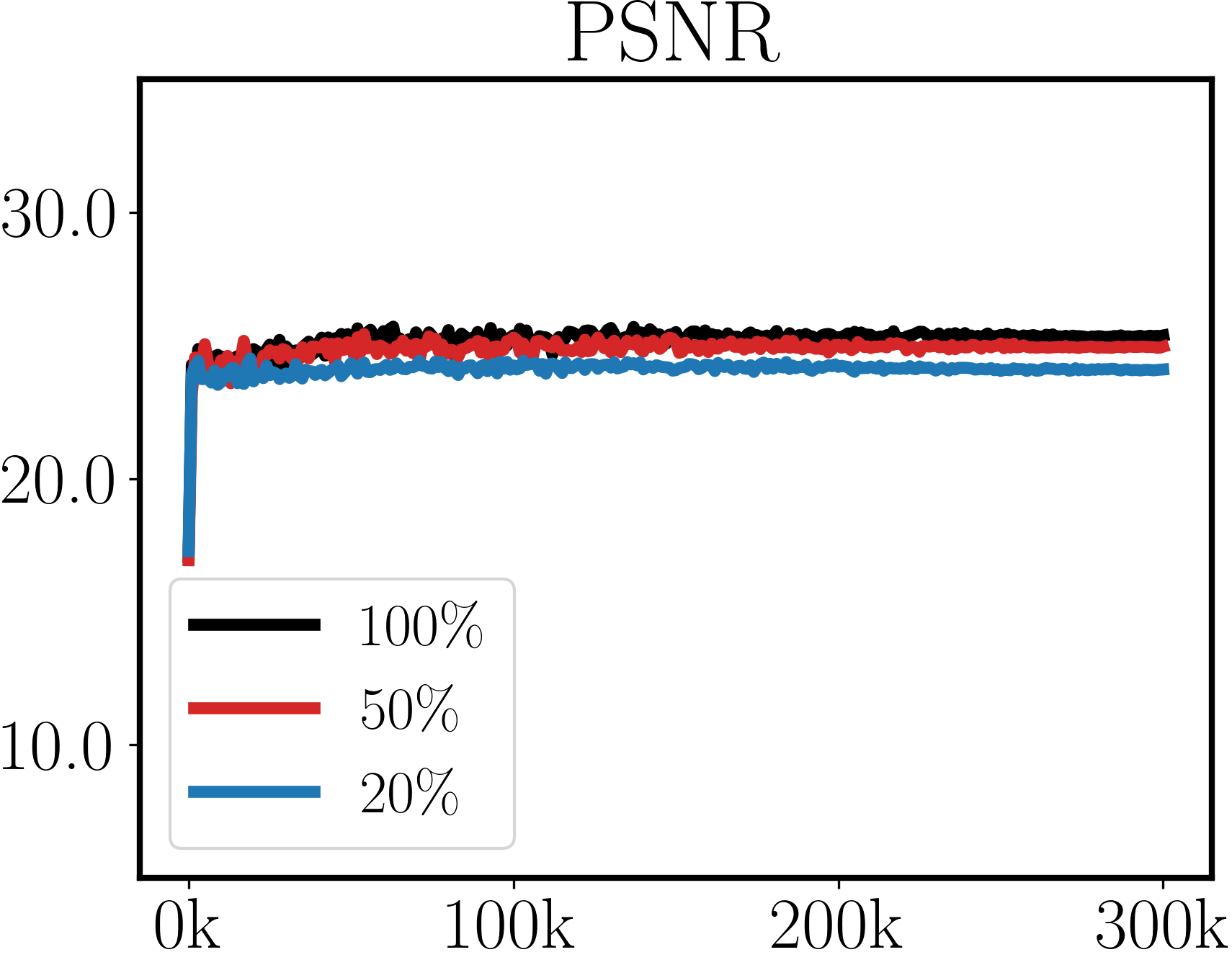}  
         &\includegraphics[width=0.24\textwidth]{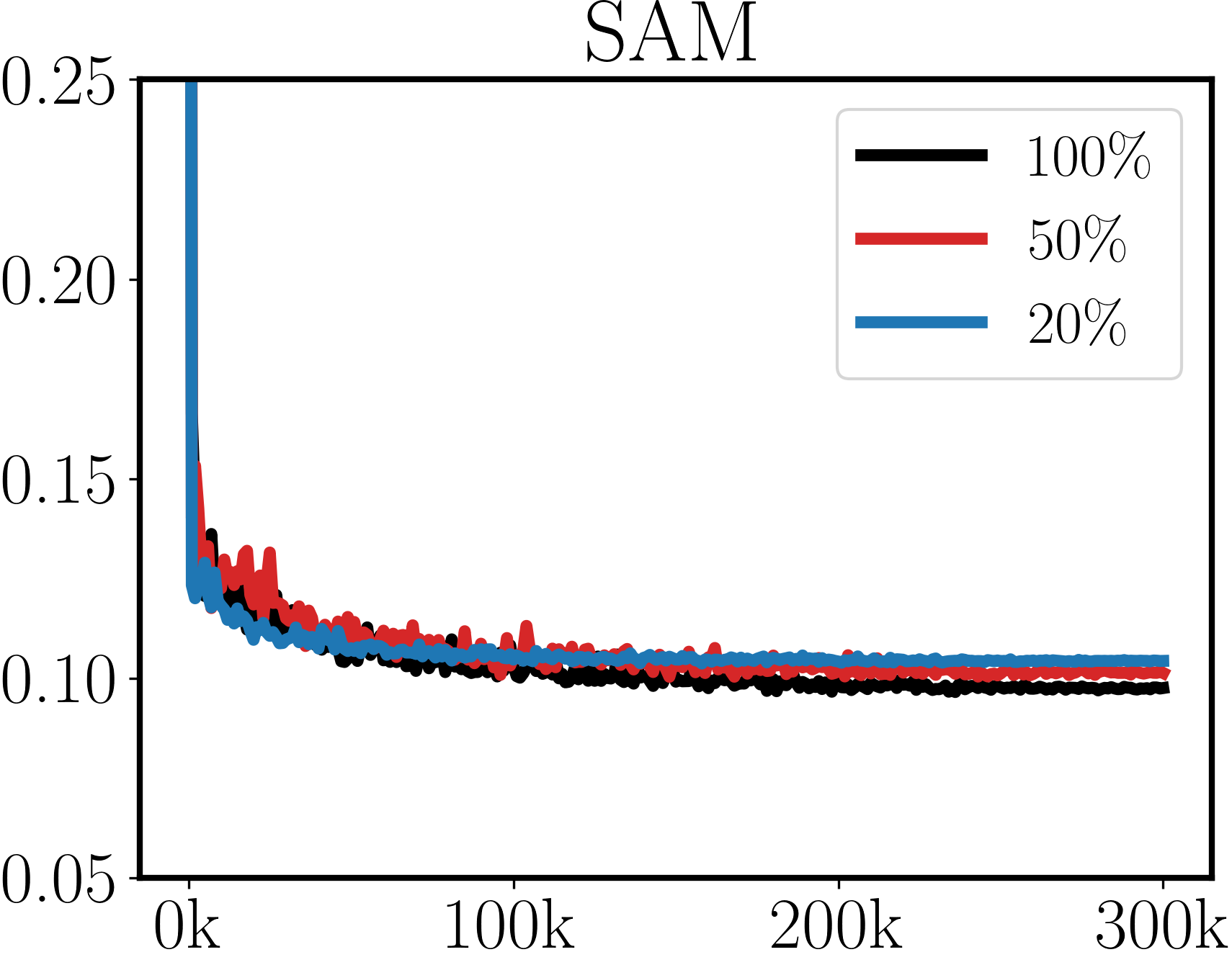}\\
    \end{tabular}
    \caption{Performance evaluation on MRAE, RMSE, PSNR, and SAM for HDNet\cite{hu2022hdnet}, AWAN\cite{li2020adaptive}, EDSR\cite{lim2017enhanced}, HRNet\cite{zhao2020hierarchical}, and HSCNN+\cite{shi2018hscnn+} with 100\%, 50\%, and 20\% of the original training data on the ARAD1K dataset.}
    \label{fig:supp_train_percent2}
\end{figure*}
% Supplementary results for training with less data, Part III
\begin{figure*}
    \centering
    \newcommand{\placeholder}{\fbox{\rule{0pt}{30mm} \rule{40mm}{0pt}}}
    \begin{tabular}{c@{\hspace{0.6em}}c@{}c@{}c@{}c}
         \raisebox{15mm}{\rotatebox[origin=c]{90}{HySAT}}
         &\includegraphics[width=0.24\textwidth]{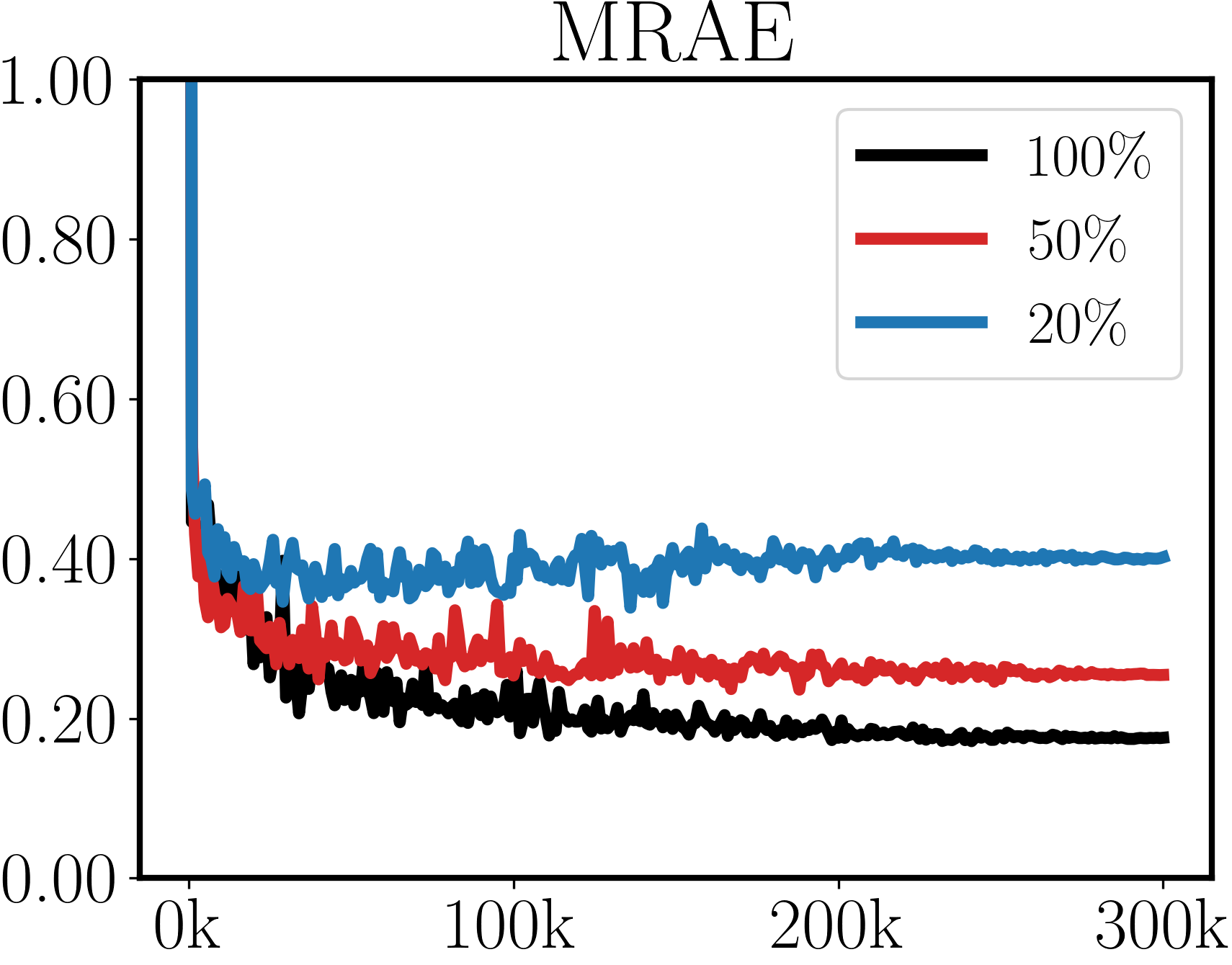}  
         &\includegraphics[width=0.24\textwidth]{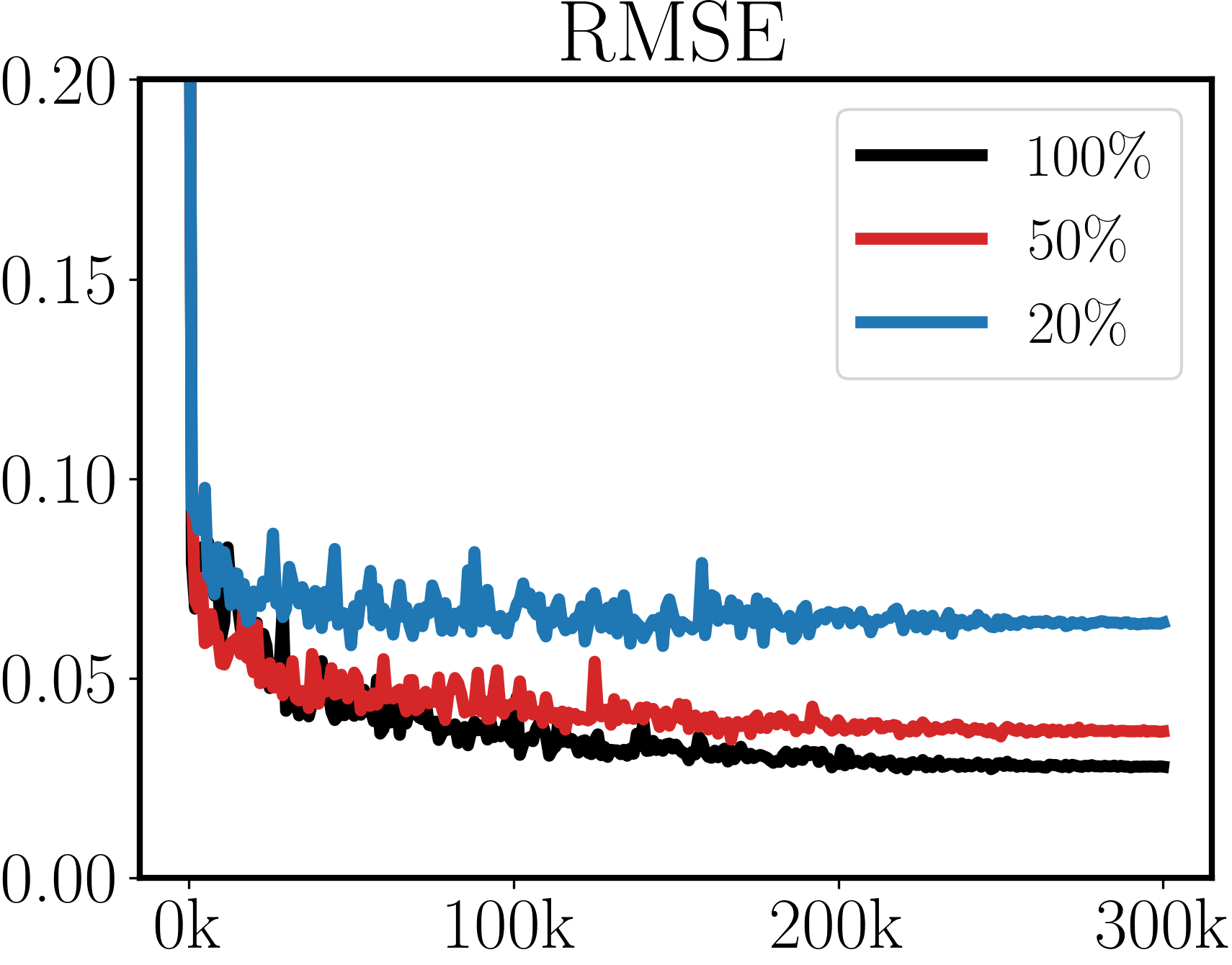} 
         &\includegraphics[width=0.24\textwidth]{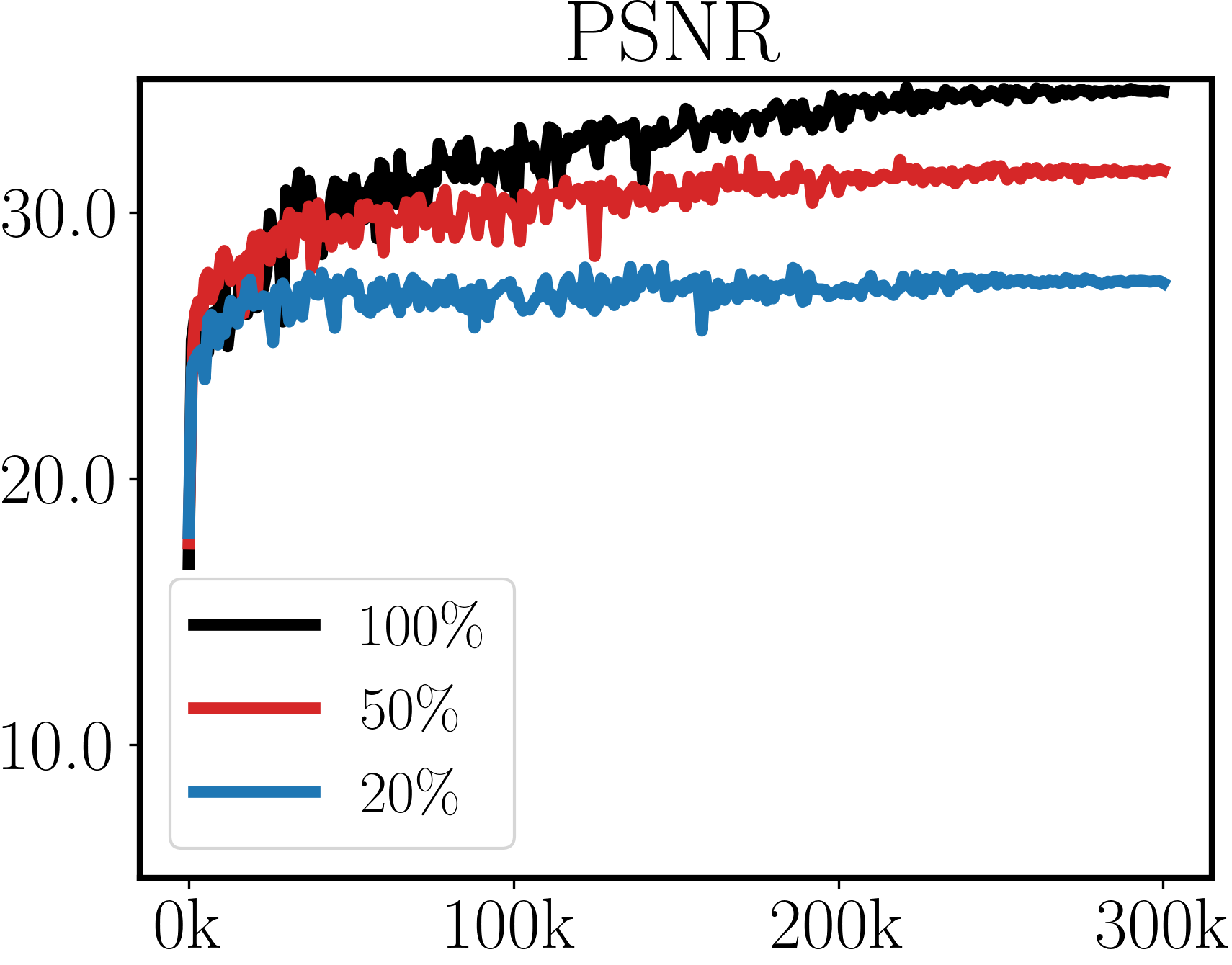} 
         &\includegraphics[width=0.24\textwidth]{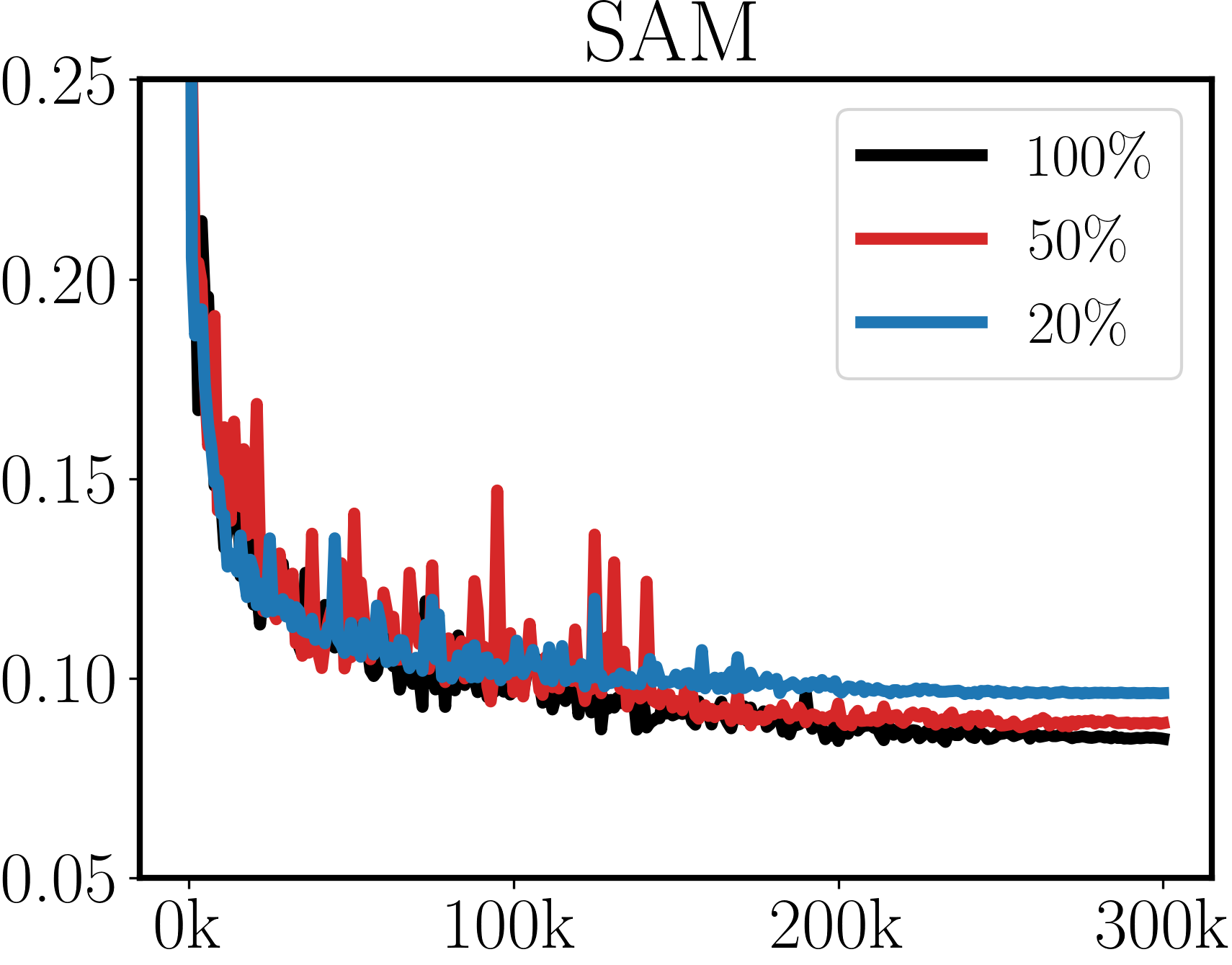}\\
         \raisebox{15mm}{\rotatebox[origin=c]{90}{HPRN*}}
         &\includegraphics[width=0.24\textwidth]{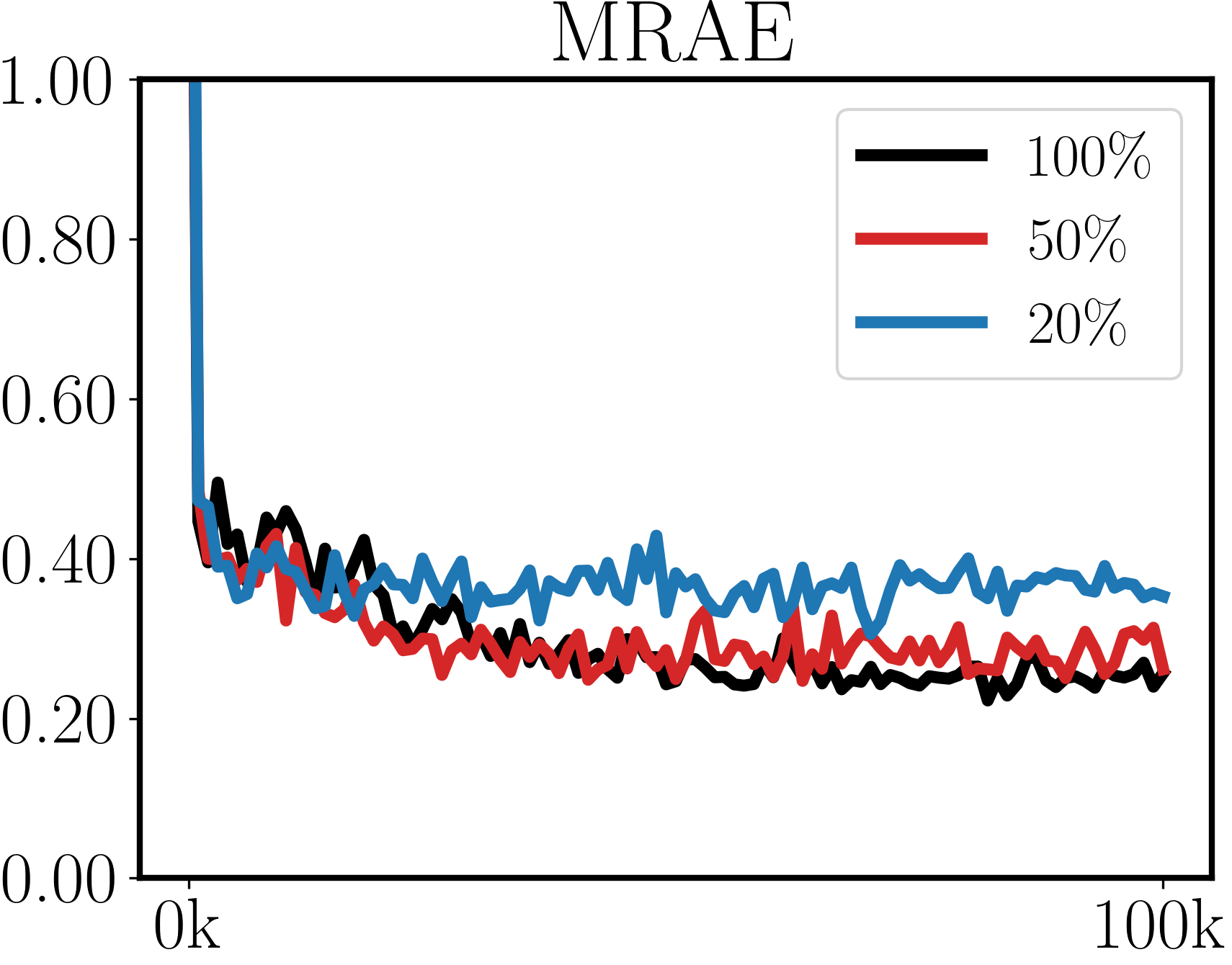}  
         &\includegraphics[width=0.24\textwidth]{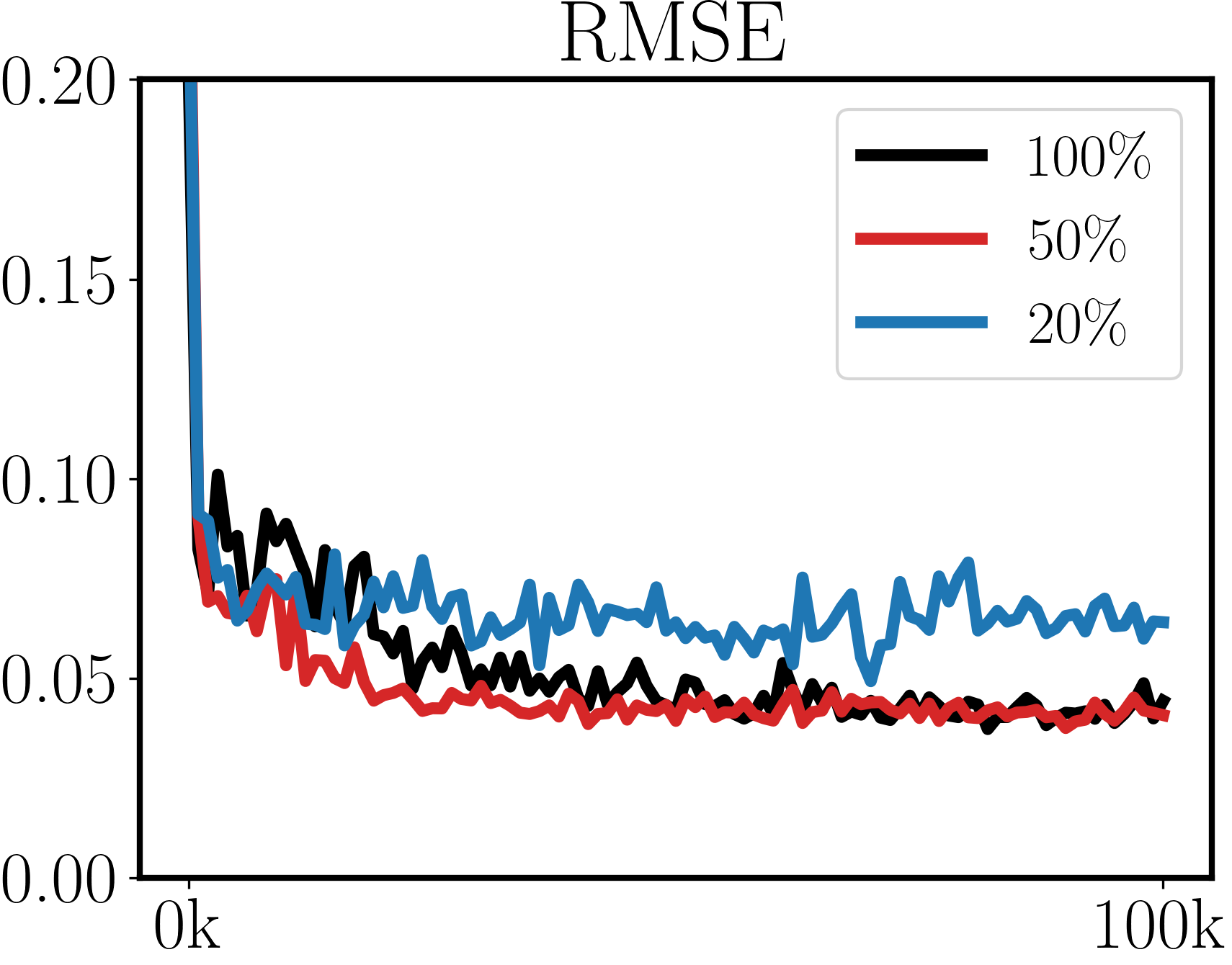} 
         &\includegraphics[width=0.24\textwidth]{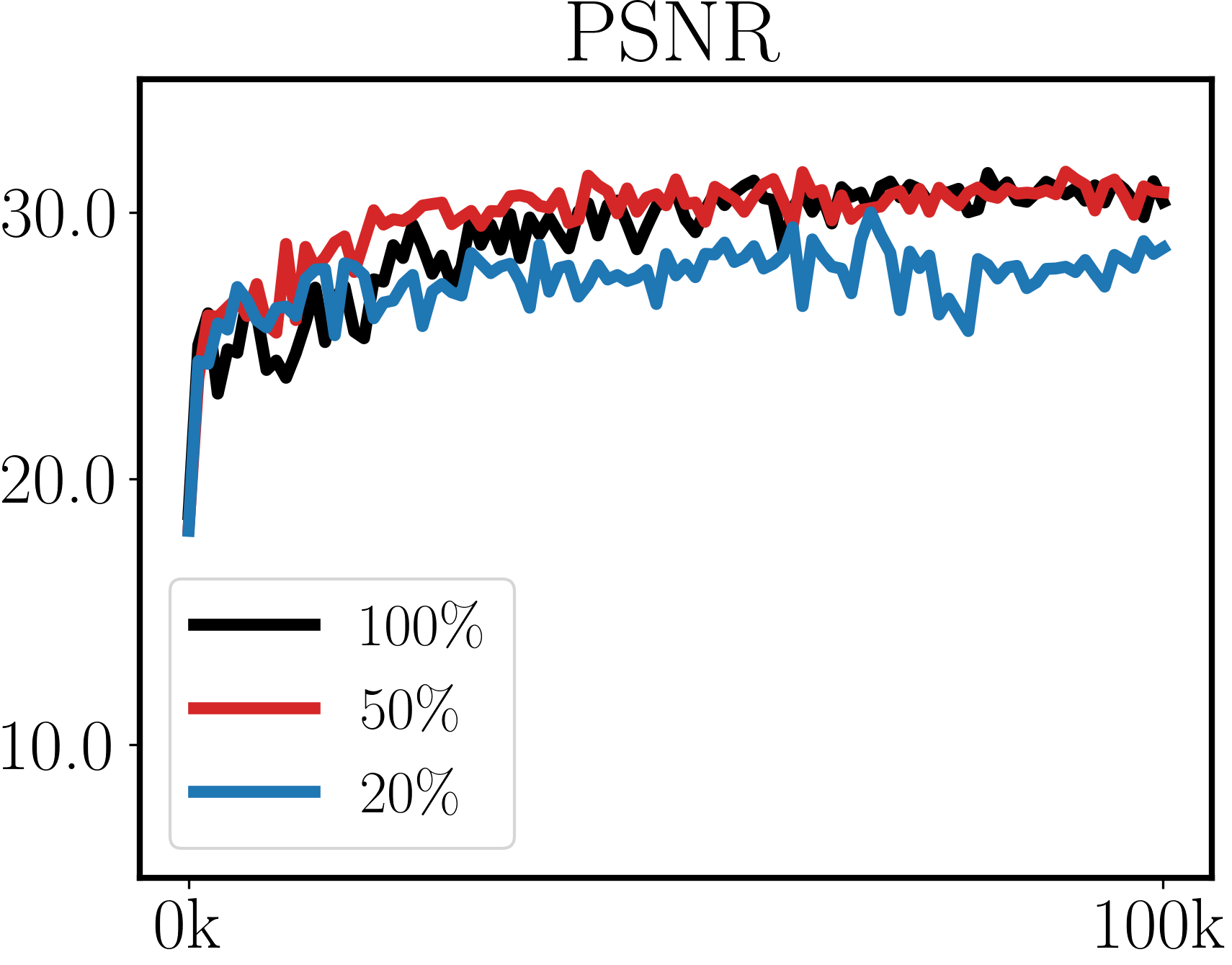} 
         &\includegraphics[width=0.24\textwidth]{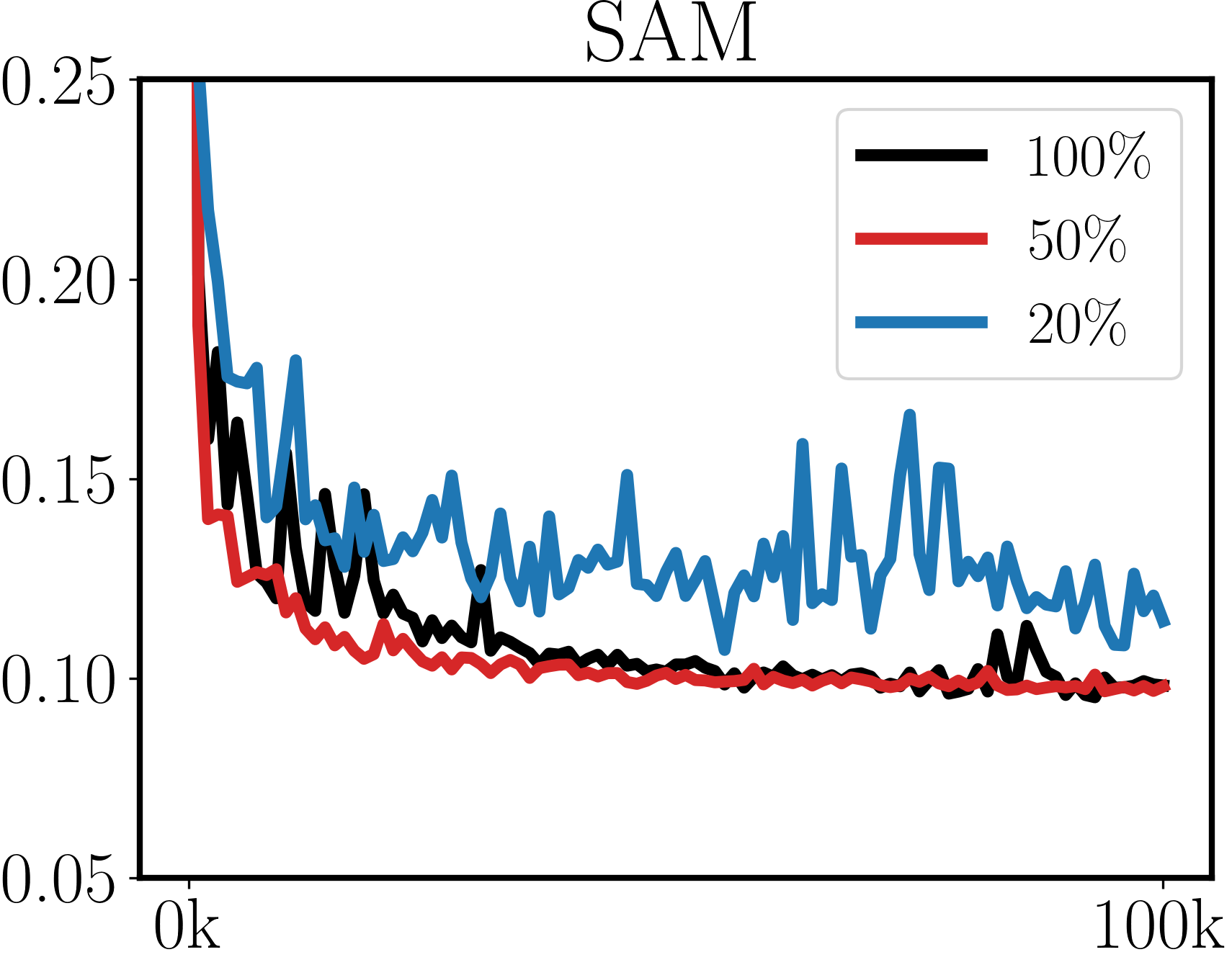}\\
         \raisebox{15mm}{\rotatebox[origin=c]{90}{SSTHyper}}
         &\includegraphics[width=0.24\textwidth]{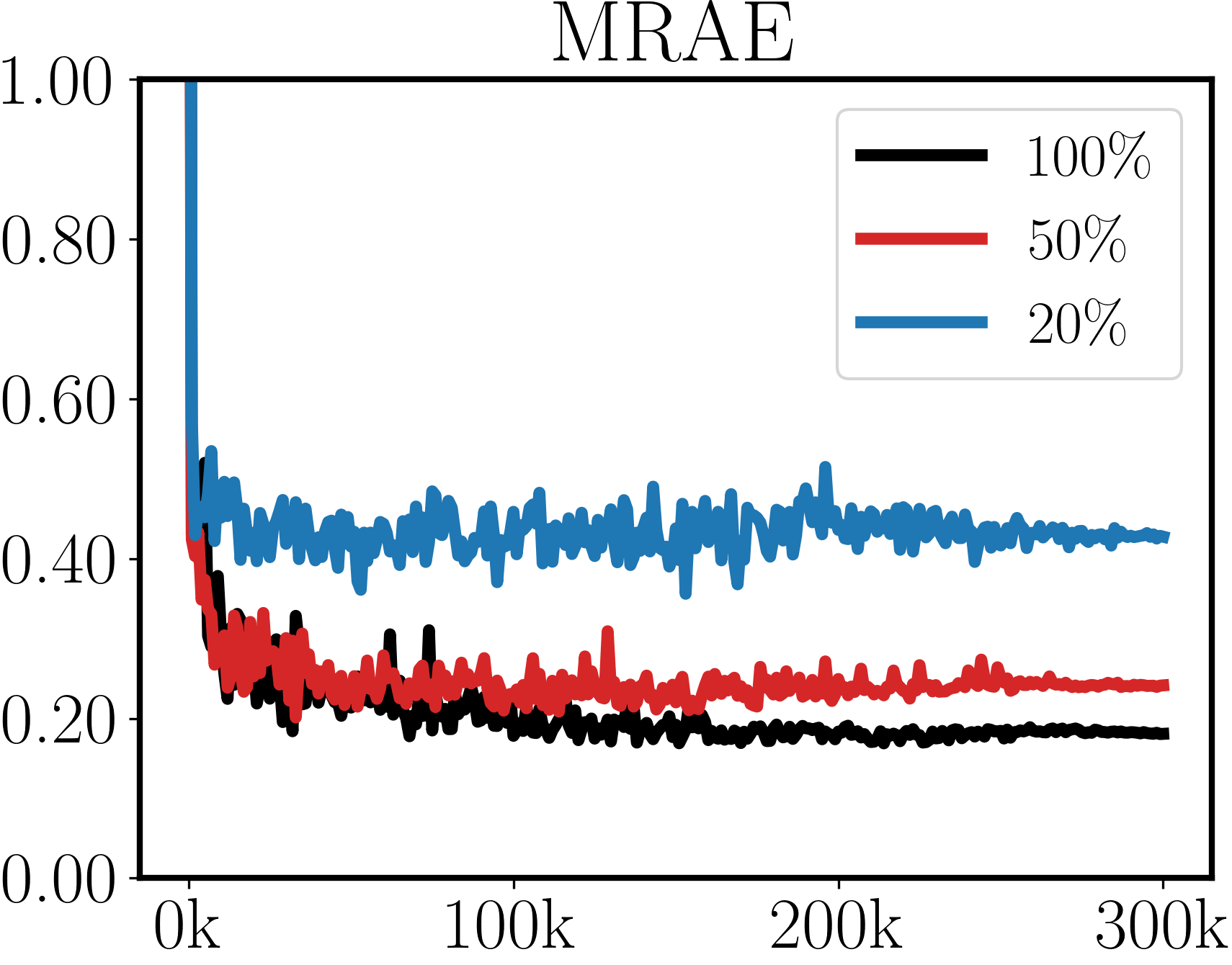}  
         &\includegraphics[width=0.24\textwidth]{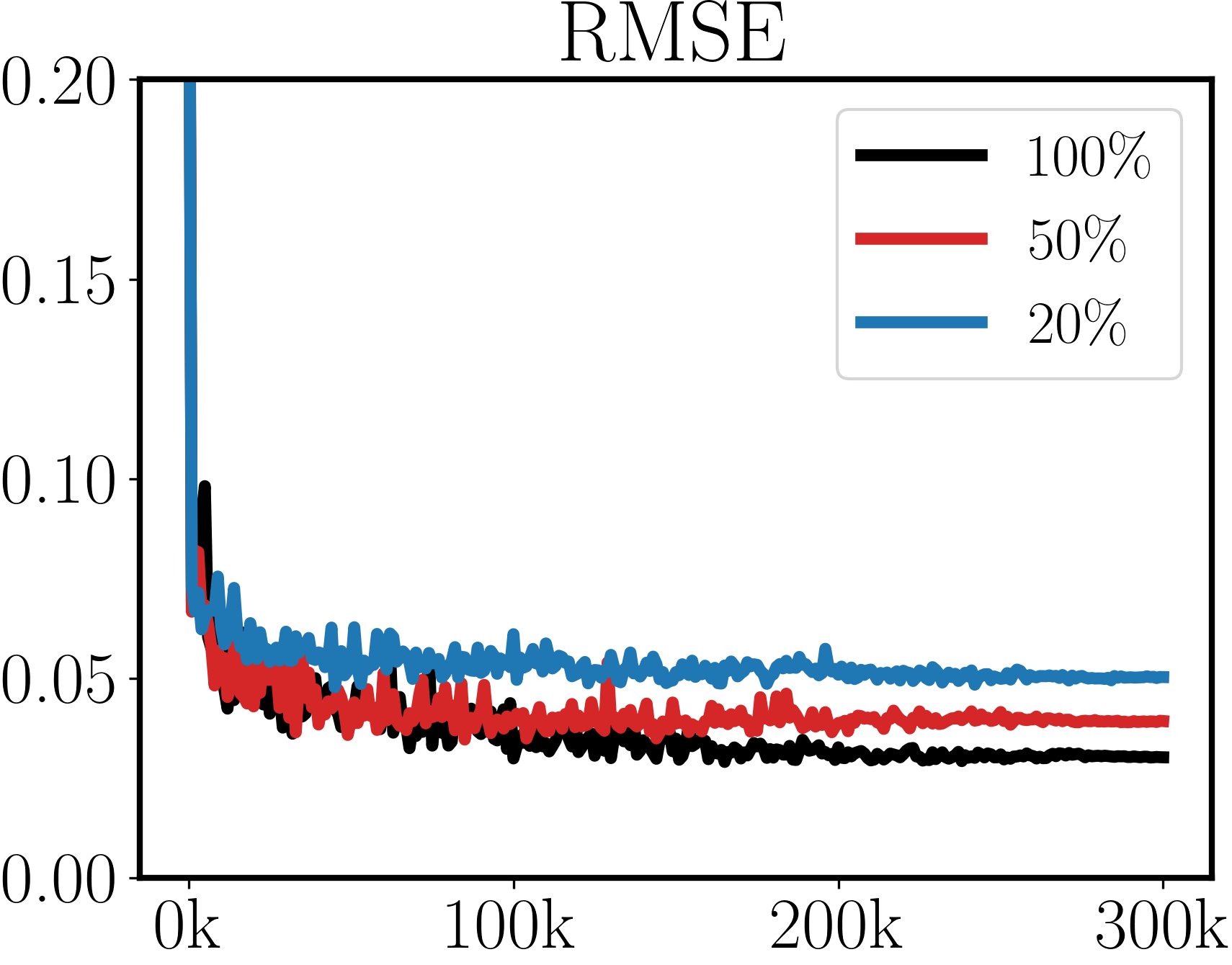} 
         &\includegraphics[width=0.24\textwidth]{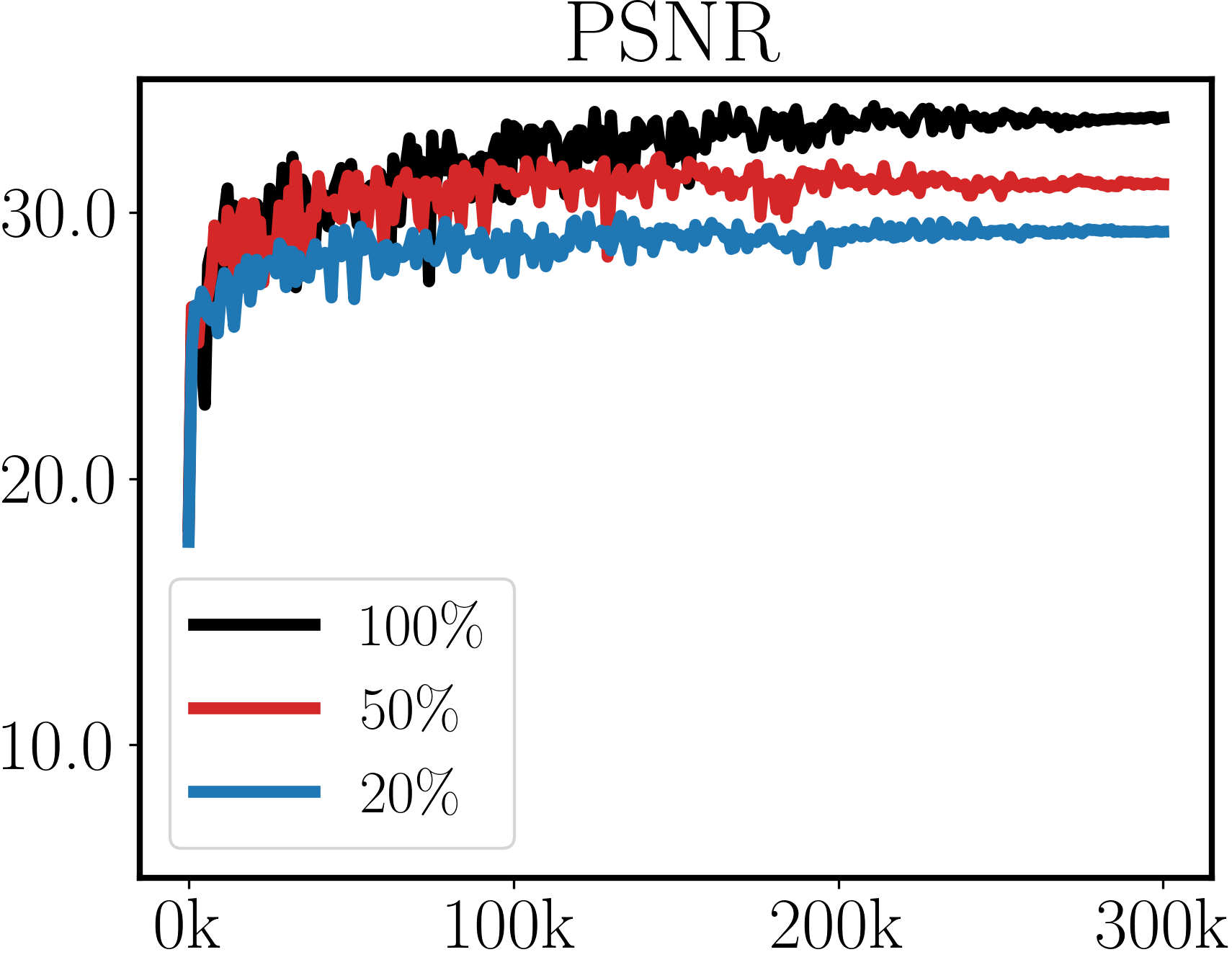} 
         &\includegraphics[width=0.24\textwidth]{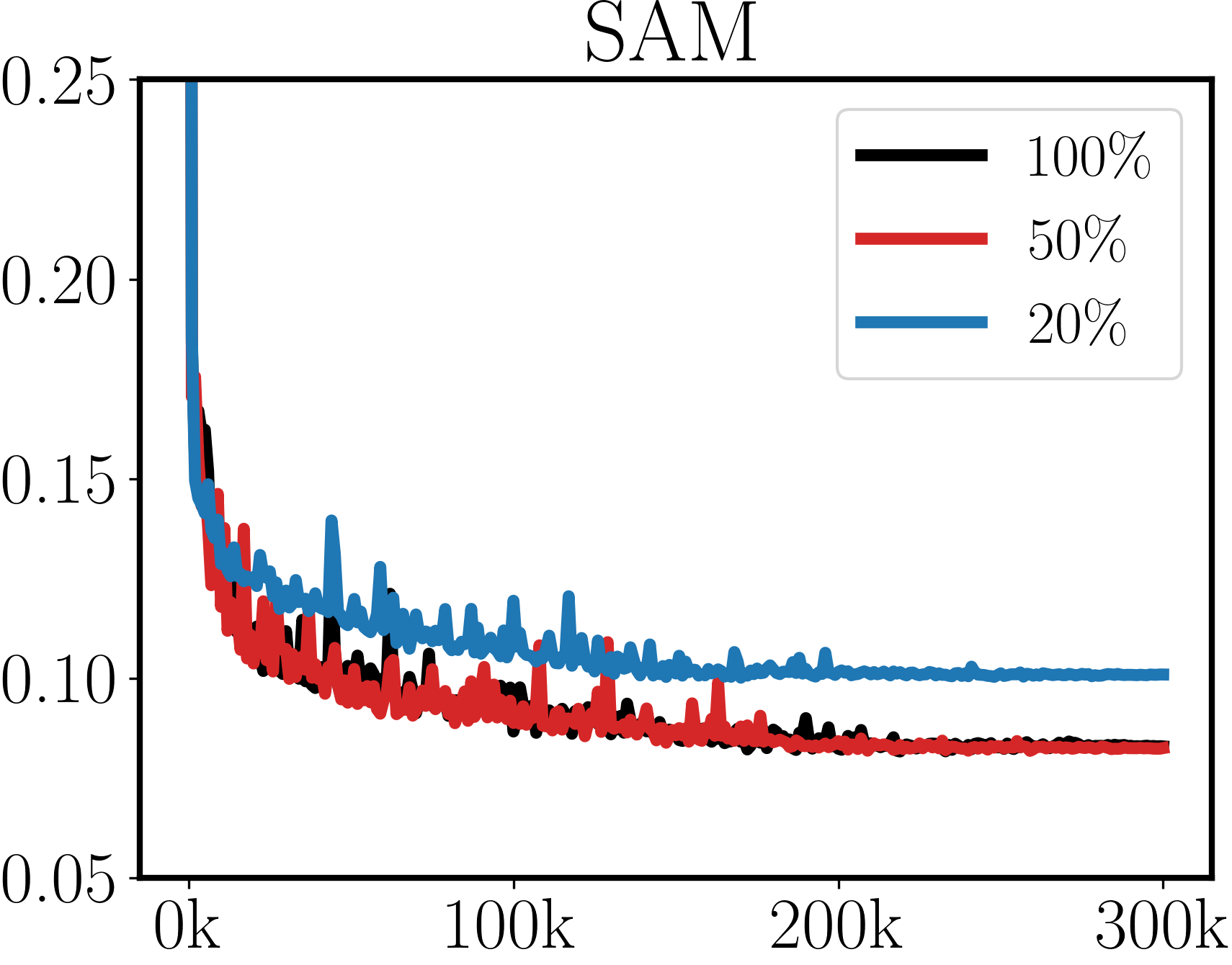}\\
         \raisebox{15mm}{\rotatebox[origin=c]{90}{MSFN}}
         &\includegraphics[width=0.24\textwidth]{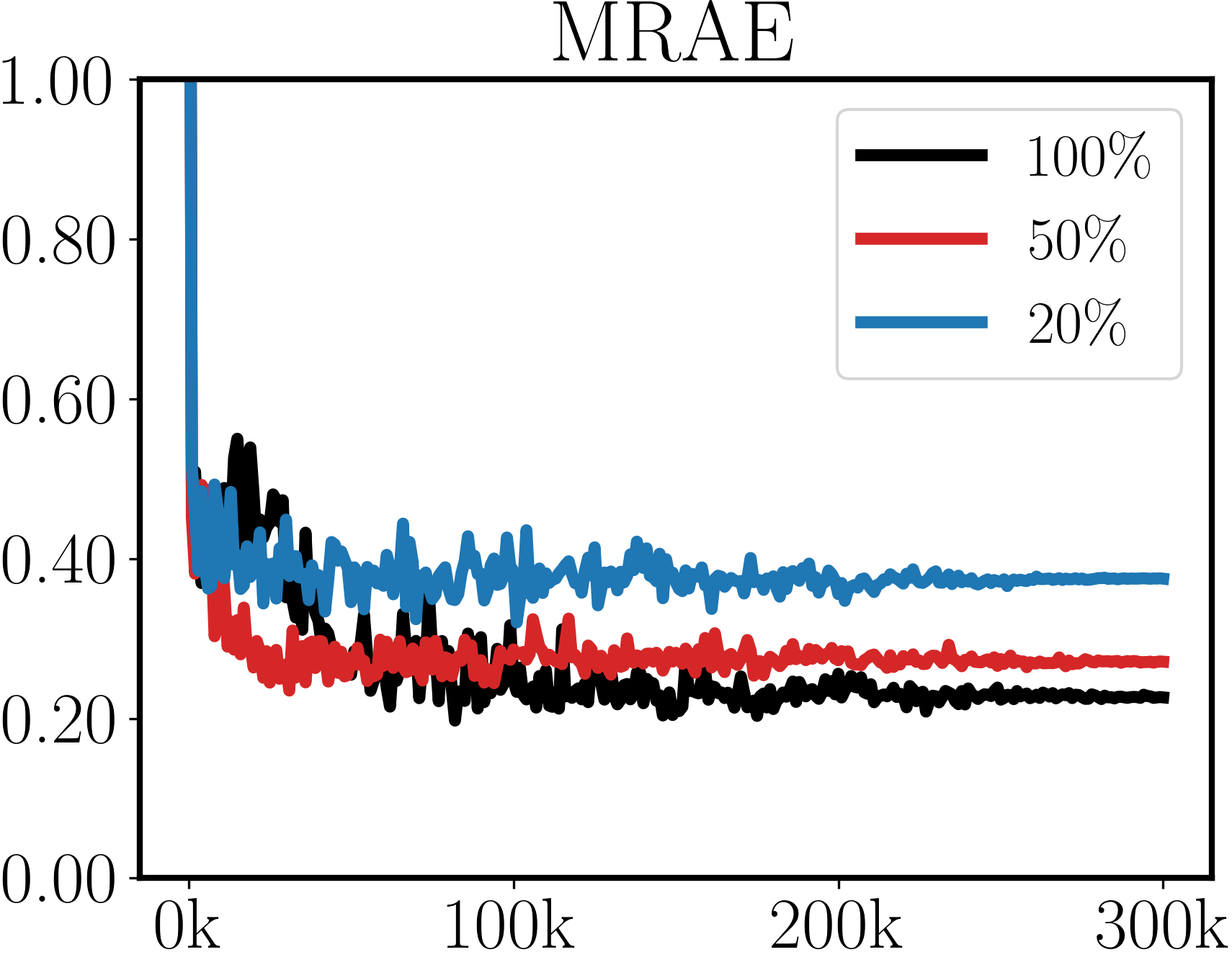}  
         &\includegraphics[width=0.24\textwidth]{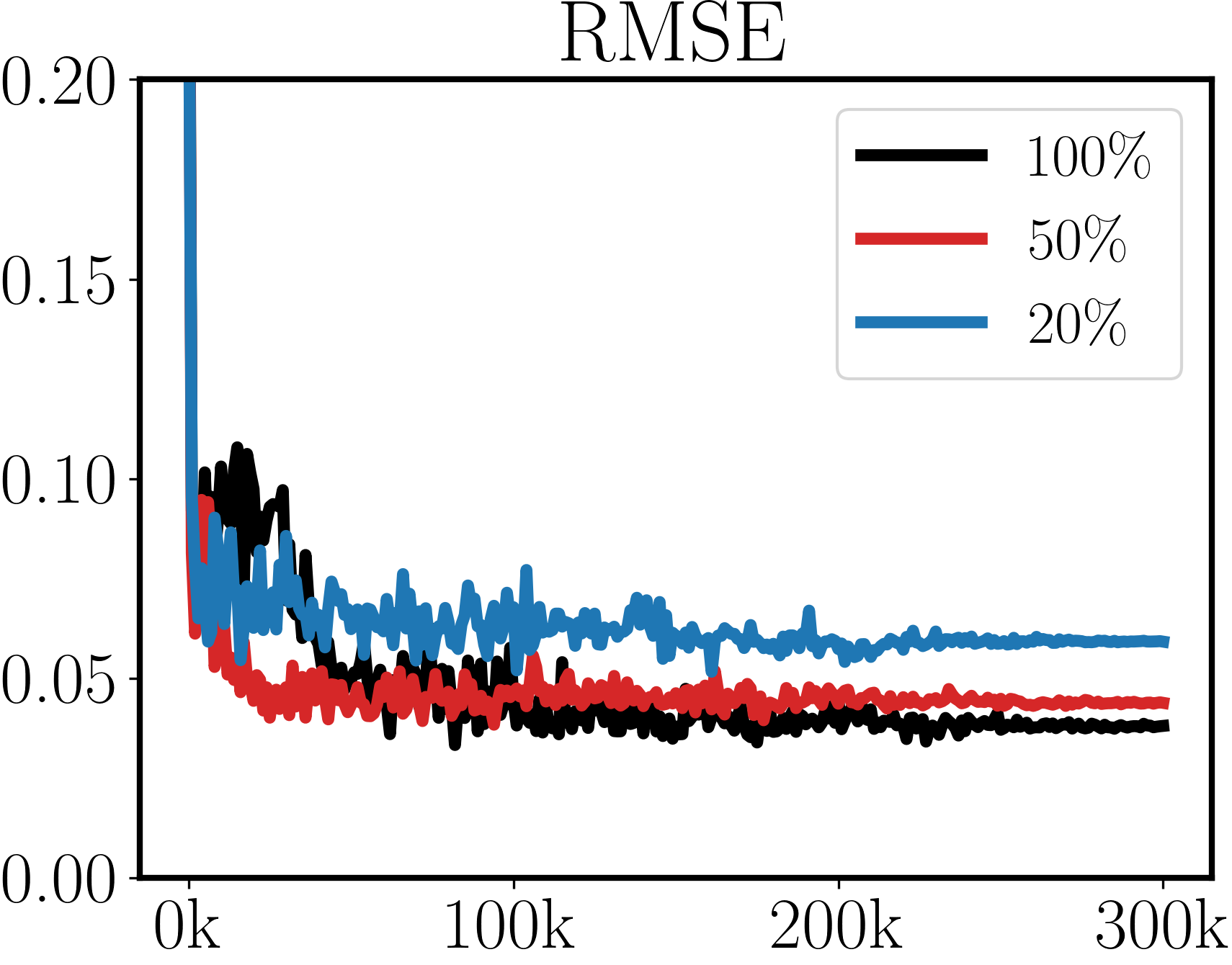} 
         &\includegraphics[width=0.24\textwidth]{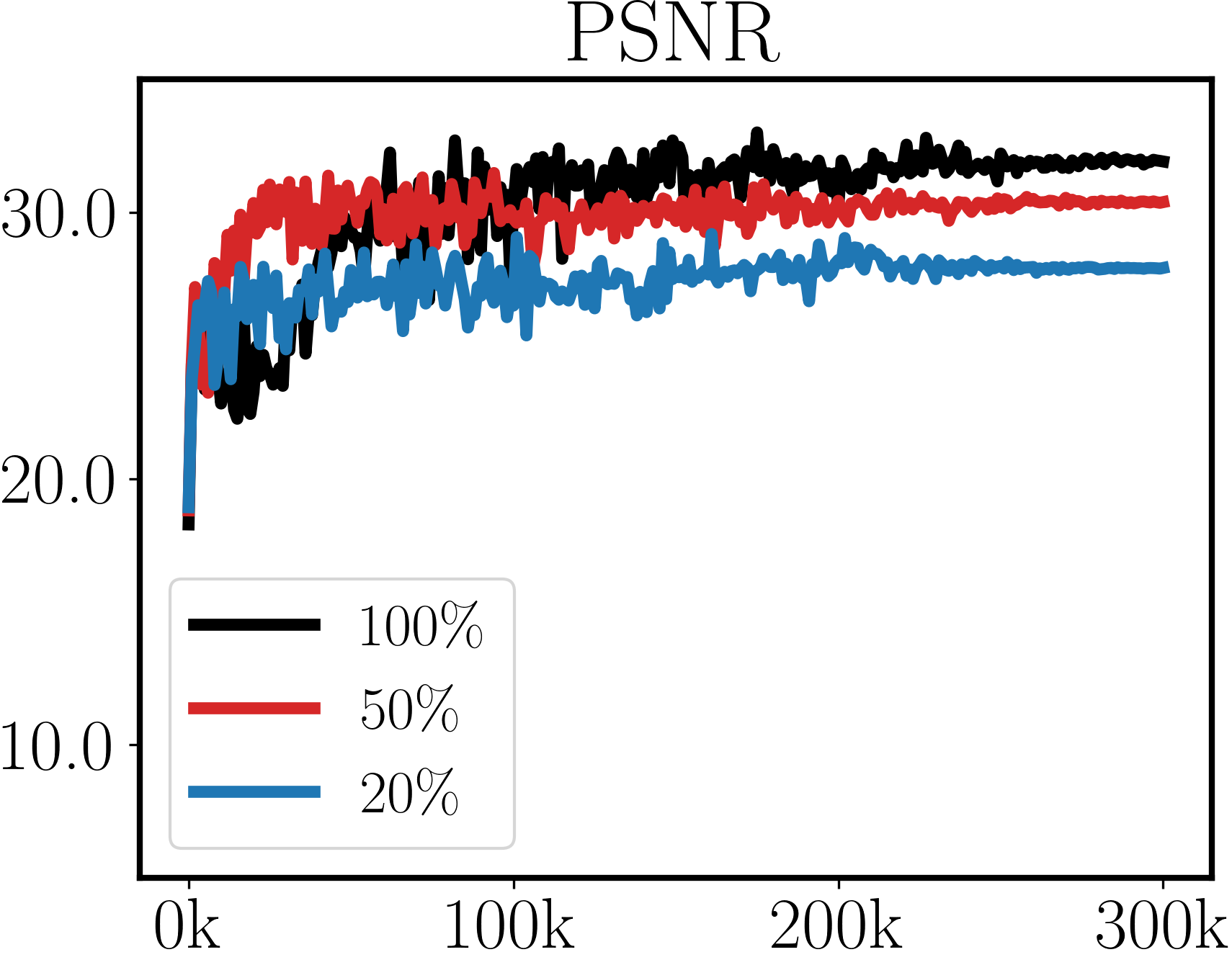} 
         &\includegraphics[width=0.24\textwidth]{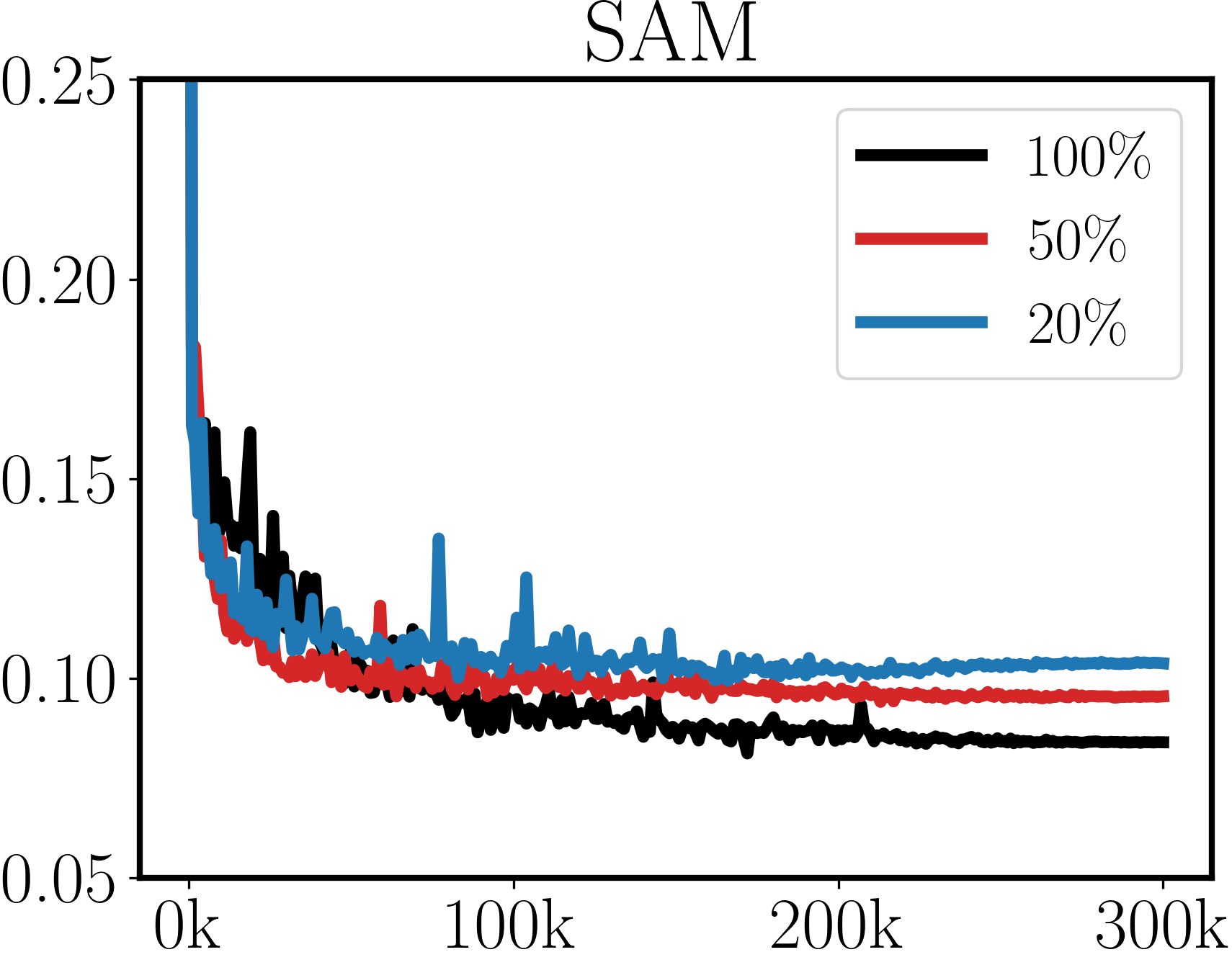}\\
         \raisebox{15mm}{\rotatebox[origin=c]{90}{GMSR}}
         &\includegraphics[width=0.24\textwidth]{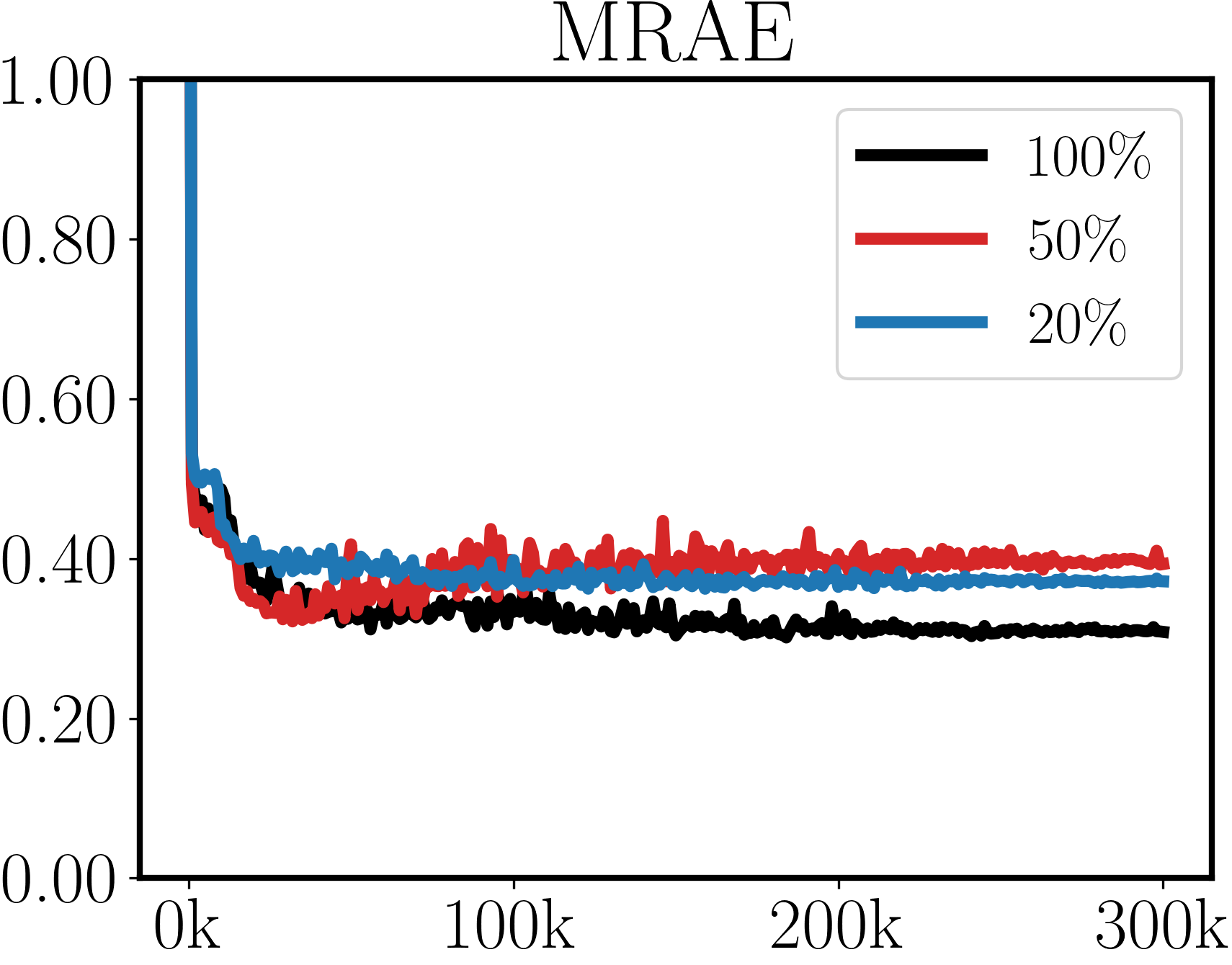}  
         &\includegraphics[width=0.24\textwidth]{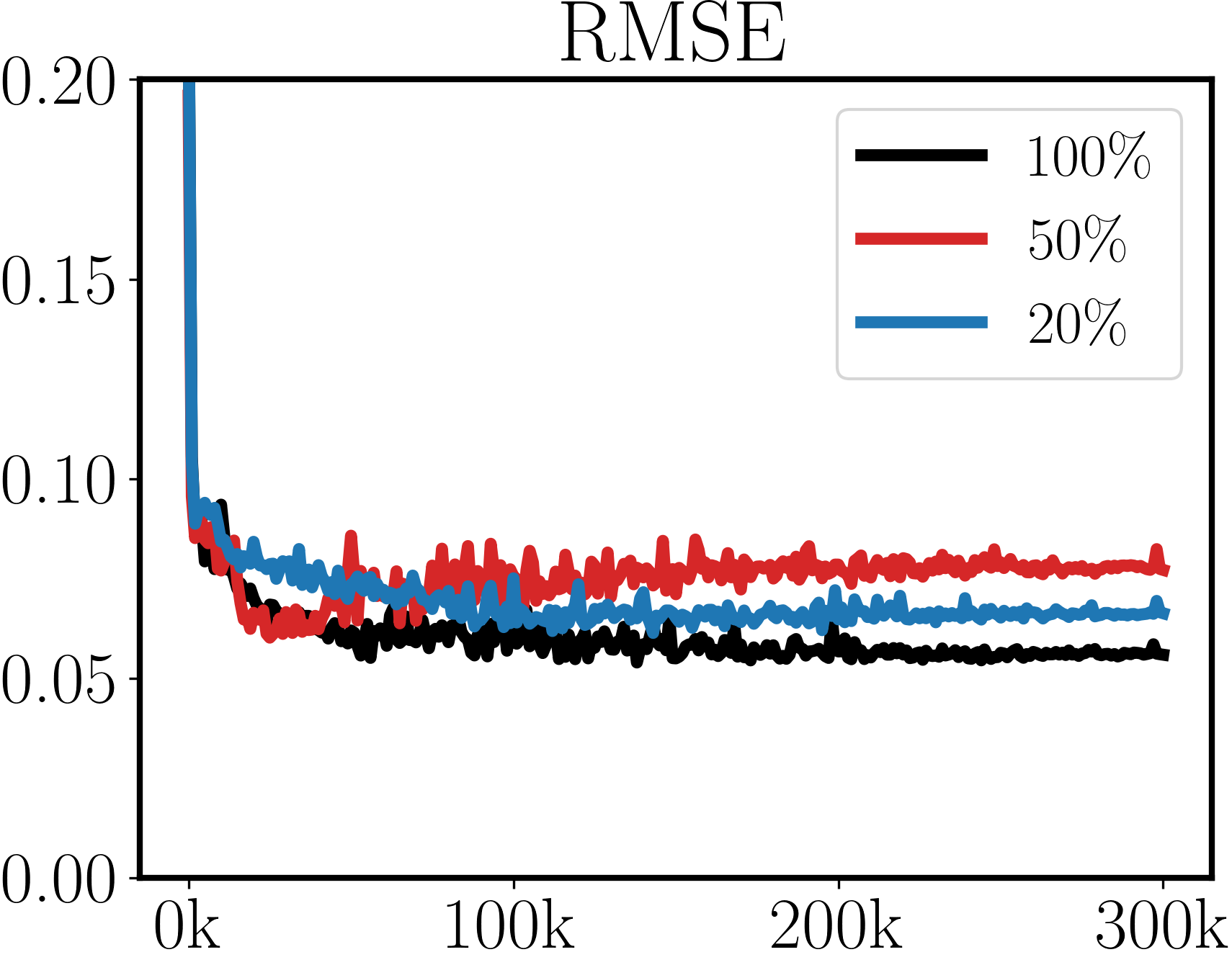} 
         &\includegraphics[width=0.24\textwidth]{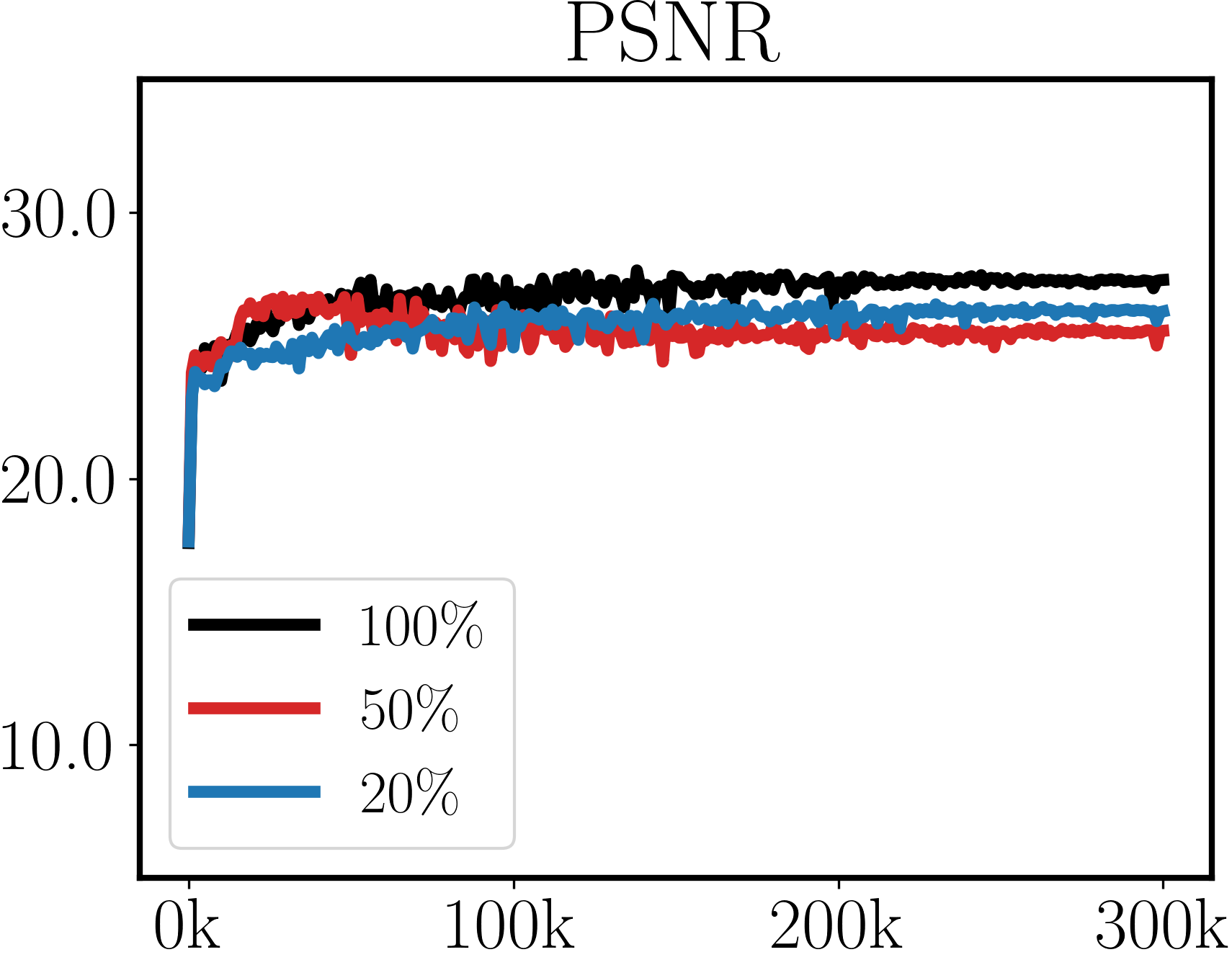} 
         &\includegraphics[width=0.24\textwidth]{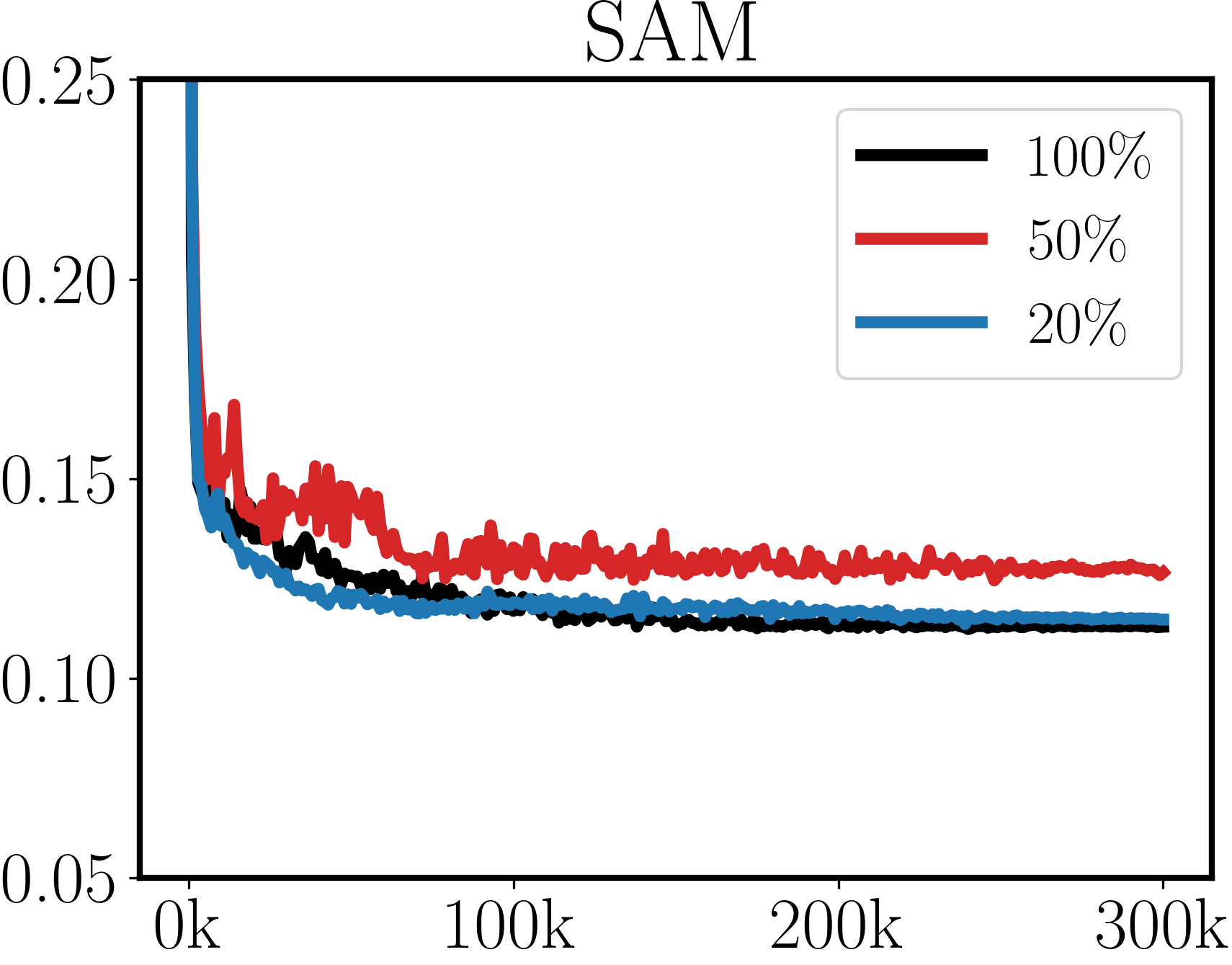}\\
         \raisebox{15mm}{\rotatebox[origin=c]{90}{SSRNet}}
         &\includegraphics[width=0.24\textwidth]{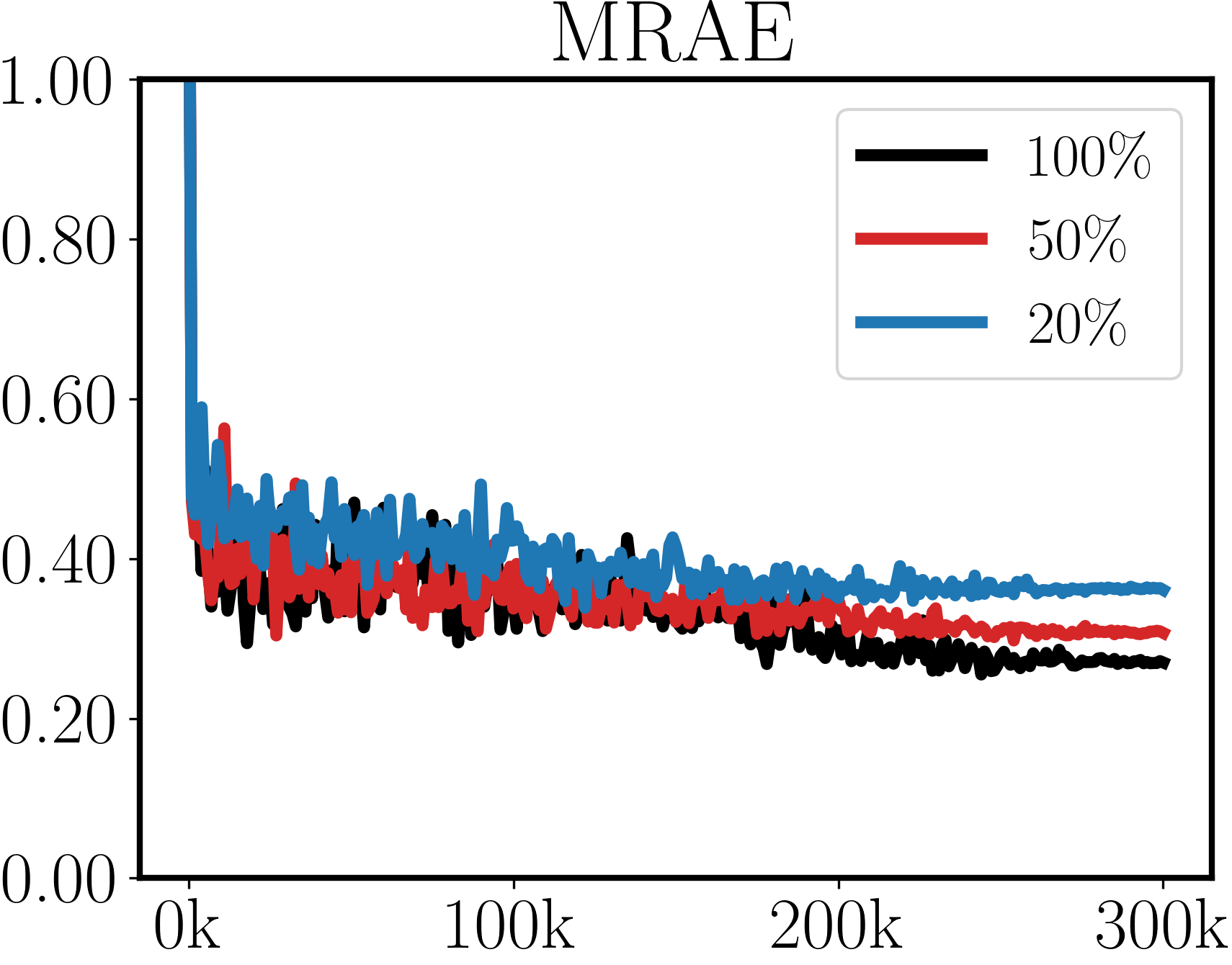}  
         &\includegraphics[width=0.24\textwidth]{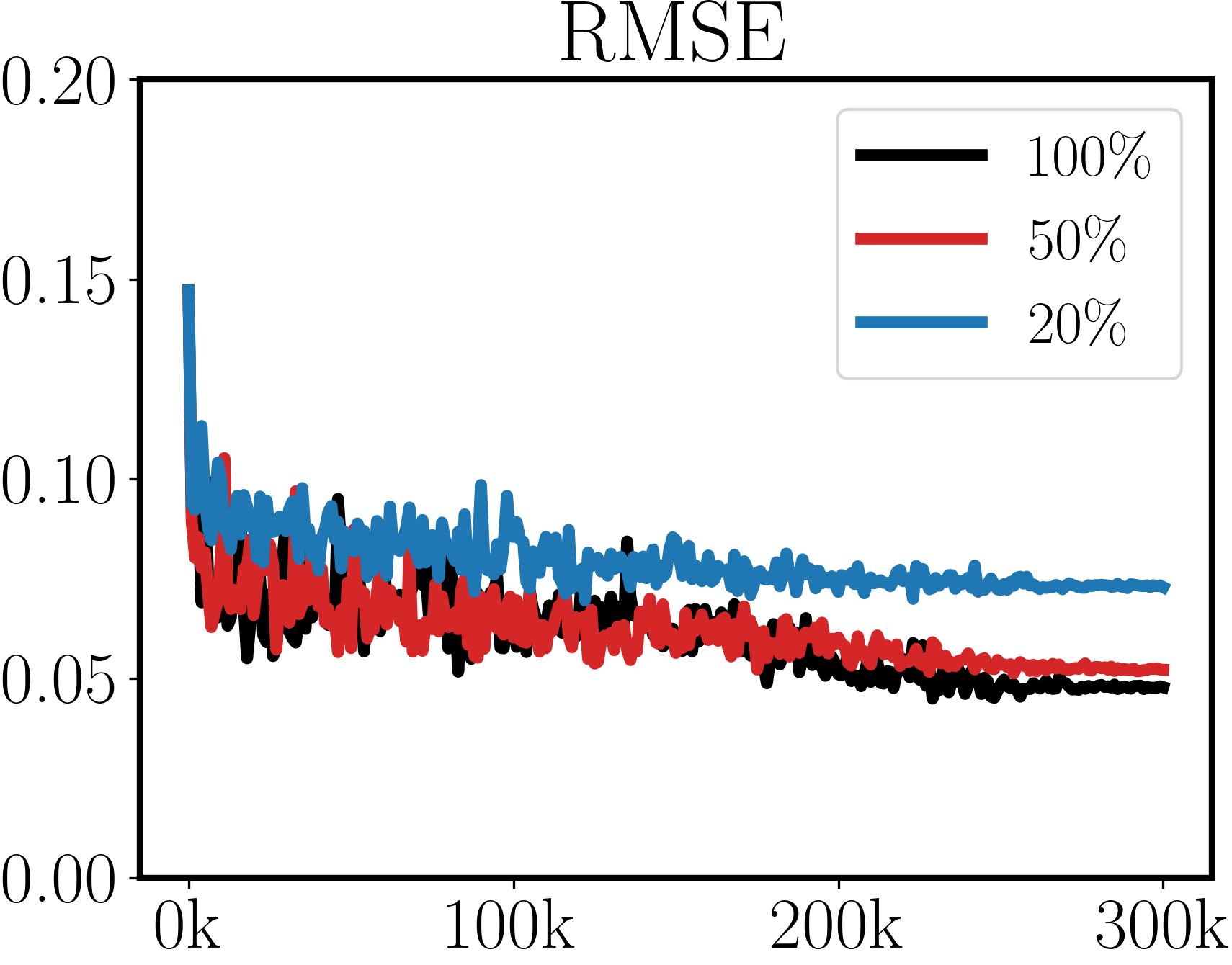} 
         &\includegraphics[width=0.24\textwidth]{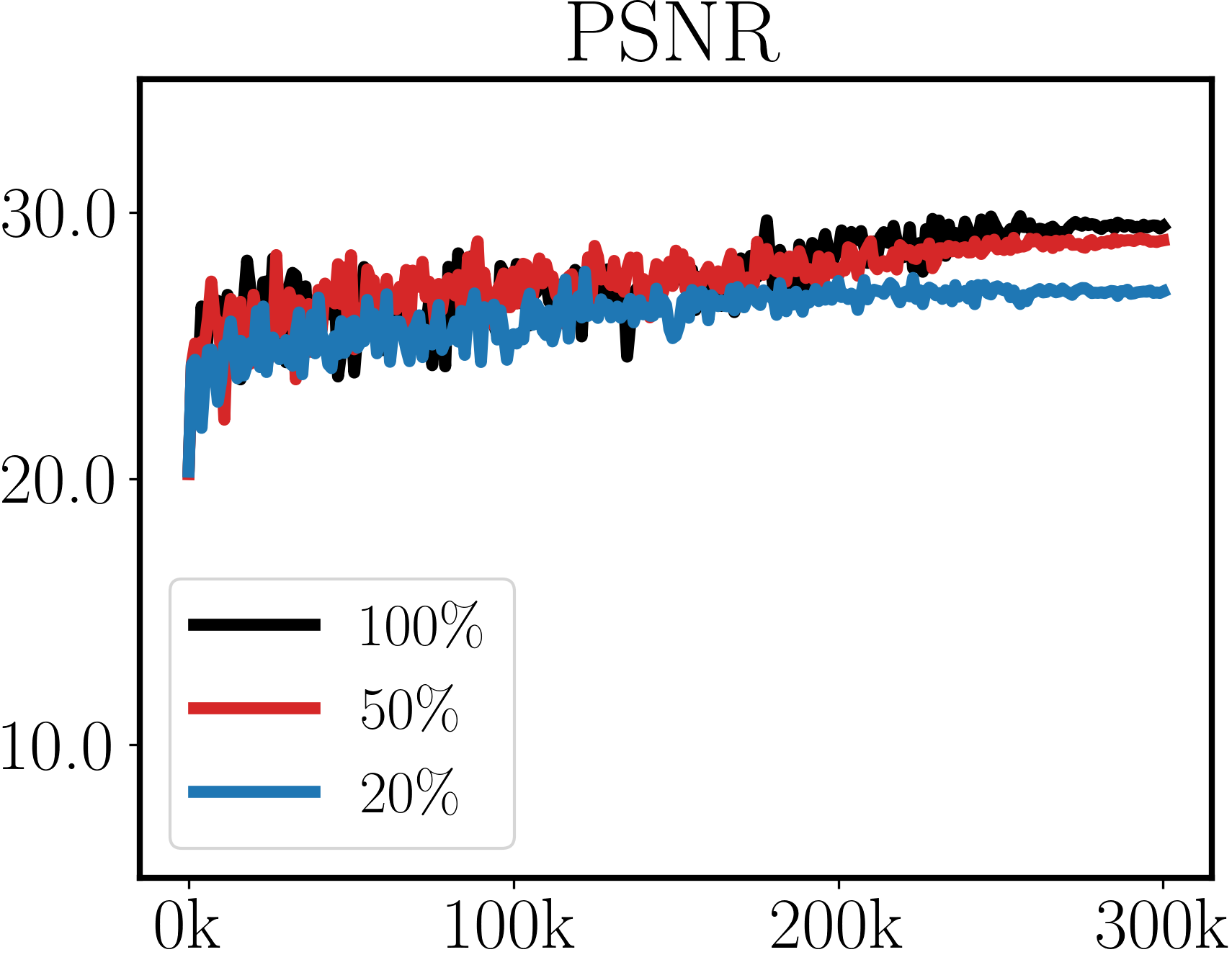} 
         &\includegraphics[width=0.24\textwidth]{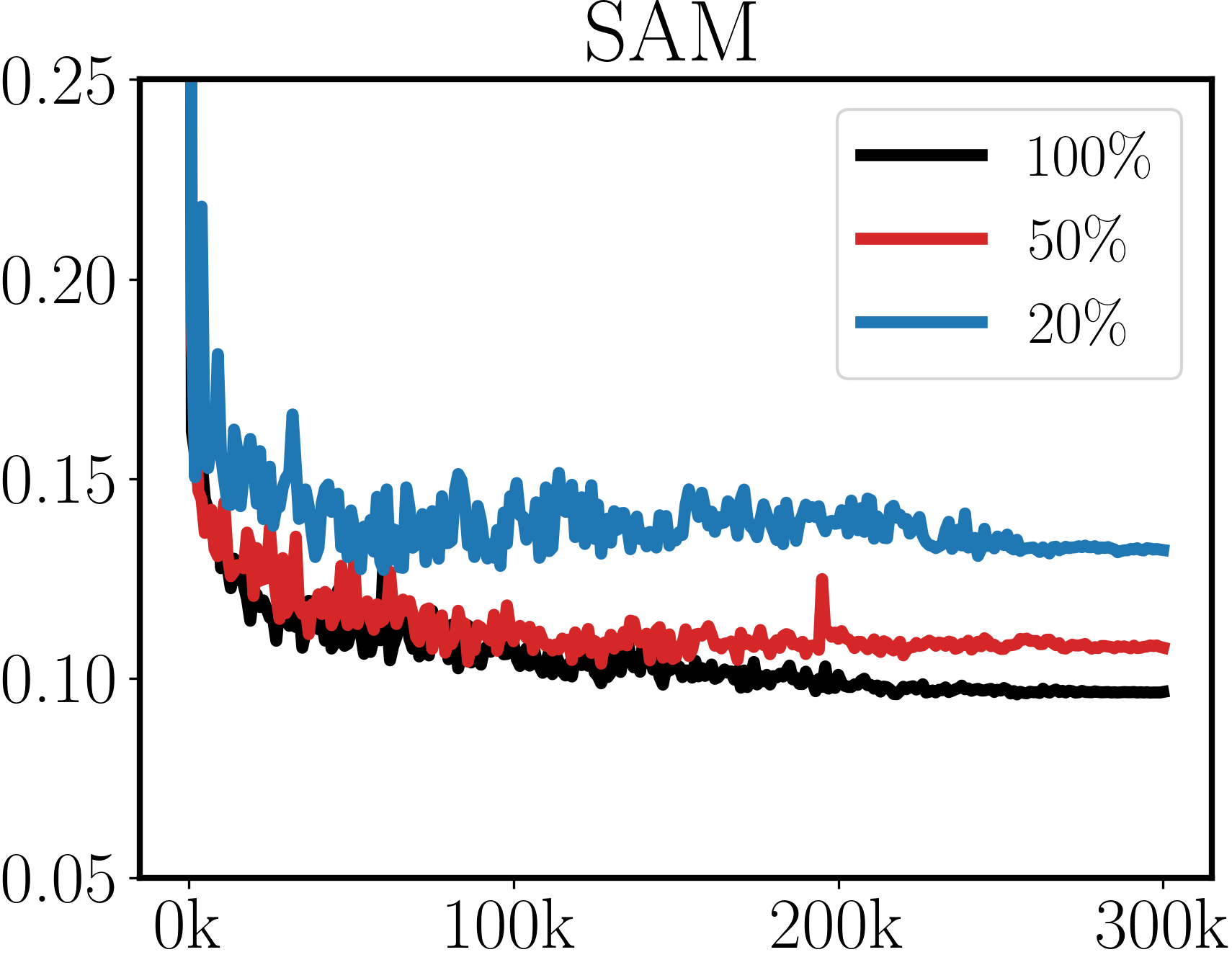}\\
    \end{tabular}
    \caption{Performance evaluation on MRAE, RMSE, PSNR, and SAM for HySAT, HPRN, SSTHyper, MSFN, GMSR, and SSRNet with 100\%, 50\%, and 20\% of the original training data on the ARAD1K dataset. * Note: HPRN was originally not trained on ARAD1K. We followed the training strategy in the original paper, but the results diverged after around 100k iterations, so we reported the results only up to 100k iterations when the convergence was observed.}
    \label{fig:supp_train_percent3}
\end{figure*}

\section{Results of Validation with Unseen Data for Other Networks}
In Table~3 of the main paper, we demonstrate the performance drop behaviour of the MST++ network on the ARAD1K dataset. To prove that this is true to other networks as well, we carry out the same experiments for all the other candidate networks. The results are summarized in Supplemental Table~\ref{tab:supp_validation_ARAD1K1} and Supplemental Table~\ref{tab:supp_validation_ARAD1K2}. As the noise levels, RGB formats, and aberration conditions asymptotically approach realistic imaging scenarios in the real world, the pre-trained models~\cite{cai2022mst++} for other networks gradually degrade, similar as the MST++. All the results consistently support our conclusion about the generalization difficulties of these methods in realistic imaging conditions.

% validation results of different networks on ARAD1K for overfitting
\begin{table*}[!htp]
\caption{Evaluation of pre-trained models on synthesized validation data generated from the ARAD1K dataset (Part I).}
\centering
\begin{tabular}{l|lccc|cccc}
\hline
\multirow{2}{*}{Network} & \multicolumn{4}{c|}{Data property} & \multirow{2}{*}{MRAE $\downarrow$} & \multirow{2}{*}{RMSE $\downarrow$} & \multirow{2}{*}{PSNR $\uparrow$} & \multirow{2}{*}{SAM $\downarrow$} \\ \cline{2-5}
 & \multicolumn{1}{c|}{Data source} & \multicolumn{1}{c|}{Noise (npe)} & \multicolumn{1}{c|}{RGB format} & Aberration &  &  &  &  \\ \hline
\multirow{4}{*}{MST++} & \multicolumn{1}{c|}{NTIRE 2022} & \multicolumn{1}{c|}{unknown} & \multicolumn{1}{c|}{jpg (Q unknown)} & None & 0.170 & 0.029 & 33.8 & 0.084 \\ \cline{2-9} 
 & \multicolumn{1}{c|}{\multirow{3}{*}{Synthesized}} & \multicolumn{1}{c|}{0} & \multicolumn{1}{c|}{jpg (Q = 65)} & None & 0.460 & 0.049 & 29.2 & 0.094 \\ \cline{3-9} 
 & \multicolumn{1}{c|}{} & \multicolumn{1}{c|}{0} & \multicolumn{1}{c|}{png (lossless)} & None & 0.362 & 0.057 & 28.7 & 0.087 \\ \cline{3-9} 
 & \multicolumn{1}{c|}{} & \multicolumn{1}{c|}{1000} & \multicolumn{1}{c|}{png (lossless)} & CA* & 0.312 & 0.055 & 28.4 & 0.118 \\ \hline \hline
\multirow{4}{*}{MST-L} & \multicolumn{1}{c|}{NTIRE 2022} & \multicolumn{1}{c|}{unknown} & \multicolumn{1}{c|}{jpg (Q unknown)} & None & 0.181 & 0.031 & 33.0 & 0.091 \\ \cline{2-9} 
 & \multicolumn{1}{l|}{\multirow{3}{*}{Synthesized}} & \multicolumn{1}{c|}{0} & \multicolumn{1}{c|}{jpg (Q = 65)} & None & 0.417 & 0.047 & 29.7 & 0.099 \\ \cline{3-9} 
 & \multicolumn{1}{l|}{} & \multicolumn{1}{c|}{0} & \multicolumn{1}{c|}{png (lossless)} & None & 0.384 & 0.058 & 28.4 & 0.096 \\ \cline{3-9} 
 & \multicolumn{1}{l|}{} & \multicolumn{1}{c|}{1000} & \multicolumn{1}{c|}{png (lossless)} & CA* & 0.327 & 0.055 & 28.0 & 0.118 \\ \hline \hline
\multirow{4}{*}{MPRNet} & \multicolumn{1}{c|}{NTIRE 2022} & \multicolumn{1}{c|}{unknown} & \multicolumn{1}{c|}{jpg (Q unknown)} & None & 0.182 & 0.032 & 32.9 & 0.088 \\ \cline{2-9} 
 & \multicolumn{1}{l|}{\multirow{3}{*}{Synthesized}} & \multicolumn{1}{c|}{0} & \multicolumn{1}{c|}{jpg (Q = 65)} & None & 0.453 & 0.048 & 29.1 & 0.092 \\ \cline{3-9} 
 & \multicolumn{1}{l|}{} & \multicolumn{1}{c|}{0} & \multicolumn{1}{c|}{png (lossless)} & None & 0.359 & 0.051 & 29.5 & 0.086 \\ \cline{3-9} 
 & \multicolumn{1}{l|}{} & \multicolumn{1}{c|}{1000} & \multicolumn{1}{c|}{png (lossless)} & CA* & 0.661 & 0.066 & 26.1 & 0.125 \\ \hline \hline
\multirow{4}{*}{Restormer} & \multicolumn{1}{c|}{NTIRE 2022} & \multicolumn{1}{c|}{unknown} & \multicolumn{1}{c|}{jpg (Q unknown)} & None & 0.190 & 0.032 & 33.0 & 0.097 \\ \cline{2-9} 
 & \multicolumn{1}{l|}{\multirow{3}{*}{Synthesized}} & \multicolumn{1}{c|}{0} & \multicolumn{1}{c|}{jpg (Q = 65)} & None & 0.454 & 0.051 & 28.6 & 0.100 \\ \cline{3-9} 
 & \multicolumn{1}{l|}{} & \multicolumn{1}{c|}{0} & \multicolumn{1}{c|}{png (lossless)} & None & 0.363 & 0.053 & 28.6 & 0.098 \\ \cline{3-9} 
 & \multicolumn{1}{l|}{} & \multicolumn{1}{c|}{1000} & \multicolumn{1}{c|}{png (lossless)} & CA* & 0.510 & 0.066 & 25.5 & 0.126 \\ \hline \hline
\multirow{4}{*}{MIRNet} & \multicolumn{1}{c|}{NTIRE 2022} & \multicolumn{1}{c|}{unknown} & \multicolumn{1}{c|}{jpg (Q unknown)} & None & 0.189 & 0.032 & 33.3 & 0.091 \\ \cline{2-9} 
 & \multicolumn{1}{l|}{\multirow{3}{*}{Synthesized}} & \multicolumn{1}{c|}{0} & \multicolumn{1}{c|}{jpg (Q = 65)} & None & 0.467 & 0.051 & 28.7 & 0.096 \\ \cline{3-9} 
 & \multicolumn{1}{l|}{} & \multicolumn{1}{c|}{0} & \multicolumn{1}{c|}{png (lossless)} & None & 0.366 & 0.055 & 28.8 & 0.091 \\ \cline{3-9} 
 & \multicolumn{1}{l|}{} & \multicolumn{1}{c|}{1000} & \multicolumn{1}{c|}{png (lossless)} & CA* & 0.404 & 0.077 & 24.8 & 0.124 \\ \hline \hline
\multirow{4}{*}{HINet} & \multicolumn{1}{c|}{NTIRE 2022} & \multicolumn{1}{c|}{unknown} & \multicolumn{1}{c|}{jpg (Q unknown)} & None & 0.212 & 0.037 & 31.4 & 0.091 \\ \cline{2-9} 
 & \multicolumn{1}{l|}{\multirow{3}{*}{Synthesized}} & \multicolumn{1}{c|}{0} & \multicolumn{1}{c|}{jpg (Q = 65)} & None & 0.460 & 0.051 & 28.3 & 0.094 \\ \cline{3-9} 
 & \multicolumn{1}{l|}{} & \multicolumn{1}{c|}{0} & \multicolumn{1}{c|}{png (lossless)} & None & 0.384 & 0.055 & 28.2 & 0.094 \\ \cline{3-9} 
 & \multicolumn{1}{l|}{} & \multicolumn{1}{c|}{1000} & \multicolumn{1}{c|}{png (lossless)} & CA* & 0.450 & 0.063 & 26.5 & 0.120 \\ \hline \hline
\multirow{4}{*}{HDNet} & \multicolumn{1}{c|}{NTIRE 2022} & \multicolumn{1}{c|}{unknown} & \multicolumn{1}{c|}{jpg (Q unknown)} & None & 0.214 & 0.037 & 31.5 & 0.098 \\ \cline{2-9} 
 & \multicolumn{1}{l|}{\multirow{3}{*}{Synthesized}} & \multicolumn{1}{c|}{0} & \multicolumn{1}{c|}{jpg (Q = 65)} & None & 0.404 & 0.050 & 28.8 & 0.102 \\ \cline{3-9} 
 & \multicolumn{1}{l|}{} & \multicolumn{1}{c|}{0} & \multicolumn{1}{c|}{png (lossless)} & None & 0.395 & 0.057 & 28.0 & 0.096 \\ \cline{3-9} 
 & \multicolumn{1}{l|}{} & \multicolumn{1}{c|}{1000} & \multicolumn{1}{c|}{png (lossless)} & CA* & 0.450 & 0.082 & 23.9 & 0.126 \\ \hline \hline
\multirow{4}{*}{AWAN} & \multicolumn{1}{c|}{NTIRE 2022} & \multicolumn{1}{c|}{unknown} & \multicolumn{1}{c|}{jpg (Q unknown)} & None & 0.222 & 0.041 & 31.0 & 0.098 \\ \cline{2-9} 
 & \multicolumn{1}{l|}{\multirow{3}{*}{Synthesized}} & \multicolumn{1}{c|}{0} & \multicolumn{1}{c|}{jpg (Q = 65)} & None & 0.299 & 0.044 & 29.7 & 0.105 \\ \cline{3-9} 
 & \multicolumn{1}{l|}{} & \multicolumn{1}{c|}{0} & \multicolumn{1}{c|}{png (lossless)} & None & 0.338 & 0.060 & 28.3 & 0.090 \\ \cline{3-9} 
 & \multicolumn{1}{l|}{} & \multicolumn{1}{c|}{1000} & \multicolumn{1}{c|}{png (lossless)} & CA* & 0.424 & 0.080 & 24.6 & 0.119 \\ \hline \hline
\multirow{4}{*}{EDSR} & \multicolumn{1}{c|}{NTIRE 2022} & \multicolumn{1}{c|}{unknown} & \multicolumn{1}{c|}{jpg (Q unknown)} & None & 0.340 & 0.051 & 27.5 & 0.095 \\ \cline{2-9} 
 & \multicolumn{1}{l|}{\multirow{3}{*}{Synthesized}} & \multicolumn{1}{c|}{0} & \multicolumn{1}{c|}{jpg (Q = 65)} & None & 0.473 & 0.064 & 25.8 & 0.104 \\ \cline{3-9} 
 & \multicolumn{1}{l|}{} & \multicolumn{1}{c|}{0} & \multicolumn{1}{c|}{png (lossless)} & None & 0.474 & 0.074 & 24.5 & 0.096 \\ \cline{3-9} 
 & \multicolumn{1}{l|}{} & \multicolumn{1}{c|}{1000} & \multicolumn{1}{c|}{png (lossless)} & CA* & 0.421 & 0.066 & 25.5 & 0.132 \\ \hline \hline
\multirow{4}{*}{HRNet} & \multicolumn{1}{c|}{NTIRE 2022} & \multicolumn{1}{c|}{unknown} & \multicolumn{1}{c|}{jpg (Q unknown)} & None & 0.376 & 0.065 & 25.4 & 0.102 \\ \cline{2-9} 
 & \multicolumn{1}{l|}{\multirow{3}{*}{Synthesized}} & \multicolumn{1}{c|}{0} & \multicolumn{1}{c|}{jpg (Q = 65)} & None & 0.397 & 0.066 & 25.3 & 0.108 \\ \cline{3-9} 
 & \multicolumn{1}{l|}{} & \multicolumn{1}{c|}{0} & \multicolumn{1}{c|}{png (lossless)} & None & 0.411 & 0.070 & 25.0 & 0.101 \\ \cline{3-9} 
 & \multicolumn{1}{l|}{} & \multicolumn{1}{c|}{1000} & \multicolumn{1}{c|}{png (lossless)} & CA* & 0.514 & 0.078 & 23.9 & 0.128 \\ \hline \hline
\multirow{4}{*}{HSCNN+} & \multicolumn{1}{c|}{NTIRE 2022} & \multicolumn{1}{c|}{unknown} & \multicolumn{1}{c|}{jpg (Q unknown)} & None & 0.391 & 0.067 & 25.5 & 0.105 \\ \cline{2-9} 
 & \multicolumn{1}{l|}{\multirow{3}{*}{Synthesized}} & \multicolumn{1}{c|}{0} & \multicolumn{1}{c|}{jpg (Q = 65)} & None & 0.490 & 0.073 & 24.5 & 0.113 \\ \cline{3-9} 
 & \multicolumn{1}{l|}{} & \multicolumn{1}{c|}{0} & \multicolumn{1}{c|}{png (lossless)} & None & 0.485 & 0.080 & 23.9 & 0.101 \\ \cline{3-9} 
 & \multicolumn{1}{l|}{} & \multicolumn{1}{c|}{1000} & \multicolumn{1}{c|}{png (lossless)} & CA* & 0.508 & 0.075 & 24.4 & 0.148 \\ \hline
\end{tabular}
\flushleft{\footnotesize{\quad \quad \quad \quad \quad *CA: chromatic aberration, from a patent double Gauss lens (US20210263286A1).}}
\label{tab:supp_validation_ARAD1K1}
\end{table*}

\begin{table*}[!htp]
\caption{Evaluation of pre-trained models on synthesized validation data generated from the ARAD1K dataset (Part II).}
\centering
\begin{tabular}{l|lccc|cccc}
\hline
\multirow{2}{*}{Network} & \multicolumn{4}{c|}{Data property} & \multirow{2}{*}{MRAE $\downarrow$} & \multirow{2}{*}{RMSE $\downarrow$} & \multirow{2}{*}{PSNR $\uparrow$} & \multirow{2}{*}{SAM $\downarrow$} \\ \cline{2-5}
 & \multicolumn{1}{c|}{Data source} & \multicolumn{1}{c|}{Noise (npe)} & \multicolumn{1}{c|}{RGB format} & Aberration &  &  &  &  \\ \hline
 \multirow{4}{*}{HySAT} & \multicolumn{1}{c|}{NTIRE 2022} & \multicolumn{1}{c|}{unknown} & \multicolumn{1}{c|}{jpg (Q unknown)} & None & 0.176 & 0.028 & 34.6 & 0.085 \\ \cline{2-9} 
 & \multicolumn{1}{l|}{\multirow{3}{*}{Synthesized}} & \multicolumn{1}{c|}{0} & \multicolumn{1}{c|}{jpg (Q = 65)} & None & 0.438 & 0.047 & 29.5 & 0.087 \\ \cline{3-9} 
 & \multicolumn{1}{l|}{} & \multicolumn{1}{c|}{0} & \multicolumn{1}{c|}{png (lossless)} & None & 0.355 & 0.055 & 29.1 & 0.083 \\ \cline{3-9} 
 & \multicolumn{1}{l|}{} & \multicolumn{1}{c|}{1000} & \multicolumn{1}{c|}{png (lossless)} & CA* & 0.326 & 0.047 & 29.4 & 0.127 \\ \hline \hline
 \multirow{4}{*}{HPRN} & \multicolumn{1}{c|}{NTIRE 2022} & \multicolumn{1}{c|}{unknown} & \multicolumn{1}{c|}{jpg (Q unknown)} & None & 0.257 & 0.044 & 30.4 & 0.098 \\ \cline{2-9} 
 & \multicolumn{1}{l|}{\multirow{3}{*}{Synthesized}} & \multicolumn{1}{c|}{0} & \multicolumn{1}{c|}{jpg (Q = 65)} & None & 0.456 & 0.056 & 28.1 & 0.106 \\ \cline{3-9} 
 & \multicolumn{1}{l|}{} & \multicolumn{1}{c|}{0} & \multicolumn{1}{c|}{png (lossless)} & None & 0.383 & 0.067 & 27.1 & 0.102 \\ \cline{3-9} 
 & \multicolumn{1}{l|}{} & \multicolumn{1}{c|}{1000} & \multicolumn{1}{c|}{png (lossless)} & CA* & 0.524 & 0.104 & 22.0 & 0.130 \\ \hline \hline 
 \multirow{4}{*}{SSTHyper} & \multicolumn{1}{c|}{NTIRE 2022} & \multicolumn{1}{c|}{unknown} & \multicolumn{1}{c|}{jpg (Q unknown)} & None & 0.181 & 0.030 & 33.6 & 0.083 \\ \cline{2-9} 
 & \multicolumn{1}{l|}{\multirow{3}{*}{Synthesized}} & \multicolumn{1}{c|}{0} & \multicolumn{1}{c|}{jpg (Q = 65)} & None & 0.439 & 0.047 & 29.7 & 0.088 \\ \cline{3-9} 
 & \multicolumn{1}{l|}{} & \multicolumn{1}{c|}{0} & \multicolumn{1}{c|}{png (lossless)} & None & 0.367 & 0.055 & 29.0 & 0.082 \\ \cline{3-9} 
 & \multicolumn{1}{l|}{} & \multicolumn{1}{c|}{1000} & \multicolumn{1}{c|}{png (lossless)} & CA* & 0.314 & 0.058 & 27.7 & 0.117 \\ \hline \hline
 \multirow{4}{*}{MSFN} & \multicolumn{1}{c|}{NTIRE 2022} & \multicolumn{1}{c|}{unknown} & \multicolumn{1}{c|}{jpg (Q unknown)} & None & 0.226 & 0.038 & 31.9 & 0.084 \\ \cline{2-9} 
 & \multicolumn{1}{l|}{\multirow{3}{*}{Synthesized}} & \multicolumn{1}{c|}{0} & \multicolumn{1}{c|}{jpg (Q = 65)} & None & 0.366 & 0.043 & 30.2 & 0.090 \\ \cline{3-9} 
 & \multicolumn{1}{l|}{} & \multicolumn{1}{c|}{0} & \multicolumn{1}{c|}{png (lossless)} & None & 0.360 & 0.056 & 28.6 & 0.085 \\ \cline{3-9} 
 & \multicolumn{1}{l|}{} & \multicolumn{1}{c|}{1000} & \multicolumn{1}{c|}{png (lossless)} & CA* & 0.328 & 0.055 & 28.3 & 0.119 \\ \hline \hline
 \multirow{4}{*}{GMSR} & \multicolumn{1}{c|}{NTIRE 2022} & \multicolumn{1}{c|}{unknown} & \multicolumn{1}{c|}{jpg (Q unknown)} & None & 0.308 & 0.056 & 27.5 & 0.113 \\ \cline{2-9} 
 & \multicolumn{1}{l|}{\multirow{3}{*}{Synthesized}} & \multicolumn{1}{c|}{0} & \multicolumn{1}{c|}{jpg (Q = 65)} & None & 0.366 & 0.043 & 30.2 & 0.090 \\ \cline{3-9} 
 & \multicolumn{1}{l|}{} & \multicolumn{1}{c|}{0} & \multicolumn{1}{c|}{png (lossless)} & None & 0.342 & 0.062 & 26.9 & 0.110 \\ \cline{3-9} 
 & \multicolumn{1}{l|}{} & \multicolumn{1}{c|}{1000} & \multicolumn{1}{c|}{png (lossless)} & CA* & 0.484 & 0.075 & 24.9 & 0.138 \\ \hline \hline
 \multirow{4}{*}{SSRNet} & \multicolumn{1}{c|}{NTIRE 2022} & \multicolumn{1}{c|}{unknown} & \multicolumn{1}{c|}{jpg (Q unknown)} & None & 0.270 & 0.048 & 29.5 & 0.097 \\ \cline{2-9} 
 & \multicolumn{1}{l|}{\multirow{3}{*}{Synthesized}} & \multicolumn{1}{c|}{0} & \multicolumn{1}{c|}{jpg (Q = 65)} & None & 0.301 & 0.051 & 29.1 & 0.103 \\ \cline{3-9} 
 & \multicolumn{1}{l|}{} & \multicolumn{1}{c|}{0} & \multicolumn{1}{c|}{png (lossless)} & None & 0.356 & 0.058 & 27.8 & 0.096 \\ \cline{3-9} 
 & \multicolumn{1}{l|}{} & \multicolumn{1}{c|}{1000} & \multicolumn{1}{c|}{png (lossless)} & CA* & 0.419 & 0.075 & 25.8 & 0.130 \\ \hline \hline
\end{tabular}
\flushleft{\footnotesize{\quad \quad \quad \quad \quad *CA: chromatic aberration, from a patent double Gauss lens (US20210263286A1).}}
\label{tab:supp_validation_ARAD1K2}
\end{table*}

\section{Results of Cross-Dataset Validation for Other Networks}

As shown in Table~4 in the main paper, we demonstrate that the MST++ network has difficulties in keeping high performance when it is trained on one dataset and validated on another dataset. In Supplemental Table~\ref{tab:supp_cross_validation1} and Supplemental Table~\ref{tab:supp_cross_validation2}, we show with extended experimental results that the effects of cross-dataset validation are true for all other networks as well. Similar cross-dataset failure can be observed for all the candidate networks. These results consistently support our conclusion about the important roles of scene content and acquisition devices in different datasets.

\begin{table*}[!htp]
\caption{Cross-dataset validation for all networks (Part I).}
\centering
\begin{tabular}{ll|cccc|cccc}
\hline
% MST++ and MST-L
 &  & \multicolumn{4}{c|}{MST++\cite{cai2022mst++}} & \multicolumn{4}{c}{MST-L\cite{cai2022mask}} \\ \hline
\multicolumn{1}{l|}{Trained on} & Validated on & MRAE$\downarrow$   & RMSE$\downarrow$ & PSNR$\uparrow$ & SAM$\downarrow$ & MRAE$\downarrow$   & RMSE$\downarrow$ & PSNR$\uparrow$ & SAM$\downarrow$\\ \hline
\multicolumn{1}{l|}{CAVE}     & CAVE     &0.237 &0.034 &31.9  &0.194 &0.234 &0.031 &32.0 &0.187 \\ \hline
\multicolumn{1}{l|}{ARAD1K}   & CAVE     &1.626 &0.074 &24.4  &0.376 &2.055 &0.070 &25.3 &0.367 \\ \hline \hline
\multicolumn{1}{l|}{ICVL}     & ICVL     &0.079 &0.019 &38.3  &0.024 &0.063 &0.015 &41.1 &0.023 \\ \hline
\multicolumn{1}{l|}{ARAD1K}   & ICVL     &1.032 &0.188 &19.3  &0.924 &0.349 &0.052 &27.8 &0.100 \\ \hline \hline
\multicolumn{1}{l|}{KAUST}    & KAUST    &0.069 &0.013 &44.4  &0.061 &0.082 &0.016 &43.8 &0.070 \\ \hline
\multicolumn{1}{l|}{ARAD1K}   & KAUST    &1.042 &0.100 &22.0  &0.370 &1.114 &0.115 &21.8 &0.370 \\ \hline
% MPRNet and Restormer
 &  & \multicolumn{4}{c|}{MPRNet\cite{zamir2021multi}} & \multicolumn{4}{c}{Restormer\cite{zamir2022restormer}} \\ \hline
 \multicolumn{1}{l|}{Trained on} & Validated on & MRAE$\downarrow$   & RMSE$\downarrow$ & PSNR$\uparrow$ & SAM$\downarrow$ & MRAE$\downarrow$   & RMSE$\downarrow$ & PSNR$\uparrow$ & SAM$\downarrow$\\ \hline
\multicolumn{1}{l|}{CAVE}     & CAVE     &0.295 &0.045 &29.4 &0.173 &0.246 &0.036 &30.0 &0.177 \\ \hline
\multicolumn{1}{l|}{ARAD1K}   & CAVE     &2.063 &0.060 &26.4 &0.378 &1.689 &0.060 &26.5 &0.375 \\ \hline \hline
\multicolumn{1}{l|}{ICVL}     & ICVL     &0.077 &0.018 &39.9 &0.024 &0.084 &0.020 &37.4 &0.026 \\ \hline
\multicolumn{1}{l|}{ARAD1K}   & ICVL     &0.349 &0.050 &27.5 &0.100 &0.347 &0.052 &27.4 &0.097 \\ \hline \hline
\multicolumn{1}{l|}{KAUST}    & KAUST    &0.170 &0.022 &35.8 &0.071 &0.067 &0.014 &44.9 &0.066 \\ \hline
\multicolumn{1}{l|}{ARAD1K}   & KAUST    &0.885 &0.101 &23.1 &0.350 &1.496 &0.144 &20.0 &0.363 \\ \hline
% MIRNet and HINet
 &  & \multicolumn{4}{c|}{MIRNet\cite{zamir2020learning}} & \multicolumn{4}{c}{HINet\cite{chen2021hinet}} \\ \hline
 \multicolumn{1}{l|}{Trained on} & Validated on & MRAE$\downarrow$   & RMSE$\downarrow$ & PSNR$\uparrow$ & SAM$\downarrow$ & MRAE$\downarrow$   & RMSE$\downarrow$ & PSNR$\uparrow$ & SAM$\downarrow$\\ \hline
\multicolumn{1}{l|}{CAVE}     & CAVE     &0.214 &0.027 &33.6 &0.177 &0.283 &0.041 &29.4 &0.191 \\ \hline
\multicolumn{1}{l|}{ARAD1K}   & CAVE     &2.039 &0.075 &24.9 &0.406 &1.512 &0.084 &23.3 &0.393 \\ \hline \hline
\multicolumn{1}{l|}{ICVL}     & ICVL     &0.060 &0.013 &40.8 &0.023 &0.087 &0.021 &36.2 &0.028 \\ \hline
\multicolumn{1}{l|}{ARAD1K}   & ICVL     &0.365 &0.052 &27.5 &0.114 &0.387 &0.058 &26.2 &0.127 \\ \hline \hline
\multicolumn{1}{l|}{KAUST}    & KAUST    &0.079 &0.015 &42.9 &0.070 &0.089 &0.017 &42.2 &0.074 \\ \hline
\multicolumn{1}{l|}{ARAD1K}   & KAUST    &1.642 &0.154 &19.4 &0.357 &1.393 &0.140 &19.8 &0.362 \\ \hline
 % HDNet and AWAN
 &  & \multicolumn{4}{c|}{HDNet\cite{hu2022hdnet}} & \multicolumn{4}{c}{AWAN\cite{li2020adaptive}} \\ \hline
 \multicolumn{1}{l|}{Trained on} & Validated on & MRAE$\downarrow$   & RMSE$\downarrow$ & PSNR$\uparrow$ & SAM$\downarrow$ & MRAE$\downarrow$   & RMSE$\downarrow$ & PSNR$\uparrow$ & SAM$\downarrow$\\ \hline
\multicolumn{1}{l|}{CAVE}     & CAVE     &0.266 &0.041 &29.1 &0.199 &0.305 &0.057 &28.0 &0.237 \\ \hline
\multicolumn{1}{l|}{ARAD1K}   & CAVE     &1.564 &0.078 &23.9 &0.406 &1.743 &0.077 &24.7 &0.397 \\ \hline \hline
\multicolumn{1}{l|}{ICVL}     & ICVL     &0.076 &0.018 &37.4 &0.028 &0.083 &0.018 &38.6 &0.026 \\ \hline
\multicolumn{1}{l|}{ARAD1K}   & ICVL     &0.598 &0.085 &22.3 &0.129 &0.408 &0.060 &26.5 &0.108 \\ \hline \hline
\multicolumn{1}{l|}{KAUST}    & KAUST    &0.076 &0.015 &42.0 &0.070 &0.083 &0.015 &41.0 &0.063 \\ \hline
\multicolumn{1}{l|}{ARAD1K}   & KAUST    &1.389 &0.142 &19.6 &0.394 &1.844 &0.174 &18.7 &0.370 \\ \hline
 % EDSR and HRNet
 &  & \multicolumn{4}{c|}{EDSR\cite{lim2017enhanced}} & \multicolumn{4}{c}{HRNet\cite{zhao2020hierarchical}} \\ \hline
 \multicolumn{1}{l|}{Trained on} & Validated on & MRAE$\downarrow$   & RMSE$\downarrow$ & PSNR$\uparrow$ & SAM$\downarrow$ & MRAE$\downarrow$   & RMSE$\downarrow$ & PSNR$\uparrow$ & SAM$\downarrow$\\ \hline
\multicolumn{1}{l|}{CAVE}     & CAVE     &0.308 &0.058 &26.1 &0.194 &0.317 &0.058 &26.7 &0.196 \\ \hline
\multicolumn{1}{l|}{ARAD1K}   & CAVE     &1.757 &0.078 &23.4 &0.416 &1.170 &0.076 &23.8 &0.389 \\ \hline \hline
\multicolumn{1}{l|}{ICVL}     & ICVL     &0.111 &0.030 &33.0 &0.027 &0.103 &0.026 &33.8 &0.028 \\ \hline
\multicolumn{1}{l|}{ARAD1K}   & ICVL     &0.474 &0.067 &24.2 &0.119 &0.531 &0.075 &23.3 &0.115 \\ \hline \hline
\multicolumn{1}{l|}{KAUST}    & KAUST    &0.215 &0.037 &31.6 &0.081 &0.094 &0.020 &39.2 &0.069 \\ \hline
\multicolumn{1}{l|}{ARAD1K}   & KAUST    &1.391 &0.145 &19.4 &0.408 &1.202 &0.127 &20.6 &0.412 \\ \hline
% HSCNN+
 &  & \multicolumn{4}{c|}{HSCNN+\cite{shi2018hscnn+}} & \multicolumn{4}{c}{}\\ \cline{1-6}
 \multicolumn{1}{l|}{Trained on} & Validated on & MRAE$\downarrow$   & RMSE$\downarrow$ & PSNR$\uparrow$ & SAM$\downarrow$ & & & & \\ \cline{1-6}
\multicolumn{1}{l|}{CAVE}     & CAVE     &0.328 &0.067 &25.4 &0.222 & & & & \\ \cline{1-6}
\multicolumn{1}{l|}{ARAD1K}   & CAVE     &2.522 &0.098 &21.1 &0.419 & & & & \\ \cline{1-6} \cline{1-6}
\multicolumn{1}{l|}{ICVL}     & ICVL     &0.223 &0.042 &28.9 &0.029 & & & & \\ \cline{1-6}
\multicolumn{1}{l|}{ARAD1K}   & ICVL     &0.528 &0.074 &23.3 &0.117 & & & & \\ \cline{1-6} \cline{1-6}
\multicolumn{1}{l|}{KAUST}    & KAUST    &2.093 &0.192 &19.5 &0.075 & & & & \\ \cline{1-6}
\multicolumn{1}{l|}{ARAD1K}   & KAUST    &1.281 &0.135 &19.9 &0.351 & & & & \\ \hline
\end{tabular}
\label{tab:supp_cross_validation1}
\end{table*}

\begin{table*}[!htp]
\caption{Cross-dataset validation for all networks (Part II).}
\centering
\begin{tabular}{ll|cccc|cccc}
\hline
% HySAT and HPRN
 &  & \multicolumn{4}{c|}{HySAT} & \multicolumn{4}{c}{HPRN} \\ \hline
\multicolumn{1}{l|}{Trained on} & Validated on & MRAE$\downarrow$   & RMSE$\downarrow$ & PSNR$\uparrow$ & SAM$\downarrow$ & MRAE$\downarrow$   & RMSE$\downarrow$ & PSNR$\uparrow$ & SAM$\downarrow$\\ \hline
\multicolumn{1}{l|}{CAVE}     & CAVE     &0.286 &0.043 &28.7  &0.202 &0.265 &0.039 &30.4 &0.181 \\ \hline
\multicolumn{1}{l|}{ARAD1K}   & CAVE     &1.250 &0.088 &23.7  &0.376 &1.057 &0.111 &21.0 &0.424 \\ \hline \hline
\multicolumn{1}{l|}{ICVL}     & ICVL     &0.071 &0.017 &39.1  &0.026 &0.089 &0.019 &39.6 &0.026 \\ \hline
\multicolumn{1}{l|}{ARAD1K}   & ICVL     &0.428 &0.054 &26.6  &0.114 &0.418 &0.056 &26.6 &0.130 \\ \hline \hline
\multicolumn{1}{l|}{KAUST}    & KAUST    &0.066 &0.013 &44.9  &0.058 &0.078 &0.015 &42.5 &0.067 \\ \hline
\multicolumn{1}{l|}{ARAD1K}   & KAUST    &1.057 &0.107 &21.8  &0.366 &1.357 &0.144 &19.8 &0.370 \\ \hline
% SSTHyper and MSFN
 &  & \multicolumn{4}{c|}{SSTHyper} & \multicolumn{4}{c}{MSFN} \\ \hline
 \multicolumn{1}{l|}{Trained on} & Validated on & MRAE$\downarrow$   & RMSE$\downarrow$ & PSNR$\uparrow$ & SAM$\downarrow$ & MRAE$\downarrow$   & RMSE$\downarrow$ & PSNR$\uparrow$ & SAM$\downarrow$\\ \hline
\multicolumn{1}{l|}{CAVE}     & CAVE     &0.232 &0.035 &31.2 &0.176 &0.24 &0.035 &31.3 &0.191 \\ \hline
\multicolumn{1}{l|}{ARAD1K}   & CAVE     &1.036 &0.091 &22.9 &0.399 &0.948 &0.093 &22.9 &0.387 \\ \hline \hline
\multicolumn{1}{l|}{ICVL}     & ICVL     &0.056 &0.011 &41.7 &0.022 &0.074 &0.017 &39.5 &0.023 \\ \hline
\multicolumn{1}{l|}{ARAD1K}   & ICVL     &0.430 &0.054 &26.5 &0.112 &0.397 &0.053 &27.8 &0.124 \\ \hline \hline
\multicolumn{1}{l|}{KAUST}    & KAUST    &0.079 &0.015 &43.7 &0.064 &0.076 &0.014 &44.7 &0.064 \\ \hline
\multicolumn{1}{l|}{ARAD1K}   & KAUST    &1.052 &0.106 &22.2 &0.400 &1.213 &0.122 &20.9 &0.383 \\ \hline
% GMSR and SSRNet
 &  & \multicolumn{4}{c|}{GMSR} & \multicolumn{4}{c}{SSRNet} \\ \hline
 \multicolumn{1}{l|}{Trained on} & Validated on & MRAE$\downarrow$   & RMSE$\downarrow$ & PSNR$\uparrow$ & SAM$\downarrow$ & MRAE$\downarrow$   & RMSE$\downarrow$ & PSNR$\uparrow$ & SAM$\downarrow$\\ \hline
\multicolumn{1}{l|}{CAVE}     & CAVE     &0.309 &0.052 &26.9 &0.199 &0.250 &0.032 &31.9 &0.195 \\ \hline
\multicolumn{1}{l|}{ARAD1K}   & CAVE     &0.971 &0.120 &19.5 &0.441 &1.097 &0.072 &25.8 &0.394 \\ \hline \hline
\multicolumn{1}{l|}{ICVL}     & ICVL     &0.083 &0.022 &36.0 &0.033 &0.102 &0.024 &37.1 &0.030 \\ \hline
\multicolumn{1}{l|}{ARAD1K}   & ICVL     &0.574 &0.079 &23.1 &0.125 &0.478 &0.064 &25.3 &0.149 \\ \hline \hline
\multicolumn{1}{l|}{KAUST}    & KAUST    &0.090 &0.018 &39.3 &0.083 &0.083 &0.016 &42.0 &0.082 \\ \hline
\multicolumn{1}{l|}{ARAD1K}   & KAUST    &0.883 &0.087 &23.0 &0.357 &1.039 &0.106 &21.6 &0.378 \\ \hline
\end{tabular}
\label{tab:supp_cross_validation2}
\end{table*}

\section{Results of Metamer Failure for Other Datasets}
In Table~6 of the main paper, we compare the performance of the candidate networks for the standard data (no metamers), fixed metamers ($\alpha = 0$), and on-the-fly metamers ($\alpha$ varies in the range of [-1, 2] during training) synthesized from the ARAD1K dataset. All the metrics degrade significantly in the presence of metamers. In Supplemental Table~\ref{tab:supp_train_metamers_other_datasets1}, Table~\ref{tab:supp_train_metamers_other_datasets2} and Table~\ref{tab:supp_train_metamers_other_datasets3}, we further show that this is also true for all the networks on the CAVE~\cite{yasuma2010generalized}, ICVL~\cite{arad2016sparse}, and KAUST~\cite{li2021multispectral} datasets. All the results consistently support our conclusion that existing methods cannot distinguish metamers, regardless of the network architectures and datasets.

\begin{table*}[!htp]
\caption{Training with metamers on the CAVE, ICVL, and KAUST datasets. Part I: {MST++}, {MST-L}, MPRNet, Restormer, MIRNet, and HINet.}
\centering
\begin{tabular}{ll|cccc|cccc}
\hline
% MST++ and MST-L
 &  & \multicolumn{4}{c|}{MST++\cite{cai2022mst++}} & \multicolumn{4}{c}{MST-L\cite{cai2022mask}} \\ \hline
Dataset & Metamer & MRAE$\downarrow$ & RMSE$\downarrow$ & PSNR$\uparrow$ & SAM$\downarrow$ & MRAE$\downarrow$ & RMSE$\downarrow$ & PSNR$\uparrow$ & SAM$\downarrow$ \\ \hline
\multirow{3}{*}{CAVE\cite{yasuma2010generalized}}& no &1.014 &0.038  &29.9  &0.192 &0.932 &0.057 &26.1 &0.195 \\ \cline{2-10} 
                                        & fixed &38.26 &0.053  &29.6  &0.229 &66.470 &0.062 &27.6 &0.295 \\ \cline{2-10} 
                                        & on-the-fly &226.0 &0.078  &25.2  &0.451 &286.557 &0.085 &24.7 &0.504  \\ \hline
\multirow{3}{*}{ICVL\cite{arad2016sparse}} & no &0.067  &0.016  &40.1  &0.027 &0.067 &0.015 &39.8 &0.025 \\ \cline{2-10} 
                                        & fixed &1.454  &0.041  &34.8 &0.229  &2.080 &0.040 &34.5 &0.28 \\ \cline{2-10} 
                                        & on-the-fly &2.615  &0.087  &24.3  &0.268 &3.281 &0.087 &23.7 &0.261 \\ \hline
\multirow{3}{*}{KAUST\cite{li2021multispectral}} & no &0.082  &0.016  &43.2  &0.076 &0.097 &0.017 &42.0 &0.074 \\ \cline{2-10} 
                                        & fixed &2.033  &0.022  &39.0 &0.217 &1.920 &0.022 &38.8 &0.219  \\ \cline{2-10} 
                                        & on-the-fly &1.874  &0.032  &33.7  &0.245 &4.235 &0.023 &39.8 &0.236 \\ \hline
% MPRNet and Restormer
 &  & \multicolumn{4}{c|}{MPRNet\cite{zamir2021multi}} & \multicolumn{4}{c}{Restormer\cite{zamir2022restormer}} \\ \hline
Dataset & Metamer & MRAE$\downarrow$ & RMSE$\downarrow$ & PSNR$\uparrow$ & SAM$\downarrow$ & MRAE$\downarrow$ & RMSE$\downarrow$ & PSNR$\uparrow$ & SAM$\downarrow$ \\ \hline
\multirow{3}{*}{CAVE\cite{yasuma2010generalized}}& no &1.110 &0.041 &30.3 &0.177 &0.987 &0.040 &29.5 &0.175 \\ \cline{2-10} 
                                        & fixed &34.272 &0.046 &32.3 &0.209 &93.697 &0.047 &34.7 &0.276 \\ \cline{2-10} 
                                        & on-the-fly &355.958 &0.112 &22.8 &0.626 &379.277 &0.093 &25.7 &0.501  \\ \hline
\multirow{3}{*}{ICVL\cite{arad2016sparse}} & no &0.084 &0.019 &38.2 &0.025 &0.083 &0.020 &38.2 &0.026 \\ \cline{2-10} 
                                        & fixed &1.693 &0.041 &33.4 &0.228 &1.730 &0.040 &34.7 &0.228 \\ \cline{2-10} 
                                        & on-the-fly &2.584 &0.094 &22.9 &0.259 &2.254 &0.099 &22.7 &0.272 \\ \hline
\multirow{3}{*}{KAUST\cite{li2021multispectral}} & no &0.071 &0.013 &43.6 &0.066 &0.063 &0.013 &44.6 &0.063 \\ \cline{2-10} 
                                        & fixed &2.668 &0.024 &35.7 &0.216 &2.077 &0.020 &40.3 &0.213  \\ \cline{2-10} 
                                        & on-the-fly &2.431 &0.037 &32.2 &0.251 &2.623 &0.032 &34.1 &0.281 \\ \hline
% MIRNet and HINet
 &  & \multicolumn{4}{c|}{MIRNet\cite{zamir2020learning}} & \multicolumn{4}{c}{HINet\cite{chen2021hinet}} \\ \hline
Dataset & Metamer & MRAE$\downarrow$ & RMSE$\downarrow$ & PSNR$\uparrow$ & SAM$\downarrow$ & MRAE$\downarrow$ & RMSE$\downarrow$ & PSNR$\uparrow$ & SAM$\downarrow$ \\ \hline
\multirow{3}{*}{CAVE\cite{yasuma2010generalized}}& no &1.167 &0.035 &31.5 &0.190 &1.064 &0.050 &27.3 &0.196 \\ \cline{2-10} 
                                        & fixed &29.483 &0.044 &31.7 &0.212 &55.259 &0.056 &29.2 &0.223 \\ \cline{2-10} 
                                        & on-the-fly &131.464 &0.083 &25.0 &0.405 &141.691 &0.072 &27.5 &0.362  \\ \hline
\multirow{3}{*}{ICVL\cite{arad2016sparse}} & no &0.070 &0.015 &39.7 &0.025 &0.071 &0.017 &38.9 &0.027 \\ \cline{2-10} 
                                        & fixed &3.794 &0.038 &35.0 &0.227 &2.445 &0.041 &34.8 &0.228 \\ \cline{2-10} 
                                        & on-the-fly &2.895 &0.095 &23.0 &0.265 &3.600 &0.092 &23.4 &0.271 \\ \hline
\multirow{3}{*}{KAUST\cite{li2021multispectral}} & no &0.078 &0.015 &42.2 &0.069 &0.097 &0.017 &41.9 &0.080 \\ \cline{2-10} 
                                        & fixed &2.072 &0.021 &38.7 &0.216 &2.481 &0.023 &37.5 &0.218  \\ \cline{2-10} 
                                        & on-the-fly &2.312 &0.037 &33.1 &0.257 &4.323 &0.023 &38.9 &0.229 \\ \hline
\end{tabular}
\label{tab:supp_train_metamers_other_datasets1}
\end{table*}

\begin{table*}[!htp]
\caption{Training with metamers on the CAVE, ICVL, and KAUST datasets. Part II: HDNet, AWAN, MIRNet, HINet, and HSCNN+.}
\centering
% \small
\begin{tabular}{ll|cccc|cccc}
\hline
% HDNet and AWAN
 &  & \multicolumn{4}{c|}{HDNet\cite{hu2022hdnet}} & \multicolumn{4}{c}{AWAN\cite{li2020adaptive}} \\ \hline
Dataset & Metamer & MRAE$\downarrow$ & RMSE$\downarrow$ & PSNR$\uparrow$ & SAM$\downarrow$ & MRAE$\downarrow$ & RMSE$\downarrow$ & PSNR$\uparrow$ & SAM$\downarrow$ \\ \hline
\multirow{3}{*}{CAVE\cite{yasuma2010generalized}}& no &1.071 &0.043 &29.4 &0.197 &1.045 &0.069 &24.7 &0.208 \\ \cline{2-10} 
                                        & fixed &56.996 &0.071 &26.8 &0.298 &88.236 &0.072 &28.8 &0.291 \\ \cline{2-10} 
                                        & on-the-fly &78.524 &0.093 &23.8 &0.274 &259.101 &0.079 &26.7 &0.366  \\ \hline
\multirow{3}{*}{ICVL\cite{arad2016sparse}} & no &0.076 &0.019 &37.2 &0.027 &0.100 &0.020 &37.9 &0.028 \\ \cline{2-10} 
                                        & fixed &2.786 &0.039 &34.0 &0.226 &3.034 &0.040 &34.3 &0.230 \\ \cline{2-10} 
                                        & on-the-fly &2.891 &0.091 &23.3 &0.259 &2.707 &0.094 &22.8 &0.280 \\ \hline
\multirow{3}{*}{KAUST\cite{li2021multispectral}} & no &0.085 &0.017 &40.8 &0.082 &0.101 &0.017 &39.6 &0.105 \\ \cline{2-10} 
                                        & fixed &2.891 &0.022 &37.9 &0.217 &2.152 &0.021 &38.3 &0.221  \\ \cline{2-10} 
                                        & on-the-fly &3.789 &0.023 &38.9 &0.233 &3.855 &0.024 &38.2 &0.233 \\ \hline
% EDSR and HRNet
 &  & \multicolumn{4}{c|}{EDSR\cite{lim2017enhanced}} & \multicolumn{4}{c}{HRNet\cite{zhao2020hierarchical}} \\ \hline
Dataset & Metamer & MRAE$\downarrow$ & RMSE$\downarrow$ & PSNR$\uparrow$ & SAM$\downarrow$ & MRAE$\downarrow$ & RMSE$\downarrow$ & PSNR$\uparrow$ & SAM$\downarrow$ \\ \hline
\multirow{3}{*}{CAVE\cite{yasuma2010generalized}}& no &1.207 &0.057 &26.3 &0.202 &1.087 &0.056 &26.6 &0.198 \\ \cline{2-10} 
                                        & fixed &54.125 &0.066 &27.0 &0.292 &106.082 &0.065 &27.3 &0.310 \\ \cline{2-10} 
                                        & on-the-fly &199.916 &0.106 &21.2 &0.371 &151.295 &0.100 &21.8 &0.356  \\ \hline
\multirow{3}{*}{ICVL\cite{arad2016sparse}} & no &0.112 &0.029 &32.9 &0.028 &0.106 &0.027 &33.3 &0.029 \\ \cline{2-10} 
                                        & fixed &2.275 &0.045 &30.5 &0.227 &2.182 &0.045 &30.8 &0.228 \\ \cline{2-10} 
                                        & on-the-fly &2.721 &0.093 &23.0 &0.262 &2.703 &0.083 &24.0 &0.244 \\ \hline
\multirow{3}{*}{KAUST\cite{li2021multispectral}} & no &0.260 &0.044 &30.5 &0.085 &0.097 &0.021 &37.8 &0.070 \\ \cline{2-10} 
                                        & fixed &2.002 &0.033 &32.2 &0.218 &2.819 &0.024 &37.5 &0.218  \\ \cline{2-10} 
                                        & on-the-fly &3.883 &0.025 &36.1 &0.221 &3.373 &0.026 &36.9 &0.223 \\ \hline
% HSCNN+ 
 &  & \multicolumn{4}{c|}{HSCNN+\cite{shi2018hscnn+}} & \\ \cline{2-6}
Dataset & Metamer & MRAE$\downarrow$ & RMSE$\downarrow$ & PSNR$\uparrow$ & SAM$\downarrow$ & & & & \\ \cline{2-6}
\multirow{3}{*}{CAVE\cite{yasuma2010generalized}}& no &1.027 &0.065 &25.3 &0.229 & & & & \\ \cline{2-6} 
                                        & fixed &55.611 &0.076 &25.1 &0.313 & & & & \\ \cline{2-6} 
                                        & on-the-fly &154.416 &0.104 &21.8 &0.353 & & & &  \\ \cline{2-6}
\multirow{3}{*}{ICVL\cite{arad2016sparse}} & no &0.226 &0.043 &28.6 &0.030 & & & & \\ \cline{2-6} 
                                        & fixed &1.908 &0.052 &28.1 &0.229 & & & & \\ \cline{2-6} 
                                        & on-the-fly &1.627 &0.102 &21.9 &0.253 & & & & \\ \cline{2-6}
\multirow{3}{*}{KAUST\cite{li2021multispectral}} & no &1.832 &0.171 &19.9 &0.077 & & & & \\ \cline{2-6} 
                                        & fixed &6.240 &0.166 &20.6 &0.217 & & & &  \\ \cline{2-6} 
                                        & on-the-fly &2.669 &0.057 &25.9 &0.222 & & & & \\ \hline
\end{tabular}
\label{tab:supp_train_metamers_other_datasets2}
\end{table*}

\begin{table*}[!htp]
\caption{Training with metamers on the CAVE, ICVL, and KAUST datasets. Part III: HySAT, HPRN, SSTHyper, MSFN, GMSR, and SSRNet.}
\centering
\begin{tabular}{ll|cccc|cccc}
\hline
% HySAT and HPRN
 &  & \multicolumn{4}{c|}{HySAT} & \multicolumn{4}{c}{HPRN} \\ \hline
Dataset & Metamer & MRAE$\downarrow$ & RMSE$\downarrow$ & PSNR$\uparrow$ & SAM$\downarrow$ & MRAE$\downarrow$ & RMSE$\downarrow$ & PSNR$\uparrow$ & SAM$\downarrow$ \\ \hline
\multirow{3}{*}{CAVE\cite{yasuma2010generalized}}& no &1.088 &0.035 &31.8 &0.184 &0.900 &0.043 &29.3 &0.196 \\ \cline{2-10} 
                                        & fixed &218.925 &0.065 &28.1 &0.372 &76.216 &0.062 &27.7 &0.314 \\ \cline{2-10} 
                                        & on-the-fly &338.654 &0.087 &25.0 &0.487 &352.391 &0.106 &23.2 &0.512  \\ \hline
\multirow{3}{*}{ICVL\cite{arad2016sparse}} & no &0.080 &0.020 &38.9 &0.026 &0.081 &0.018 &38.4 &0.026 \\ \cline{2-10} 
                                        & fixed &3.100 &0.042 &32.9 &0.228 &2.543 &0.042 &34.1 &0.232 \\ \cline{2-10} 
                                        & on-the-fly &3.044 &0.092 &23.1 &0.260 &2.367 &0.098 &22.7 &0.265 \\ \hline
\multirow{3}{*}{KAUST\cite{li2021multispectral}} & no &0.070 &0.013 &43.7 &0.073 &0.109 &0.020 &40.9 &0.078 \\ \cline{2-10} 
                                        & fixed &2.895 &0.021 &39.1 &0.216 &2.989 &0.025 &37.5 &0.221  \\ \cline{2-10} 
                                        & on-the-fly &4.553 &0.026 &36.9 &0.241 &3.836 &0.025 &38.3 &0.246 \\ \hline
% SSTHyper and MSFN
 &  & \multicolumn{4}{c|}{SSTHyper} & \multicolumn{4}{c}{MSFN} \\ \hline
Dataset & Metamer & MRAE$\downarrow$ & RMSE$\downarrow$ & PSNR$\uparrow$ & SAM$\downarrow$ & MRAE$\downarrow$ & RMSE$\downarrow$ & PSNR$\uparrow$ & SAM$\downarrow$ \\ \hline
\multirow{3}{*}{CAVE\cite{yasuma2010generalized}}& no &1.190 &0.032 &32.8 &0.182 &0.863 &0.037 &30.8 &0.189 \\ \cline{2-10} 
                                        & fixed &98.149 &0.066 &28.2 &0.313 &96.891 &0.060 &31.2 &0.303 \\ \cline{2-10} 
                                        & on-the-fly &398.022 &0.097 &24.2 &0.558 &44.497 &0.067 &26.9 &0.224  \\ \hline
\multirow{3}{*}{ICVL\cite{arad2016sparse}} & no &0.058 &0.014 &41.3 &0.025 &0.078 &0.019 &38.1 &0.025 \\ \cline{2-10} 
                                        & fixed &3.080 &0.039 &35.1 &0.225 &2.141 &0.041 &34.3 &0.226 \\ \cline{2-10} 
                                        & on-the-fly &2.160 &0.094 &23.1 &0.256 &2.801 &0.084 &24.3 &0.250 \\ \hline
\multirow{3}{*}{KAUST\cite{li2021multispectral}} & no &0.075 &0.014 &43.6 &0.063 &0.075 &0.014 &43.8 &0.069 \\ \cline{2-10} 
                                        & fixed &1.965 &0.021 &39.5 &0.216 &2.395 &0.021 &39.1 &0.216  \\ \cline{2-10} 
                                        & on-the-fly &4.441 &0.036 &33.8 &0.235 &3.971 &0.021 &39.9 &0.227 \\ \hline
% GMSR and SSRNet
 &  & \multicolumn{4}{c|}{GMSR} & \multicolumn{4}{c}{SSRNet} \\ \hline
Dataset & Metamer & MRAE$\downarrow$ & RMSE$\downarrow$ & PSNR$\uparrow$ & SAM$\downarrow$ & MRAE$\downarrow$ & RMSE$\downarrow$ & PSNR$\uparrow$ & SAM$\downarrow$ \\ \hline
\multirow{3}{*}{CAVE\cite{yasuma2010generalized}}& no &1.302 &0.035 &31.2 &0.193 &1.114 &0.042 &29.8 &0.203 \\ \cline{2-10} 
                                        & fixed &105.385 &0.052 &30.4 &0.328 &117.309 &0.055 &31.0 &0.294 \\ \cline{2-10} 
                                        & on-the-fly &195.193 &0.073 &24.9 &0.319 &56.450 &0.053 &29.0 &0.231  \\ \hline
\multirow{3}{*}{ICVL\cite{arad2016sparse}} & no &0.093 &0.024 &35.0 &0.034 &0.098 &0.022 &37.6 &0.030 \\ \cline{2-10} 
                                        & fixed &1.845 &0.043 &32.3 &0.230 &2.023 &0.040 &34.2 &0.229 \\ \cline{2-10} 
                                        & on-the-fly &2.770 &0.071 &25.9 &0.238 &2.526 &0.075 &25.0 &0.237 \\ \hline
\multirow{3}{*}{KAUST\cite{li2021multispectral}} & no &0.152 &0.025 &36.8 &0.094 &0.114 &0.019 &41.7 &0.085 \\ \cline{2-10} 
                                        & fixed &2.411 &0.024 &35.9 &0.223 &2.155 &0.022 &38.7 &0.216  \\ \cline{2-10} 
                                        & on-the-fly &3.810 &0.023 &38.5 &0.221 &4.099 &0.022 &39.6 &0.221 \\ \hline
\end{tabular}
\label{tab:supp_train_metamers_other_datasets3}
\end{table*}

\section{Results of the Aberration Advantage for Other Networks and Other Datasets}
In Fig.~4 of the main paper, we demonstrate that it is beneficial to incorporate the realistic chromatic aberrations of the optical system into the training pipeline, such that the spectral accuracy can be improved. To prove that this phenomenon is regardless of the network architectures, we perform the same experiment for the other candidate networks on the ARAD1K dataset. The results are shown in Supplemental Fig.~\ref{fig:supp_aberration_advantage1}, Fig.~\ref{fig:supp_aberration_advantage2} and Fig.~\ref{fig:supp_aberration_advantage3}.

We also demonstrate that the aberration advantage applies to other datasets. Since the MST++ network performs consistently among the top-performing architectures, we conduct the same experiments with this network on the CAVE, ICVL, and KAUST datasets. The results are shown in Supplemental Fig.~\ref{fig:supp_aberration_advantage_all_datasets}. All the results clearly support our conclusion that the chromatic aberrations encode spectral information into the RGB images for the networks to effectively learn the embedded spectra.

% aberration advantage
\begin{figure*}[!htp]
    \centering
    \begin{tabular}{c@{}c@{\hspace{5mm}}c@{}c}
    \multicolumn{2}{c}{MST++} & \multicolumn{2}{c}{MST-L} \\
    % MST++ fixed
    \includegraphics[width=0.49\columnwidth]{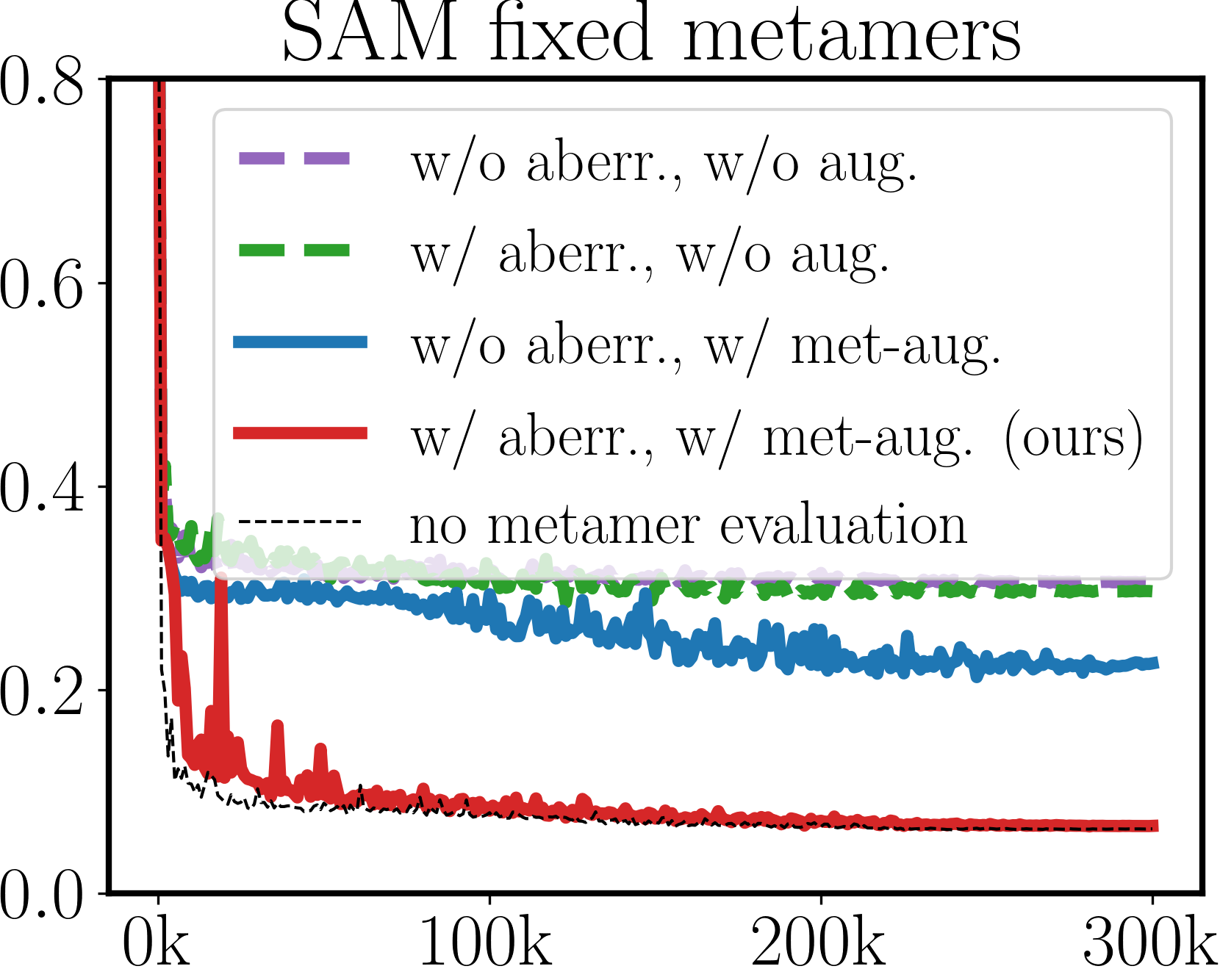} 
    % MST++ on-the-fly
    & \includegraphics[width=0.49\columnwidth]{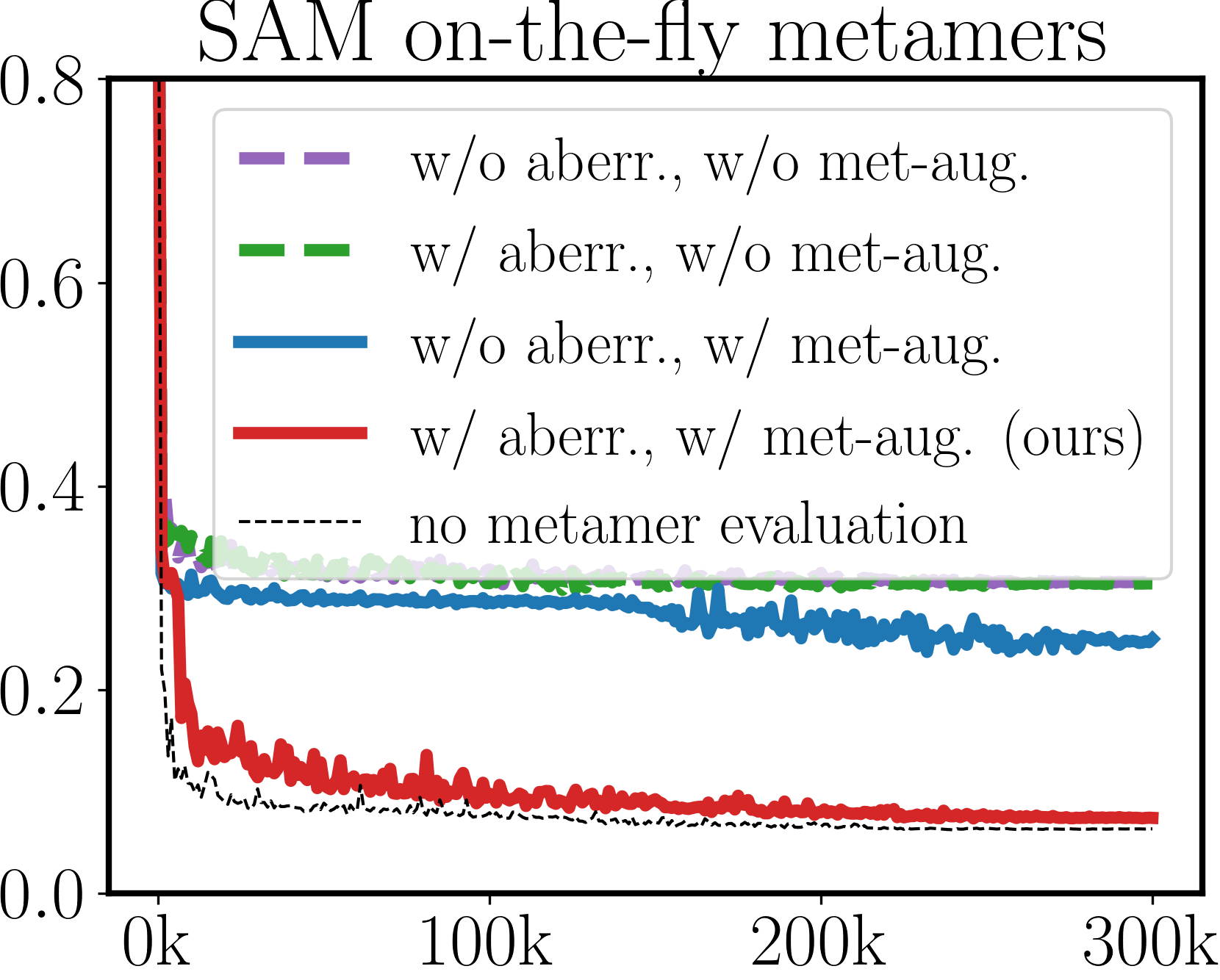}
    % MST fixed
    & \includegraphics[width=0.49\columnwidth]{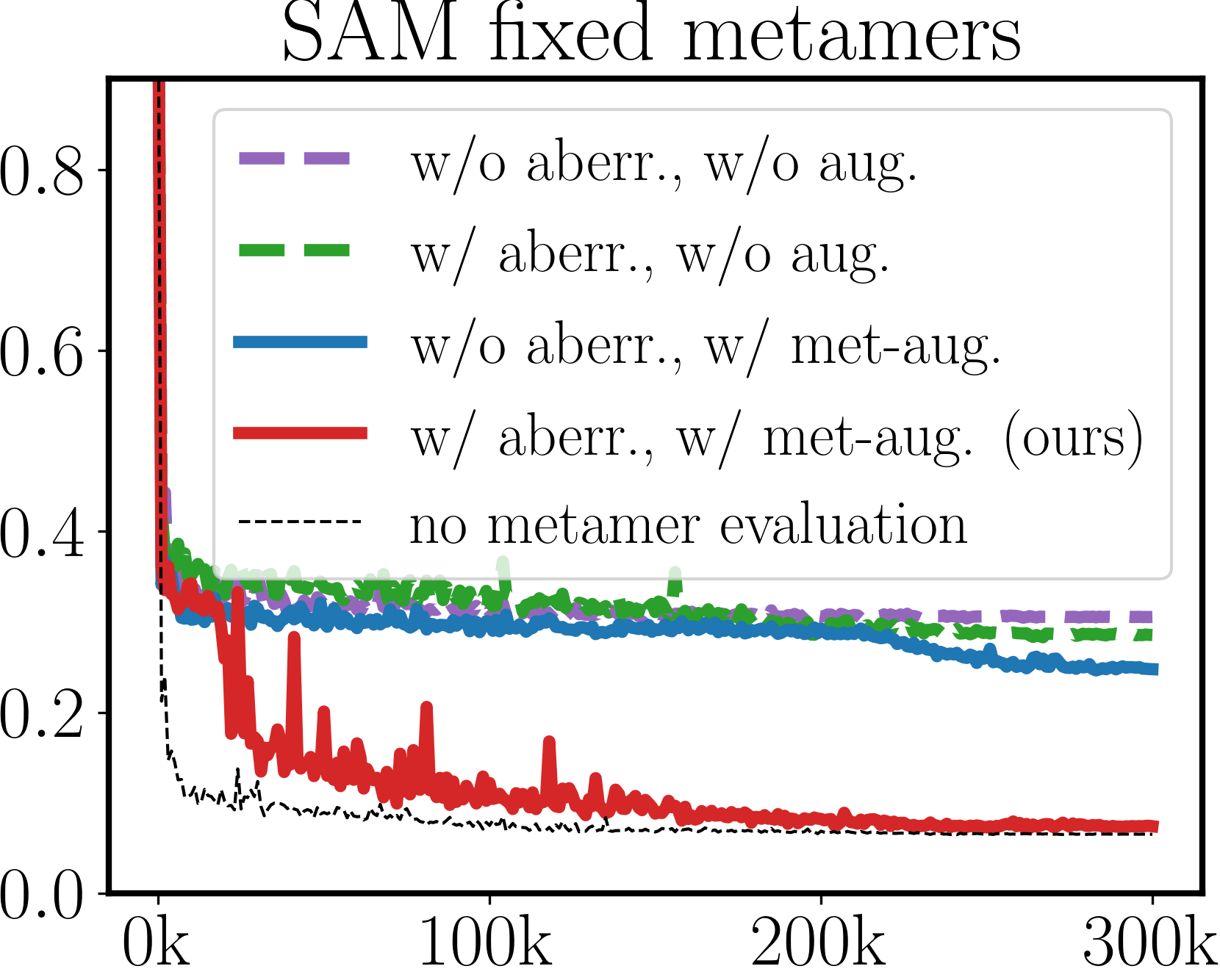}
    % MST on-the-fly
    & \includegraphics[width=0.49\columnwidth]{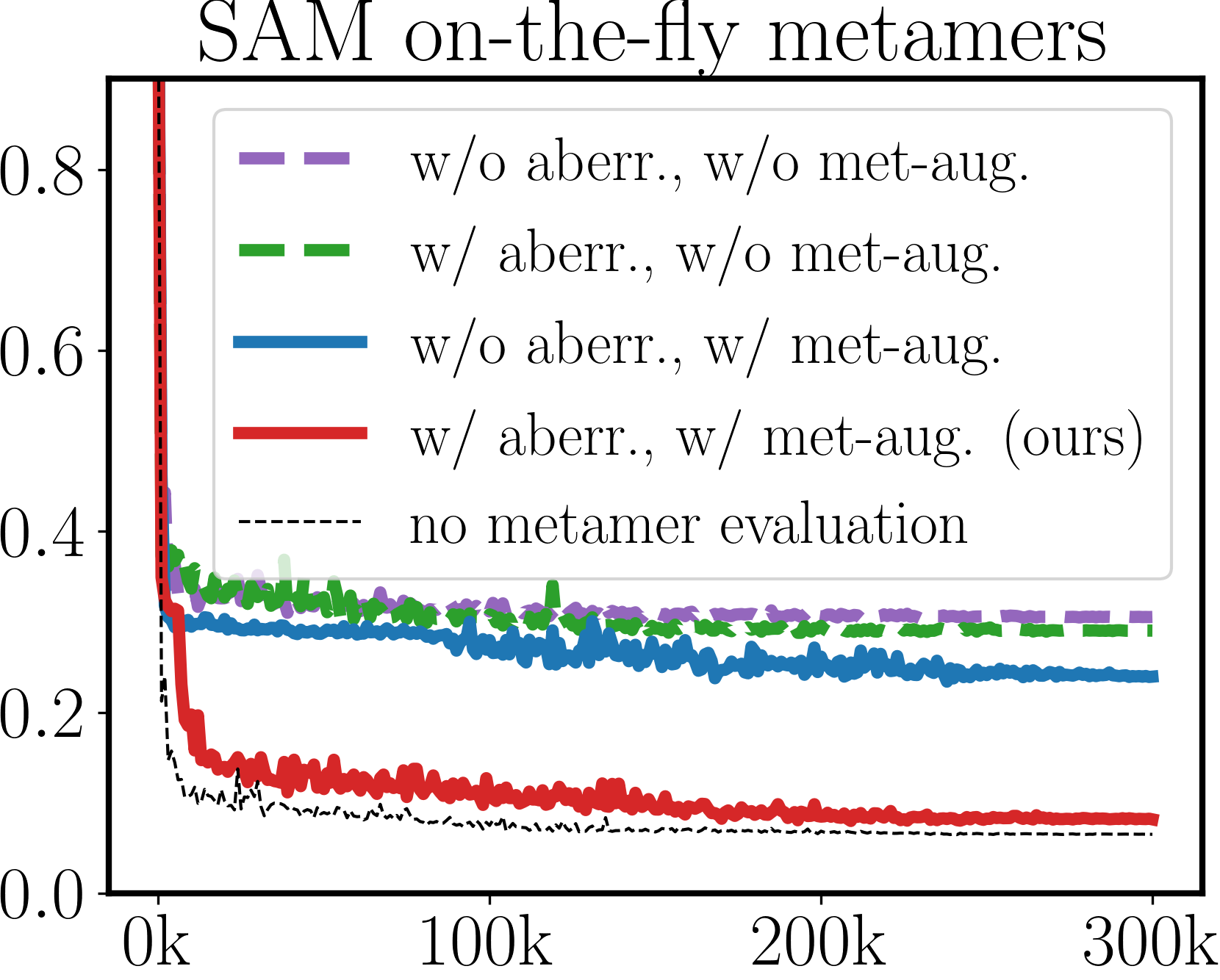} \\
    \multicolumn{2}{c}{MPRNet} & \multicolumn{2}{c}{Restormer} \\
    % MPTNet fixed
    \includegraphics[width=0.49\columnwidth]{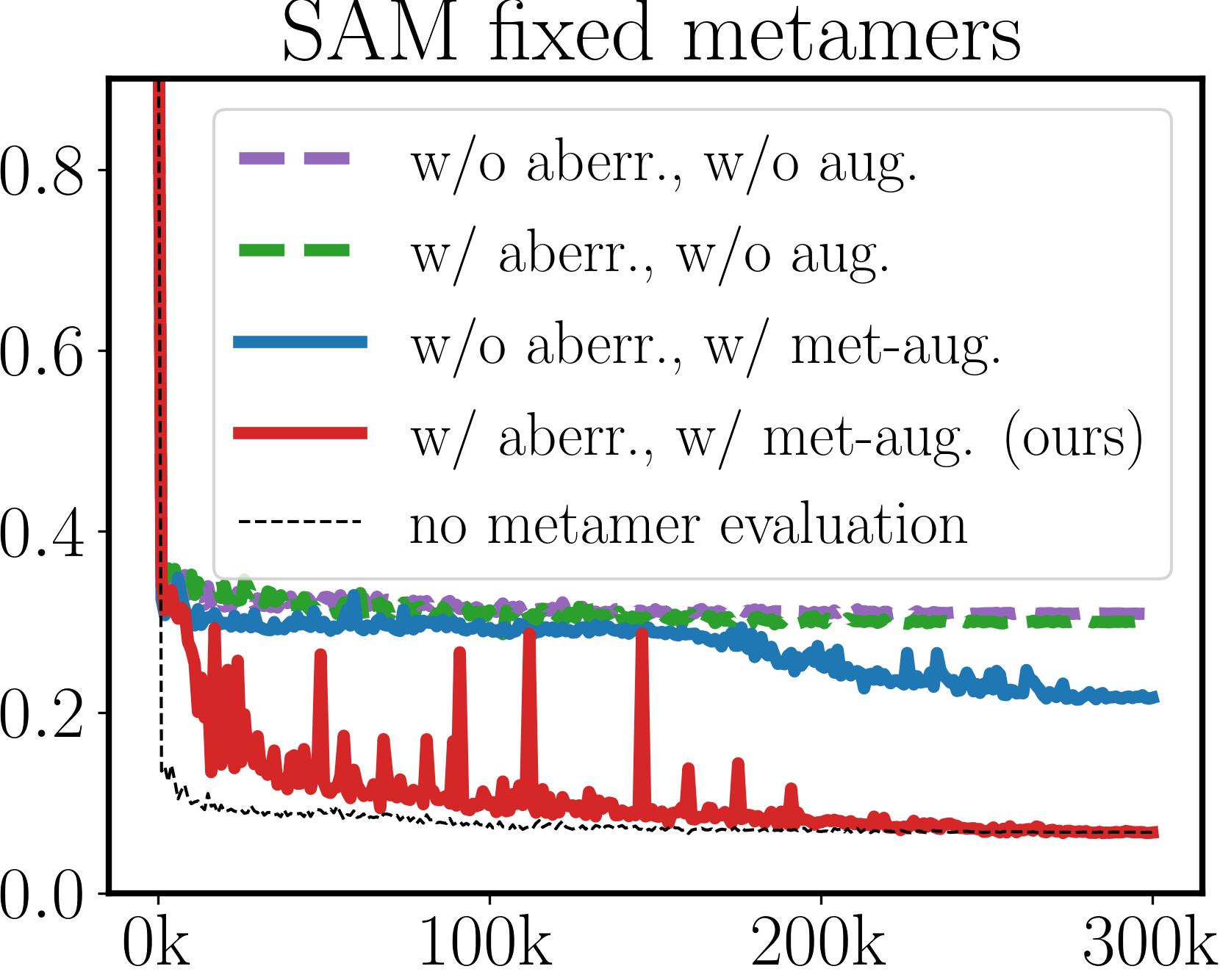}
    % MPRNet on-the-fly
    &\includegraphics[width=0.49\columnwidth]{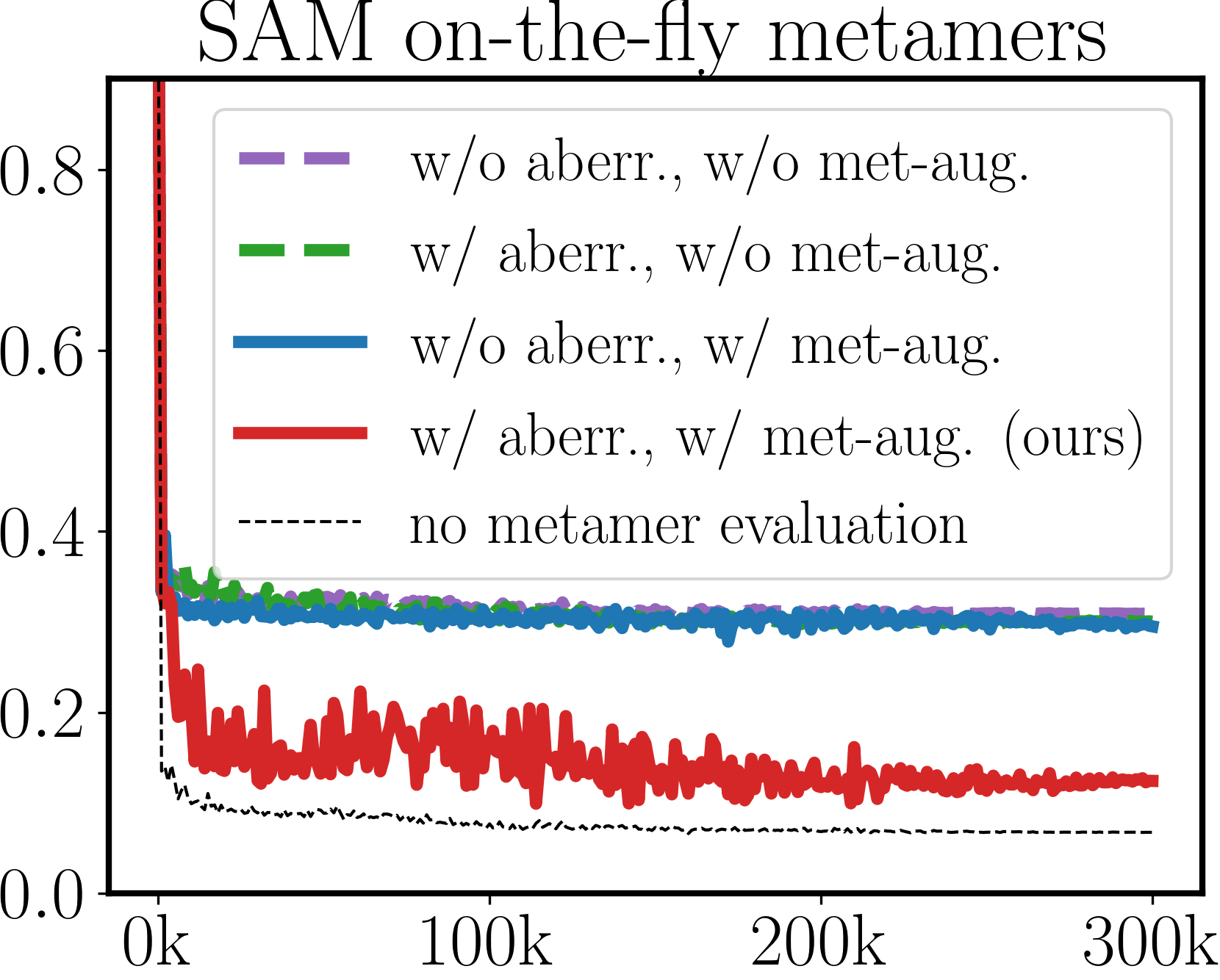}
    % Resotrmer fixed
    &\includegraphics[width=0.49\columnwidth]{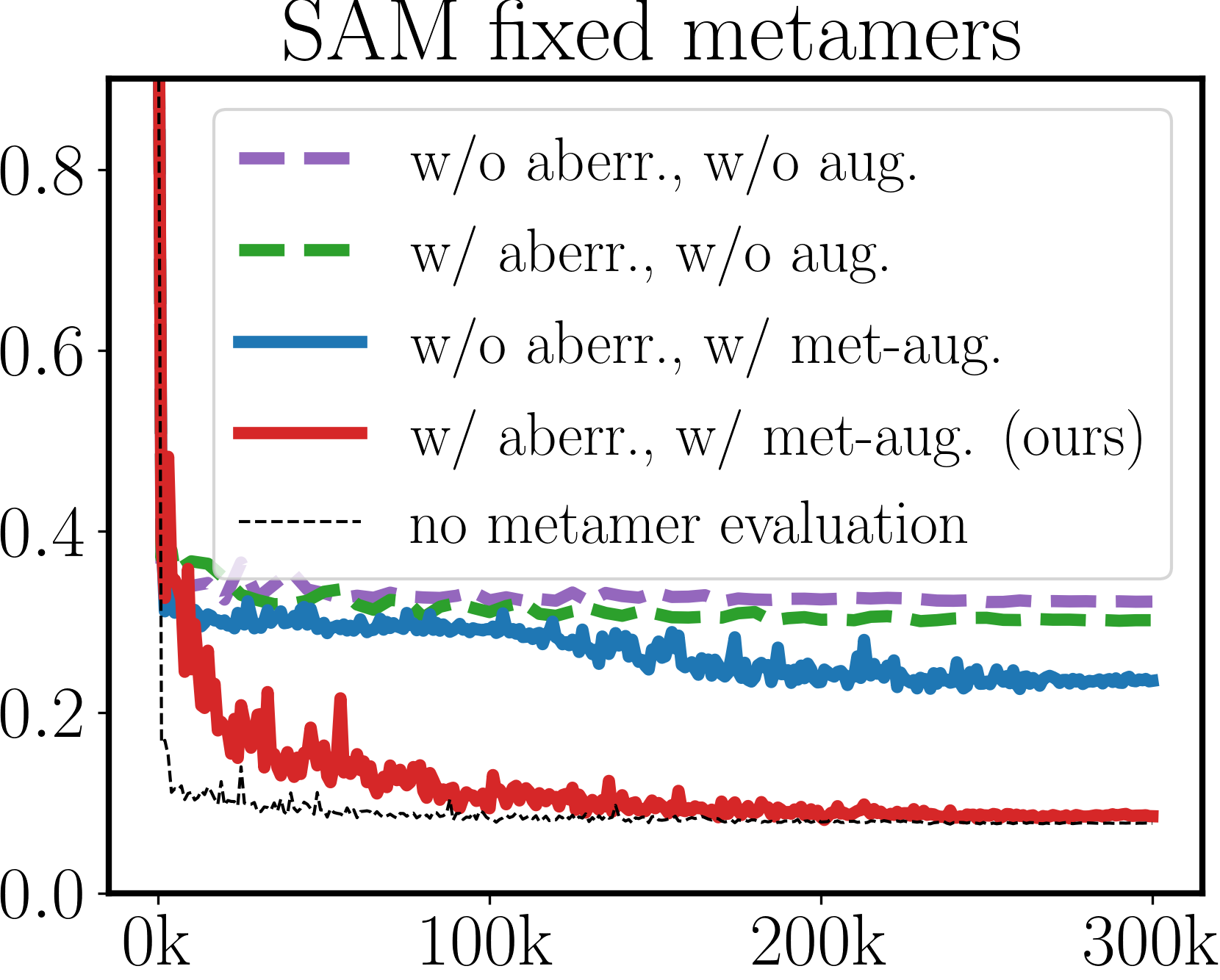}
    % Restormer on-the-fly
    & \includegraphics[width=0.49\columnwidth]{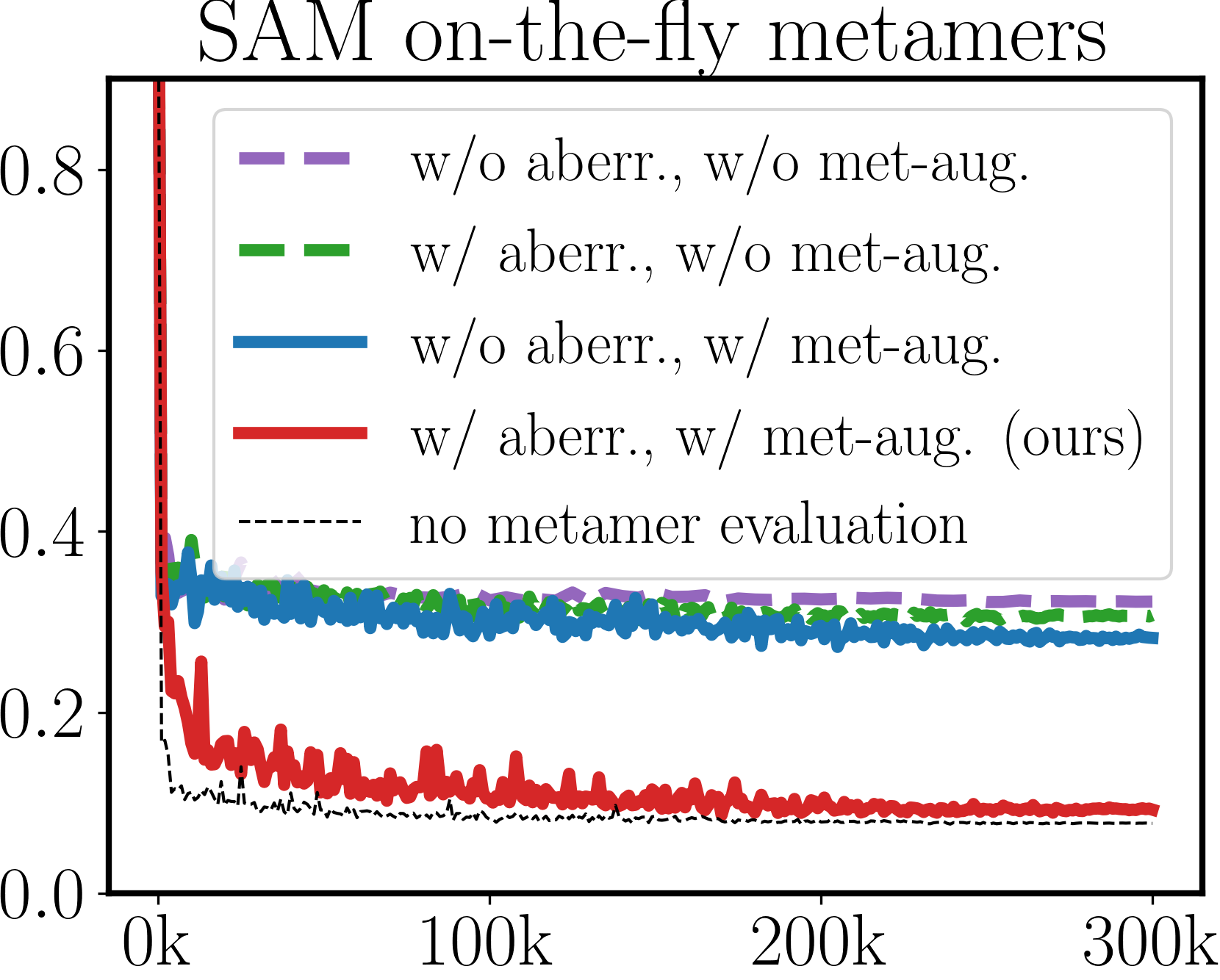}\\
    \multicolumn{2}{c}{MIRNet} & \multicolumn{2}{c}{HINet} \\
    % MIRNet fixed
    \includegraphics[width=0.49\columnwidth]{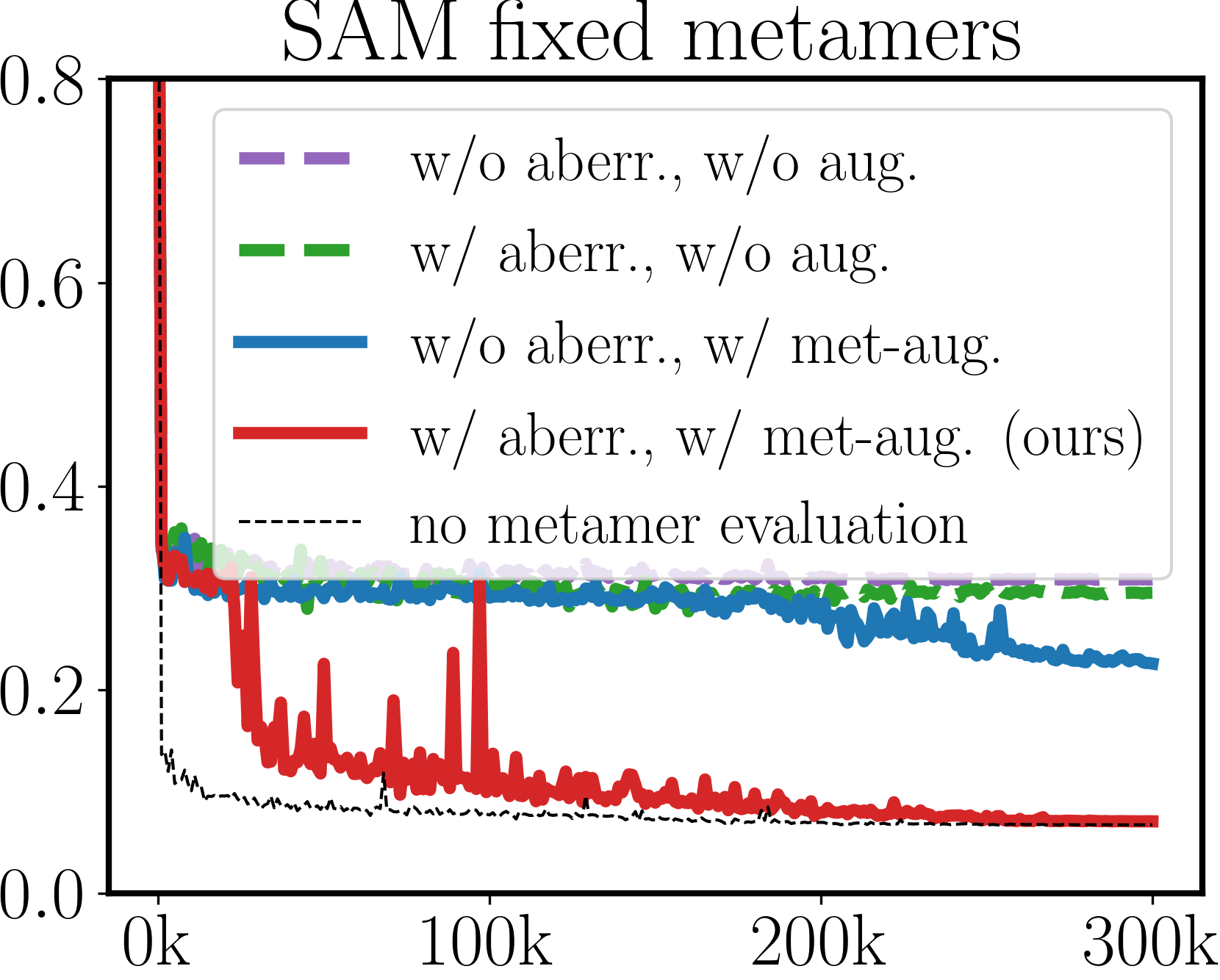}
    % MIRNet on-the-fly
    &\includegraphics[width=0.49\columnwidth]{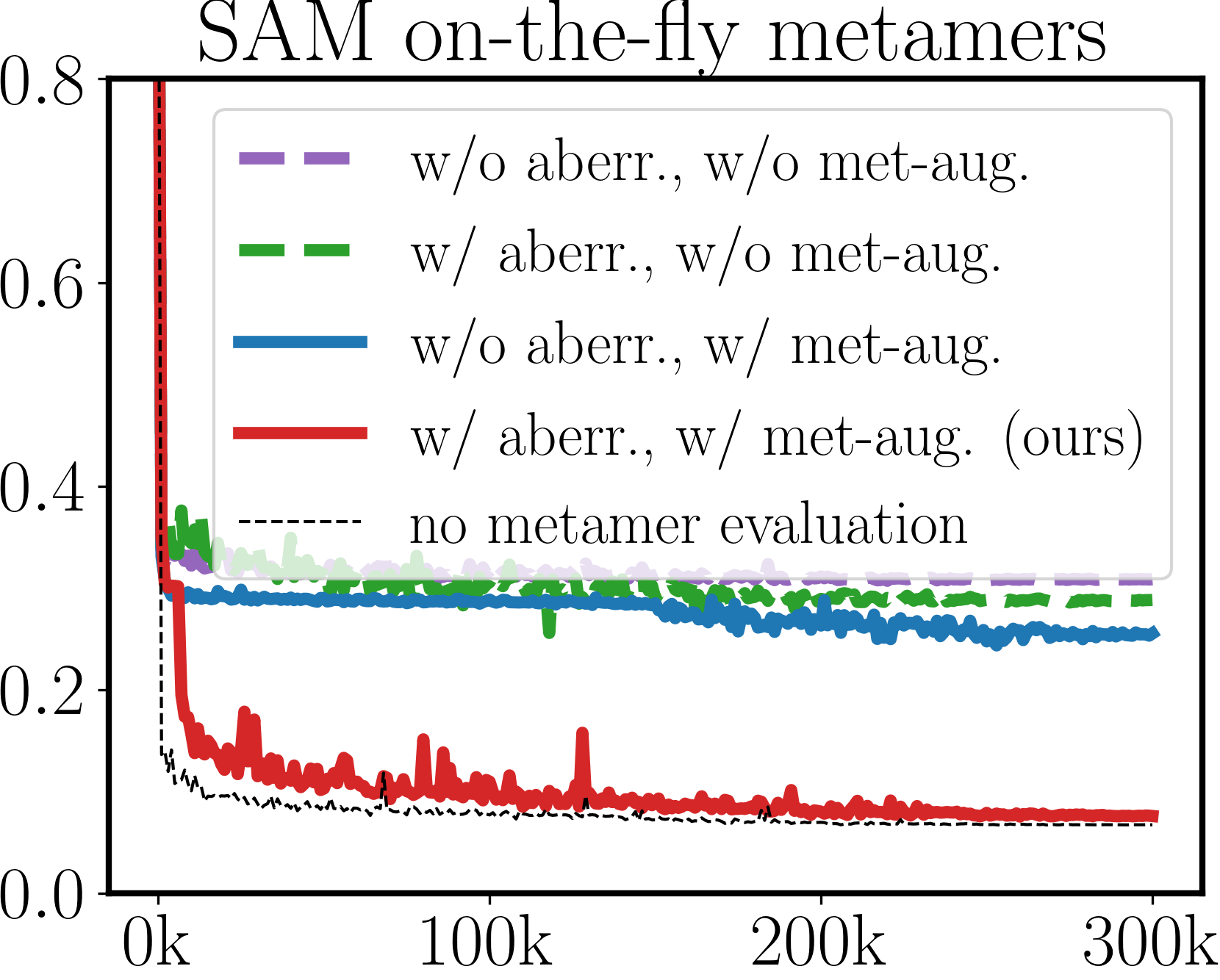}
    % HINet fixed
    &\includegraphics[width=0.49\columnwidth]{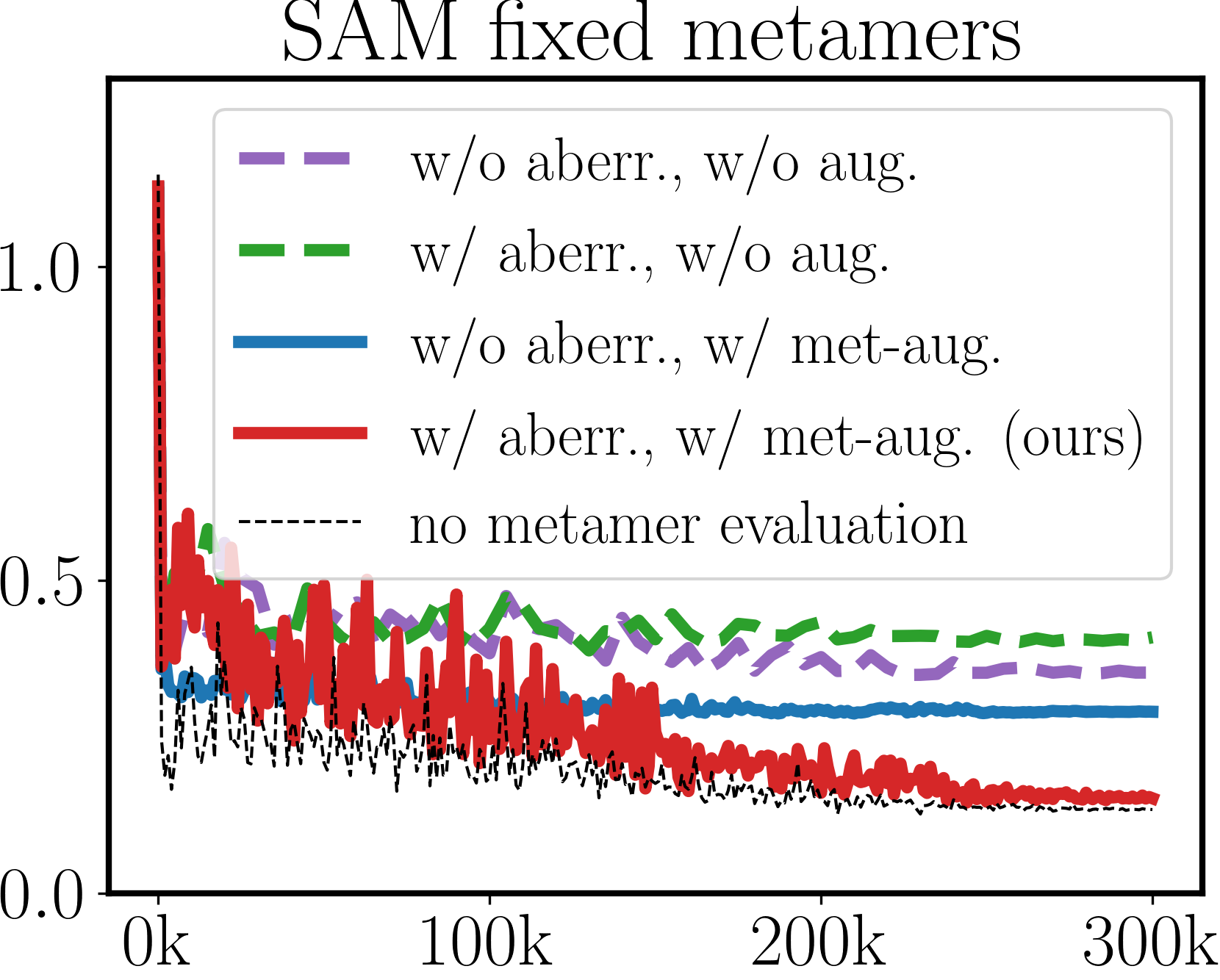}
    % HINet on-the-fly
    &\includegraphics[width=0.49\columnwidth]{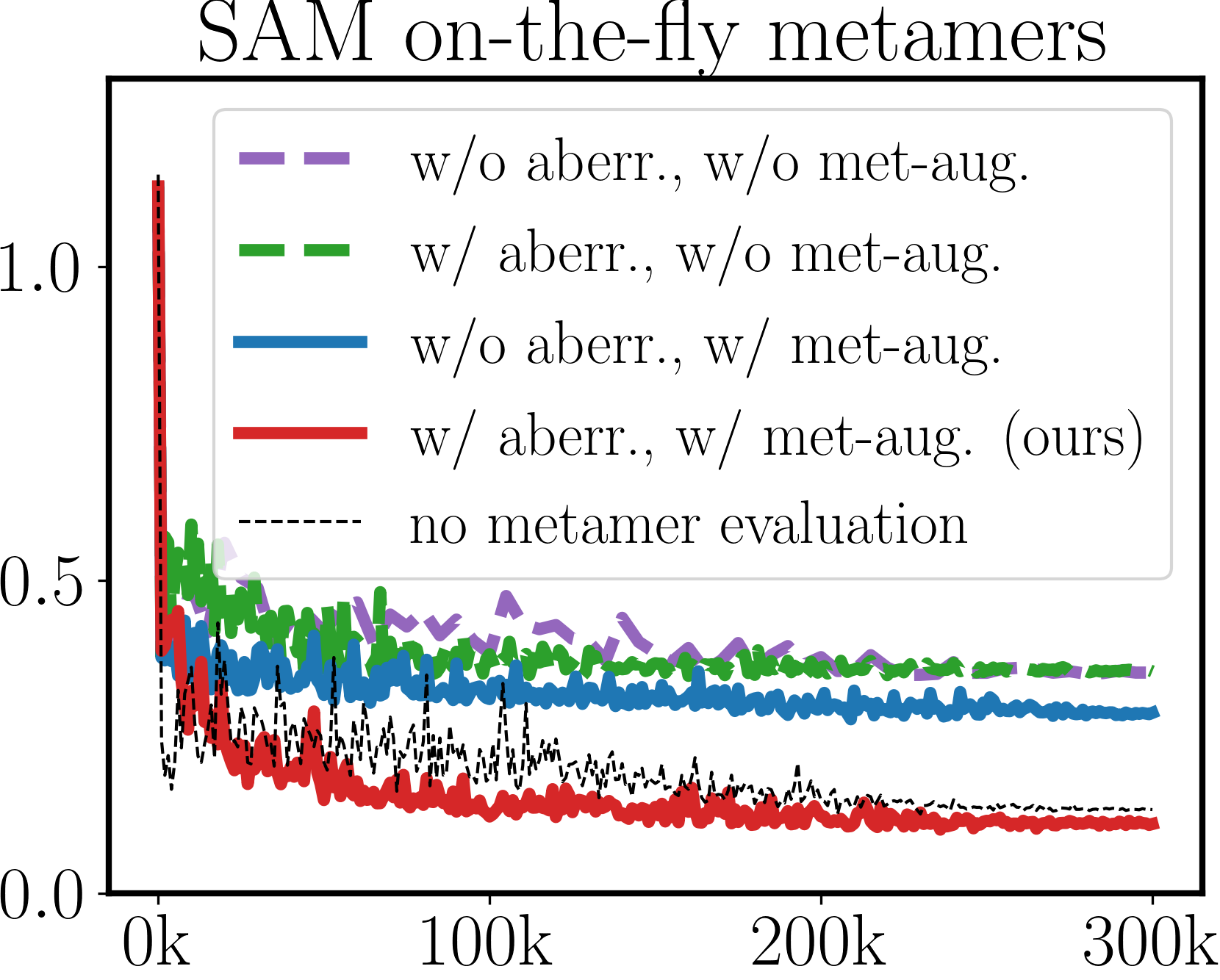} \\
    \end{tabular}
    \caption{Chromatic aberrations improve spectral accuracy for MST++, MST-L, MPRNet, Restormer, MIRNet, and HINet. In each group, left: fixed metamers, and right: on-the-fly metamers.}
    \label{fig:supp_aberration_advantage1}
\end{figure*}

\begin{figure*}[!htp]
    \centering
    \begin{tabular}{c@{}c@{\hspace{5mm}}c@{}c}
    \multicolumn{2}{c}{HDNet} & \multicolumn{2}{c}{AWAN} \\
    % HDNet fixed
    \includegraphics[width=0.49\columnwidth]{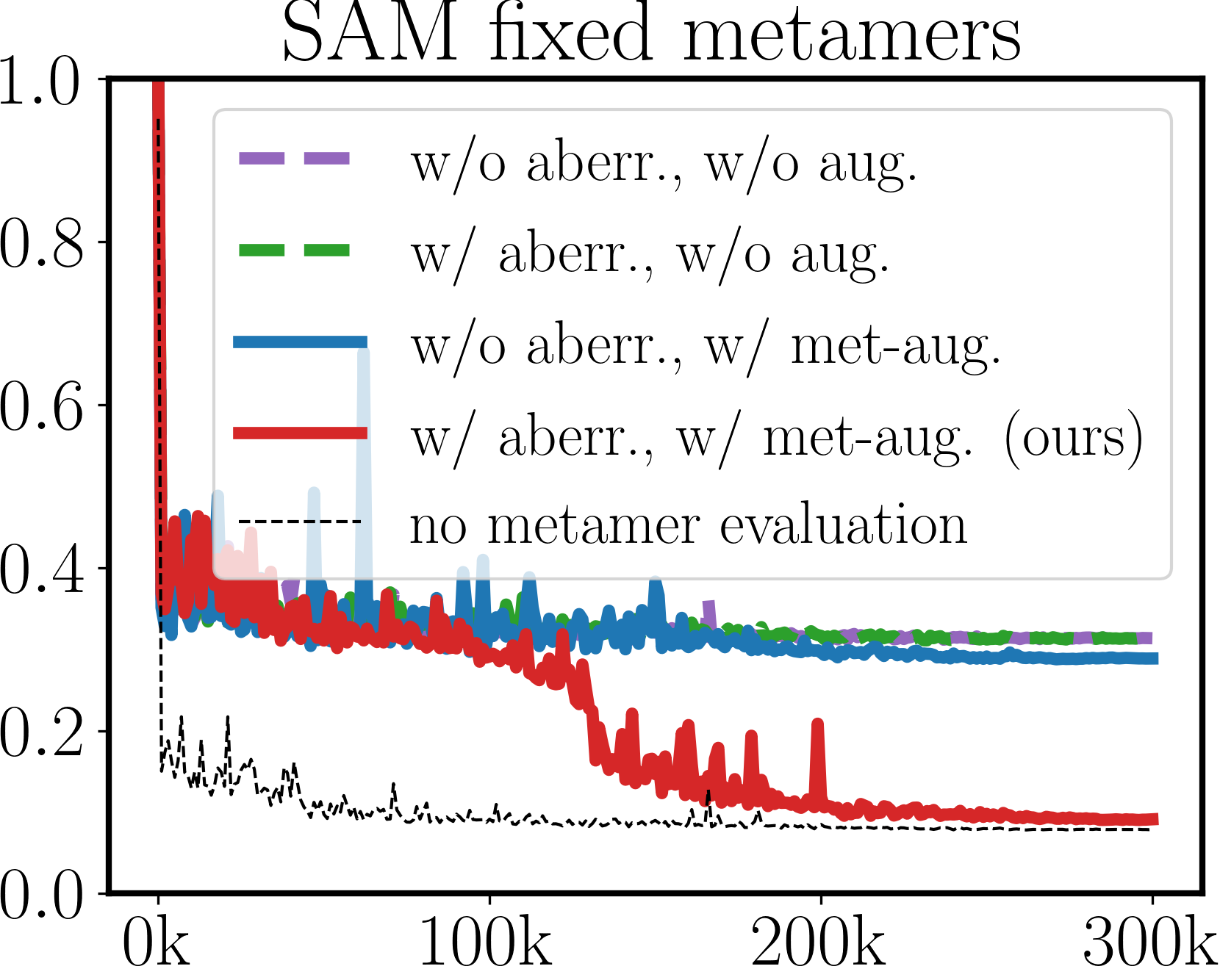}
    % HDNet on-the-fly
    &\includegraphics[width=0.49\columnwidth]{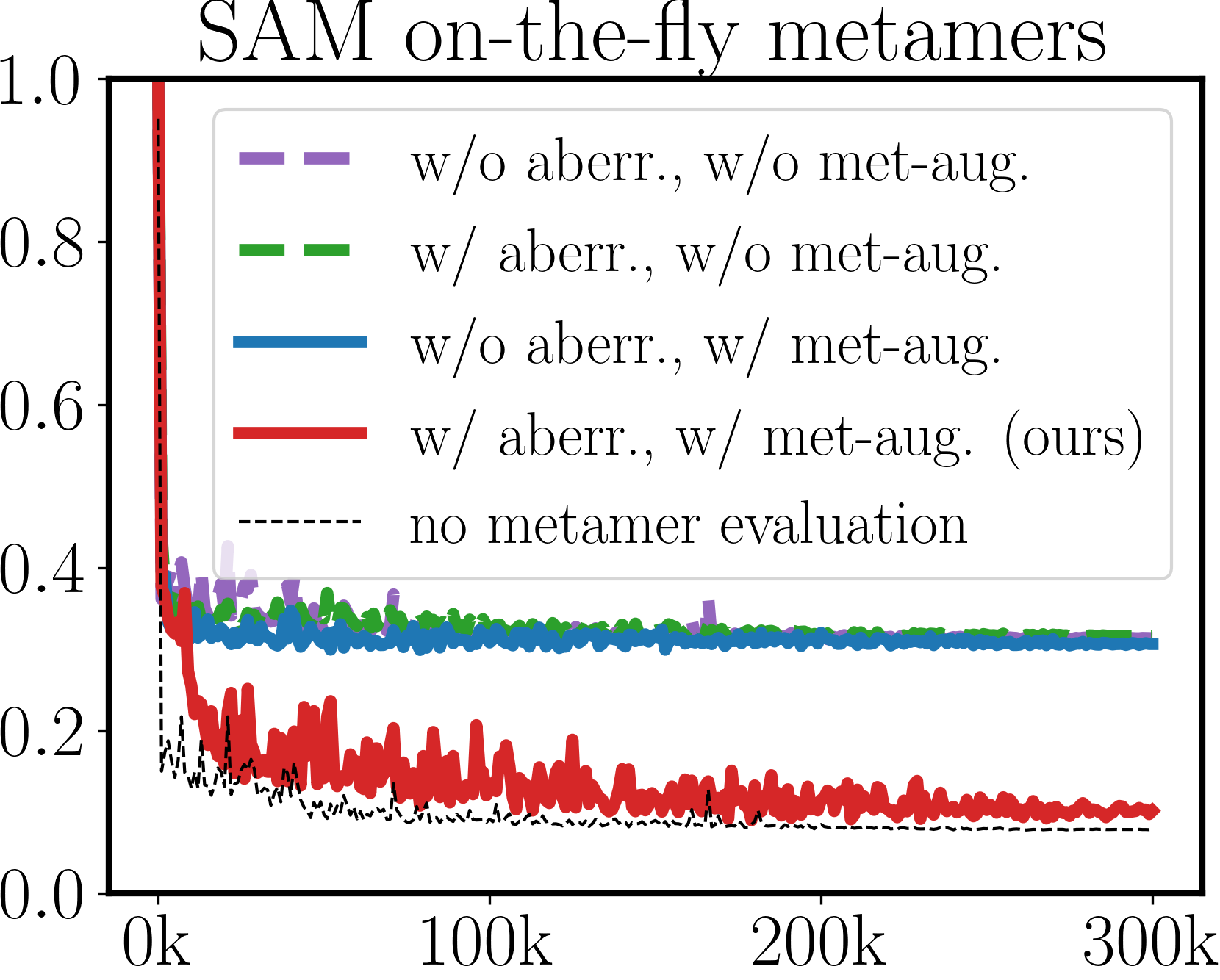}
    % AWAN fixed
    &\includegraphics[width=0.49\columnwidth]{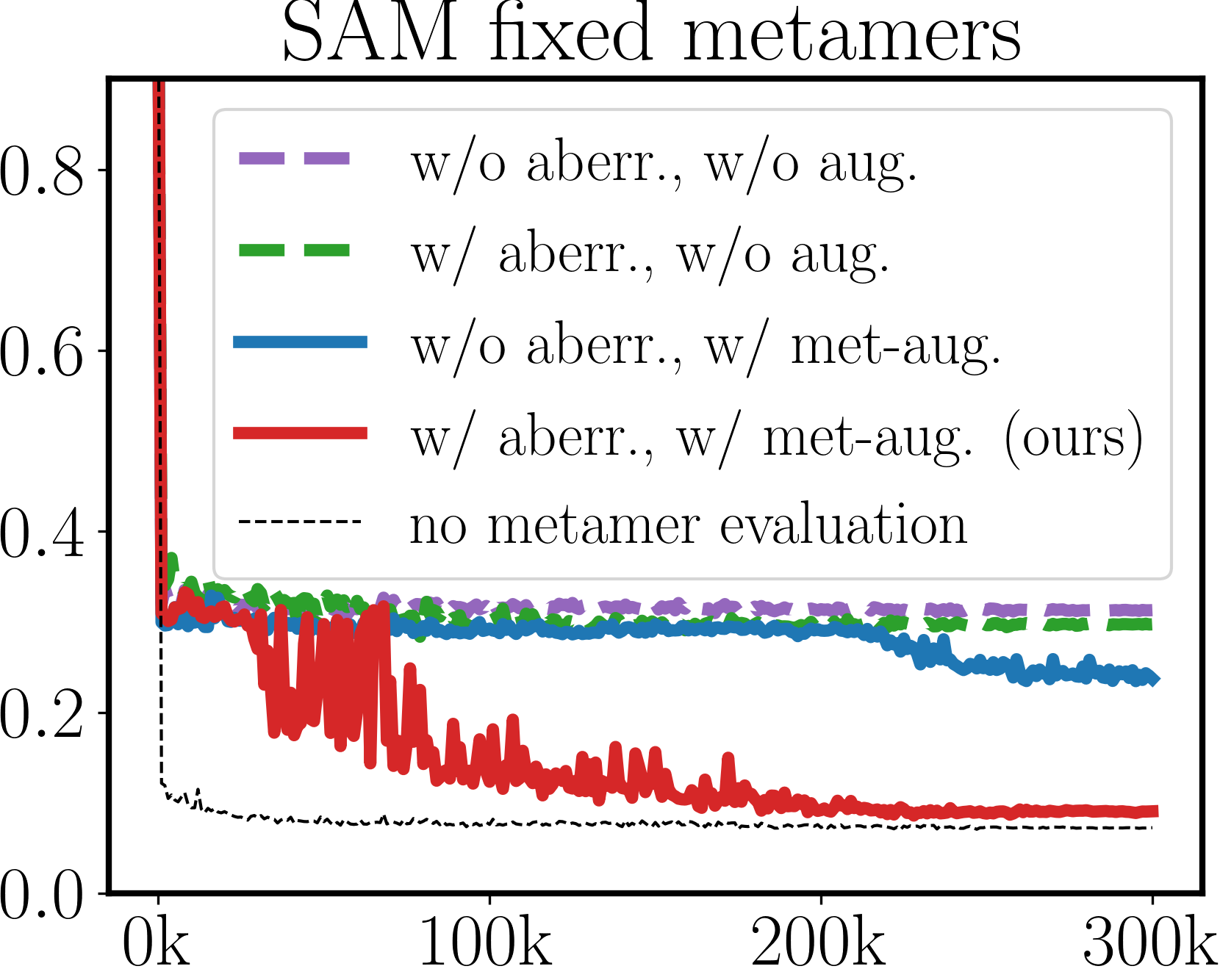}
    % AWAN on-the-fly
    & \includegraphics[width=0.49\columnwidth]{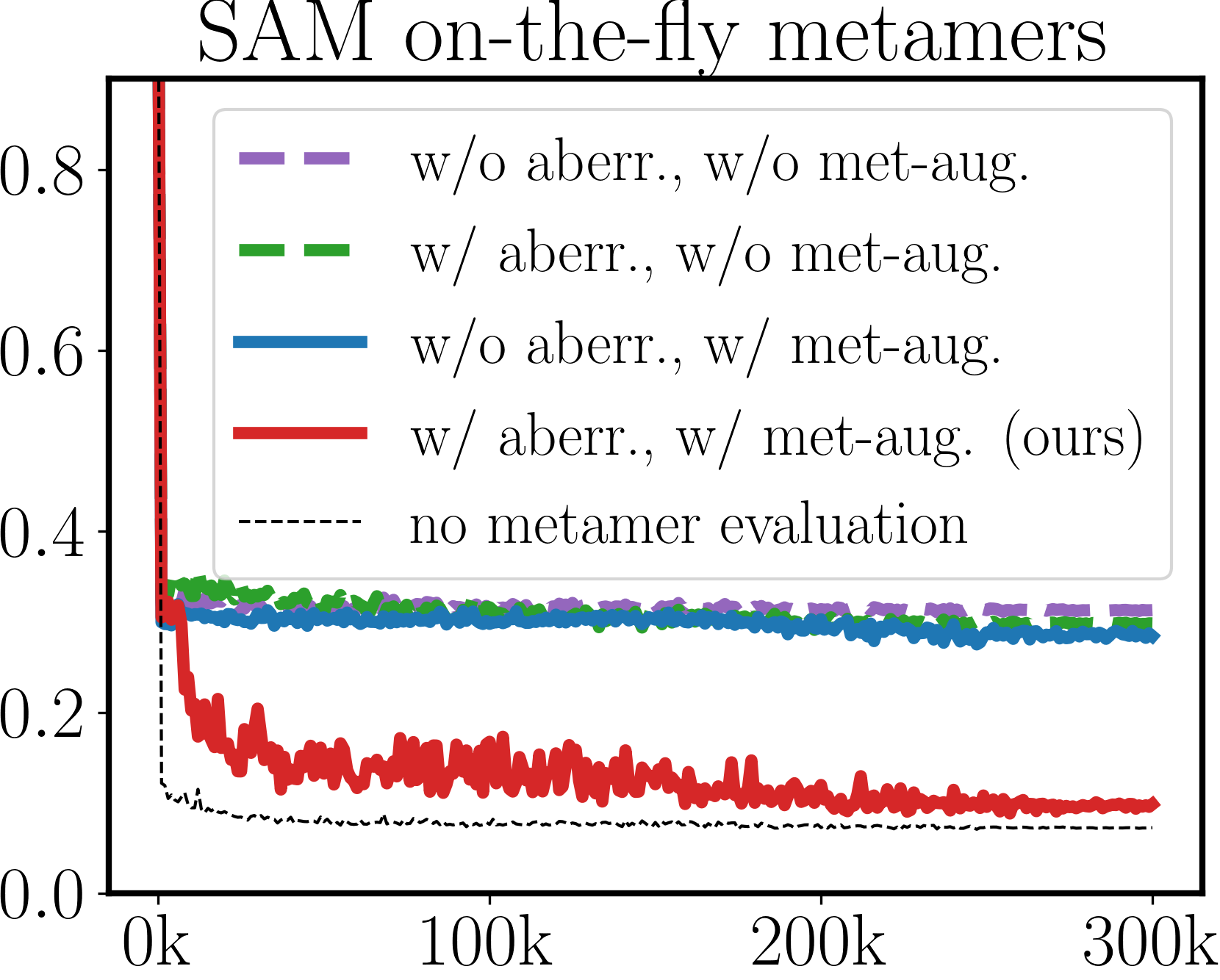}\\
    \multicolumn{2}{c}{EDSR} & \multicolumn{2}{c}{HRNet} \\
    % EDSR fixed
    \includegraphics[width=0.49\columnwidth]{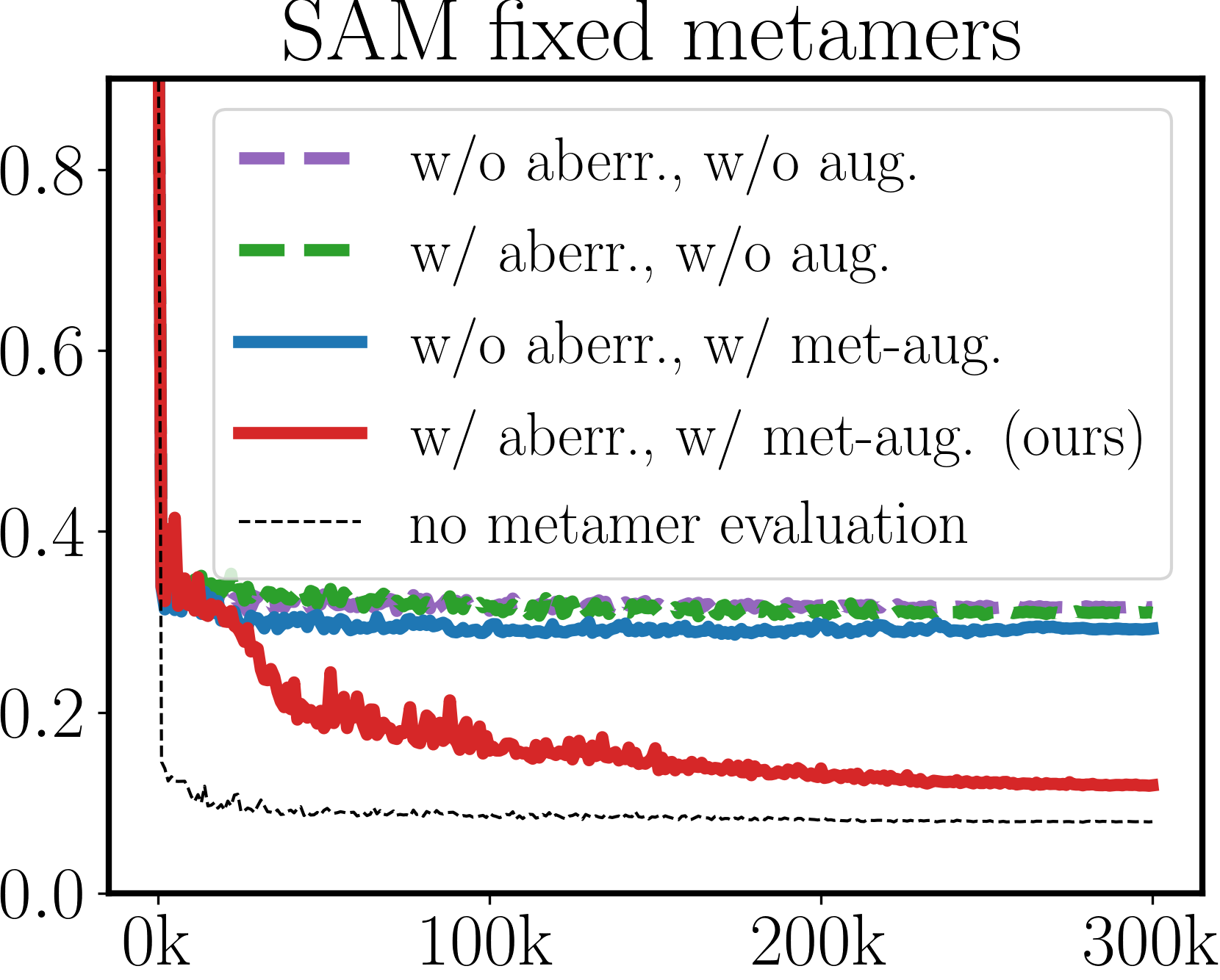}
    % EDSR on-the-fly
    &\includegraphics[width=0.49\columnwidth]{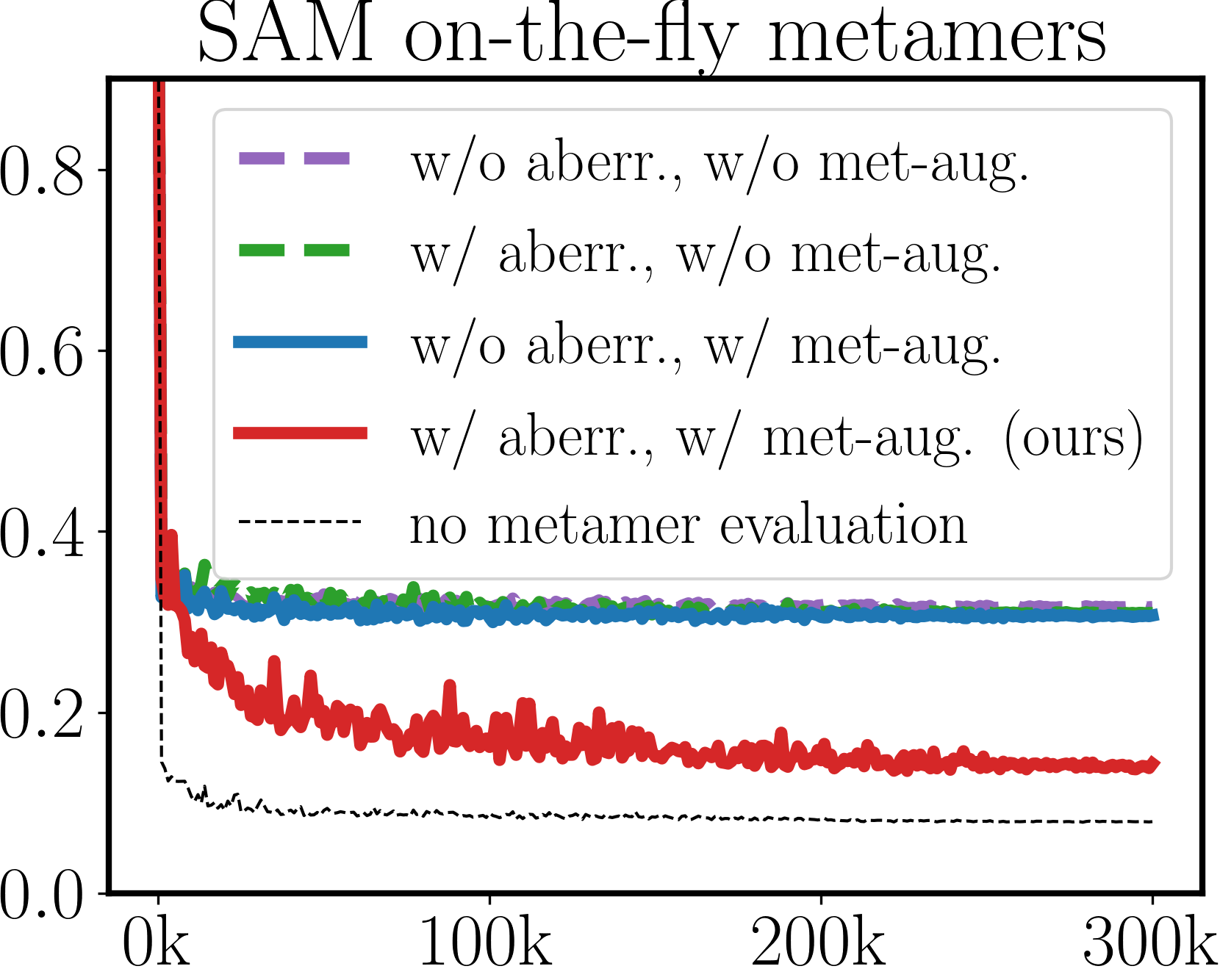}
    % HRNet fixed
    &\includegraphics[width=0.49\columnwidth]{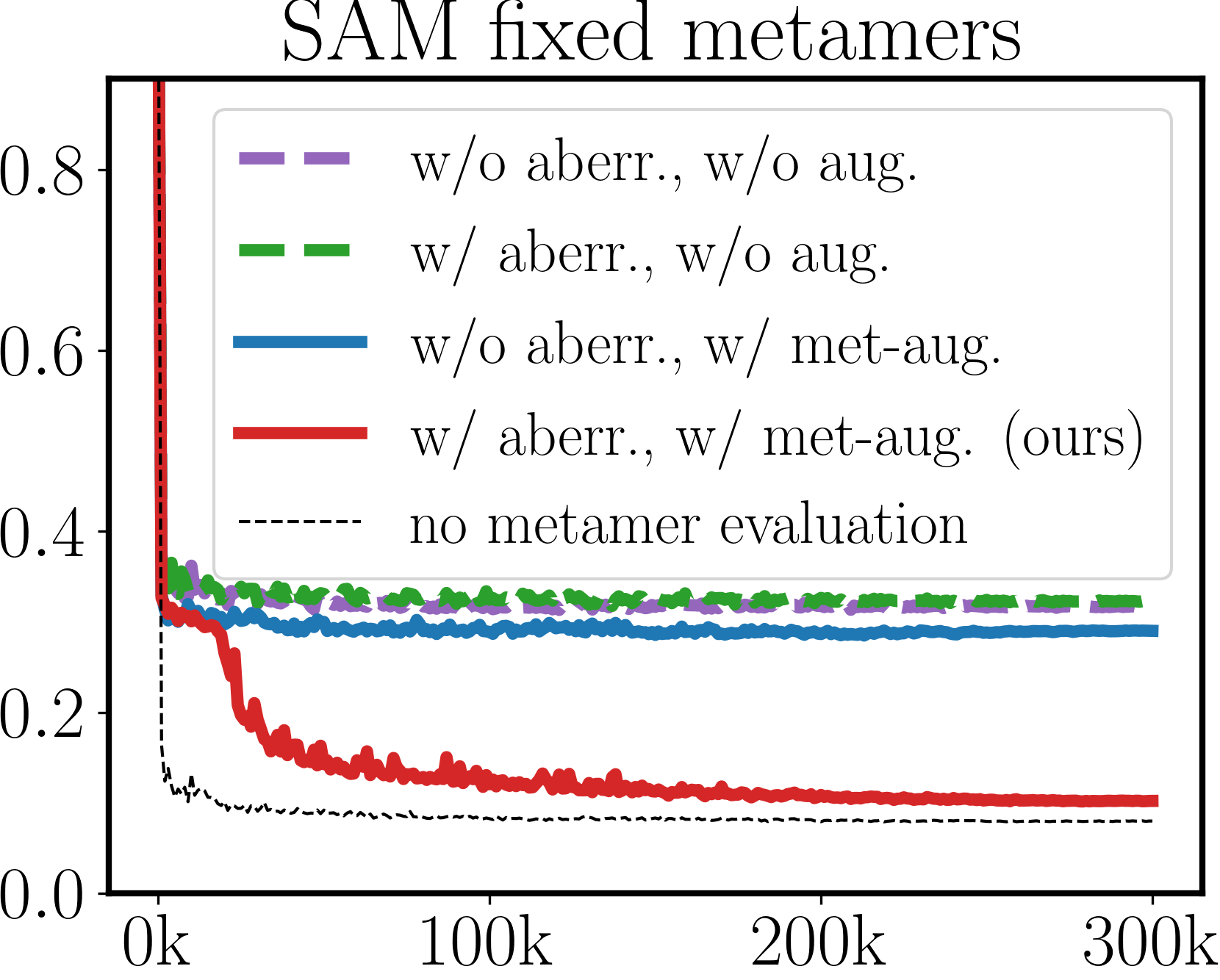}
    % HRNet on-the-fly
    & \includegraphics[width=0.49\columnwidth]{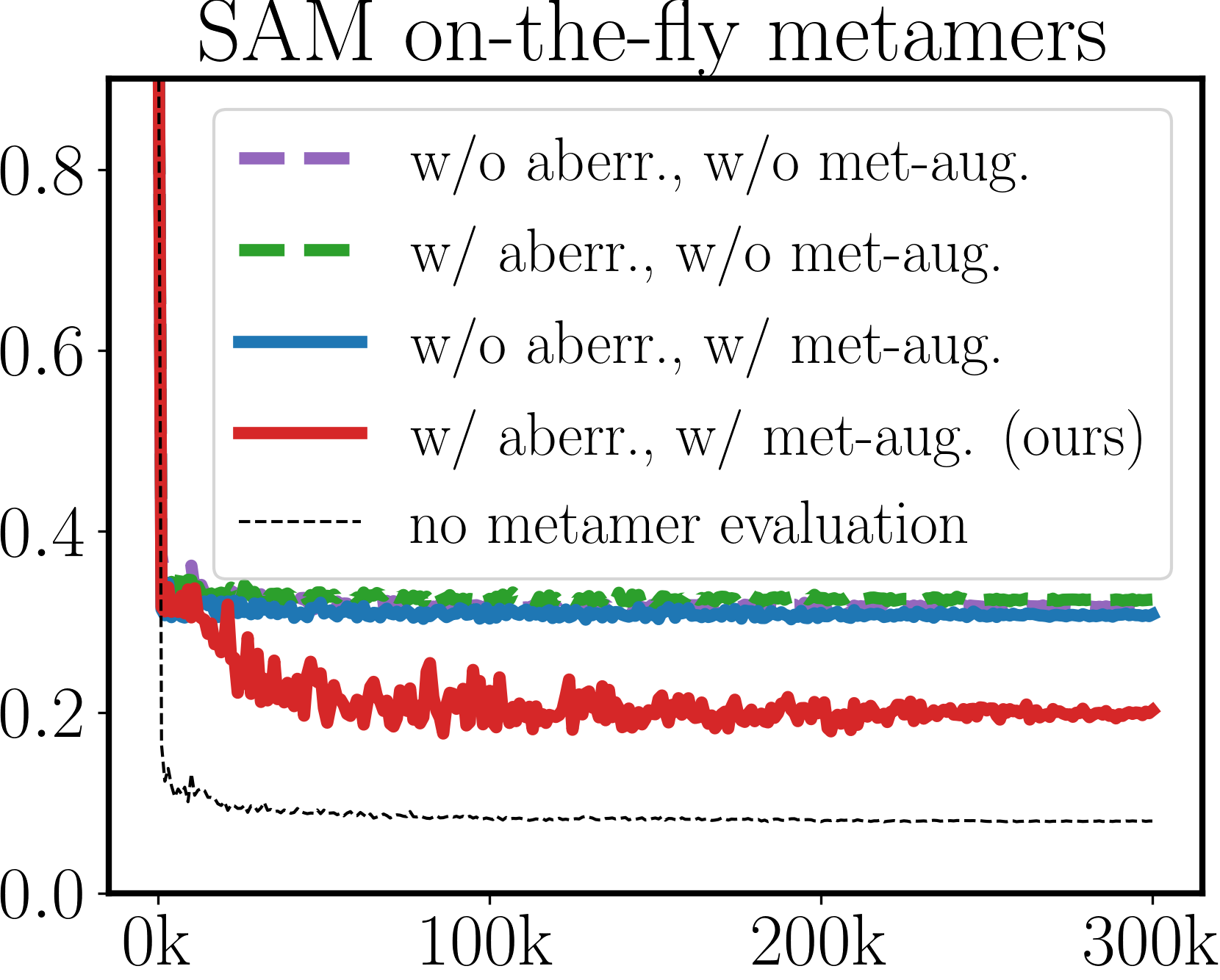}\\
    \multicolumn{2}{c}{HSCNN+} &  \\
    % HSCNN+ fixed
    \includegraphics[width=0.49\columnwidth]{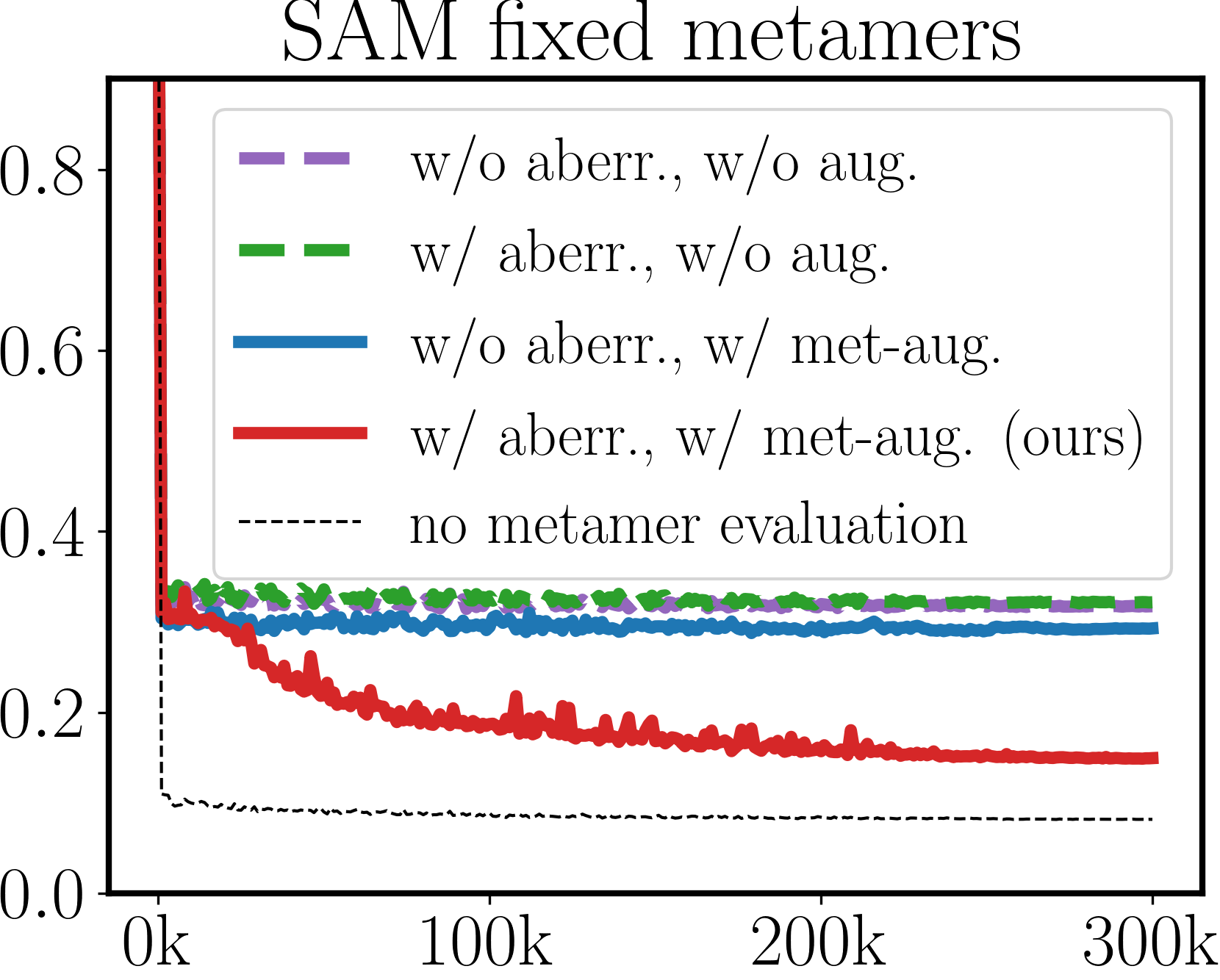}
    % HSCNN+ on-the-fly
    &\includegraphics[width=0.49\columnwidth]{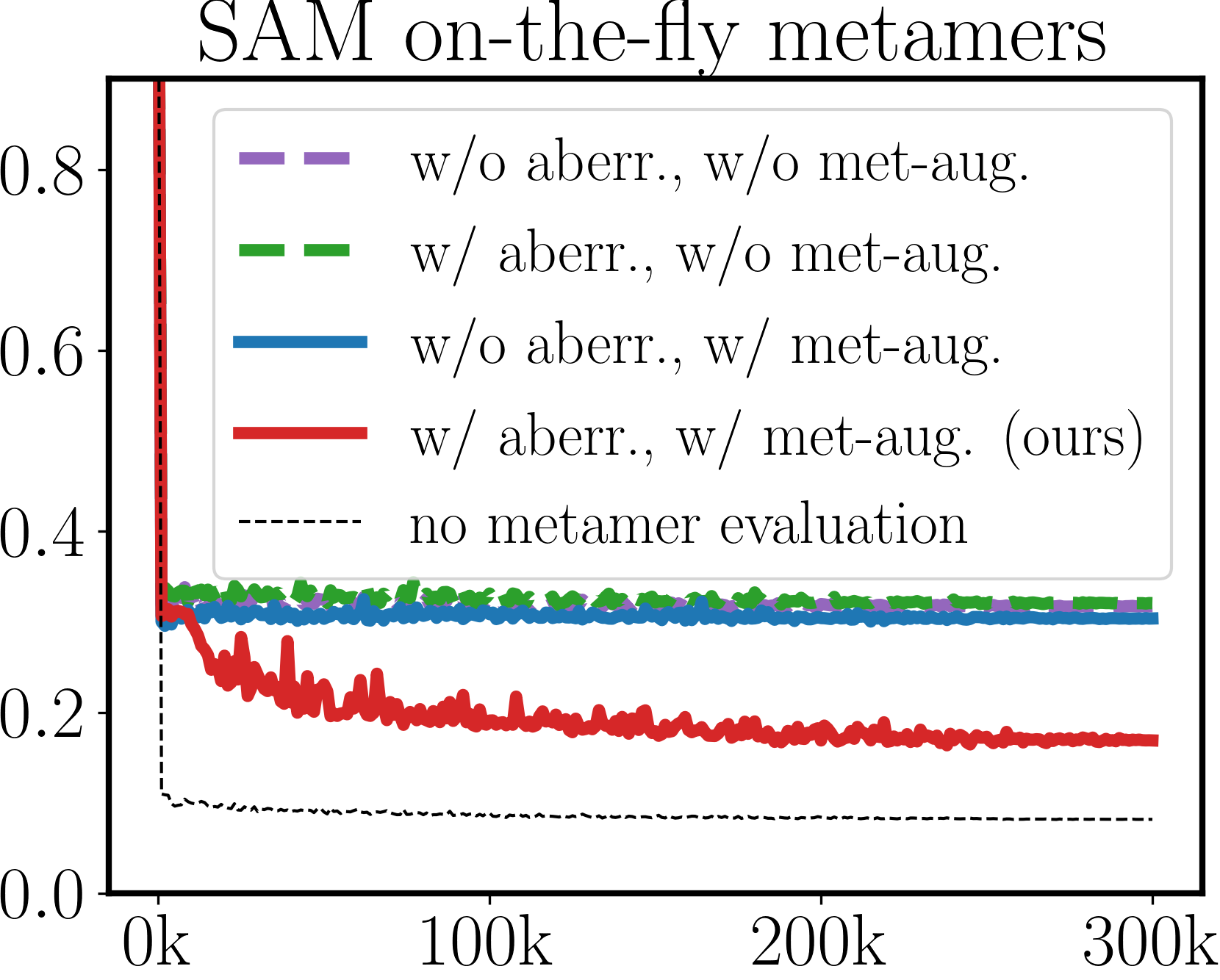}
    &    & \\
    \end{tabular}
    \caption{The aberration advantage results for HDNet, AWAN, EDSR, HRNet, and HSCNN+ on ARAD1K. In each group, left is for fixed metamers, and right is for on-the-fly metamers.}
    \label{fig:supp_aberration_advantage2}
\end{figure*}

\begin{figure*}[!htp]
    \centering
    \newcommand{\placeholder}{\fbox{\rule{0pt}{30mm} \rule{40mm}{0pt}}}
    \begin{tabular}{c@{}c@{\hspace{5mm}}c@{}c}
    \multicolumn{2}{c}{HySAT} & \multicolumn{2}{c}{HPRN} \\
    % HySAT fixed
    \includegraphics[width=0.49\columnwidth]{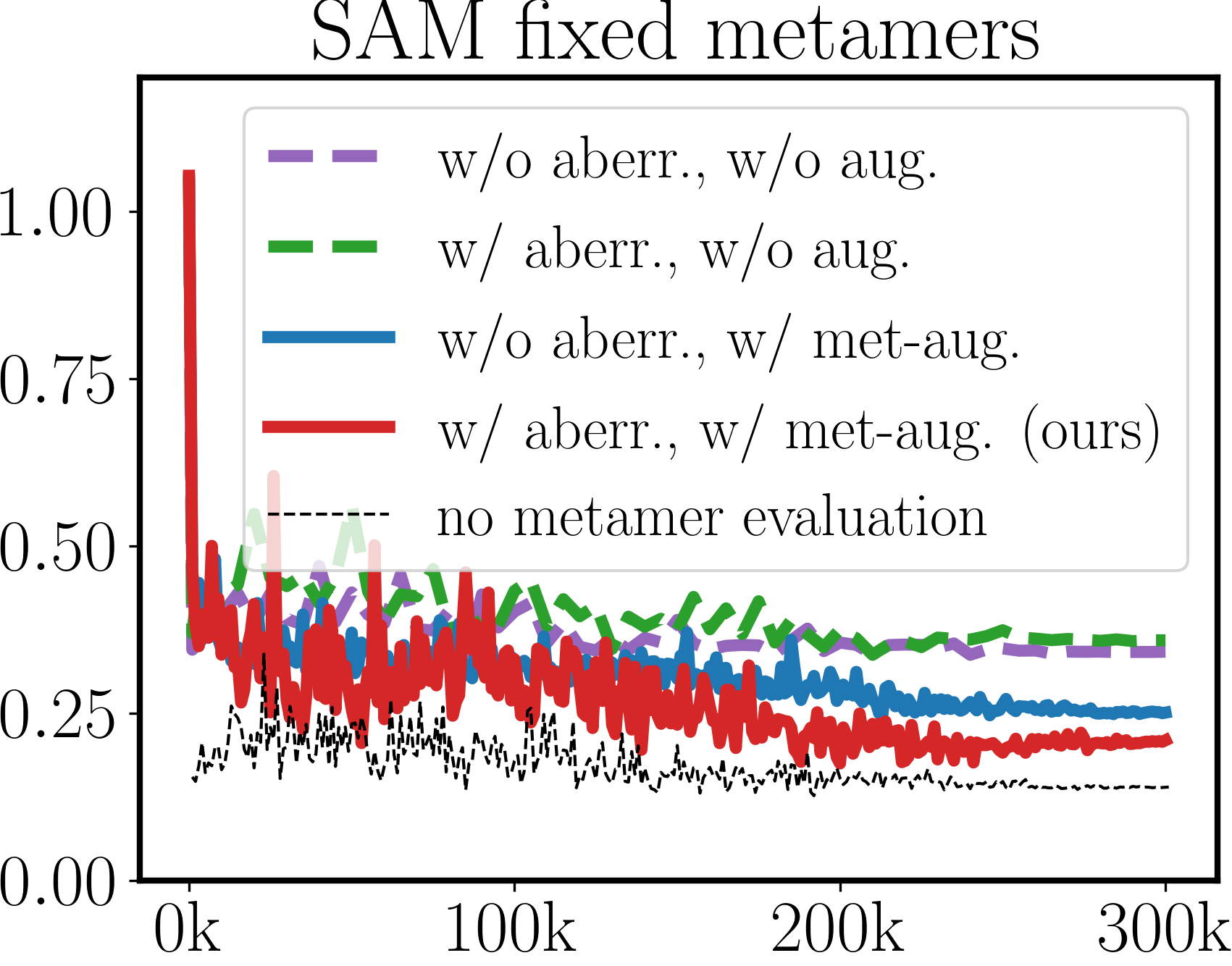} 
    % HySAT on-the-fly
    &\includegraphics[width=0.49\columnwidth]{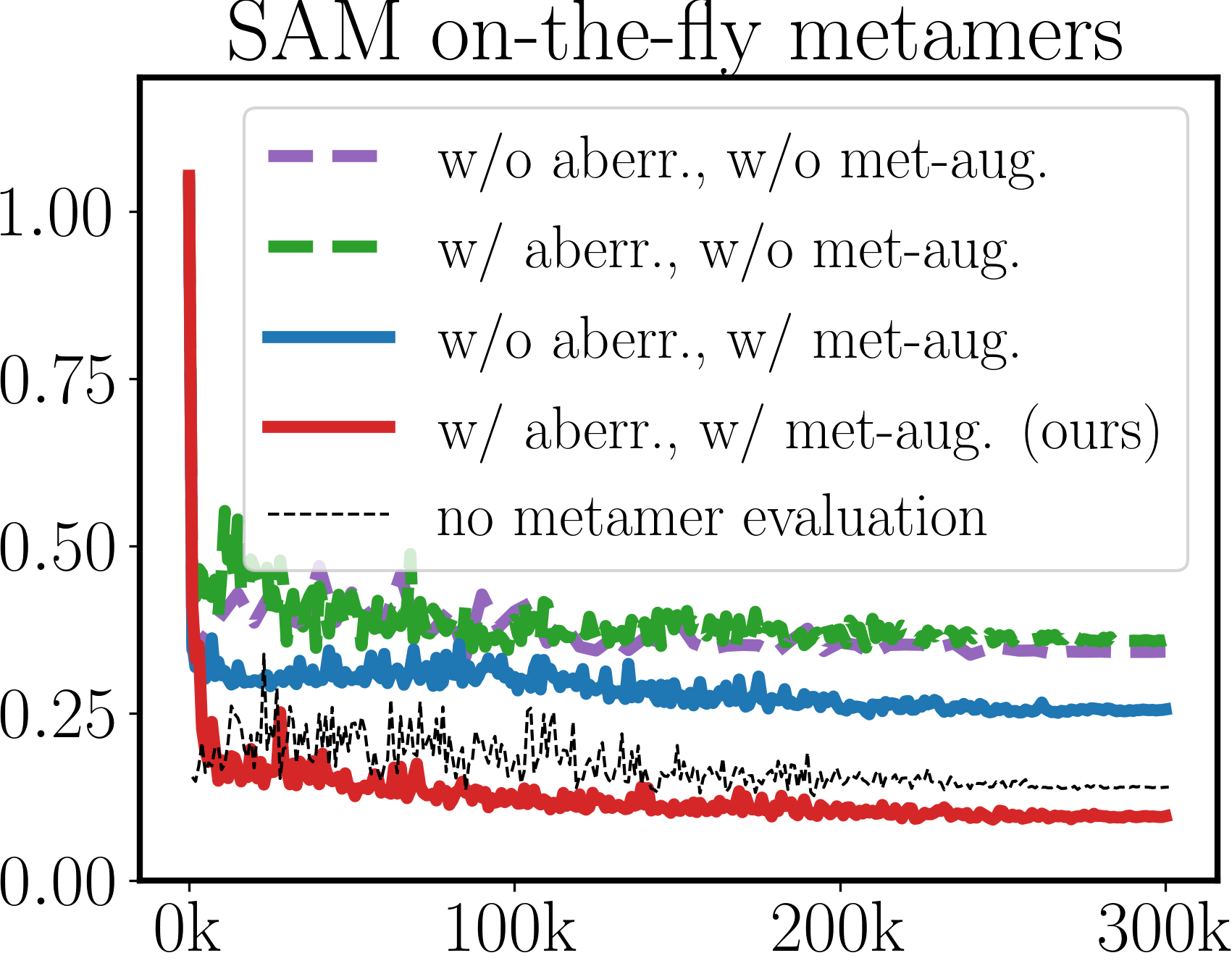}
    % HPRN fixed
    &\includegraphics[width=0.49\columnwidth]{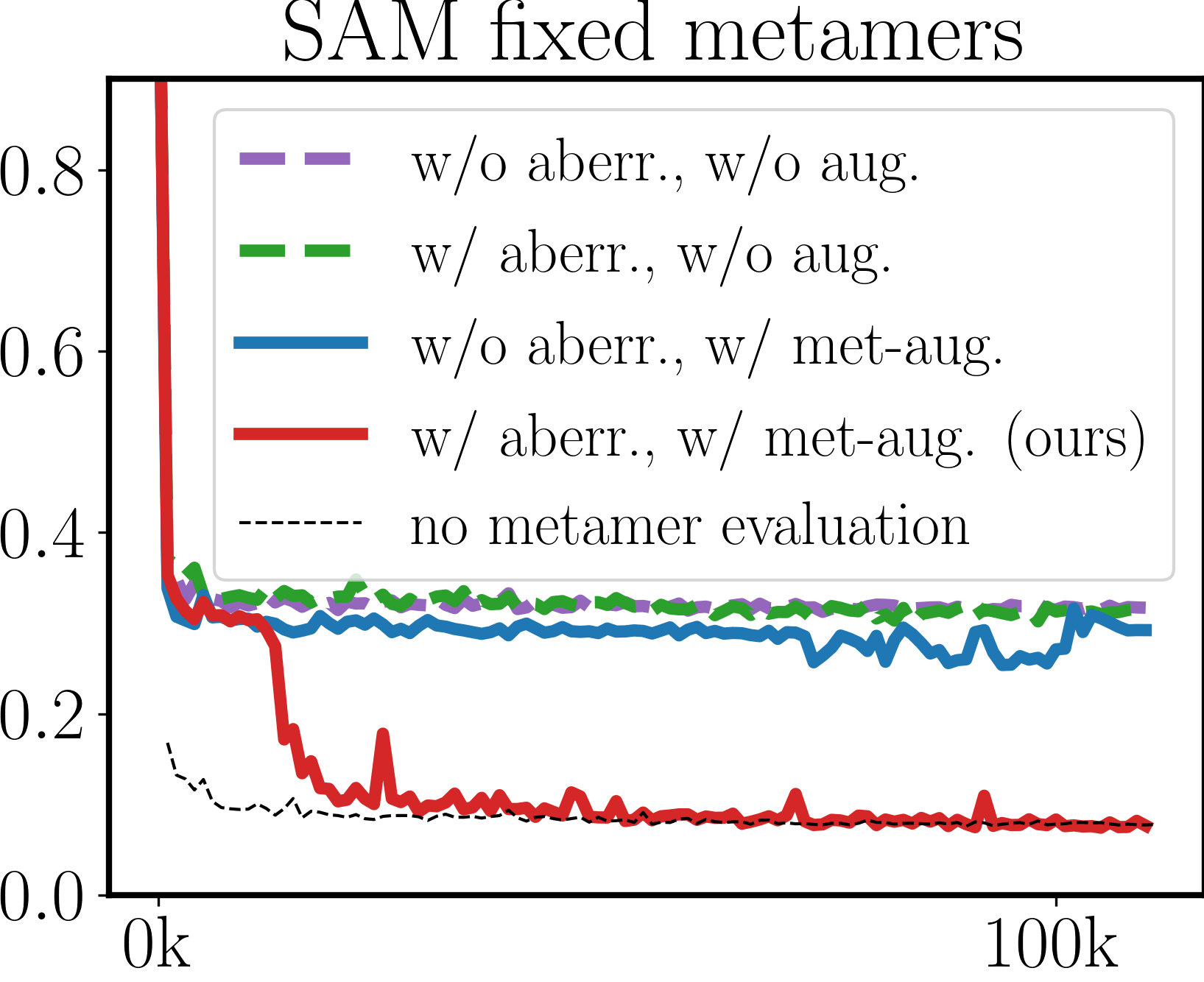}
    % HPRN on-the-fly
    &\includegraphics[width=0.49\columnwidth]{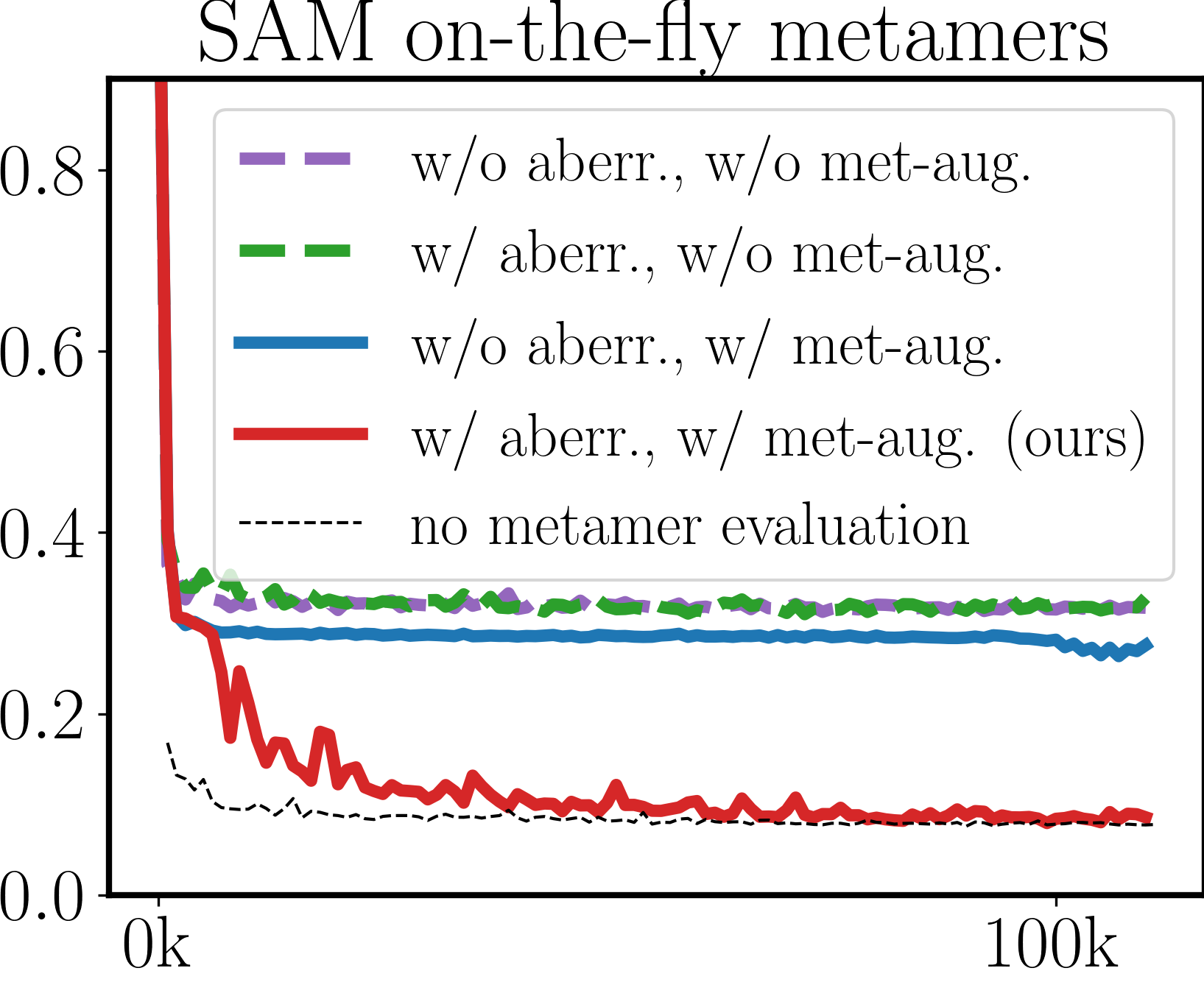}\\
    \multicolumn{2}{c}{SSTHyper} & \multicolumn{2}{c}{MSFN} \\
    \includegraphics[width=0.49\columnwidth]{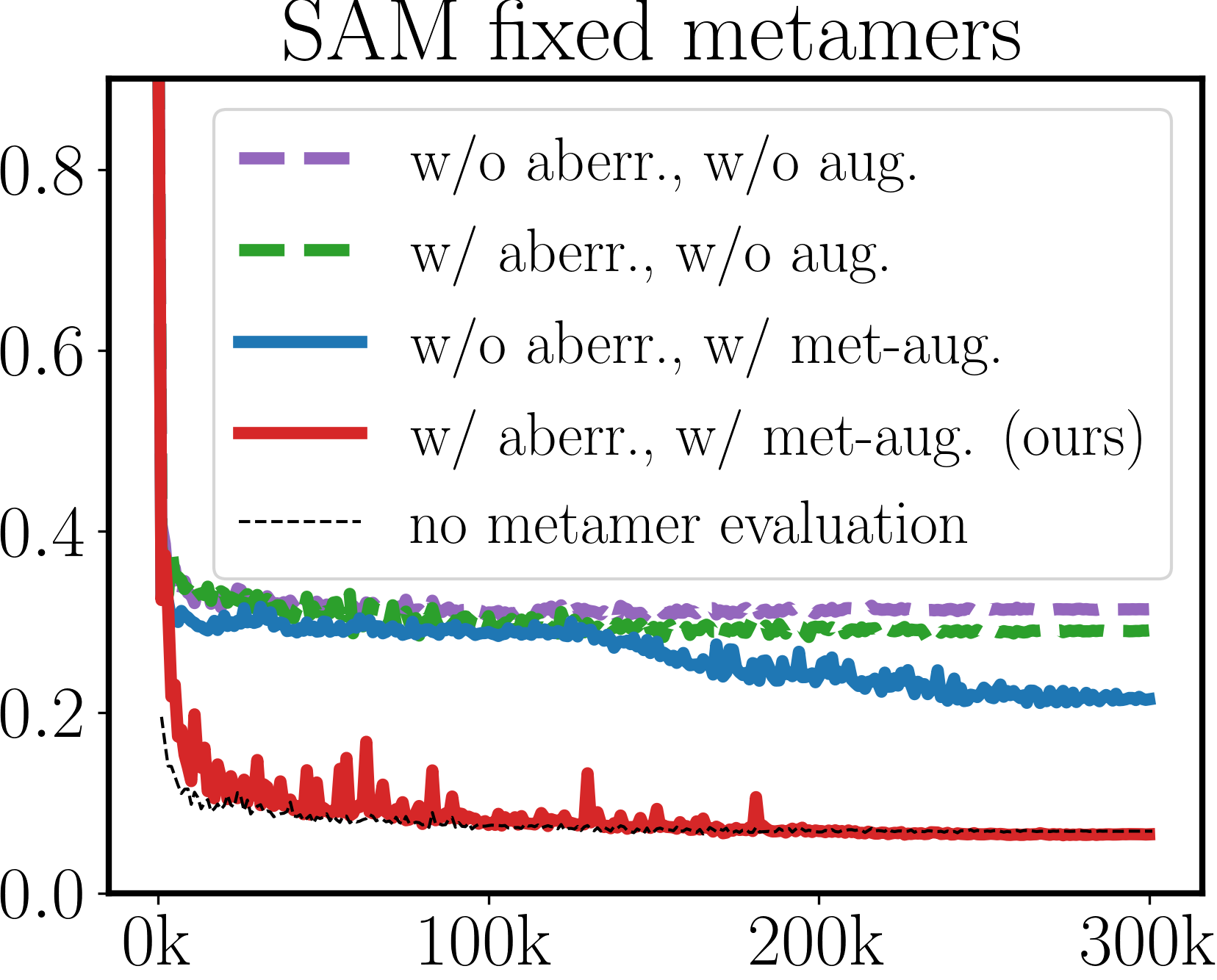}
    % SSTHyper on-the-fly
    &\includegraphics[width=0.49\columnwidth]{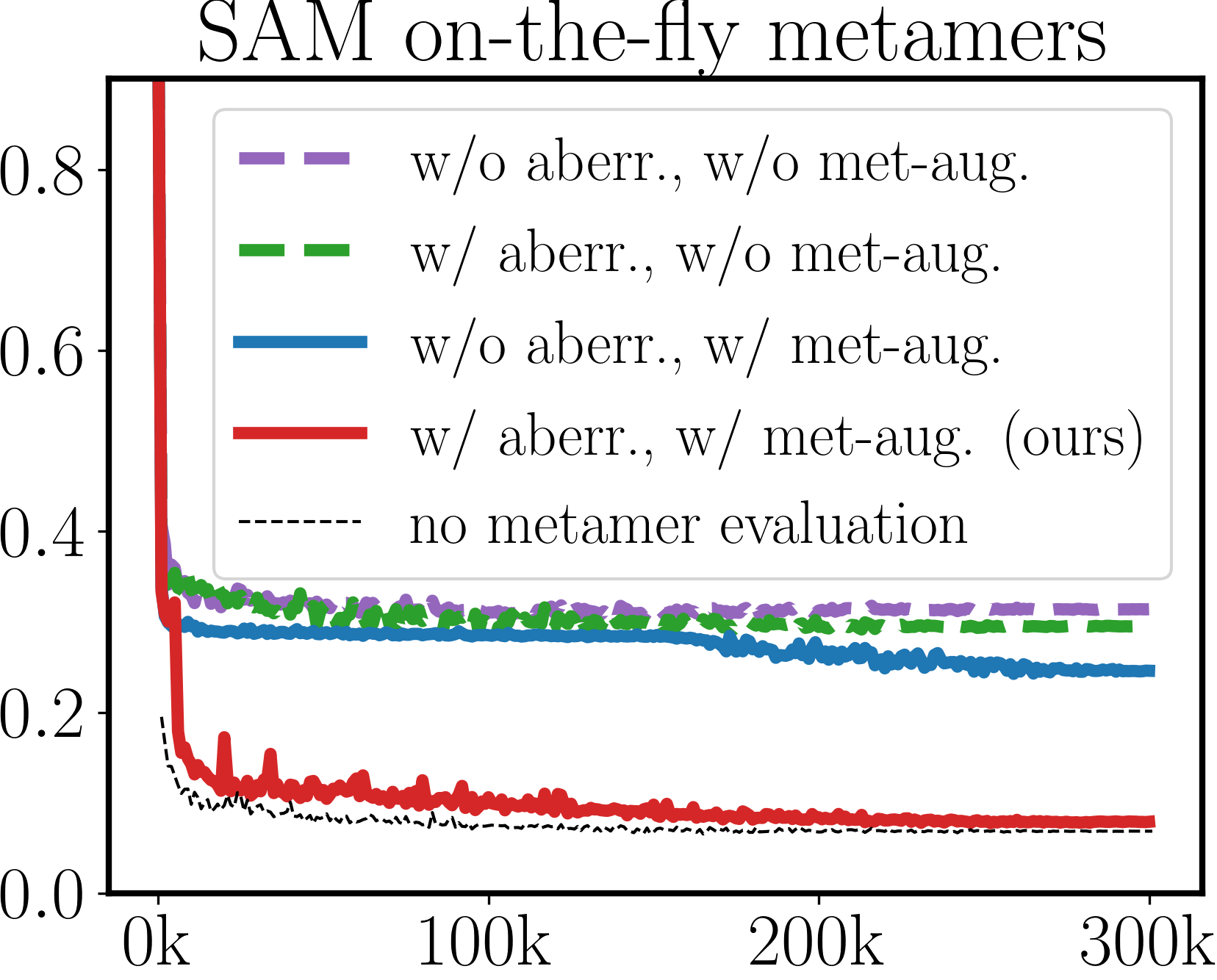}
    % MSFN fixed
    &\includegraphics[width=0.49\columnwidth]{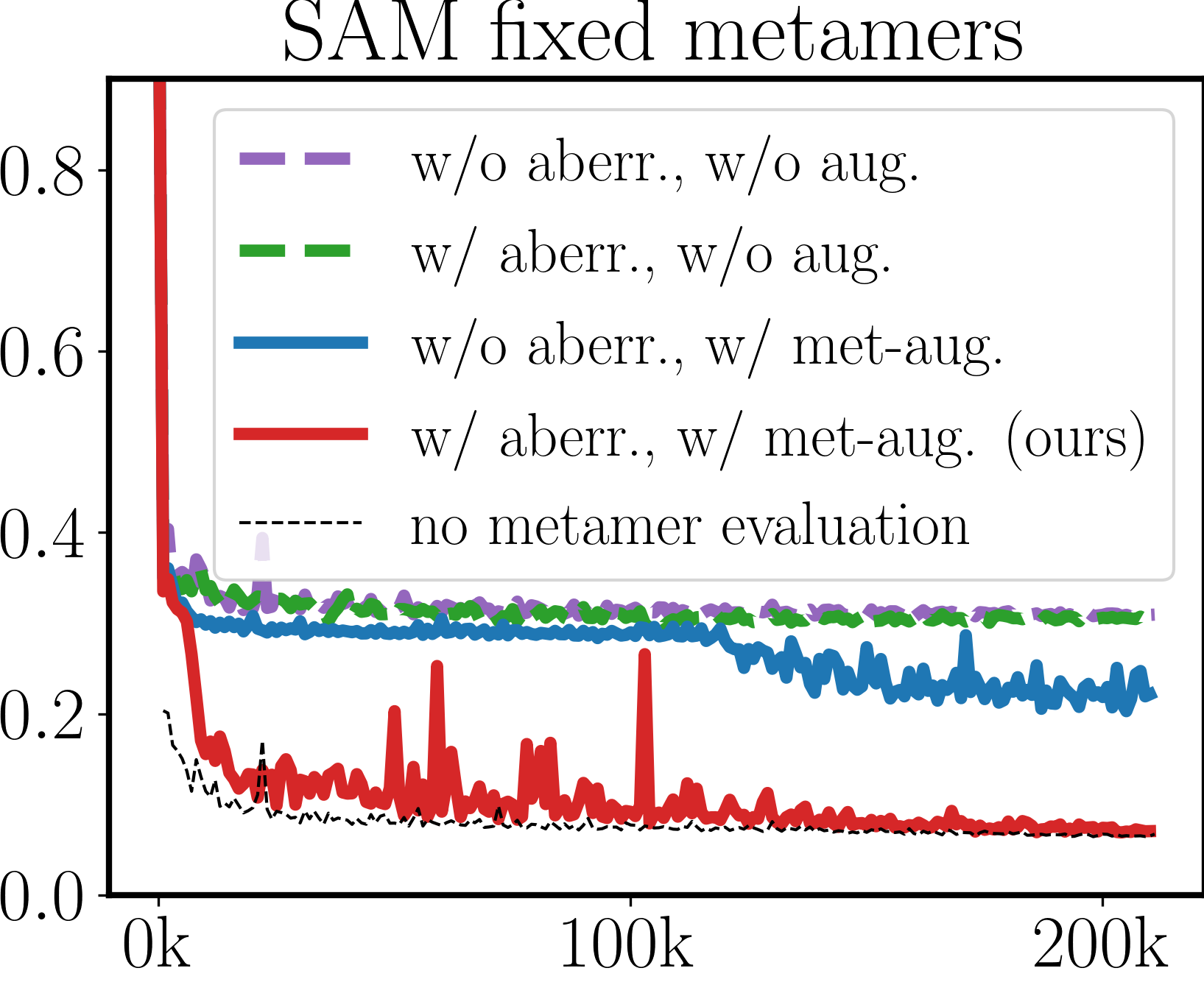}
    % MSFN on-the-fly
    &\includegraphics[width=0.49\columnwidth]{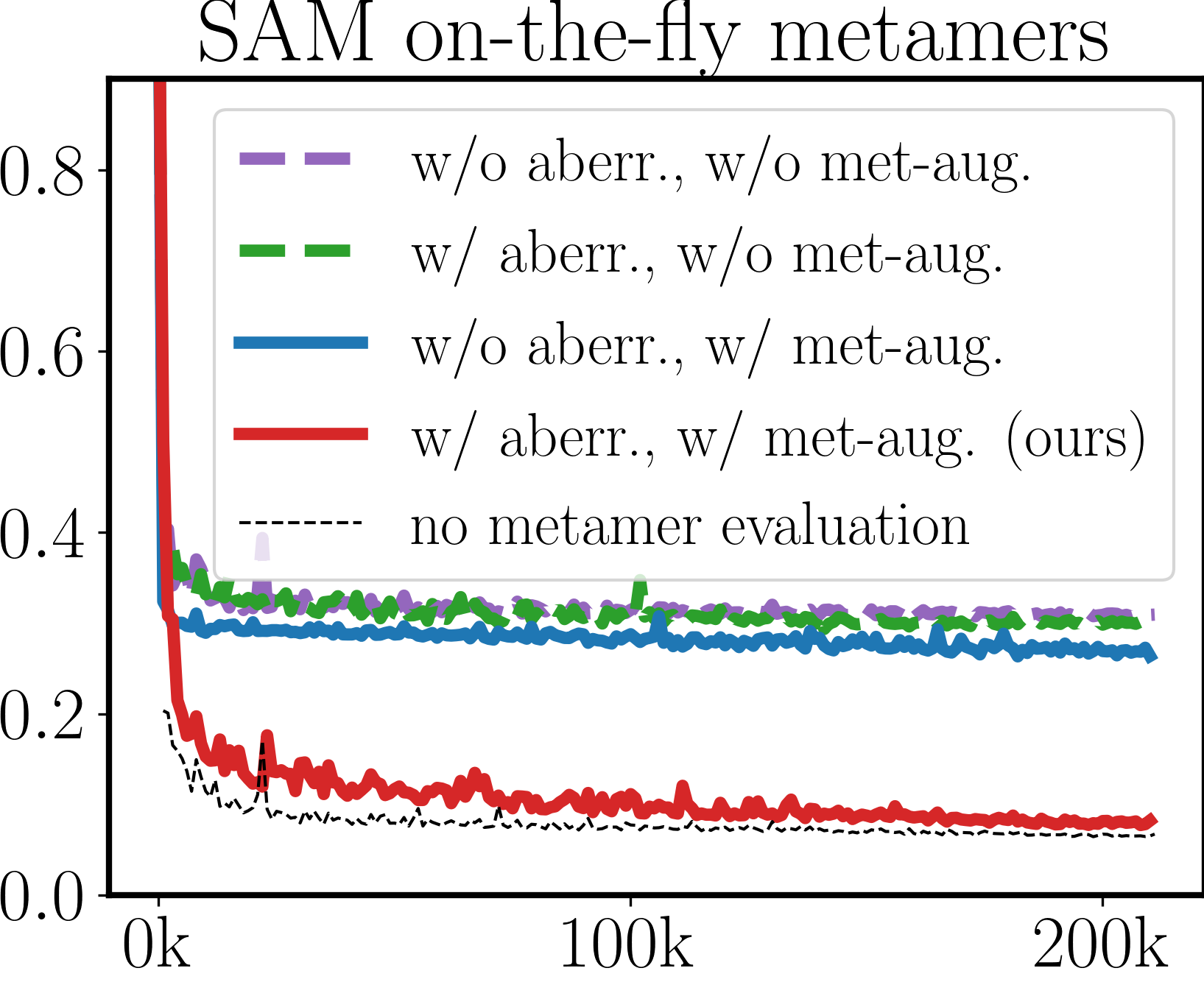} \\
    \multicolumn{2}{c}{GMSR} & \multicolumn{2}{c}{SSRNet} \\
    % GMSR fixed
    \includegraphics[width=0.49\columnwidth]{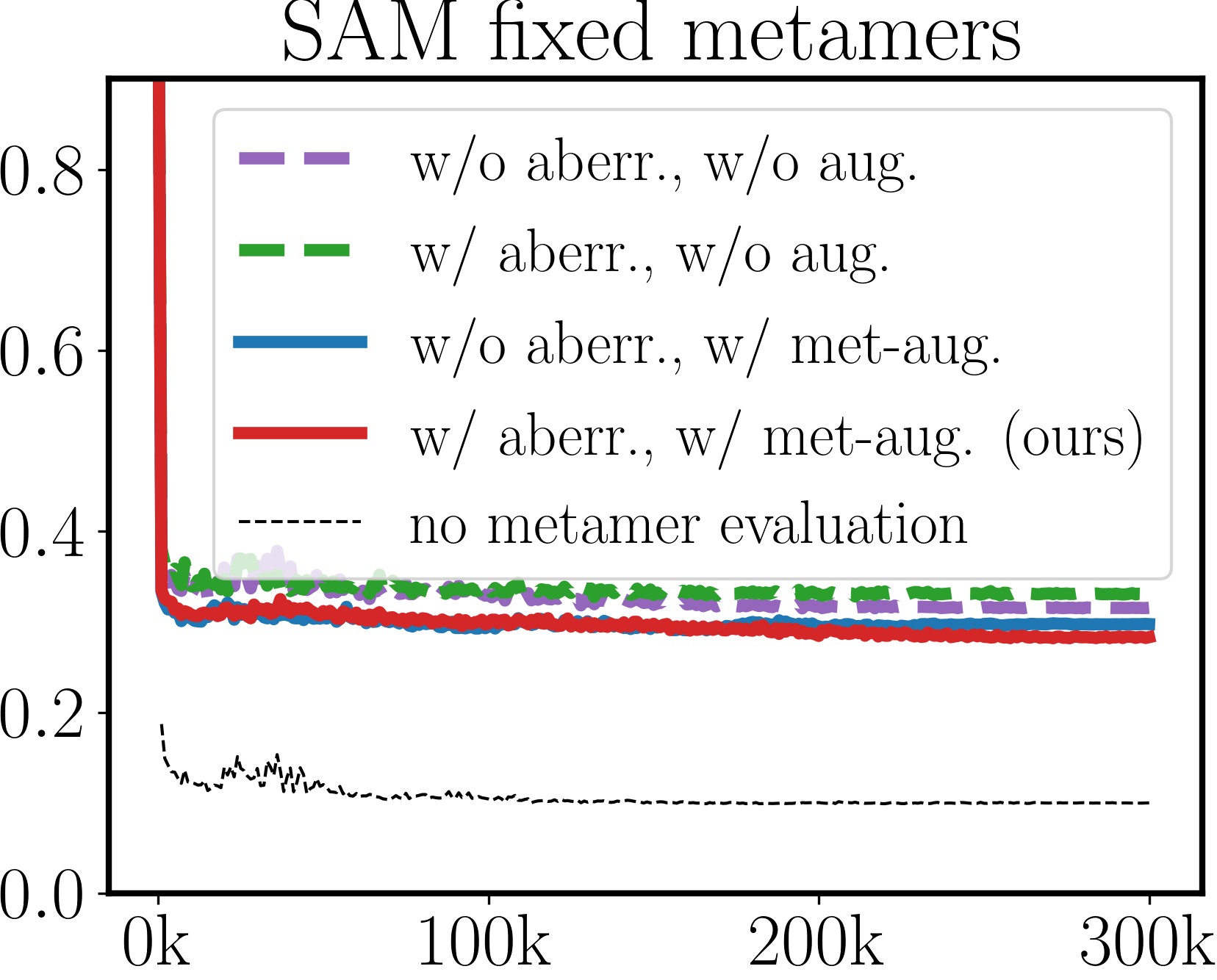}
    % GMSR on-the-fly
    &\includegraphics[width=0.49\columnwidth]{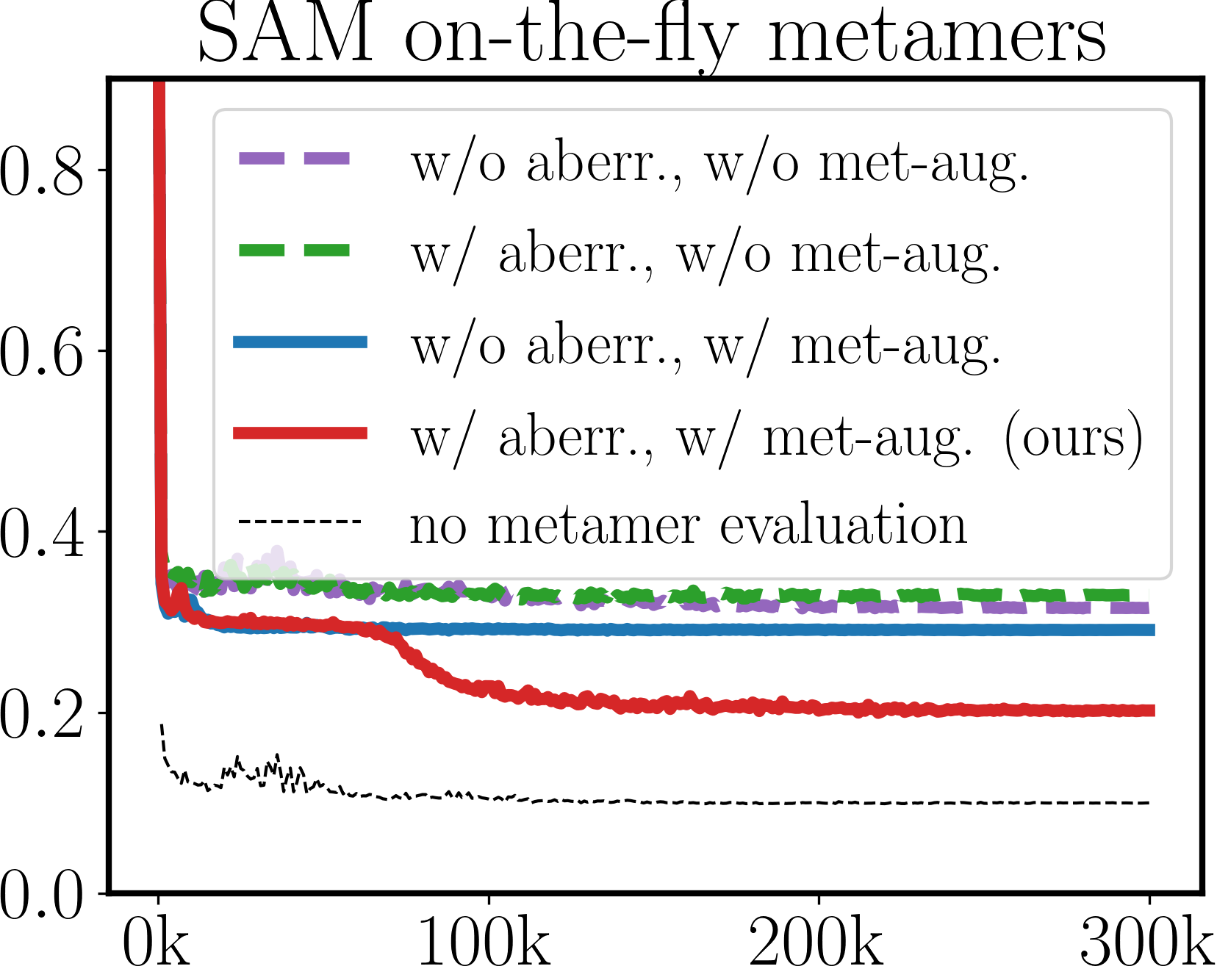}
    % SSRNet fixed
    &\includegraphics[width=0.49\columnwidth]{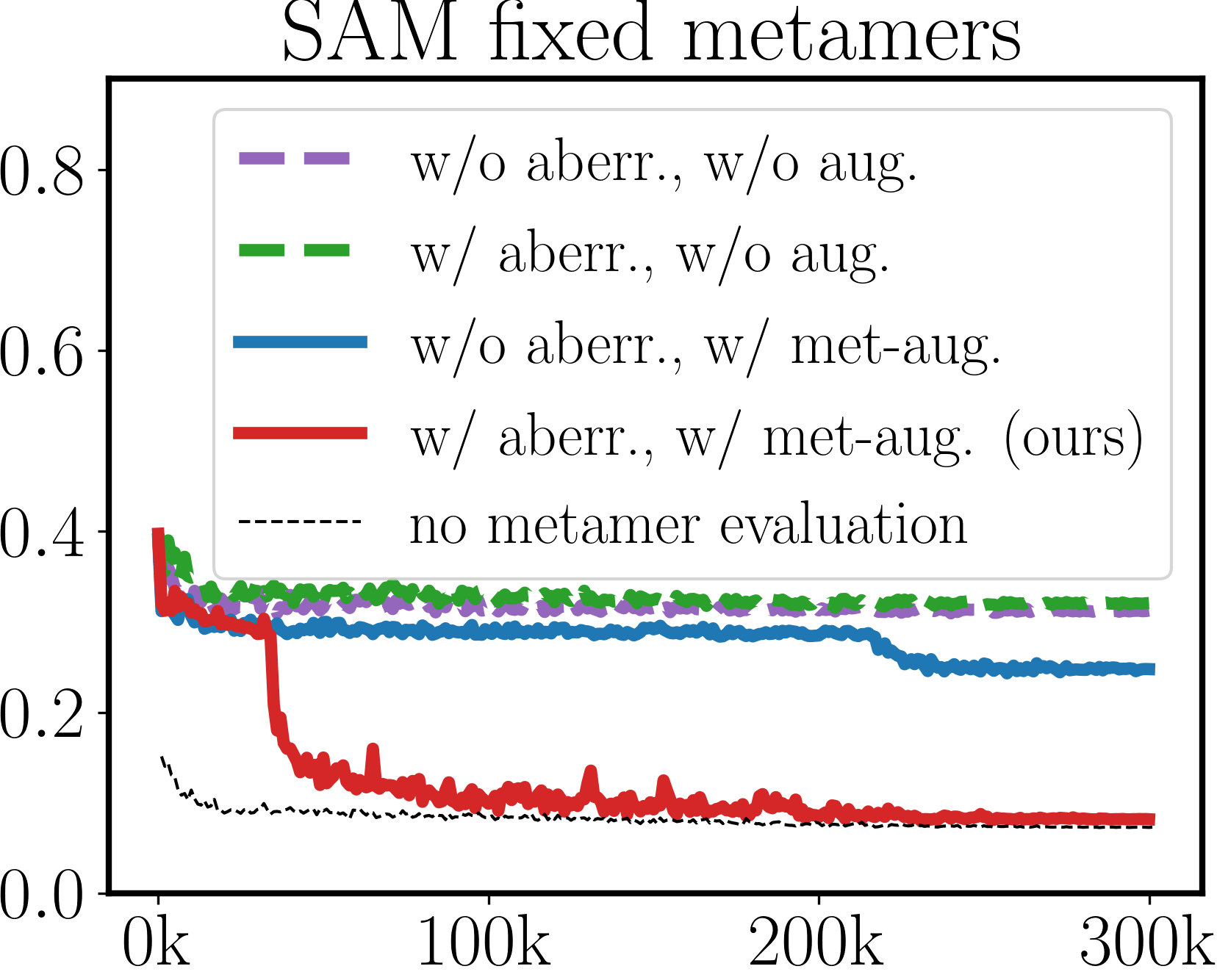}
    % SSRNet on-the-fly
    &\includegraphics[width=0.49\columnwidth]{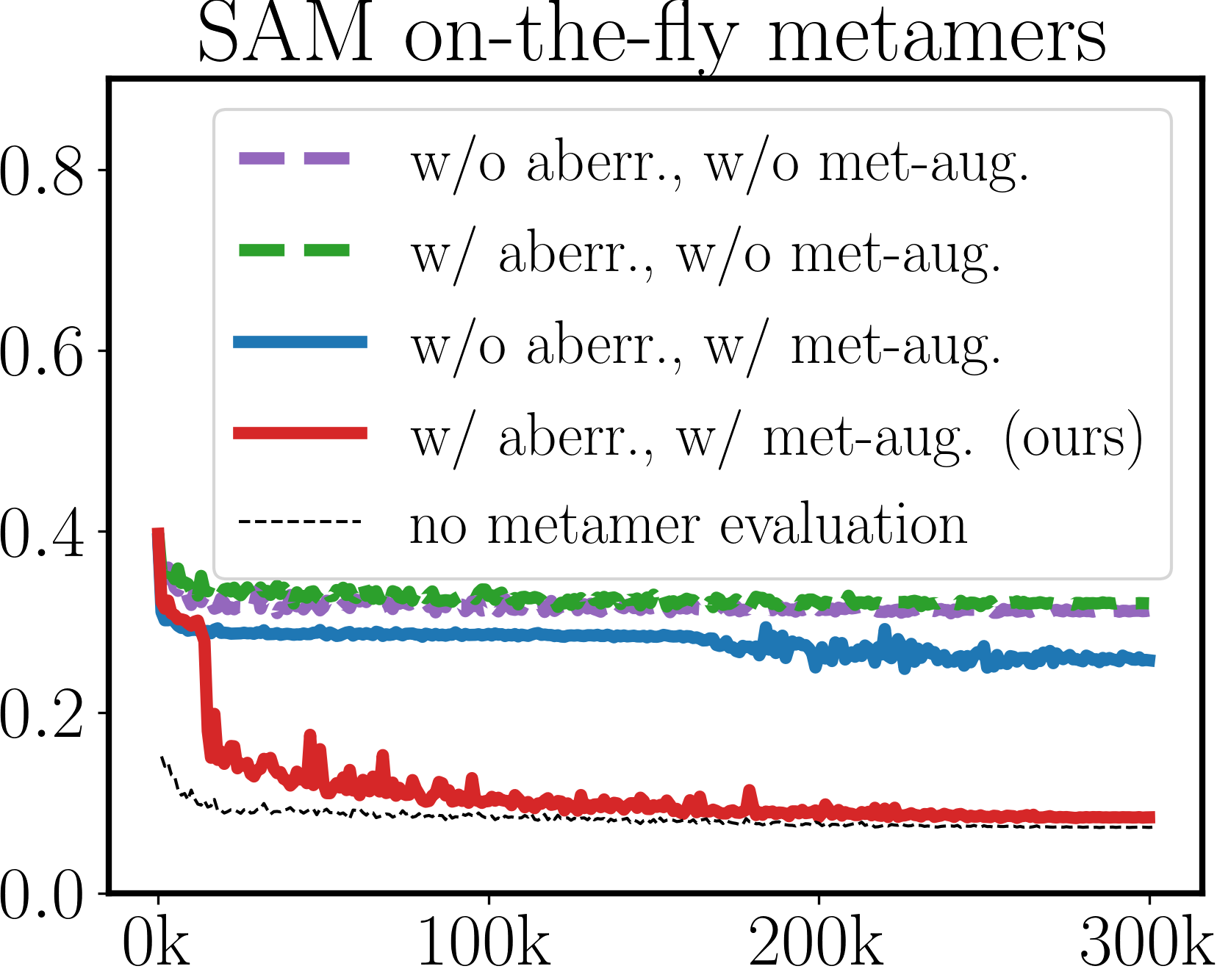} \\
    \end{tabular}
    \caption{Chromatic aberrations improve spectral accuracy for HySAT, HPRN*, SSTHyper, MSFN**, GMSR, and SSRNet. In each group, left: fixed metamers, and right: on-the-fly metamers. * For HPRN, we followed the training strategy and observed divergence after 100k iterations. Converged results are reported here at 100k iterations. ** For MSFN, it takes $>$72h to reach 200k, but we have observed convergence, so the results are reported at 200k iterations.}
    \label{fig:supp_aberration_advantage3}
\end{figure*}

\begin{figure*}[!htp]
    \centering
    \begin{tabular}{cccc}
    \multicolumn{2}{c}{ARAD1K} & \multicolumn{2}{c}{CAVE} \\
    \includegraphics[width=0.49\columnwidth]{supp_figures/fixed_metamer_aberrations_mst_plus_plus_sam.png} 
    & \includegraphics[width=0.49\columnwidth]{supp_figures/onthefly_metamer_aberrations_mst_plus_plus_sam.png}
    & \includegraphics[width=0.49\columnwidth]{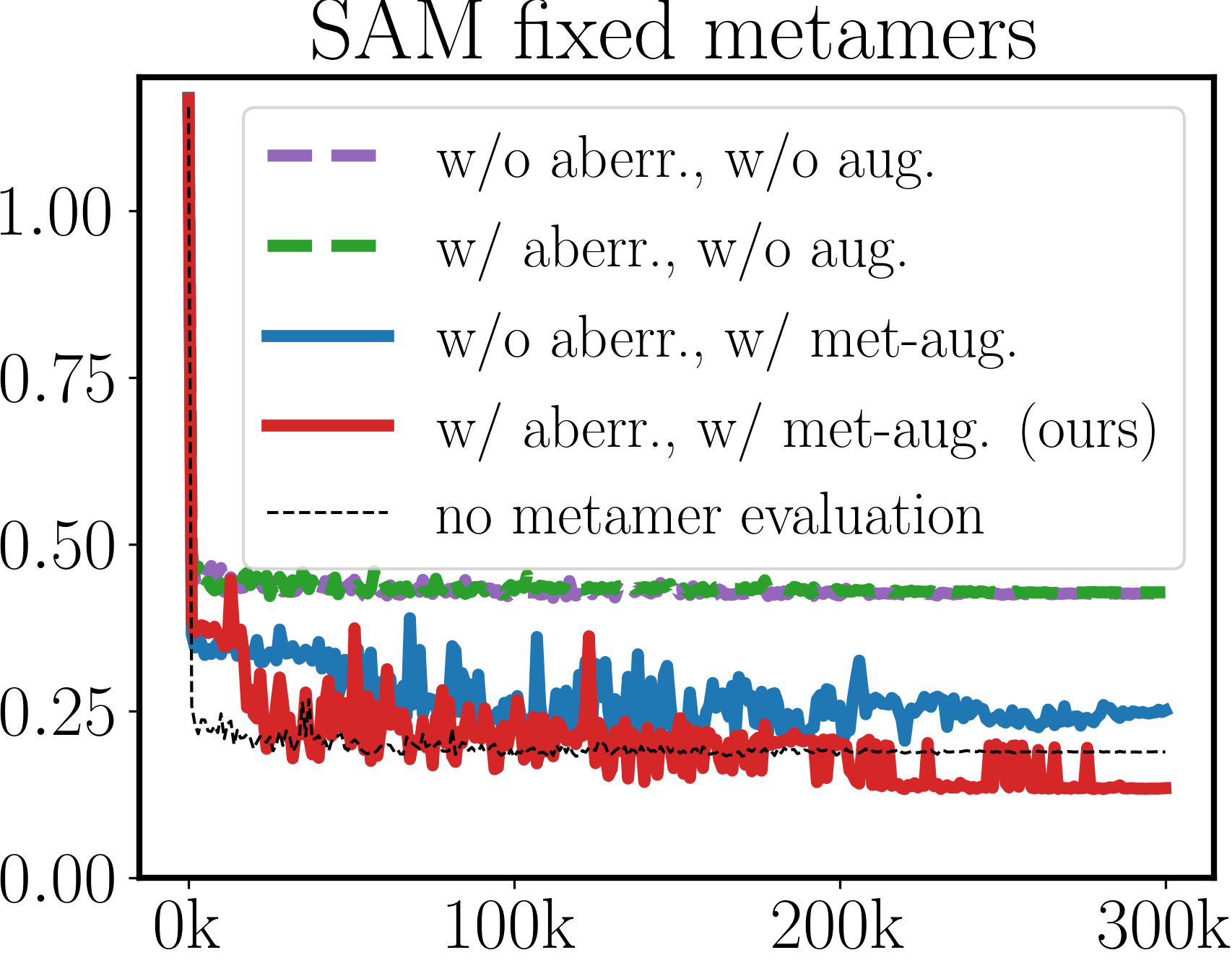}
    & \includegraphics[width=0.49\columnwidth]{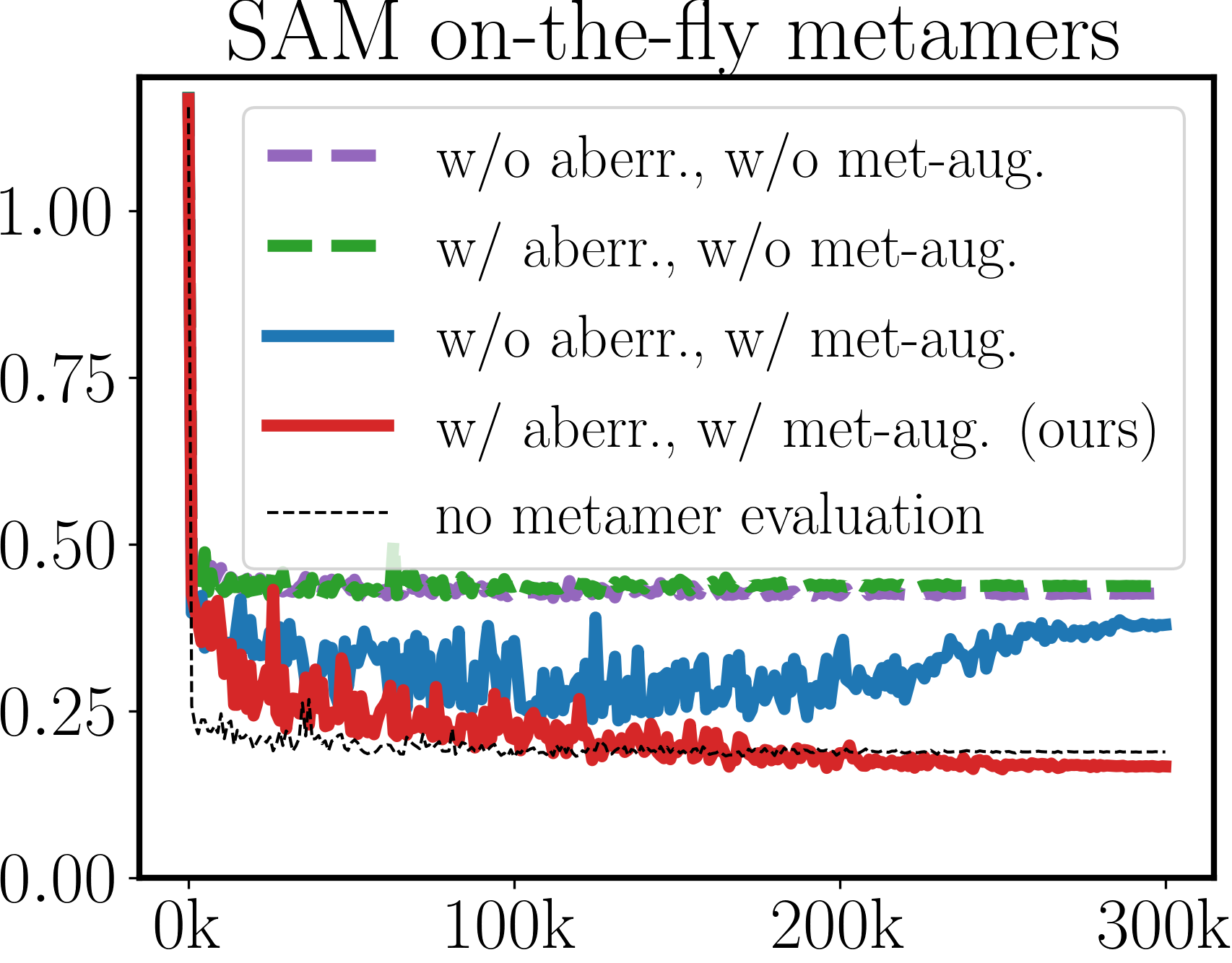} \\
    \multicolumn{2}{c}{ICVL} & \multicolumn{2}{c}{KAUST} \\
    \includegraphics[width=0.49\columnwidth]{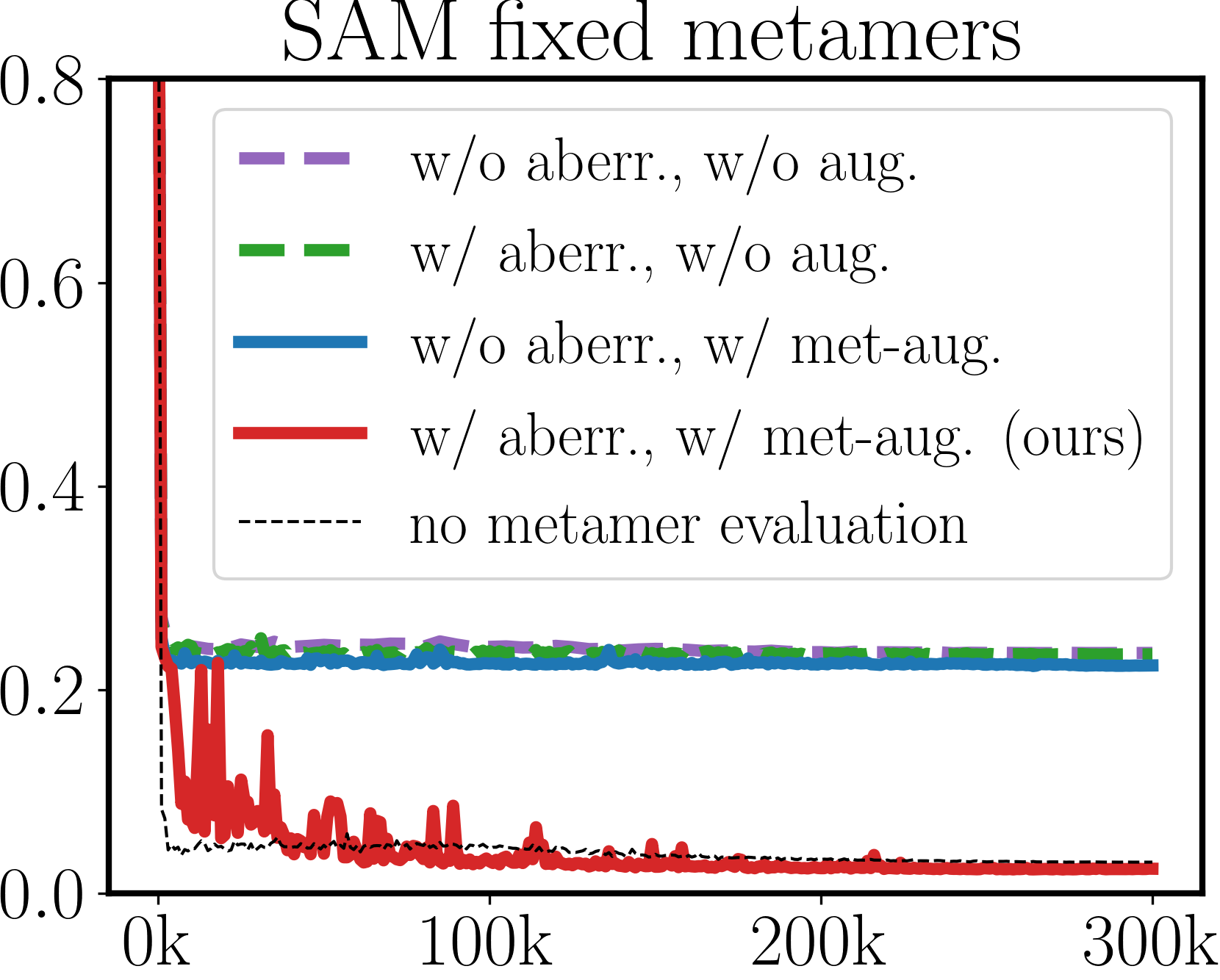} 
    & \includegraphics[width=0.49\columnwidth]{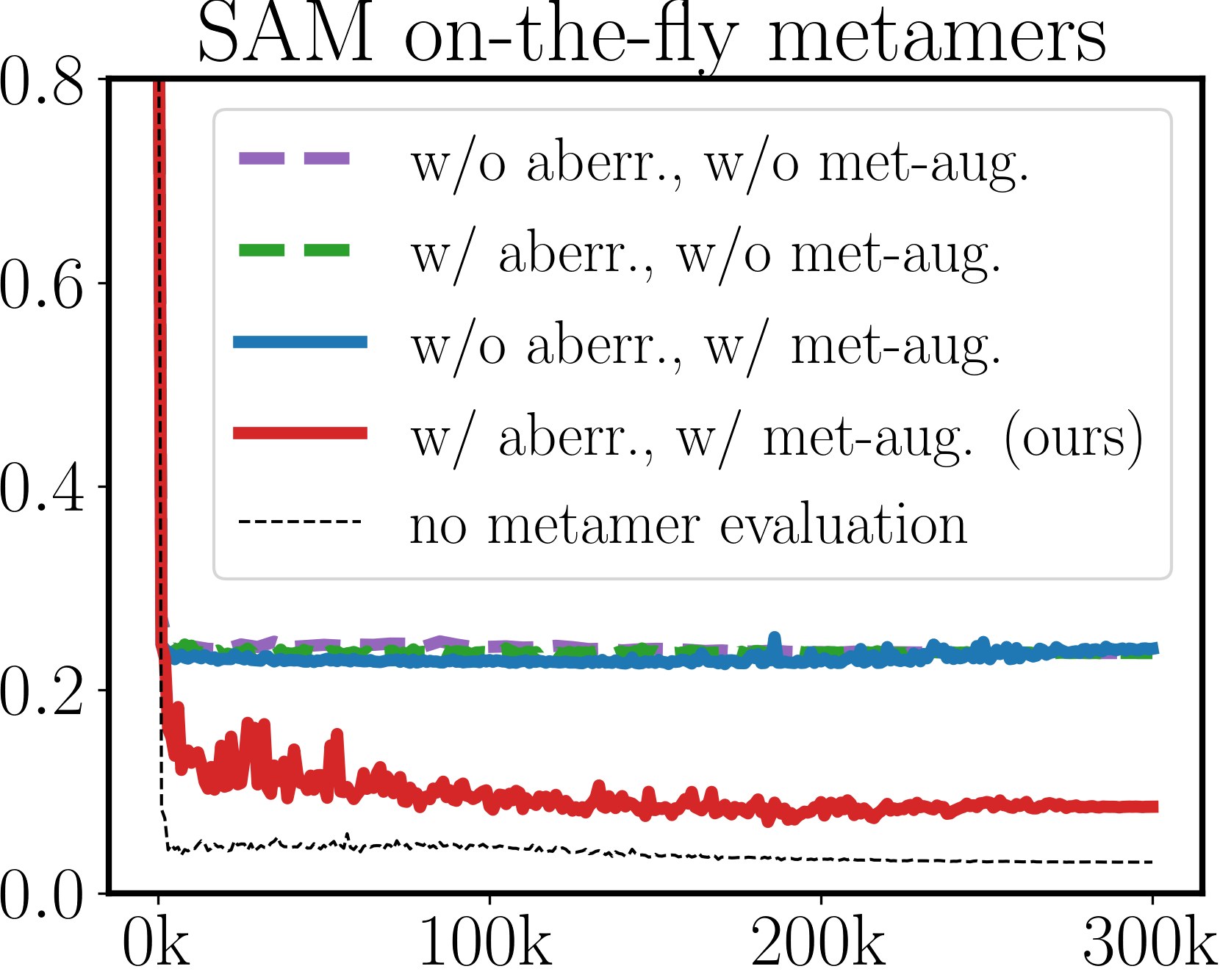}
    & \includegraphics[width=0.49\columnwidth]{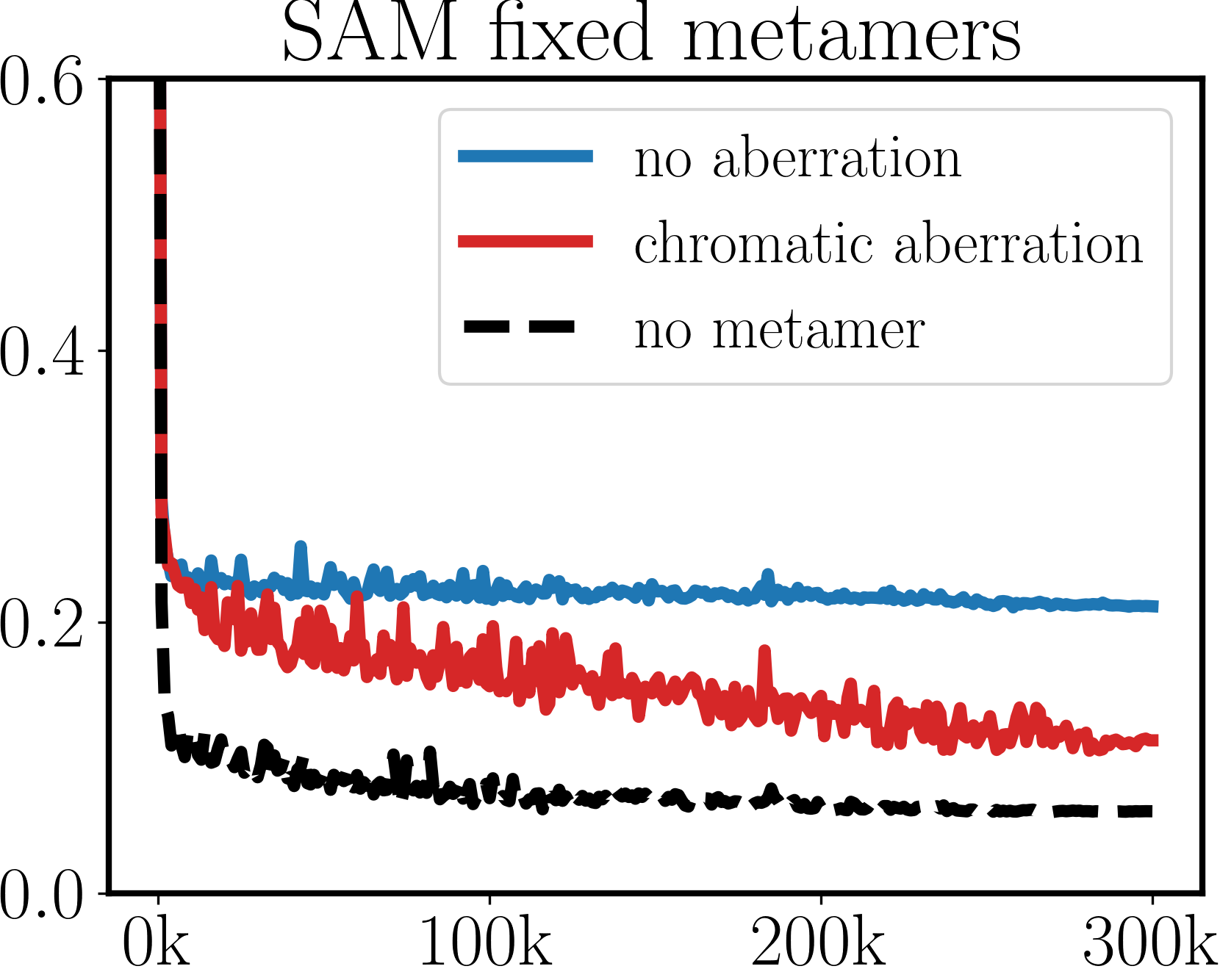}
    & \includegraphics[width=0.49\columnwidth]{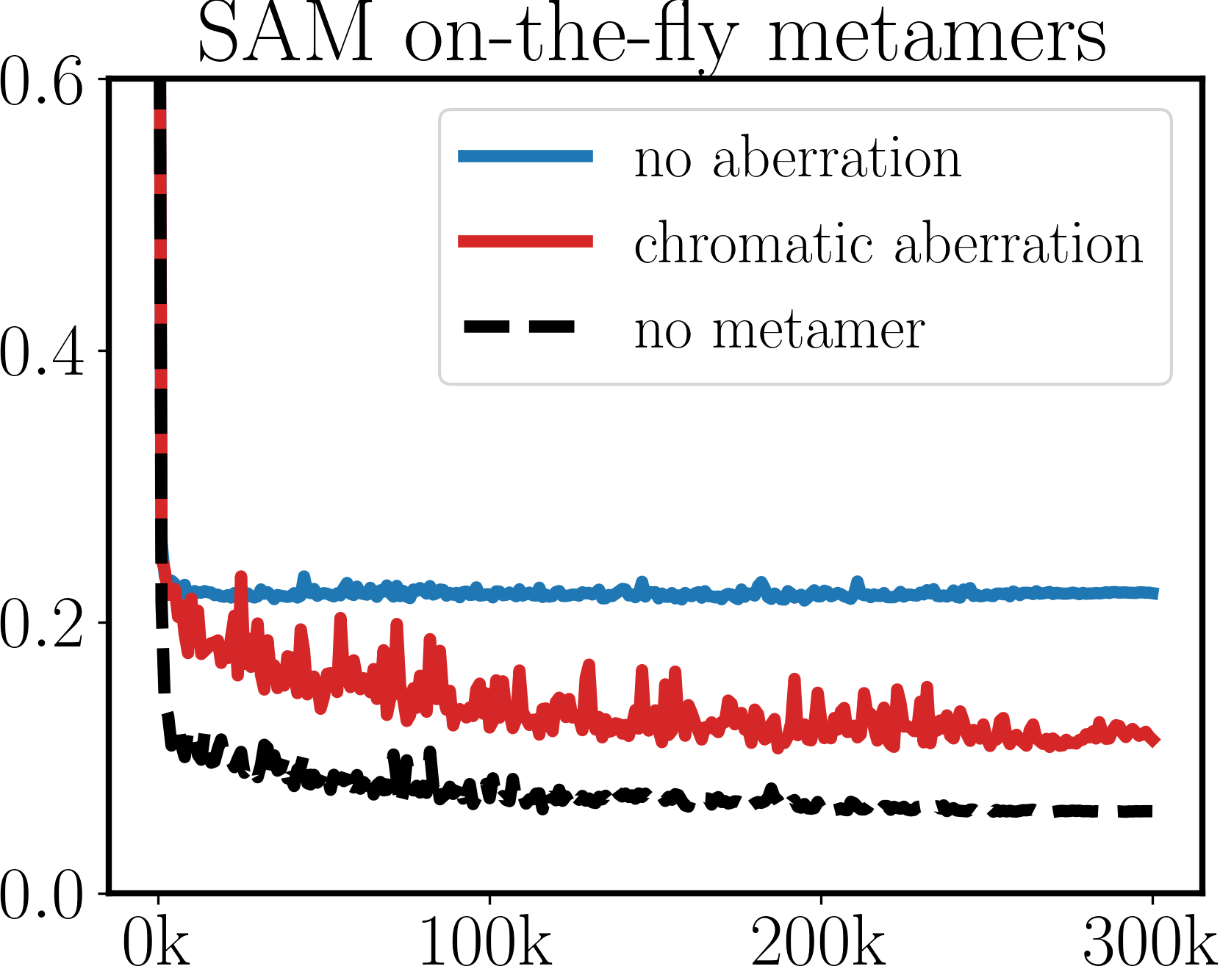} \\
    \end{tabular}
    \caption{The aberration advantage results for MST++ on all datasets. In each group, left is for fixed metamers, and right is for on-the-fly metamers.}
    \label{fig:supp_aberration_advantage_all_datasets}
\end{figure*}

%---------- main texts ----------%

\bibliographystyle{plain}
\bibliography{main}